\pdfoutput=1

\documentclass[11pt]{article}

\usepackage[final]{acl}

\usepackage{times}
\usepackage{latexsym}

\usepackage[T1]{fontenc}

\usepackage[utf8]{inputenc}

\usepackage{microtype}

\usepackage{inconsolata}

\usepackage{graphicx}

\usepackage{latexsym}
\usepackage{amsmath}
\usepackage{graphicx}
\usepackage{amssymb}
\usepackage{bbm}
\usepackage{booktabs}
\usepackage{multirow}
\usepackage[inkscapelatex=false]{svg}
\usepackage{adjustbox}
\usepackage{arabtex}
\usepackage{utf8}
\usepackage{subcaption}
\setcode{utf8}

\renewenvironment{abstract}%
         {\centerline{\large\bf Abstract}%
          \begin{list}{}%
             {\setlength{\rightmargin}{0.6cm}%
              \setlength{\leftmargin}{0.6cm}}%
           \item[]\ignorespaces}%
         {\unskip\end{list}}

\title{Are Knowledge and Reference in Multilingual Language Models Cross-Lingually Consistent?}

\author{
\textbf{Xi Ai  \thanks{Equal contributions.} \textsuperscript{1}},
\textbf{Mahardika Krisna Ihsani \footnotemark[1] \thanks{Works conducted during the internship in WING@NUS} \textsuperscript{2}},
\textbf{Min-Yen Kan \textsuperscript{1}}
\\
\textsuperscript{1} Web IR / NLP Group (WING), National University of Singapore
\\
\textsuperscript{2} Mohamed bin Zayed University of Artificial Intelligence (MBZUAI)
\\
\texttt{barid.x.ai@gmail.com}, \texttt{mahardika.ihsani@mbzuai.ac.ae}, \texttt{kanmy@comp.nus.edu}
}

\begin{document}
\maketitle
\begin{abstract}
Cross-lingual consistency should be considered to assess cross-lingual transferability, maintain the factuality of the model knowledge across languages, and preserve the parity of language model performance. We are thus interested in analyzing, evaluating, and interpreting cross-lingual consistency for factual knowledge.
To facilitate our study, we examine multiple pretrained models and tuned models with code-mixed coreferential statements that convey identical knowledge across languages. Interpretability approaches are leveraged to analyze the behavior of a model in cross-lingual contexts, showing different levels of consistency in multilingual models, subject to language families, linguistic factors, scripts, and a bottleneck in cross-lingual consistency on a particular layer. Code-switching training and cross-lingual word alignment objectives show the most promising results, emphasizing the worthiness of cross-lingual alignment supervision and code-switching strategies for both multilingual performance and cross-lingual consistency enhancement. In addition, experimental results suggest promising result for calibrating consistency in the test time via activation patching.  
\end{abstract}
\section{Introduction}
\begin{figure*}[ht]
\centering
\includegraphics[trim=0cm 1.3cm 0cm 1.3cm,clip=true,width=0.8\linewidth]{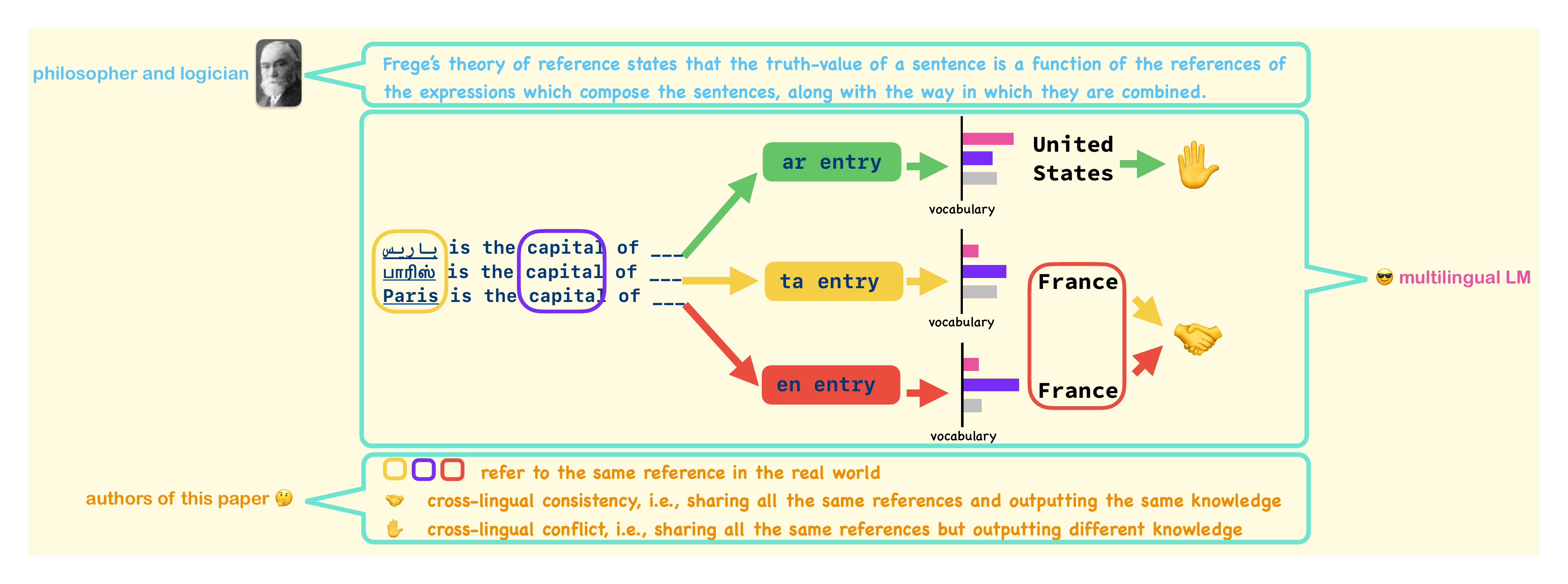}
\caption{\label{kc-illust} Illustration of Cross-lingual knowledge consistency. Frege’s theory of reference defines the reference of a sub-sentential expression as the object singled out by the name. A salient aspect of humanity is that, they can understand knowledge based on references regardless of language.} 
\end{figure*} 
 Frege’s theory of reference \citep{frege1892sense} indicates that the knowledge conveyed by a sentence depends on the references of the expressions that make up the sentence. A salient aspect of humanity is that, while people may speak different languages, they can share common references and knowledge. Thus, references and knowledge must be consistent across languages, and a multilingual model serving as a knowledge base \citep{x-fact,mLAMA,hu2024towards} should provide consistent knowledge when consulted in different languages. Not only does this theory contribute to cross-lingual performance and maintain knowledge between languages, but it also ensures parity and self-consistency of model performance \citep{god_consistency_paper,wang2023selfconsistency}. This motivates us to evaluate the knowledge consistency of multilingual language models in all languages when sharing the same references. 

 Few recent works \citep{mLAMA,fierro-sogaard-2022-factual,qi-etal-2023-cross} focused on translation pair consistency and reported that multilingual models may output knowledge for a particular query that differs with knowledge obtained from the query's translation. We argue that multilingual models show different language biases, leaving a non-trivial confounding factor when evaluating consistency with translation pairs. We hypothesize that for a consistent multilingual model, references, regardless of the surface language, provide energy to constrain the degree of freedom in knowledge recalling. To evaluate this hypothesis (Figure \ref{kc-illust}), we examine the difference in the output distribution between the original monolingual statement and the corresponding code-mixed coreferential statements, which takes a different angle and is orthogonal to existing works \citep{mLAMA, qi-etal-2023-cross} that rely on translation pairs and the output candidates. This examination explicitly instantiates Frege’s theory of reference to check the consistency of knowledge across languages that the same references for sub-sentential expressions, e.g., entities, should result in the identical knowledge. We attempt to answer three questions: \textit{1)} do multilingual language models recall factual knowledge for the coreferential statements in a similar manner, \textit{2)} how does the mechanism of multilingual language models work on the incorporation between entities or references to convey knowledge in cross-lingual settings, and \textit{3)} which factors prevent model consistency in multilingual settings? 

 In addition to model consistency in cross-lingual settings, our study is related to a broader linguistic phenomenon of entity-level code-switching and language interference: an entity code-switches between two languages without changing the reference, as we create code-mixed coreferential statements from monolingual statements by substituting a subject entity with an equivalent one in another language that shares the same reference. More recently, we share a similar goal with knowledge incorporation and editing \citep{beniwal-etal-2024-cross,li2024unveiling}, since we incorporate a coreferential entity from other languages to recall factual knowledge in cross-lingual settings. Our main findings are as follows.
  
\begin{itemize}
    \item We present a code-mixed coreferential task to observe implicit consistency across languages within a sentence. In our experiments, observations and findings are also transferable to explicit cross-lingual consistency across translation pairs. 
    \item We discover consistency bottlenecks and issues tied to language characteristics, scripts, and training biases through layer-wise analyses and interpretability approaches, which potentially prevent cross-lingual consistency improvements and gains from scaling.
    \item There is a partial causality from adding language biases (of high-resource languages) to improving cross-lingual knowledge consistency. Directly adding bias via representation patching could be a potential method to calibrate consistency in the test time.
    \item Shared language scripts contribute to cross-lingual consistency, especially for encoder and decoder models, but it is not a necessary condition to achieve it. Reducing script overlaps by expanding vocabulary size slightly improves the consistency yet it helps to improve the consistency for some low-resource languages. 
    \item Cross-lingual supervision can alleviate the consistency bottleneck to enhance alignments between coreferential entities, which can be achieved by training with an explicit alignment objective or a code-switching objective. On the other hand, parallel samples providing cross-lingual generalization supervision offer limited gains to consistency.
\end{itemize}
Our contribution is to offer an understanding of multilingual language models' limitations under cross-lingual settings and highlight potential research directions to address such issues.

\section{Methodology}
\subsection{Task Definition}
We focus on a code-mixed, generative task that forces the multilingual model to condition on coreferential entities across languages to recall a factual answer from its internal knowledge base\footnote{See limitation in \S \ref{sssec:limitation}.}. We show an example in Figure \ref{kc-illust} where en entry "Paris is the capital of \_\_\_" is evaluated with its possible code-mixed coreferential statements (ar entry \& ta entry). Readers can refer to Appendix \S \ref{apx_probabilistic_task_definition} and \S \ref{apx_input_format}  for details and implementations. 

Let $I = \{S^{l1},\cdots,O,\cdots\} \in {l1}$\footnote{The surface structure is not restricted. We use the common subject--object structure as an example.} be a statement, where $l1$ stands for matrix language (the predominant language), $S^{l1}=\{s_1,\cdots,s_k\} \in l1$ are subject sub-tokens, and $O=\{o_1,o_2,\cdots,o_j\} \in l1$ denote object sub-tokens. This statement is used to format an input $I_{mono}$ by removing $O$ to elicit the internal knowledge $K^*_{\theta}$ and instruct the model to output the n-gram $Cand(O_{\in V}|I_{mono})$ over the model's vocabulary $V$, where $Cand(O_{\in V}|I_{mono}) = P(O_{\in V} | K^*_{\theta})$$P(K^*_{\theta}| S^{l1}, I_{\setminus (S^{l1} \cap O)})$. Similarly, we create a code-mixed coreferential statement $I_{cm}$ by replacing $S^{l1}$ with a coreferential subject $S^{l2}$ in the embedded language $l2$ to obtain $Cand(O_{\in V}|I_{cm}) = P(O_{\in V} |K_{\theta})$$P(K_{\theta}| S^{l2}, I_{\setminus (S^{l1} \cap O) })$. $I_{cm}$ and $I_{mono}$ with coreferential subjects $S^{l1}$ and $S^{l2}$ condition the model for recalling knowledge. To measure the knowledge consistency between $K^*_{\theta}$ and $K_{\theta}$, we calculate the difference between the output $Cand(O_{\in V}|I_{cm})$ and $Cand(O_{\in V}|I_{mono})$ as $K^*_{\theta}$ and $K_{\theta}$ provide energies to constrain the degree of freedom in generation. Additionally, we also evaluated the baseline setting of $I_{cm}$ by removing the subject entities to obtain the model's default outputs with no references for comparison.

We analyze the \textbf{consistency evolution} as the layer goes deeper to trace the consistency and understand the models' behavior. 
Specifically, we apply LogitLens \citep{lesswrong2024interpreting} for encoder and encoder-decoder models or DecoderLens \citep{langedijk2023decoderlens} for decoder models to computing the layer-wise output distributions from the layer representations, retrieving layer-wise $Cand(O_{\in V}|I_{cm})$ and $Cand(O_{\in V}|I_{mono})$.

\subsection{Metric Function and Interpretability}
\label{sssec:interpret-approaches}
Readers can refer to Appendix \S \ref{apx_consistency_metric} for more details, e.g., equations.
\paragraph{Output Distributions Consistency.}
Top@1 Accuracy and RankC (i.e., weighted Precision@5) \citep{qi-etal-2023-cross} are used to capture the difference between two output distributions, $Cand(O_{\in V}|I_{mono})$ and $Cand(O_{\in V}|I_{cm})$. In contrast to the previous works \citep{mLAMA,qi-etal-2023-cross}, we do not constrain the output candidate or domain. Instead, the output distribution over the full vocabulary is examined. Since the experimental results in Top@1 and RankC are similar, Top@1 are moved to the Appendix \S \ref{sess:findings_in_details}. 

 \paragraph{Cross-lingual Representations Similarity.} We hypothesize that cross-lingual generalization across languages results in cross-lingual consistency to some extent. To evaluate this hypothesis, we examine the contextualized representation similarity for our correferential statements by computing batch-wise CKA similarity scores \citep{kornblith2019similarity} between them over each layer.  


\paragraph{$IG^2$ Score.} We adapted $IG^2$ \citep{liu2024devil} to interpret the impact of each feed-forward neuron on the output where the higher the value is, the more critical the neuron is to predict the ground truth object. This examination is used to analyze the correlation between cross-lingual consistency and shared neurons across languages. 

\subsection{Dataset and model}
\paragraph{Dataset.} We use mLAMA dataset \citep{mLAMA} that provides parallel triples (object, predicate, subject) in 53 languages written in cloze, completion task format (e.g., “Paris is the capital of ”) to query knowledge in zero-shot settings. In our experiments, $l1$ is set to English for all pairs, and $l2$ is the other 52 languages to report an overall result, where $l2$ languages are categorized into two separate categories for each of the three factors (geographics, writing scripts, and language family) using ISO-639 language codes information from "localizely"\footnote{\url{https://localizely.com/language-code}}. For an in-depth analysis, we examine 2 similar $l2$ languages (De, Id) and 2 dissimilar $l2$ languages (Ar, Ta) to observe the consistency evolution from early layers to later ones\footnote{While Id does not belong to the same language family as En, it has many loanwords from En \citep{indodic_english_indonesian}. Ar and Ta are not considered as Indo-European languages and also do not use latin scripts.}.
\paragraph{Models.} We examine distinct model families: encoder models (xlm-r from 0.3B to 10B) \citep{Conneau2020}), encoder-decoder models (mT0 from 0.6B to 3.7B \citep{mt0_paper}, mT5 from 0.6B to 3.7B) \citep{xue2020mt5}), and decoder models (Llama3-instruct 1B \& 8B) \citep{grattafiori2024llama3herdmodels}). In our experiments, we obtain consistent findings across model families and sizes. Therefore, we show essential results in the main text and move the rest to the Appendix \S \ref{sess:findings_in_details}.

\section{Observing Consistency}
\subsection{Consistency on All Languages}
\begin{figure}[ht]
    \centering
        \includegraphics[ width=0.9\linewidth]{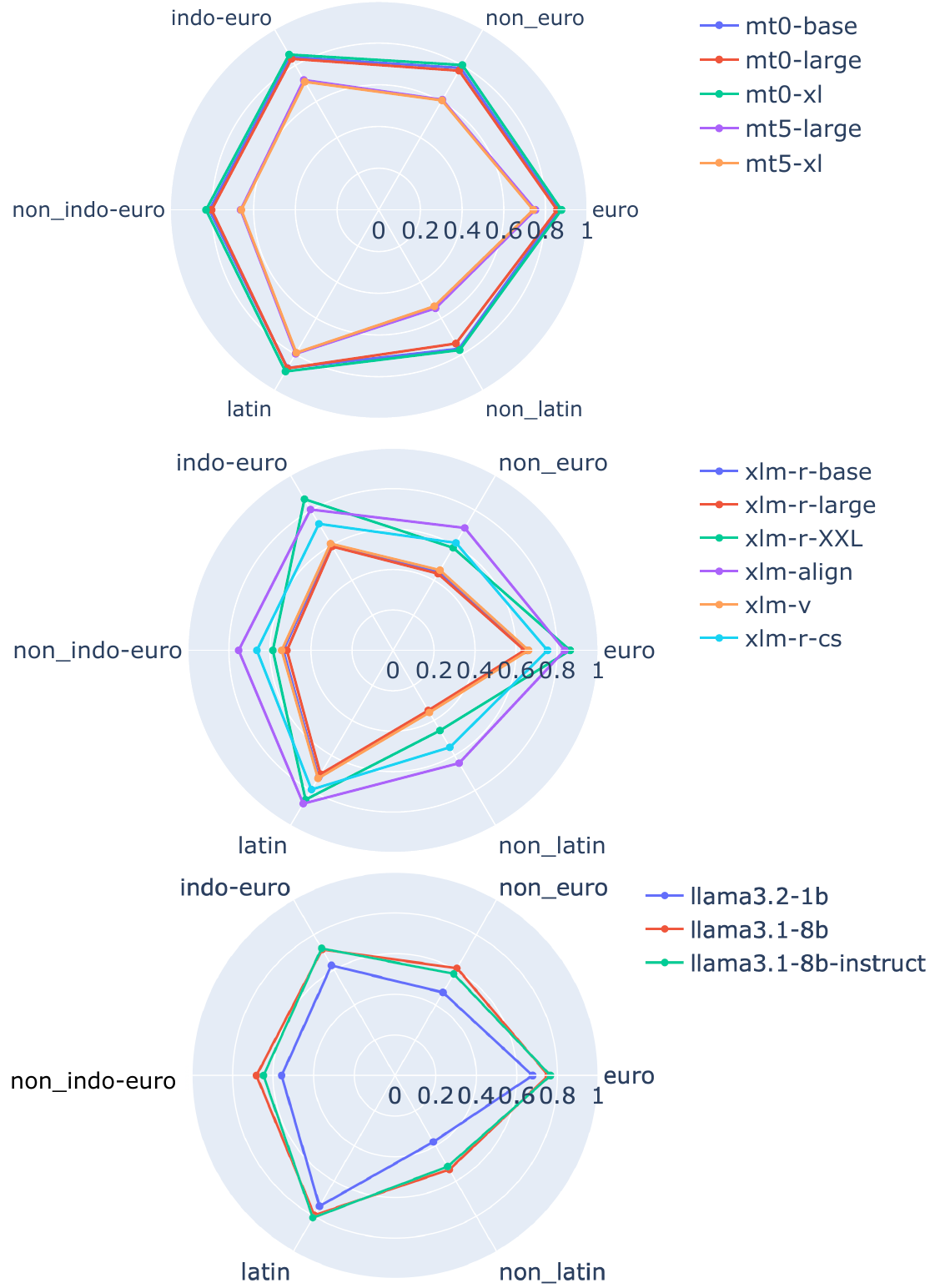}
    \caption{Cross-lingual consistency of output distribution in different model types (top: encoder-decoder, middle: encoder, bottom: decoder)  grouped by 3 factors (geographics: europe \& non\_europe, language family: indo-european \& non\_indo-european, writing scripts: latin \& non\_latin). ( \textit{cf.} \S \ref{sess:findings_in_details_Overall_Consistency}. )}
    \label{main_fig:overall-encoder-encoder-decoder-crosslingual-factors-rankc}
\end{figure}

From Figure \ref{main_fig:overall-encoder-encoder-decoder-crosslingual-factors-rankc}, dissimilar $l2$ tends to have lower consistency than those similar to $l1$ across all factors. The difference in writing scripts plays the most important role in both encoder and decoder models. However, surprisingly, encoder-decoder models are more tolerant to any kind of factors. Another intriguing finding is that geographic factor affects consistency, and this could be attributed to common culture and vocabulary \citep{zhao2024tracing}. On the other hand, we suppose that other linguistic factors contributing to cross-lingual performance, such as similarity in linguistic characteristics \citep{chronopoulou-etal-2023-language}, or borrowing \citep{tsvetkov2016cross}, could also affect cross-lingual knowledge consistency. However, such factors are difficult to quantify, leaving such analyses for future work. Note that language families and writing scripts have an impact on vocabulary, and we will confirm it in a later section. 

\subsection{Consistency Evolution across Layers}
\begin{figure*}[ht!]
        \centering
        \includegraphics[width=0.9\linewidth]{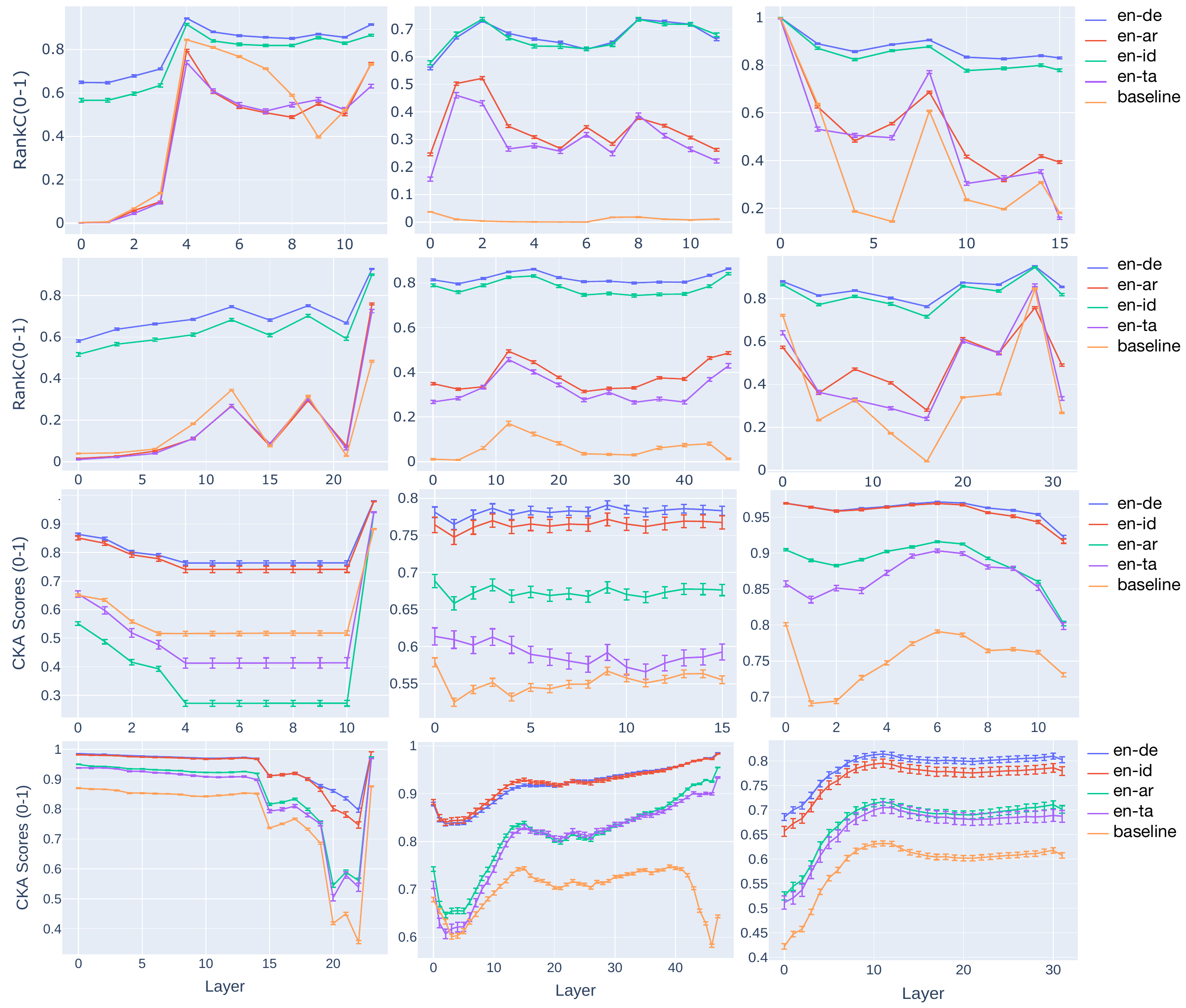}
    \label{main_fig:model-cka}
    \caption{Consistency evolution (1st and 2nd row: consistency score for small and large models, 3rd and 4th row: CKA score for small and large models) in different model types (L: encoder-decoder, M: encoder, R: decoder). For each model family, scaling models is not a promising strategy in general to mitigate consistency bottlenecks when observing 1st row vs 2nd row and 3rd row vs 4th row (except for the xlm-r-xxl CKA similarity). (\textit{cf.} \S \ref{sess:findings_in_details_Layer-wise_Consistency})}
    \label{main_fig:mt0-kc}
\end{figure*}

To better understand the cross-lingual consistency bottleneck, we examine the layer-wise consistency patterns in different model sizes and types, as presented in 1st and 2nd Row of Figure \ref{main_fig:mt0-kc}. For encoder and encoder-decoder models, the noticeable difference lies in the initial consistency, whereby dissimilar language pairs have low consistency scores. The consistency gap between dissimilar and similar languages starts to close at some specific layer while widening again later. Meanwhile, for decoder models, the pattern is more distinct, where there is a consistent degradation for smaller model in dissimilar language pairs and baseline, as for the larger model, it interestingly manages to recover the consistency starting from middle layer yet we can notice a bottleneck in last layer. This observation provides evidence for empirical studies that scaling benefits downstream task performance \citep{Conneau2020}, for example, XNLI, but offers limited gain for cross-lingual consistency, as we can observe in Figure \ref{main_fig:overall-encoder-encoder-decoder-crosslingual-factors-rankc}.

Layer-wise analyses help us to understand the model behaviors. However, the question remains as to why such behaviors could happen. To answer that question, we analyze contextualized representation similarity across layers from the 3rd and 4th Row of Figure \ref{main_fig:mt0-kc}, which shows different patterns from the cross-lingual consistency. In general, for encoder-decoder and encoder models, there is a degradation of similarity scores until the middle layer (except for xlm-r-base, where the growth is slightly fluctuating). In contrast, the small decoder model shows more stable similarity over the layers, and there is also a monotonic increase until the middle layer for the larger decoder model. This finding suggests that the cross-lingual representation similarity improved via model scaling might be a necessary condition rather than a sufficient condition to achieve cross-lingual consistency. Some other factors, such as isotropy and contextualization \cite{ethayarajh-2019-contextual}, might impact cross-lingual consistency other than the cross-lingual representation similarity. In addition, dissimilar languages have low similarity scores that are quite similar to the baseline setting, which is also observed in the layer-wise consistency scores. 

\subsection{Correlation and Interpretability}
\begin{figure*}[ht]
    \centering
\includegraphics[width=0.9\linewidth]{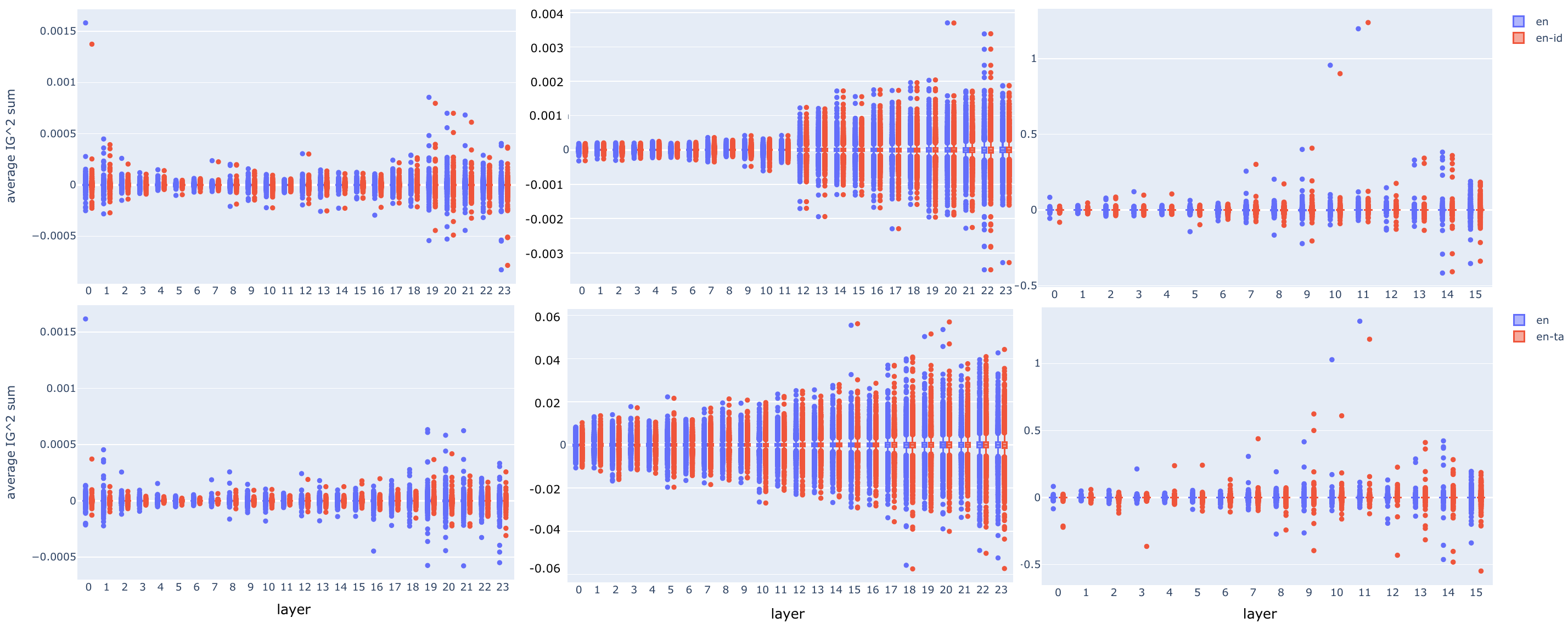}
    \caption{$IG^2$ scores in across different model types (L: encoder-decoder, M: encoder, R: decoder) for en--id (1st row) and en--ta (2nd row). The distribution is more contrastive on dissimilar languages (en--ta) than the similar languages (en--id). (\textit{cf.} \S \ref{sess:findings_in_details_Feed-Forward Neurons}. )}
    \label{main_fig:mt0-ffn-bias-de-ta-en}
\end{figure*}

To understand the model behaviors,  we analyze the contribution of every neuron within MLP on the correferential statements based on findings from \citep{geva2020transformer}. Specifically, we inspect the $IG^2$ scores of all the feed-forward neurons at all the layers. Our analysis for this factor could show a moderate correlation with the cross-lingual consistency, as shown in Table \ref{tab:corr-ig2-consistency}. In Figure \ref{main_fig:mt0-ffn-bias-de-ta-en}, the $IG^2$ scores for similar language pairs are almost the same, while there is a subtle difference for the dissimilar language pairs. This disparity on neurons could explain why the multilingual model is only highly consistent for certain language pairs.
\begin{table}
\centering
\tiny
\begin{tabular}{@{}lll@{}}
\toprule
 &\multicolumn{2}{c}{$IG^2$} \\
 \midrule
Model  & RankC & Acc \\ \midrule
mT0-base  & $0.528^*$ & $0.519^*$\\
mT0-large & $0.705^*$ & $0.699^*$\\
xlm-r-base & $0.400^*$ &  $0.397^*$\\
xlm-r-large & $0.508^*$ & $0.481^*$\\
llama3.2-1B-Instruct & $0.544^*$ & $0.489^*$\\
\bottomrule 
\end{tabular}
\caption{Statistical spearman $\rho$ correlation ($\alpha = 0.05$) between average scores of layers with the patterns on each language model's $IG^2$ absolute difference.}
\label{tab:corr-ig2-consistency}
\end{table}

\section{Correlation between Consistency, Language Bias, and Cross-lingual Bias}


\subsection{Can Language Bias Calibrate Consistency in The Test Time?}
From previous findings, we think of one question: \textit{ Can we add biases from $I_{mono}$ to the feed-forward layers for consistency calibration in the test time?} Considering that two different patterns (on $IG^2$ scores) are discovered from our experiments and $IG^2$ score is moderately correlated with the consistency score, we perform one causal intervention on the feed-forward network to align the output of $I_{cm}$ closer to the output of $I_{mono}$ by patching $I_{mono}$'s activations of all tokens to $I_{cm}$ in selected feedforward neurons based on $IG^2$ \citep{vig2020causal, geiger2021causal}. This experiment measures whether each pattern has a causal relationship with cross-lingual consistency. 

\begin{table}[]
\centering
\tiny
\begin{tabular}{@{}lll@{}}
\toprule
Model & Codemixing Language & Patched FFN Layers \\ \midrule
\multirow{2}{*}{mt0-base} & en--ta  & [0,3,10,11]  \\
 & en--ar & [0,1,9,10]  \\
\hline 
\multirow{2}{*}{mt0-large} & en--ta & [0,1,19,20,21]  \\
 & en--ar& [0,1,19,20,21]  \\
\hline 
\multirow{2}{*}{xlmr-base} & en--ta & [5,8,9,10]  \\
 & en--ar & [5,7,8,10]  \\
\hline 
\multirow{2}{*}{xlmr-large} & en--ta & [0,2,5,19,20]  \\
 & en--ar & [17,18,19,20,21]  \\
 \hline
 \multirow{2}{*}{Llama 3.2-1B} & en--ta & [2,5,10,12]  \\
 & en--ar & [2,5,10,12]  \\
 \hline
  \multirow{2}{*}{Llama 3.1-8B} & en--ta & [5,10,15,18,20]  \\
 & en--ar & [5,10,15,18,20]  \\
\bottomrule
\end{tabular}
\caption{Causal Intervention Hyperparameters Setup}
\label{tab:causal-experiment-setup}
\end{table}
Specifically, we consider $a_i^{(l,p)}$ as the activation of the $i$-th token on $I_{mono}$ produced by the $p$-th neuron in the feed-forward network of the $l$-th layer, and then the patched activation value for the $i$-th token on $I_{cm}$ is $\bar{a}_i^{(l,p)} = a_i^{(l,p)}$, in which we apply this to every new token. We intervene 4 different layers for base models and 5 different layers that have language-sensitive neurons based on $IG^2$ (i.e., layer which has noticeable $IG^2$ distribution difference between $I_{mono}$ and $I_{cm}$). Table \ref{tab:causal-experiment-setup} lists the hyperparameters used in this experiment.

\begin{figure*}[ht]
    \centering
        \includegraphics[width=0.9\linewidth]{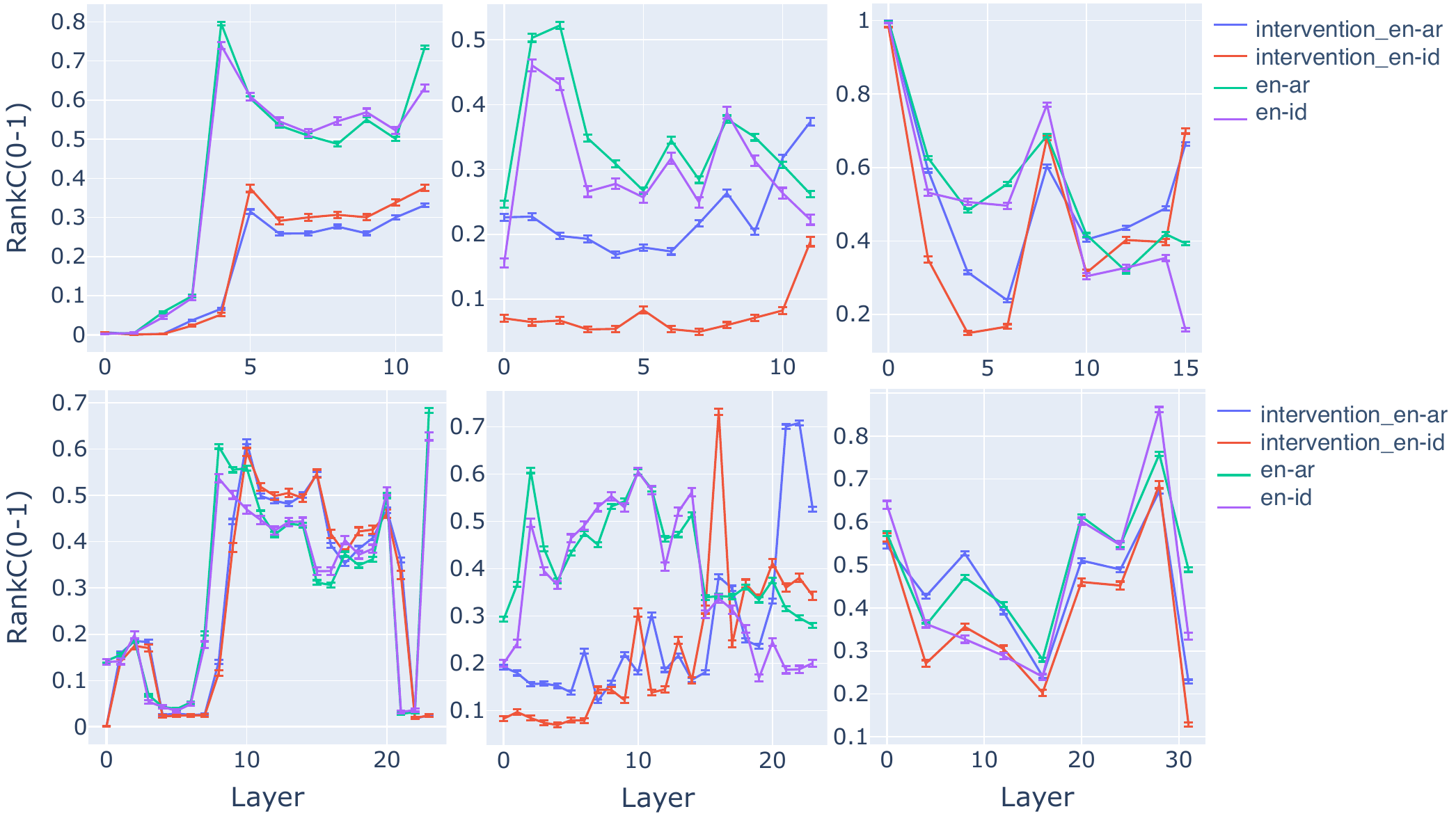}        
        \caption{FFN intervention scores in different model types (L: encoder-decoder, M: encoder, R: decoder) with different model sizes (1st row: small models, 2nd row: large models). ( \textit{cf.} \S \ref{sess:findings_in_details_Adding_Monolingual_Bias}).}
    \label{main_fig:consistency-intervention}
\end{figure*}

In Figure \ref{main_fig:consistency-intervention}, there is a potential causal relationship between the activation intervention and consistency, subject to model architectures and sizes. Specifically, for encoder-decoder models, the intervention approach can increase the consistency scores in the middle-later layers only in the larger model, while such intervention does not offer substantiate gains for the smaller model. Similarly, we observe the effectiveness of the intervention in large encoder models but not in small encoder models.
In contrast, the intervention shows effectiveness for small decoder models, but not for the large decoder models.

\subsection{Vocabulary Expansion and Script Overlapping to Cross-lingual Consistency}
\label{ssec:vocab-expansion-effect}
 \begin{figure}[h]
 \centering
 \includegraphics[width=0.9\linewidth]{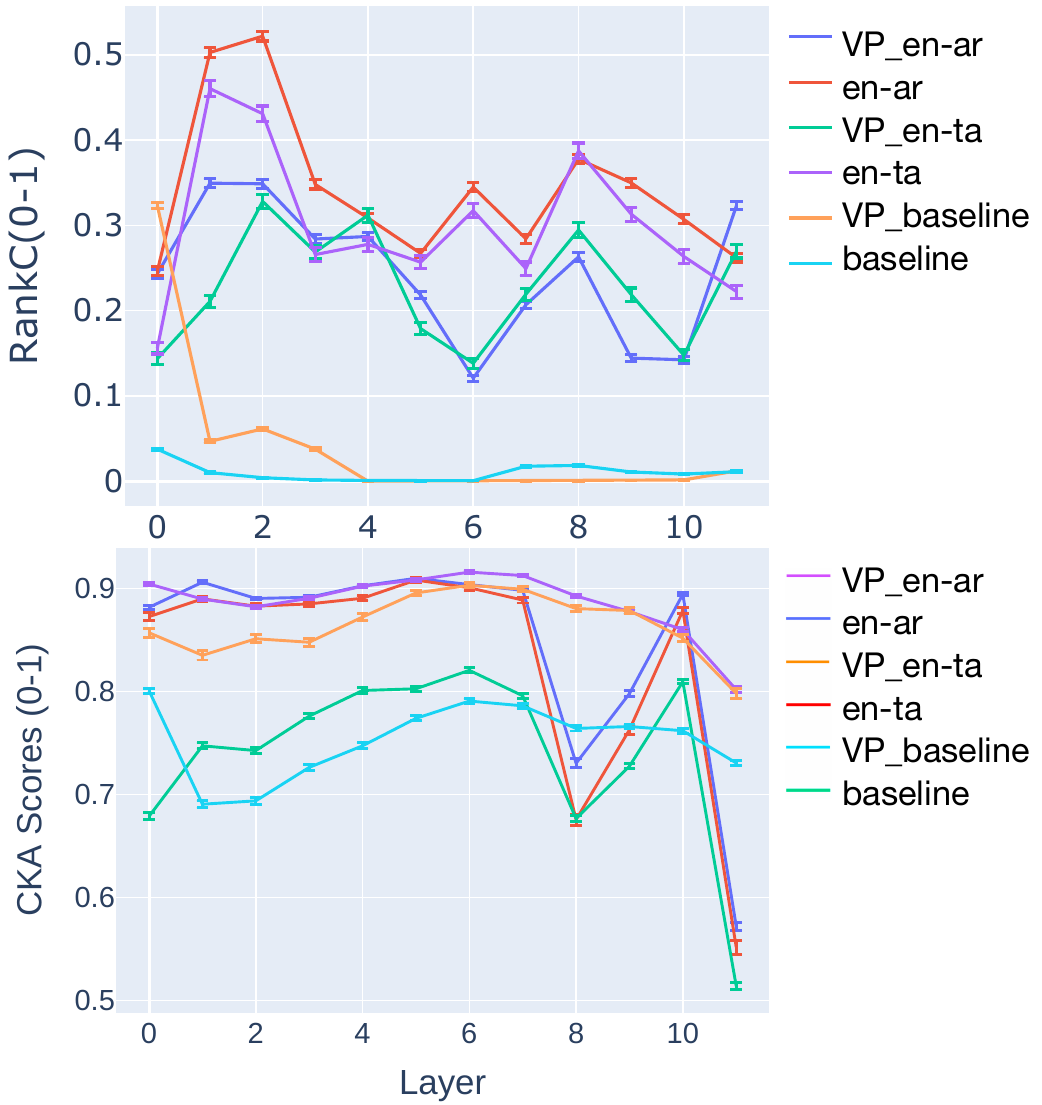}
        \caption{Effects of vocabulary expansion to consistency scores (top) and CKA scores (bottom). (\textit{cf.} \S \ref{sess:findings_in_details_Impact_of_Larger_Vocabulary})}
    \label{main_fig:xlmv-xlmr-kc-cka}
\end{figure}

We hypothesize that vocabulary size plays a crucial role in improving consistency, as 1) it allows a language model to potentially align semantics better due to preventing the model from latching onto shallow local signals or restoring words from subtokens \citep{levine2021pmimasking}\footnote{e.g., if the tokenizer splits the word "Tokyo" into ["To," "Kyo"], the token "To" is polysemous making thus the alignment of this word would be one-to-many, on the other hand, if a tokenizer keeps the word as it is, the tokenized form of the word is monosemous making it less ambiguous.} and 2) it impacts the script overlapping across language. To test this hypothesis, we consider two similar language models, xlm-r-base and xlm-v-base \citep{liang2023xlm}, where xlm-v-base has a larger vocabulary (901,629 tokens) than xlm-r-base (250,002 tokens). 

The vocabulary expansion offers a slight consistency improvement in any categorization, which is evident from the consistency difference between xlm-v-base and xlm-r-base in Figure \ref{main_fig:overall-encoder-encoder-decoder-crosslingual-factors-rankc}. This finding challenges the conclusion in previous works \citep{mLAMA,fierro-sogaard-2022-factual,qi-etal-2023-cross}, where sharing script is the key to cross-lingual consistency. Specifically, the base model shows better cross-lingual consistency in early layers due to the surface alignments via possible shared scripts. This can be observed from the study of representation similarity in Figure \ref{main_fig:xlmv-xlmr-kc-cka}, where the base model shows strong alignments in early layers before final contextualization. However, such cross-lingual consistency cannot propagate to later layers. Compared to that, vocabulary-expanded models rely on deep semantic alignments in later layers for cross-lingual consistency. In Figure \ref{main_fig:xlmv-xlmr-kc-cka}, the layer-wise consistency drops significantly in the base model's last layers but increases in the vocabulary-expanded model's last layers. On the other hand, more samples are required to generalize in the pre-training phase for the vocabulary expansion. Therefore, it alone cannot improve consistency significantly, especially for low-resource languages with limited corpora, but it still benefits dissimilar languages with lower consistency in the last layers to alleviate the consistency bottleneck to some extent because of the deep semantic alignment regardless of the script overlapping. Overall, this finding shares the insight from \citet{zhao2024tracing} where they found that the one-token P@1 of Afrikaans is higher than the Japanese due to segmentation and tokenization\footnote{See discussions about a token parity issue in Figure \ref{fig:parity-ratio-correlation-rankC}.}. Additionally, we studied whether transliteration could help, and found that such a factor does not boost consistency, which we could attribute to the lack of semantic alignments in later layers (\textit{cf.} \S \ref{app_Transliteration} ).

\subsection{Cross-lingual Supervisions to Cross-lingual Consistency}
\label{ssec:pretraining-effect}
Lastly, we analyze how different training supervisions could contribute to the cross-lingual consistency. For this factor, we evaluated several training approaches: additional cross-lingual word alignment training \citep{chi2021improving}, code-switching training \citep{whitehouse-etal-2022-entitycs}, multilingual multitask instruction tuning \citep{wei2021finetuned}, and multilingual chat instruction tuning \citep{grattafiori2024llama3herdmodels}. The former two strategies provide explicit alignments across languages, while the latter two strategies leverage the cross-lingual generalization from parallel samples for implicit alignments.  

\begin{figure*}
    \centering
        \includegraphics[width=\linewidth]{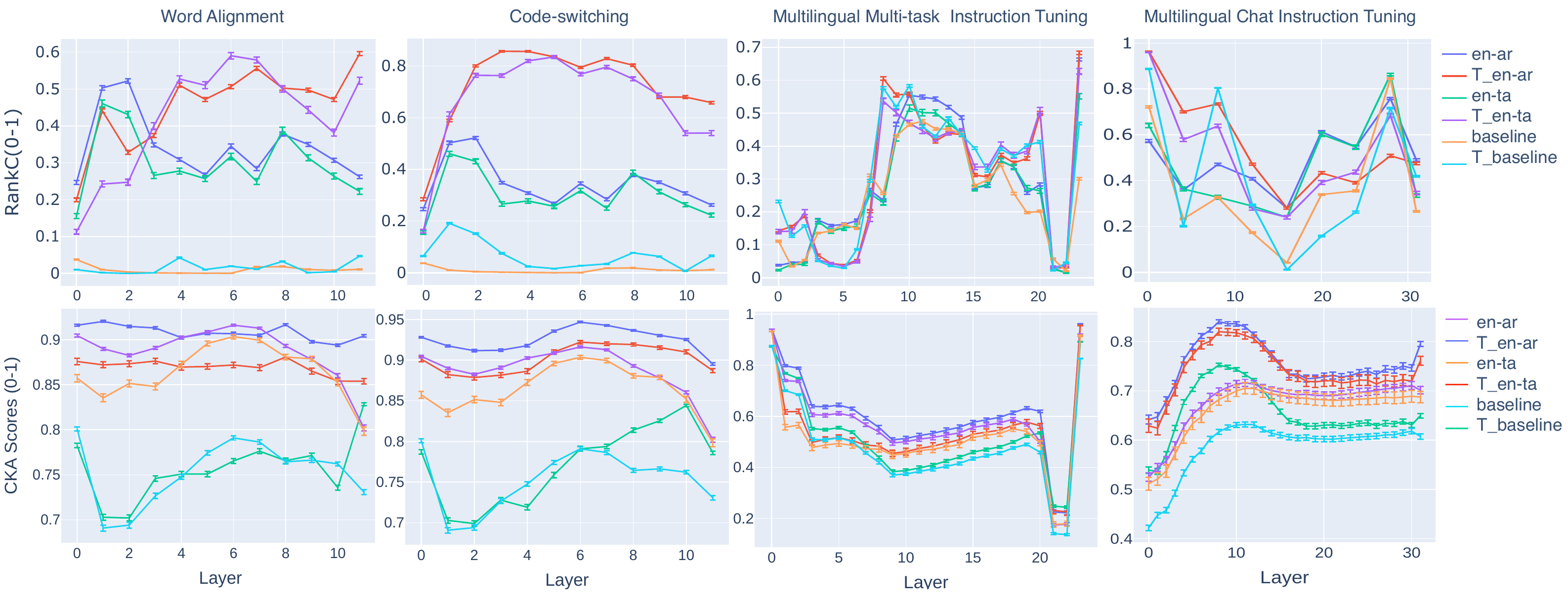}
    \caption{Consistency evolution (1st row: consistency score, 2nd row: CKA score) with different pre-training objectives (1st col: word alignment, 2nd col: code-switching training, 3rd col: instruction tuning on mt5, 4th col: instruction tuning on llama3.1-8b)}
    \label{main_fig:pretraining-layerwise-rankc}
\end{figure*}

Overall, as presented in Figure \ref{main_fig:overall-encoder-encoder-decoder-crosslingual-factors-rankc}, instruction tuning does not offer significant gains, but code-switching and word alignment training objectives improve the consistency significantly, especially for non-Latin script languages, which is not surprising as these objectives encourage models to align word knowledge and writing systems across languages. In addition, the alignment might also improve robustness for handling non-standard spellings and orthographic variations, which is observed in our case study for "transliteration vs translation" presented in \S \ref{app_Transliteration}. This finding may show the importance of adding explicit cross-lingual alignment in the training objective.

In the layer-wise analysis, from Figure \ref{main_fig:pretraining-layerwise-rankc}, code-switching (2nd column, xlm-r-cs) and word alignment (1st column, xlm-align) training objectives could contribute to alleviating the consistency bottleneck occurring in middle layer onward, with word alignment showing the best effect.  Such an attribute could cause the cross-lingual representation to be more consistently high as shown in 2nd row of Figure \ref{main_fig:pretraining-layerwise-rankc}. On the other hand, instruction tuning with parallel samples, including the 3rd column (mt0 tuned from mt5) and the 4th column (Llama 3.1-8b-instruct tuned from Llama 3.1-8b) in the figure, does not offer a universal solution to the consistency bottleneck across model types. Specifically, it manages to slightly improve the cross-lingual consistency for the encoder-decoder, as shown in the 1st row of Figure \ref{main_fig:overall-encoder-encoder-decoder-crosslingual-factors-rankc} with mt0-base showing better consistency over mt5-xl. This could be attributed to additional parallel samples used in the instruction tuning, e.g., multilingual task datasets used for mt0. However, this is not a successful strategy for decoder models, where Llama 3.1-8b-instruct is not more cross-lingually consistent than the base model Llama 3.1-8b. For further analysis of the effect of each supervision on consistency, readers can refer to \S  \ref{ssecsec:word-align-effect}, \S \ref{ssecsec:cs-effect}, and \S \ref{ssecsec:mutlitask-effect}.

\section{Transferable Findings to Other Language Bias}
\begin{figure*}
    \centering
        \includegraphics[width=0.9\linewidth]{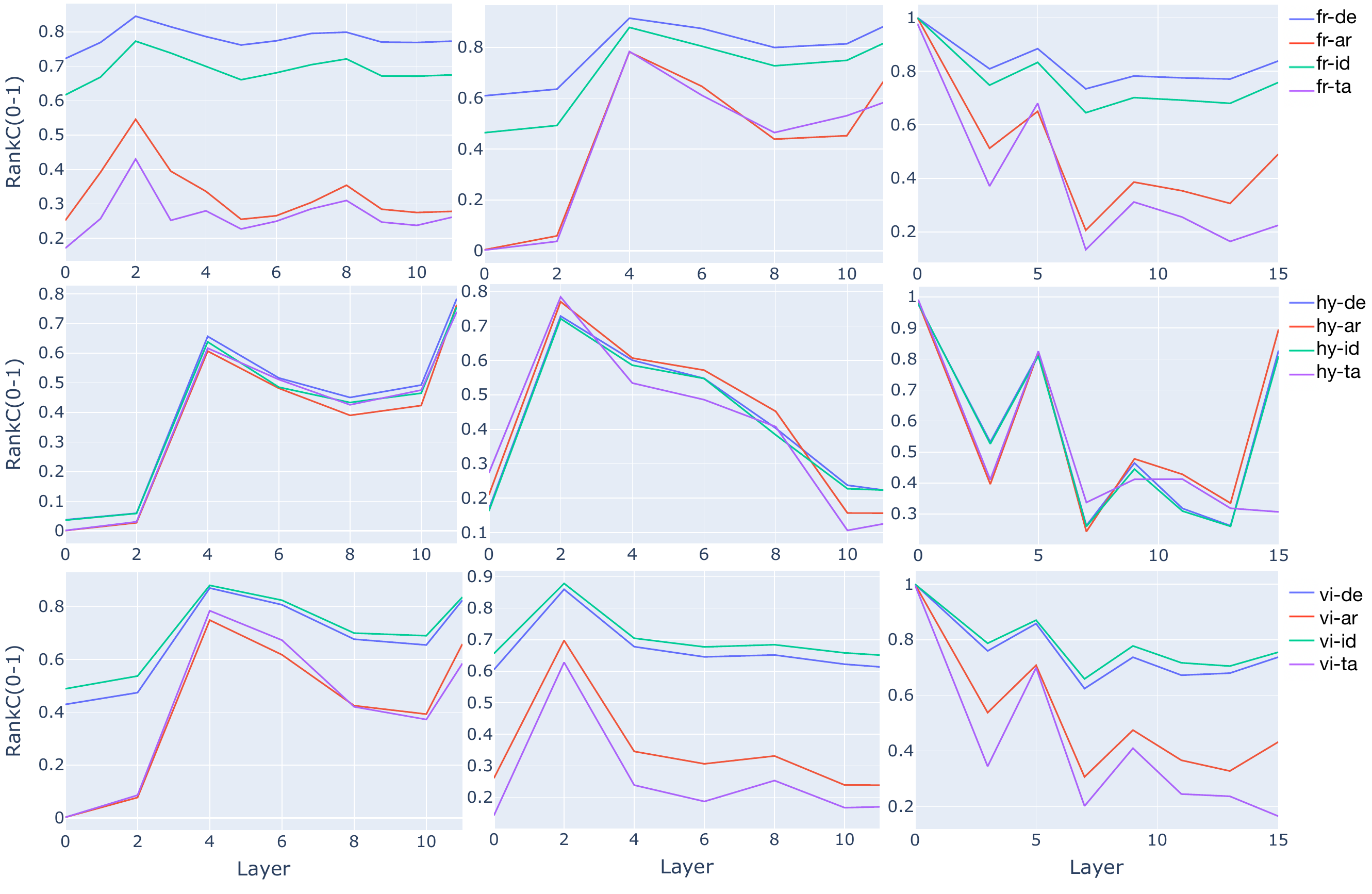}
    \caption{Consistency evolution across different model types (1st row: fr, 2nd row: hy, 3rd row:vi) on different non-english matrix languages (L: encoder-decoder, M: encoder, R: decoder). (\textit{cf.} Figure \ref{app_fig:layerwise-non-english-acc})}
    \label{main:layerwise-non-english-rankc}
\end{figure*}

Throughout this paper, studies are conducted on code-mixed coreferential statements between English and other languages. However, references and knowledge are universal. This raises a question: are all findings transferable to other coreferential entities and statements in non-English-centric scenarios? To answer this question, we conduct experiments for Llama-3.1-1B-Instruct, mt0-base, and xlm-r-base, using fr, vi and hy as the matrix ($L_1$) languages. Experimental results in Figure \ref{main:layerwise-non-english-rankc} are consistent with our main findings, providing evidence that our findings can be transferred to other language bias.   

\section{Related Work}
\citet{mLAMA} extended LAMA \citep{petroni2019language} to a multilingual version multilingual version, mLAMA, and discovered that the language's relational knowledge capability varies in different languages, sharing similar findings with \citet{schott-etal-2023-polyglot, zhao2024tracing} and other benchmarks \citep{wang-etal-2024-seaeval,qi-etal-2023-cross}. \citet{fierro-sogaard-2022-factual,zhao2024tracing} studied the final predictions in different languages and reported inconsistencies across languages, especially for low-resource languages. \citet{mousi-etal-2024-exploring} quantified the entity alignment in the shared space for the consistency goal, and \citet{gao-etal-2024-multilingual,hua-etal-2024-mothello} further traced the alignments emerged from multilingual training. We take a different angle from those works in which we evaluate the consistency against code-mixed coreferential statements in cross-lingual settings.

\citet{bhattacharya2023unveiling,kojima-etal-2024-multilingual, tan2024neuron,miao2025discursive} discovered that a considerable portion of language-agnostic neurons encode universal concepts and utilize the latent language (in this case English). \citet{zhao2024large, wang2024sharing,zhang2024unveiling} further showed that the cross-lingual downstream performance is potentially proportional to the number of language-agnostic neurons. \citet{ferrando2024similarity} discovered a shared circuit or sub-network that is responsible for subject-verb agreement task for English \& Spanish,  and \citet{stanczak-etal-2022-neurons, wang2024probing} found that morpho-syntax attributes have noticeable neuron overlapping degree over notable amount of language pairs. \citet{wang2025lost} discovered three stages of cross-lingual factual recall in which the inconsistency occurred in the last stage called translation stage happening in later layers. In addition, they discovered that language models are able to recall the correct knowledge in the middle layers using the English concept, which is consistent with \citet{wendler2024llamas, dumas2024llamas}. We trace consistent information and knowledge throughout the layers in cross-lingual settings, attempting to understand and interpret how commonly used strategies to improve multilingual models for downstream tasks could impact the cross-lingual knowledge consistency. 

\section{Conclusion}

Our analysis reveals that knowledge consistency is highly dependent on model architectures, training strategies, deep semantic alignments, and language-specific information. Our layer-by-layer analysis of multilingual models uncovers a consistency bottleneck whereby the consistency does not grow monotonically on each layer. Our work highlights promising directions in the test-time calibration and training with cross-lingual alignment objectives to achieve knowledge consistency across languages, which will better preserve parity of language model performance and also alleviate such bottleneck. Cross-lingual representations, shared scripts and parallel samples might contribute to the cross-lingual consistency but are not a sufficient condition to achieve it. 

We encourage researchers to work on representation learning approaches that induce cross-lingual alignment inductive bias explicitly to enhance alignments between coreferential entities. We also suggest test-time approaches that calibrate output distributions for knowledge consistency across languages. These methods can alleviate the consistency bottleneck and enhance alignments between coreferential entities, potentially improving both multilingual performance and cross-lingual consistency.

\section{Limitations}
\label{sssec:limitation}
A promising avenue for this work is evaluating cross-lingual knowledge consistency on other language models. Moreover, we only analyze each crucial component independently due to the time constraint and left scrutinizing the interaction between each component for future work (In particular, one can run any automatic circuit discovery algorithm \citep{syed2023attribution,conmy2023towards} to find subnetwork responsible for cross-lingual consistency and evaluate its performance). In the future, we may expand this work by analyzing how the interaction among these components could affect the cross-lingual consistency of multilingual models. Another thing is that our causal intervention method needs to be done manually, and we suspect that this method could produce a side effect on the model because the representations encoded by language models are more likely to be polysemous.
In addition, we only evaluate language models in context-independent settings. Thus, in the future, we plan to evaluate the consistency of the models' knowledge and observe whether language models utilize their parametric knowledge more or emphasize the knowledge from the given context under the cross-lingual setting. Another thing to consider is that we only evaluate our solution using some particular models due to the time constraint. One interesting thing to explore in this aspect is to see whether adversarial training and multi-agent setting could help to enhance cross-lingual consistency.
Moreover, we use an assumption that one reference is represented as a single English object entity to make the evaluation tractable; hence, we do not take into account the real-world setting where one reference can be interpreted in different ways on multiple languages (e.g., "China" is written as "ZhongGuo" in Chinese rather than "China"). Lastly, our research scope assumes that the knowledge we want to evaluate is factual and not dependent on subjective aspects (e.g., cultural context). With that assumption, we assume that references here generally have one-to-one mapping to representation in one language where the representation here is considered common knowledge.

\section{Ethics Statement}
This work aims to evaluate the consistency of the language model across different senses (particularly between a monolingual input and its code-mixed counterparts) and the impact of different factors on that metric. Doing such a study could shed light on the limitations of language models and think of the mitigations of such matters.

\section{Reproducibility Statements}
We used open-source pretrained models and also dataset for all of the reported experiments thus no undisclosed assets utilized in our work. Additionally, we also provide necessary experiments' output and codes on \url{https://github.com/baridxiai/knowledgeConsistencyAndConflict}.

\section{Acknowledgments}
We would like to thank  several group members from the Web Information Retrieval / Natural Language Processing
Group (WING) at NUS and also some group members from the Research in Text Understanding and Analysis of Language (RiTUAL) at MBZUAI for providing constructive feedbacks on our paper especially Prof. Thamar Solorio. We also grateful to NUS and MBZUAI for providing computing resources to run experiments on this paper.

\bibliography{myREF}

\begin{thebibliography}{54}
\providecommand{\natexlab}[1]{#1}

\bibitem[{Beniwal et~al.(2024)Beniwal, D, and Singh}]{beniwal-etal-2024-cross}
Himanshu Beniwal, Kowsik D, and Mayank Singh. 2024.
\newblock \href {https://aclanthology.org/2024.findings-eacl.140} {Cross-lingual editing in multilingual language models}.
\newblock In \emph{Findings of the Association for Computational Linguistics: EACL 2024}, pages 2078--2128, St. Julian{'}s, Malta. Association for Computational Linguistics.

\bibitem[{Bhattacharya and Bojar(2023)}]{bhattacharya2023unveiling}
Sunit Bhattacharya and Ondrej Bojar. 2023.
\newblock Unveiling multilinguality in transformer models: Exploring language specificity in feed-forward networks.
\newblock \emph{arXiv preprint arXiv:2310.15552}.

\bibitem[{Chi et~al.(2021)Chi, Dong, Zheng, Huang, Mao, Huang, and Wei}]{chi2021improving}
Zewen Chi, Li~Dong, Bo~Zheng, Shaohan Huang, Xian-Ling Mao, Heyan Huang, and Furu Wei. 2021.
\newblock Improving pretrained cross-lingual language models via self-labeled word alignment.
\newblock \emph{arXiv preprint arXiv:2106.06381}.

\bibitem[{Chronopoulou et~al.(2023)Chronopoulou, Stojanovski, and Fraser}]{chronopoulou-etal-2023-language}
Alexandra Chronopoulou, Dario Stojanovski, and Alexander Fraser. 2023.
\newblock \href {https://doi.org/10.18653/v1/2023.loresmt-1.5} {Language-family adapters for low-resource multilingual neural machine translation}.
\newblock In \emph{Proceedings of the Sixth Workshop on Technologies for Machine Translation of Low-Resource Languages (LoResMT 2023)}, pages 59--72, Dubrovnik, Croatia. Association for Computational Linguistics.

\bibitem[{Conmy et~al.(2023)Conmy, Mavor-Parker, Lynch, Heimersheim, and Garriga-Alonso}]{conmy2023towards}
Arthur Conmy, Augustine Mavor-Parker, Aengus Lynch, Stefan Heimersheim, and Adri{\`a} Garriga-Alonso. 2023.
\newblock Towards automated circuit discovery for mechanistic interpretability.
\newblock \emph{Advances in Neural Information Processing Systems}, 36:16318--16352.

\bibitem[{Conneau et~al.(2020)Conneau, Khandelwal, Goyal, Chaudhary, Wenzek, Guzm{\'{a}}n, Grave, Ott, Zettlemoyer, and Stoyanov}]{Conneau2020}
Alexis Conneau, Kartikay Khandelwal, Naman Goyal, Vishrav Chaudhary, Guillaume Wenzek, Francisco Guzm{\'{a}}n, Edouard Grave, Myle Ott, Luke Zettlemoyer, and Veselin Stoyanov. 2020.
\newblock \href {https://arxiv.org/abs/1911.02116v2} {{Unsupervised Cross-lingual Representation Learning at Scale}}.
\newblock In \emph{Proceedings of the 58th Annual Meeting of the Association for Computational Linguistics}. Association for Computational Linguistics.

\bibitem[{Dumas et~al.(2024)Dumas, Veselovsky, Monea, West, and Wendler}]{dumas2024llamas}
Cl{\'e}ment Dumas, Veniamin Veselovsky, Giovanni Monea, Robert West, and Chris Wendler. 2024.
\newblock How do llamas process multilingual text? a latent exploration through activation patching.
\newblock In \emph{ICML 2024 Workshop on Mechanistic Interpretability}.

\bibitem[{Elhage et~al.(2022)Elhage, Hume, Olsson, Schiefer, Henighan, Kravec, Hatfield-Dodds, Lasenby, Drain, Chen et~al.}]{elhage2022toy}
Nelson Elhage, Tristan Hume, Catherine Olsson, Nicholas Schiefer, Tom Henighan, Shauna Kravec, Zac Hatfield-Dodds, Robert Lasenby, Dawn Drain, Carol Chen, et~al. 2022.
\newblock Toy models of superposition.
\newblock \emph{arXiv preprint arXiv:2209.10652}.

\bibitem[{Ethayarajh(2019)}]{ethayarajh-2019-contextual}
Kawin Ethayarajh. 2019.
\newblock \href {https://doi.org/10.18653/v1/D19-1006} {How contextual are contextualized word representations? {C}omparing the geometry of {BERT}, {ELM}o, and {GPT}-2 embeddings}.
\newblock In \emph{Proceedings of the 2019 Conference on Empirical Methods in Natural Language Processing and the 9th International Joint Conference on Natural Language Processing (EMNLP-IJCNLP)}, pages 55--65, Hong Kong, China. Association for Computational Linguistics.

\bibitem[{Ferrando and Costa-juss{\`a}(2024)}]{ferrando2024similarity}
Javier Ferrando and Marta~R Costa-juss{\`a}. 2024.
\newblock On the similarity of circuits across languages: a case study on the subject-verb agreement task.
\newblock \emph{arXiv preprint arXiv:2410.06496}.

\bibitem[{Fierro and S{\o}gaard(2022)}]{fierro-sogaard-2022-factual}
Constanza Fierro and Anders S{\o}gaard. 2022.
\newblock \href {https://doi.org/10.18653/v1/2022.findings-acl.240} {Factual consistency of multilingual pretrained language models}.
\newblock In \emph{Findings of the Association for Computational Linguistics: ACL 2022}, pages 3046--3052, Dublin, Ireland. Association for Computational Linguistics.

\bibitem[{Frege(1892)}]{frege1892sense}
Gottlob Frege. 1892.
\newblock On sense and reference.

\bibitem[{Gao et~al.(2024)Gao, Hu, Hu, Chen, Li, and Huang}]{gao-etal-2024-multilingual}
Changjiang Gao, Hongda Hu, Peng Hu, Jiajun Chen, Jixing Li, and Shujian Huang. 2024.
\newblock \href {https://doi.org/10.18653/v1/2024.naacl-long.339} {Multilingual pretraining and instruction tuning improve cross-lingual knowledge alignment, but only shallowly}.
\newblock In \emph{Proceedings of the 2024 Conference of the North American Chapter of the Association for Computational Linguistics: Human Language Technologies (Volume 1: Long Papers)}, pages 6101--6117, Mexico City, Mexico. Association for Computational Linguistics.

\bibitem[{Geiger et~al.(2021)Geiger, Lu, Icard, and Potts}]{geiger2021causal}
Atticus Geiger, Hanson Lu, Thomas Icard, and Christopher Potts. 2021.
\newblock Causal abstractions of neural networks.
\newblock \emph{Advances in Neural Information Processing Systems}, 34:9574--9586.

\bibitem[{Geva et~al.(2020)Geva, Schuster, Berant, and Levy}]{geva2020transformer}
Mor Geva, Roei Schuster, Jonathan Berant, and Omer Levy. 2020.
\newblock Transformer feed-forward layers are key-value memories.
\newblock \emph{arXiv preprint arXiv:2012.14913}.

\bibitem[{Grattafiori et~al.(2024)Grattafiori, Dubey, Jauhri, Pandey, Kadian, Al-Dahle, Letman, Mathur, Schelten, Vaughan, Yang, Fan, Goyal, Hartshorn, Yang, Mitra, Sravankumar, Korenev, Hinsvark, Rao, Zhang, Rodriguez, Gregerson, Spataru, Roziere, Biron, Tang, Chern, Caucheteux, Nayak, Bi, Marra, McConnell, Keller, Touret, Wu, Wong, Ferrer, Nikolaidis, Allonsius, Song, Pintz, Livshits, Wyatt, Esiobu, Choudhary, Mahajan, Garcia-Olano, Perino, Hupkes, Lakomkin, AlBadawy, Lobanova, Dinan, Smith, Radenovic, Guzmán, Zhang, Synnaeve, Lee, Anderson, Thattai, Nail, Mialon, Pang, Cucurell, Nguyen, Korevaar, Xu, Touvron, Zarov, Ibarra, Kloumann, Misra, Evtimov, Zhang, Copet, Lee, Geffert, Vranes, Park, Mahadeokar, Shah, van~der Linde, Billock, Hong, Lee, Fu, Chi, Huang, Liu, Wang, Yu, Bitton, Spisak, Park, Rocca, Johnstun, Saxe, Jia, Alwala, Prasad, Upasani, Plawiak, Li, Heafield, Stone, El-Arini, Iyer, Malik, Chiu, Bhalla, Lakhotia, Rantala-Yeary, van~der Maaten, Chen, Tan, Jenkins, Martin, Madaan, Malo, Blecher,
  Landzaat, de~Oliveira, Muzzi, Pasupuleti, Singh, Paluri, Kardas, Tsimpoukelli, Oldham, Rita, Pavlova, Kambadur, Lewis, Si, Singh, Hassan, Goyal, Torabi, Bashlykov, Bogoychev, Chatterji, Zhang, Duchenne, Çelebi, Alrassy, Zhang, Li, Vasic, Weng, Bhargava, Dubal, Krishnan, Koura, Xu, He, Dong, Srinivasan, Ganapathy, Calderer, Cabral, Stojnic, Raileanu, Maheswari, Girdhar, Patel, Sauvestre, Polidoro, Sumbaly, Taylor, Silva, Hou, Wang, Hosseini, Chennabasappa, Singh, Bell, Kim, Edunov, Nie, Narang, Raparthy, Shen, Wan, Bhosale, Zhang, Vandenhende, Batra, Whitman, Sootla, Collot, Gururangan, Borodinsky, Herman, Fowler, Sheasha, Georgiou, Scialom, Speckbacher, Mihaylov, Xiao, Karn, Goswami, Gupta, Ramanathan, Kerkez, Gonguet, Do, Vogeti, Albiero, Petrovic, Chu, Xiong, Fu, Meers, Martinet, Wang, Wang, Tan, Xia, Xie, Jia, Wang, Goldschlag, Gaur, Babaei, Wen, Song, Zhang, Li, Mao, Coudert, Yan, Chen, Papakipos, Singh, Srivastava, Jain, Kelsey, Shajnfeld, Gangidi, Victoria, Goldstand, Menon, Sharma, Boesenberg,
  Baevski, Feinstein, Kallet, Sangani, Teo, Yunus, Lupu, Alvarado, Caples, Gu, Ho, Poulton, Ryan, Ramchandani, Dong, Franco, Goyal, Saraf, Chowdhury, Gabriel, Bharambe, Eisenman, Yazdan, James, Maurer, Leonhardi, Huang, Loyd, Paola, Paranjape, Liu, Wu, Ni, Hancock, Wasti, Spence, Stojkovic, Gamido, Montalvo, Parker, Burton, Mejia, Liu, Wang, Kim, Zhou, Hu, Chu, Cai, Tindal, Feichtenhofer, Gao, Civin, Beaty, Kreymer, Li, Adkins, Xu, Testuggine, David, Parikh, Liskovich, Foss, Wang, Le, Holland, Dowling, Jamil, Montgomery, Presani, Hahn, Wood, Le, Brinkman, Arcaute, Dunbar, Smothers, Sun, Kreuk, Tian, Kokkinos, Ozgenel, Caggioni, Kanayet, Seide, Florez, Schwarz, Badeer, Swee, Halpern, Herman, Sizov, Guangyi, Zhang, Lakshminarayanan, Inan, Shojanazeri, Zou, Wang, Zha, Habeeb, Rudolph, Suk, Aspegren, Goldman, Zhan, Damlaj, Molybog, Tufanov, Leontiadis, Veliche, Gat, Weissman, Geboski, Kohli, Lam, Asher, Gaya, Marcus, Tang, Chan, Zhen, Reizenstein, Teboul, Zhong, Jin, Yang, Cummings, Carvill, Shepard, McPhie,
  Torres, Ginsburg, Wang, Wu, U, Saxena, Khandelwal, Zand, Matosich, Veeraraghavan, Michelena, Li, Jagadeesh, Huang, Chawla, Huang, Chen, Garg, A, Silva, Bell, Zhang, Guo, Yu, Moshkovich, Wehrstedt, Khabsa, Avalani, Bhatt, Mankus, Hasson, Lennie, Reso, Groshev, Naumov, Lathi, Keneally, Liu, Seltzer, Valko, Restrepo, Patel, Vyatskov, Samvelyan, Clark, Macey, Wang, Hermoso, Metanat, Rastegari, Bansal, Santhanam, Parks, White, Bawa, Singhal, Egebo, Usunier, Mehta, Laptev, Dong, Cheng, Chernoguz, Hart, Salpekar, Kalinli, Kent, Parekh, Saab, Balaji, Rittner, Bontrager, Roux, Dollar, Zvyagina, Ratanchandani, Yuvraj, Liang, Alao, Rodriguez, Ayub, Murthy, Nayani, Mitra, Parthasarathy, Li, Hogan, Battey, Wang, Howes, Rinott, Mehta, Siby, Bondu, Datta, Chugh, Hunt, Dhillon, Sidorov, Pan, Mahajan, Verma, Yamamoto, Ramaswamy, Lindsay, Lindsay, Feng, Lin, Zha, Patil, Shankar, Zhang, Zhang, Wang, Agarwal, Sajuyigbe, Chintala, Max, Chen, Kehoe, Satterfield, Govindaprasad, Gupta, Deng, Cho, Virk, Subramanian, Choudhury,
  Goldman, Remez, Glaser, Best, Koehler, Robinson, Li, Zhang, Matthews, Chou, Shaked, Vontimitta, Ajayi, Montanez, Mohan, Kumar, Mangla, Ionescu, Poenaru, Mihailescu, Ivanov, Li, Wang, Jiang, Bouaziz, Constable, Tang, Wu, Wang, Wu, Gao, Kleinman, Chen, Hu, Jia, Qi, Li, Zhang, Zhang, Adi, Nam, Yu, Wang, Zhao, Hao, Qian, Li, He, Rait, DeVito, Rosnbrick, Wen, Yang, Zhao, and Ma}]{grattafiori2024llama3herdmodels}
Aaron Grattafiori, Abhimanyu Dubey, Abhinav Jauhri, Abhinav Pandey, Abhishek Kadian, Ahmad Al-Dahle, Aiesha Letman, Akhil Mathur, Alan Schelten, Alex Vaughan, Amy Yang, Angela Fan, Anirudh Goyal, Anthony Hartshorn, Aobo Yang, Archi Mitra, Archie Sravankumar, Artem Korenev, Arthur Hinsvark, Arun Rao, Aston Zhang, Aurelien Rodriguez, Austen Gregerson, Ava Spataru, Baptiste Roziere, Bethany Biron, Binh Tang, Bobbie Chern, Charlotte Caucheteux, Chaya Nayak, Chloe Bi, Chris Marra, Chris McConnell, Christian Keller, Christophe Touret, Chunyang Wu, Corinne Wong, Cristian~Canton Ferrer, Cyrus Nikolaidis, Damien Allonsius, Daniel Song, Danielle Pintz, Danny Livshits, Danny Wyatt, David Esiobu, Dhruv Choudhary, Dhruv Mahajan, Diego Garcia-Olano, Diego Perino, Dieuwke Hupkes, Egor Lakomkin, Ehab AlBadawy, Elina Lobanova, Emily Dinan, Eric~Michael Smith, Filip Radenovic, Francisco Guzmán, Frank Zhang, Gabriel Synnaeve, Gabrielle Lee, Georgia~Lewis Anderson, Govind Thattai, Graeme Nail, Gregoire Mialon, Guan Pang,
  Guillem Cucurell, Hailey Nguyen, Hannah Korevaar, Hu~Xu, Hugo Touvron, Iliyan Zarov, Imanol~Arrieta Ibarra, Isabel Kloumann, Ishan Misra, Ivan Evtimov, Jack Zhang, Jade Copet, Jaewon Lee, Jan Geffert, Jana Vranes, Jason Park, Jay Mahadeokar, Jeet Shah, Jelmer van~der Linde, Jennifer Billock, Jenny Hong, Jenya Lee, Jeremy Fu, Jianfeng Chi, Jianyu Huang, Jiawen Liu, Jie Wang, Jiecao Yu, Joanna Bitton, Joe Spisak, Jongsoo Park, Joseph Rocca, Joshua Johnstun, Joshua Saxe, Junteng Jia, Kalyan~Vasuden Alwala, Karthik Prasad, Kartikeya Upasani, Kate Plawiak, Ke~Li, Kenneth Heafield, Kevin Stone, Khalid El-Arini, Krithika Iyer, Kshitiz Malik, Kuenley Chiu, Kunal Bhalla, Kushal Lakhotia, Lauren Rantala-Yeary, Laurens van~der Maaten, Lawrence Chen, Liang Tan, Liz Jenkins, Louis Martin, Lovish Madaan, Lubo Malo, Lukas Blecher, Lukas Landzaat, Luke de~Oliveira, Madeline Muzzi, Mahesh Pasupuleti, Mannat Singh, Manohar Paluri, Marcin Kardas, Maria Tsimpoukelli, Mathew Oldham, Mathieu Rita, Maya Pavlova, Melanie Kambadur,
  Mike Lewis, Min Si, Mitesh~Kumar Singh, Mona Hassan, Naman Goyal, Narjes Torabi, Nikolay Bashlykov, Nikolay Bogoychev, Niladri Chatterji, Ning Zhang, Olivier Duchenne, Onur Çelebi, Patrick Alrassy, Pengchuan Zhang, Pengwei Li, Petar Vasic, Peter Weng, Prajjwal Bhargava, Pratik Dubal, Praveen Krishnan, Punit~Singh Koura, Puxin Xu, Qing He, Qingxiao Dong, Ragavan Srinivasan, Raj Ganapathy, Ramon Calderer, Ricardo~Silveira Cabral, Robert Stojnic, Roberta Raileanu, Rohan Maheswari, Rohit Girdhar, Rohit Patel, Romain Sauvestre, Ronnie Polidoro, Roshan Sumbaly, Ross Taylor, Ruan Silva, Rui Hou, Rui Wang, Saghar Hosseini, Sahana Chennabasappa, Sanjay Singh, Sean Bell, Seohyun~Sonia Kim, Sergey Edunov, Shaoliang Nie, Sharan Narang, Sharath Raparthy, Sheng Shen, Shengye Wan, Shruti Bhosale, Shun Zhang, Simon Vandenhende, Soumya Batra, Spencer Whitman, Sten Sootla, Stephane Collot, Suchin Gururangan, Sydney Borodinsky, Tamar Herman, Tara Fowler, Tarek Sheasha, Thomas Georgiou, Thomas Scialom, Tobias Speckbacher,
  Todor Mihaylov, Tong Xiao, Ujjwal Karn, Vedanuj Goswami, Vibhor Gupta, Vignesh Ramanathan, Viktor Kerkez, Vincent Gonguet, Virginie Do, Vish Vogeti, Vítor Albiero, Vladan Petrovic, Weiwei Chu, Wenhan Xiong, Wenyin Fu, Whitney Meers, Xavier Martinet, Xiaodong Wang, Xiaofang Wang, Xiaoqing~Ellen Tan, Xide Xia, Xinfeng Xie, Xuchao Jia, Xuewei Wang, Yaelle Goldschlag, Yashesh Gaur, Yasmine Babaei, Yi~Wen, Yiwen Song, Yuchen Zhang, Yue Li, Yuning Mao, Zacharie~Delpierre Coudert, Zheng Yan, Zhengxing Chen, Zoe Papakipos, Aaditya Singh, Aayushi Srivastava, Abha Jain, Adam Kelsey, Adam Shajnfeld, Adithya Gangidi, Adolfo Victoria, Ahuva Goldstand, Ajay Menon, Ajay Sharma, Alex Boesenberg, Alexei Baevski, Allie Feinstein, Amanda Kallet, Amit Sangani, Amos Teo, Anam Yunus, Andrei Lupu, Andres Alvarado, Andrew Caples, Andrew Gu, Andrew Ho, Andrew Poulton, Andrew Ryan, Ankit Ramchandani, Annie Dong, Annie Franco, Anuj Goyal, Aparajita Saraf, Arkabandhu Chowdhury, Ashley Gabriel, Ashwin Bharambe, Assaf Eisenman, Azadeh
  Yazdan, Beau James, Ben Maurer, Benjamin Leonhardi, Bernie Huang, Beth Loyd, Beto~De Paola, Bhargavi Paranjape, Bing Liu, Bo~Wu, Boyu Ni, Braden Hancock, Bram Wasti, Brandon Spence, Brani Stojkovic, Brian Gamido, Britt Montalvo, Carl Parker, Carly Burton, Catalina Mejia, Ce~Liu, Changhan Wang, Changkyu Kim, Chao Zhou, Chester Hu, Ching-Hsiang Chu, Chris Cai, Chris Tindal, Christoph Feichtenhofer, Cynthia Gao, Damon Civin, Dana Beaty, Daniel Kreymer, Daniel Li, David Adkins, David Xu, Davide Testuggine, Delia David, Devi Parikh, Diana Liskovich, Didem Foss, Dingkang Wang, Duc Le, Dustin Holland, Edward Dowling, Eissa Jamil, Elaine Montgomery, Eleonora Presani, Emily Hahn, Emily Wood, Eric-Tuan Le, Erik Brinkman, Esteban Arcaute, Evan Dunbar, Evan Smothers, Fei Sun, Felix Kreuk, Feng Tian, Filippos Kokkinos, Firat Ozgenel, Francesco Caggioni, Frank Kanayet, Frank Seide, Gabriela~Medina Florez, Gabriella Schwarz, Gada Badeer, Georgia Swee, Gil Halpern, Grant Herman, Grigory Sizov, Guangyi, Zhang, Guna
  Lakshminarayanan, Hakan Inan, Hamid Shojanazeri, Han Zou, Hannah Wang, Hanwen Zha, Haroun Habeeb, Harrison Rudolph, Helen Suk, Henry Aspegren, Hunter Goldman, Hongyuan Zhan, Ibrahim Damlaj, Igor Molybog, Igor Tufanov, Ilias Leontiadis, Irina-Elena Veliche, Itai Gat, Jake Weissman, James Geboski, James Kohli, Janice Lam, Japhet Asher, Jean-Baptiste Gaya, Jeff Marcus, Jeff Tang, Jennifer Chan, Jenny Zhen, Jeremy Reizenstein, Jeremy Teboul, Jessica Zhong, Jian Jin, Jingyi Yang, Joe Cummings, Jon Carvill, Jon Shepard, Jonathan McPhie, Jonathan Torres, Josh Ginsburg, Junjie Wang, Kai Wu, Kam~Hou U, Karan Saxena, Kartikay Khandelwal, Katayoun Zand, Kathy Matosich, Kaushik Veeraraghavan, Kelly Michelena, Keqian Li, Kiran Jagadeesh, Kun Huang, Kunal Chawla, Kyle Huang, Lailin Chen, Lakshya Garg, Lavender A, Leandro Silva, Lee Bell, Lei Zhang, Liangpeng Guo, Licheng Yu, Liron Moshkovich, Luca Wehrstedt, Madian Khabsa, Manav Avalani, Manish Bhatt, Martynas Mankus, Matan Hasson, Matthew Lennie, Matthias Reso, Maxim
  Groshev, Maxim Naumov, Maya Lathi, Meghan Keneally, Miao Liu, Michael~L. Seltzer, Michal Valko, Michelle Restrepo, Mihir Patel, Mik Vyatskov, Mikayel Samvelyan, Mike Clark, Mike Macey, Mike Wang, Miquel~Jubert Hermoso, Mo~Metanat, Mohammad Rastegari, Munish Bansal, Nandhini Santhanam, Natascha Parks, Natasha White, Navyata Bawa, Nayan Singhal, Nick Egebo, Nicolas Usunier, Nikhil Mehta, Nikolay~Pavlovich Laptev, Ning Dong, Norman Cheng, Oleg Chernoguz, Olivia Hart, Omkar Salpekar, Ozlem Kalinli, Parkin Kent, Parth Parekh, Paul Saab, Pavan Balaji, Pedro Rittner, Philip Bontrager, Pierre Roux, Piotr Dollar, Polina Zvyagina, Prashant Ratanchandani, Pritish Yuvraj, Qian Liang, Rachad Alao, Rachel Rodriguez, Rafi Ayub, Raghotham Murthy, Raghu Nayani, Rahul Mitra, Rangaprabhu Parthasarathy, Raymond Li, Rebekkah Hogan, Robin Battey, Rocky Wang, Russ Howes, Ruty Rinott, Sachin Mehta, Sachin Siby, Sai~Jayesh Bondu, Samyak Datta, Sara Chugh, Sara Hunt, Sargun Dhillon, Sasha Sidorov, Satadru Pan, Saurabh Mahajan,
  Saurabh Verma, Seiji Yamamoto, Sharadh Ramaswamy, Shaun Lindsay, Shaun Lindsay, Sheng Feng, Shenghao Lin, Shengxin~Cindy Zha, Shishir Patil, Shiva Shankar, Shuqiang Zhang, Shuqiang Zhang, Sinong Wang, Sneha Agarwal, Soji Sajuyigbe, Soumith Chintala, Stephanie Max, Stephen Chen, Steve Kehoe, Steve Satterfield, Sudarshan Govindaprasad, Sumit Gupta, Summer Deng, Sungmin Cho, Sunny Virk, Suraj Subramanian, Sy~Choudhury, Sydney Goldman, Tal Remez, Tamar Glaser, Tamara Best, Thilo Koehler, Thomas Robinson, Tianhe Li, Tianjun Zhang, Tim Matthews, Timothy Chou, Tzook Shaked, Varun Vontimitta, Victoria Ajayi, Victoria Montanez, Vijai Mohan, Vinay~Satish Kumar, Vishal Mangla, Vlad Ionescu, Vlad Poenaru, Vlad~Tiberiu Mihailescu, Vladimir Ivanov, Wei Li, Wenchen Wang, Wenwen Jiang, Wes Bouaziz, Will Constable, Xiaocheng Tang, Xiaojian Wu, Xiaolan Wang, Xilun Wu, Xinbo Gao, Yaniv Kleinman, Yanjun Chen, Ye~Hu, Ye~Jia, Ye~Qi, Yenda Li, Yilin Zhang, Ying Zhang, Yossi Adi, Youngjin Nam, Yu, Wang, Yu~Zhao, Yuchen Hao, Yundi
  Qian, Yunlu Li, Yuzi He, Zach Rait, Zachary DeVito, Zef Rosnbrick, Zhaoduo Wen, Zhenyu Yang, Zhiwei Zhao, and Zhiyu Ma. 2024.
\newblock \href {https://arxiv.org/abs/2407.21783} {The llama 3 herd of models}.
\newblock \emph{Preprint}, arXiv:2407.21783.

\bibitem[{Gupta and Srikumar(2021)}]{x-fact}
Ashim Gupta and Vivek Srikumar. 2021.
\newblock \href {https://doi.org/10.18653/v1/2021.acl-short.86} {{X}-fact: A new benchmark dataset for multilingual fact checking}.
\newblock In \emph{Proceedings of the 59th Annual Meeting of the Association for Computational Linguistics and the 11th International Joint Conference on Natural Language Processing (Volume 2: Short Papers)}, pages 675--682, Online. Association for Computational Linguistics.

\bibitem[{Hu et~al.(2024)Hu, Chen, Li, Guo, Wen, Yu, and Guo}]{hu2024towards}
Xuming Hu, Junzhe Chen, Xiaochuan Li, Yufei Guo, Lijie Wen, Philip~S. Yu, and Zhijiang Guo. 2024.
\newblock \href {https://openreview.net/forum?id=9OevMUdods} {Towards understanding factual knowledge of large language models}.
\newblock In \emph{The Twelfth International Conference on Learning Representations}.

\bibitem[{Hua et~al.(2024)Hua, Yun, and Pavlick}]{hua-etal-2024-mothello}
Tianze Hua, Tian Yun, and Ellie Pavlick. 2024.
\newblock \href {https://doi.org/10.18653/v1/2024.findings-naacl.103} {m{O}thello: When do cross-lingual representation alignment and cross-lingual transfer emerge in multilingual models?}
\newblock In \emph{Findings of the Association for Computational Linguistics: NAACL 2024}, pages 1585--1598, Mexico City, Mexico. Association for Computational Linguistics.

\bibitem[{Hupkes et~al.(2023)Hupkes, Giulianelli, Dankers, Artetxe, Elazar, Pimentel, Christodoulopoulos, Lasri, Saphra, Sinclair, Ulmer, Schottmann, Batsuren, Sun, Sinha, Khalatbari, Ryskina, Frieske, Cotterell, and Jin}]{god_consistency_paper}
Dieuwke Hupkes, Mario Giulianelli, Verna Dankers, Mikel Artetxe, Yanai Elazar, Tiago Pimentel, Christos Christodoulopoulos, Karim Lasri, Naomi Saphra, Arabella Sinclair, Dennis Ulmer, Florian Schottmann, Khuyagbaatar Batsuren, Kaiser Sun, Koustuv Sinha, Leila Khalatbari, Maria Ryskina, Rita Frieske, Ryan Cotterell, and Zhijing Jin. 2023.
\newblock \href {https://doi.org/10.1038/s42256-023-00729-y} {A taxonomy and review of generalization research in nlp}.
\newblock \emph{Nature Machine Intelligence}, 5:1161--1174.

\bibitem[{Kassner et~al.(2021)Kassner, Dufter, and Sch{\"u}tze}]{mLAMA}
Nora Kassner, Philipp Dufter, and Hinrich Sch{\"u}tze. 2021.
\newblock \href {https://doi.org/10.18653/v1/2021.eacl-main.284} {Multilingual {LAMA}: Investigating knowledge in multilingual pretrained language models}.
\newblock In \emph{Proceedings of the 16th Conference of the European Chapter of the Association for Computational Linguistics: Main Volume}, pages 3250--3258, Online. Association for Computational Linguistics.

\bibitem[{Kojima et~al.(2024)Kojima, Okimura, Iwasawa, Yanaka, and Matsuo}]{kojima-etal-2024-multilingual}
Takeshi Kojima, Itsuki Okimura, Yusuke Iwasawa, Hitomi Yanaka, and Yutaka Matsuo. 2024.
\newblock \href {https://doi.org/10.18653/v1/2024.naacl-long.384} {On the multilingual ability of decoder-based pre-trained language models: Finding and controlling language-specific neurons}.
\newblock In \emph{Proceedings of the 2024 Conference of the North American Chapter of the Association for Computational Linguistics: Human Language Technologies (Volume 1: Long Papers)}, pages 6919--6971, Mexico City, Mexico. Association for Computational Linguistics.

\bibitem[{Kornblith et~al.(2019)Kornblith, Norouzi, Lee, and Hinton}]{kornblith2019similarity}
Simon Kornblith, Mohammad Norouzi, Honglak Lee, and Geoffrey Hinton. 2019.
\newblock Similarity of neural network representations revisited.
\newblock In \emph{International conference on machine learning}, pages 3519--3529. PMLR.

\bibitem[{Krause(n.d.)}]{indodic_english_indonesian}
Wayne~B. Krause. n.d.
\newblock \href {https://indodic.com/SimilaritiesDiffs.htm} {English \& indonesian similarities \& differences}.
\newblock Accessed: 2024-10-01.

\bibitem[{Langedijk et~al.(2023)Langedijk, Mohebbi, Sarti, Zuidema, and Jumelet}]{langedijk2023decoderlens}
Anna Langedijk, Hosein Mohebbi, Gabriele Sarti, Willem Zuidema, and Jaap Jumelet. 2023.
\newblock Decoderlens: Layerwise interpretation of encoder-decoder transformers.
\newblock \emph{arXiv preprint arXiv:2310.03686}.

\bibitem[{Levine et~al.(2021)Levine, Lenz, Lieber, Abend, Leyton-Brown, Tennenholtz, and Shoham}]{levine2021pmimasking}
Yoav Levine, Barak Lenz, Opher Lieber, Omri Abend, Kevin Leyton-Brown, Moshe Tennenholtz, and Yoav Shoham. 2021.
\newblock \href {https://openreview.net/forum?id=3Aoft6NWFej} {{\{}PMI{\}}-masking: Principled masking of correlated spans}.
\newblock In \emph{International Conference on Learning Representations}.

\bibitem[{Li et~al.(2024)Li, Zhang, Yao, Wang, Chen, and Chen}]{li2024unveiling}
Zhoubo Li, Ningyu Zhang, Yunzhi Yao, Mengru Wang, Xi~Chen, and Huajun Chen. 2024.
\newblock \href {https://openreview.net/forum?id=fNktD3ib16} {Unveiling the pitfalls of knowledge editing for large language models}.
\newblock In \emph{The Twelfth International Conference on Learning Representations}.

\bibitem[{Liang et~al.(2023)Liang, Gonen, Mao, Hou, Goyal, Ghazvininejad, Zettlemoyer, and Khabsa}]{liang2023xlm}
Davis Liang, Hila Gonen, Yuning Mao, Rui Hou, Naman Goyal, Marjan Ghazvininejad, Luke Zettlemoyer, and Madian Khabsa. 2023.
\newblock Xlm-v: Overcoming the vocabulary bottleneck in multilingual masked language models.
\newblock \emph{arXiv preprint arXiv:2301.10472}.

\bibitem[{Liu et~al.(2024)Liu, Liu, Chen, Chen, Zan, Kan, and Ho}]{liu2024devil}
Yan Liu, Yu~Liu, Xiaokang Chen, Pin-Yu Chen, Daoguang Zan, Min-Yen Kan, and Tsung-Yi Ho. 2024.
\newblock The devil is in the neurons: Interpreting and mitigating social biases in language models.
\newblock In \emph{The Twelfth International Conference on Learning Representations}.

\bibitem[{Miao and Kan(2025)}]{miao2025discursive}
Yisong Miao and Min-Yen Kan. 2025.
\newblock Discursive circuits: How do language models understand discourse relations?
\newblock In \emph{The 2025 Conference on Empirical Methods in Natural Language Processing}.

\bibitem[{Mousi et~al.(2024)Mousi, Durrani, Dalvi, Hawasly, and Abdelali}]{mousi-etal-2024-exploring}
Basel Mousi, Nadir Durrani, Fahim Dalvi, Majd Hawasly, and Ahmed Abdelali. 2024.
\newblock \href {https://doi.org/10.18653/v1/2024.acl-long.344} {Exploring alignment in shared cross-lingual spaces}.
\newblock In \emph{Proceedings of the 62nd Annual Meeting of the Association for Computational Linguistics (Volume 1: Long Papers)}, pages 6326--6348, Bangkok, Thailand. Association for Computational Linguistics.

\bibitem[{Muennighoff et~al.(2023)Muennighoff, Wang, Sutawika, Roberts, Biderman, Le~Scao, Bari, Shen, Yong, Schoelkopf, Tang, Radev, Aji, Almubarak, Albanie, Alyafeai, Webson, Raff, and Raffel}]{mt0_paper}
Niklas Muennighoff, Thomas Wang, Lintang Sutawika, Adam Roberts, Stella Biderman, Teven Le~Scao, M~Saiful Bari, Sheng Shen, Zheng~Xin Yong, Hailey Schoelkopf, Xiangru Tang, Dragomir Radev, Alham~Fikri Aji, Khalid Almubarak, Samuel Albanie, Zaid Alyafeai, Albert Webson, Edward Raff, and Colin Raffel. 2023.
\newblock \href {https://doi.org/10.18653/v1/2023.acl-long.891} {Crosslingual generalization through multitask finetuning}.
\newblock In \emph{Proceedings of the 61st Annual Meeting of the Association for Computational Linguistics (Volume 1: Long Papers)}, pages 15991--16111, Toronto, Canada. Association for Computational Linguistics.

\bibitem[{nostalgebraist(2019)}]{lesswrong2024interpreting}
nostalgebraist. 2019.
\newblock \href {https://www.lesswrong.com/posts/AcKRB8wDpdaN6v6ru/interpreting-gpt-the-logit-lens} {Interpreting gpt: The logit lens}.
\newblock Accessed: 2024-08-04.

\bibitem[{Petroni et~al.(2019)Petroni, Rockt{\"a}schel, Lewis, Bakhtin, Wu, Miller, and Riedel}]{petroni2019language}
Fabio Petroni, Tim Rockt{\"a}schel, Patrick Lewis, Anton Bakhtin, Yuxiang Wu, Alexander~H Miller, and Sebastian Riedel. 2019.
\newblock Language models as knowledge bases?
\newblock \emph{arXiv preprint arXiv:1909.01066}.

\bibitem[{Pratapa et~al.(2018)Pratapa, Bhat, Choudhury, Sitaram, Dandapat, and Bali}]{pratapa2018language}
Adithya Pratapa, Gayatri Bhat, Monojit Choudhury, Sunayana Sitaram, Sandipan Dandapat, and Kalika Bali. 2018.
\newblock Language modeling for code-mixing: The role of linguistic theory based synthetic data.
\newblock In \emph{Proceedings of the 56th Annual Meeting of the Association for Computational Linguistics (Volume 1: Long Papers)}, pages 1543--1553.

\bibitem[{Qi et~al.(2023)Qi, Fern{\'a}ndez, and Bisazza}]{qi-etal-2023-cross}
Jirui Qi, Raquel Fern{\'a}ndez, and Arianna Bisazza. 2023.
\newblock \href {https://doi.org/10.18653/v1/2023.emnlp-main.658} {Cross-lingual consistency of factual knowledge in multilingual language models}.
\newblock In \emph{Proceedings of the 2023 Conference on Empirical Methods in Natural Language Processing}, pages 10650--10666, Singapore. Association for Computational Linguistics.

\bibitem[{Schott et~al.(2023)Schott, Furman, and Bhat}]{schott-etal-2023-polyglot}
Tim Schott, Daniel Furman, and Shreshta Bhat. 2023.
\newblock \href {https://doi.org/10.18653/v1/2023.emnlp-main.691} {Polyglot or not? measuring multilingual encyclopedic knowledge in foundation models}.
\newblock In \emph{Proceedings of the 2023 Conference on Empirical Methods in Natural Language Processing}, pages 11238--11253, Singapore. Association for Computational Linguistics.

\bibitem[{Stanczak et~al.(2022)Stanczak, Ponti, Torroba~Hennigen, Cotterell, and Augenstein}]{stanczak-etal-2022-neurons}
Karolina Stanczak, Edoardo Ponti, Lucas Torroba~Hennigen, Ryan Cotterell, and Isabelle Augenstein. 2022.
\newblock \href {https://doi.org/10.18653/v1/2022.naacl-main.114} {Same neurons, different languages: Probing morphosyntax in multilingual pre-trained models}.
\newblock In \emph{Proceedings of the 2022 Conference of the North American Chapter of the Association for Computational Linguistics: Human Language Technologies}, pages 1589--1598, Seattle, United States. Association for Computational Linguistics.

\bibitem[{Syed et~al.(2023)Syed, Rager, and Conmy}]{syed2023attribution}
Aaquib Syed, Can Rager, and Arthur Conmy. 2023.
\newblock Attribution patching outperforms automated circuit discovery.
\newblock \emph{arXiv preprint arXiv:2310.10348}.

\bibitem[{Tan et~al.(2024)Tan, Wu, and Monz}]{tan2024neuron}
Shaomu Tan, Di~Wu, and Christof Monz. 2024.
\newblock Neuron specialization: Leveraging intrinsic task modularity for multilingual machine translation.
\newblock \emph{arXiv preprint arXiv:2404.11201}.

\bibitem[{Tsvetkov and Dyer(2016)}]{tsvetkov2016cross}
Yulia Tsvetkov and Chris Dyer. 2016.
\newblock Cross-lingual bridges with models of lexical borrowing.
\newblock \emph{Journal of Artificial Intelligence Research}, 55:63--93.

\bibitem[{Vig et~al.(2020)Vig, Gehrmann, Belinkov, Qian, Nevo, Sakenis, Huang, Singer, and Shieber}]{vig2020causal}
Jesse Vig, Sebastian Gehrmann, Yonatan Belinkov, Sharon Qian, Daniel Nevo, Simas Sakenis, Jason Huang, Yaron Singer, and Stuart Shieber. 2020.
\newblock Causal mediation analysis for interpreting neural nlp: The case of gender bias.
\newblock \emph{arXiv preprint arXiv:2004.12265}.

\bibitem[{Wang et~al.(2024{\natexlab{a}})Wang, Liu, Huang, Jiao, Ding, Aw, and Chen}]{wang-etal-2024-seaeval}
Bin Wang, Zhengyuan Liu, Xin Huang, Fangkai Jiao, Yang Ding, AiTi Aw, and Nancy Chen. 2024{\natexlab{a}}.
\newblock \href {https://doi.org/10.18653/v1/2024.naacl-long.22} {{S}ea{E}val for multilingual foundation models: From cross-lingual alignment to cultural reasoning}.
\newblock In \emph{Proceedings of the 2024 Conference of the North American Chapter of the Association for Computational Linguistics: Human Language Technologies (Volume 1: Long Papers)}, pages 370--390, Mexico City, Mexico. Association for Computational Linguistics.

\bibitem[{Wang et~al.(2024{\natexlab{b}})Wang, Minervini, and Ponti}]{wang2024probing}
Hetong Wang, Pasquale Minervini, and Edoardo~M Ponti. 2024{\natexlab{b}}.
\newblock Probing the emergence of cross-lingual alignment during llm training.
\newblock \emph{arXiv preprint arXiv:2406.13229}.

\bibitem[{Wang et~al.(2025)Wang, Adel, Lange, Liu, Nie, Str{\"o}tgen, and Sch{\"u}tze}]{wang2025lost}
Mingyang Wang, Heike Adel, Lukas Lange, Yihong Liu, Ercong Nie, Jannik Str{\"o}tgen, and Hinrich Sch{\"u}tze. 2025.
\newblock Lost in multilinguality: Dissecting cross-lingual factual inconsistency in transformer language models.
\newblock \emph{arXiv preprint arXiv:2504.04264}.

\bibitem[{Wang et~al.(2024{\natexlab{c}})Wang, Haddow, Wu, Peng, and Birch}]{wang2024sharing}
Weixuan Wang, Barry Haddow, Minghao Wu, Wei Peng, and Alexandra Birch. 2024{\natexlab{c}}.
\newblock Sharing matters: Analysing neurons across languages and tasks in llms.
\newblock \emph{arXiv preprint arXiv:2406.09265}.

\bibitem[{Wang et~al.(2023)Wang, Wei, Schuurmans, Le, Chi, Narang, Chowdhery, and Zhou}]{wang2023selfconsistency}
Xuezhi Wang, Jason Wei, Dale Schuurmans, Quoc~V Le, Ed~H. Chi, Sharan Narang, Aakanksha Chowdhery, and Denny Zhou. 2023.
\newblock \href {https://openreview.net/forum?id=1PL1NIMMrw} {Self-consistency improves chain of thought reasoning in language models}.
\newblock In \emph{The Eleventh International Conference on Learning Representations}.

\bibitem[{Wei et~al.(2021)Wei, Bosma, Zhao, Guu, Yu, Lester, Du, Dai, and Le}]{wei2021finetuned}
Jason Wei, Maarten Bosma, Vincent~Y Zhao, Kelvin Guu, Adams~Wei Yu, Brian Lester, Nan Du, Andrew~M Dai, and Quoc~V Le. 2021.
\newblock Finetuned language models are zero-shot learners.
\newblock \emph{arXiv preprint arXiv:2109.01652}.

\bibitem[{Wendler et~al.(2024)Wendler, Veselovsky, Monea, and West}]{wendler2024llamas}
Chris Wendler, Veniamin Veselovsky, Giovanni Monea, and Robert West. 2024.
\newblock Do llamas work in english? on the latent language of multilingual transformers.
\newblock \emph{arXiv preprint arXiv:2402.10588}.

\bibitem[{Whitehouse et~al.(2022)Whitehouse, Christopoulou, and Iacobacci}]{whitehouse-etal-2022-entitycs}
Chenxi Whitehouse, Fenia Christopoulou, and Ignacio Iacobacci. 2022.
\newblock \href {https://doi.org/10.18653/v1/2022.findings-emnlp.499} {{E}ntity{CS}: Improving zero-shot cross-lingual transfer with entity-centric code switching}.
\newblock In \emph{Findings of the Association for Computational Linguistics: EMNLP 2022}, pages 6698--6714, Abu Dhabi, United Arab Emirates. Association for Computational Linguistics.

\bibitem[{Xue et~al.(2020)Xue, Constant, Roberts, Kale, Al-Rfou, Siddhant, Barua, and Raffel}]{xue2020mt5}
Linting Xue, Noah Constant, Adam Roberts, Mihir Kale, Rami Al-Rfou, Aditya Siddhant, Aditya Barua, and Colin Raffel. 2020.
\newblock mt5: A massively multilingual pre-trained text-to-text transformer.
\newblock \emph{arXiv preprint arXiv:2010.11934}.

\bibitem[{Zhang et~al.(2024)Zhang, Zhao, Zhang, Gui, and Huang}]{zhang2024unveiling}
Zhihao Zhang, Jun Zhao, Qi~Zhang, Tao Gui, and Xuanjing Huang. 2024.
\newblock Unveiling linguistic regions in large language models.
\newblock \emph{arXiv preprint arXiv:2402.14700}.

\bibitem[{Zhao et~al.(2024{\natexlab{a}})Zhao, Yoshinaga, and Oba}]{zhao2024tracing}
Xin Zhao, Naoki Yoshinaga, and Daisuke Oba. 2024{\natexlab{a}}.
\newblock Tracing the roots of facts in multilingual language models: Independent, shared, and transferred knowledge.
\newblock \emph{arXiv preprint arXiv:2403.05189}.

\bibitem[{Zhao et~al.(2024{\natexlab{b}})Zhao, Zhang, Chen, Kawaguchi, and Bing}]{zhao2024large}
Yiran Zhao, Wenxuan Zhang, Guizhen Chen, Kenji Kawaguchi, and Lidong Bing. 2024{\natexlab{b}}.
\newblock How do large language models handle multilingualism?
\newblock \emph{arXiv preprint arXiv:2402.18815}.

\end{thebibliography}

\appendix

\section{Appendix}
\subsection{Our Task Definition}
\label{apx_probabilistic_task_definition}
We focus on a code-mixed context-independent cloze task that forces the multilingual model to rely on its internal knowledge base to recall the common knowledge shared by coreferential entities across languages due to cross-lingual generalization\footnote{See limitation in \S \ref{sssec:limitation}.}. In the following introduction, we will define the evaluation task mathematically. Let $I = \{S^{l1},\cdots,O,\cdots\} \in {l1}$\footnote{The surface structure is not restricted. We use the common subject--object structure as an example.} be a statement, where $l1$ stands for matrix language (the predominant language), $S^{l1}=\{s_1,\cdots,s_k\} \in l1$ are subject sub-tokens, and $O=\{o_1,o_2,\cdots,o_j\} \in l1$ denote object sub-tokens. This statement is used to create a cloze task input $I_{mono}=\{S^{l1},\cdots,M,\cdots\}$, where $M$ is the mask token used to substitute $O$ in $I$ (i.e., the mask $M=\{mask_1,\cdots, mask_j\}$ in encoder models, the sentinel token $M=<extra\_id\_0>$, or the next token in decoder models). We define n-gram prediction for $O$ from $M$, denoted as $Cand(O_{\in V}|I_{mono})$, as the top-k n-gram candidates obtained from beam search decoding over the model's vocabulary $V$. Similarly, we can create a code-mixed coreferential statement $I_{cm}$ by replacing $S^{l1}$ with a coreferential subject $S^{l2}$ in the embedded language $l2$ (the subsidiary language)  in order to obtain $Cand(O_{\in V}|I_{cm})$. Therefore, $I_{cm}$ and $I_{mono}$ are coreferential and expected to recall the same knowledge. Finally, we define the consistency of cross-lingual knowledge as $0 \leq f_{metric}(Cand(O_{\in V}|I_{mono}), Cand(O_{\in V}|I_{cm})) \leq 1$, where $f_{metric}$ is a consistency metric defined in the next subsection. If $f_{metric} = 1 $, it implies that multilingual language models recall the factual knowledge for the coreferential statements $I_{mono}$ and $I_{cm}$ in an identical manner. The coreferential statements are fully inconsistent if $f_{metric} = 0 $. Note that we do not consider whether the prediction is correct. Instead, $f_{metric}$ evaluates the parity and consistency across languages in which the model is expected to produce similar candidates for $I_{mono}$ and $I_{cm}$. 

From a probability view, we can define our task as measuring the difference between two distributions, $Cand(O_{\in V}|I_{cm}) = P(O_{\in V} |K_{\theta})$$P(K_{\theta}| S^{l2}, I_{\setminus (S^{l1} \cap O) })$ and $Cand(O_{\in V}|I_{mono}) = P(O_{\in V} | K^*_{\theta})$$P(K^*_{\theta}| S^{l1}, I_{\setminus (S^{l1} \cap O)})$, where $K_{\theta}$ is the knowledge recalled from the model given the preceding context, and $I_{\setminus (S^{l1} \cap O)}$ stands for $I$ without both the subject and the object. Then, cross-lingual knowledge consistency between $K^*_{\theta}$ and $K_{\theta}$ reflects on the measured difference. 

The high-level idea of this evaluation task is illustrated in Figure \ref{kc-illust} where en entry "Paris is the capital of \_\_\_" is evaluated with its possible code-mixed statements (ar entry \& ta entry). Additionally, we also evaluated the baseline setting of $I_{cm}$ by removing the subject entities to obtain the model's default object tokens for comparison.  In this example, $S^{l1}$, $I_{\setminus (S^{l1} \cap O) }$, and $S^{l2}$ are "Paris", "is the capital of", and the ar or ta entry for "Paris", respectively. If coreferential subject entries are trained to generalize across languages, we could observe the cross-lingual consistency. In addition, we are aware of a baseline from this probability view. Specifically, we define the baseline as the difference between $Cand(O_{\in V}|I_{mono})$ and $Cand(O_{\in V}|I_{\setminus (S^{l1} \cap O)}) = P(O_{\in V} | K^\alpha_{\theta})$$P(K^\alpha_{\theta}| I_{\setminus (S^{l1} \cap O)})$, measuring agnostic consistency without the coreferential subjects $S^{l1}$ and $S^{l2}$ in cross-lingual settings. In implementation, we mask the both subject and object entities to create the "code-mixed" counterpart as the baseline. Readers can refer to Appendix \S \ref{apx_input_format} for our implementation.

\subsection{Input Format}
\label{apx_input_format}
In our task definition, we introduce our evaluation task in both intuition and math perspective. Here is the input sample in Table \ref{tab_apx_input_sampel_encoder}, \ref{tab_apx_input_sampel_encoder-decoder}, \ref{tab_apx_input_sampel_decoder}. Meanwhile, as presented in the task definition, we do not consider whether predictions are true but focus on the same prediction distributions regardless of languages. Note that we did not perturb the surface structure in order to minimize variables to affect factual knowledge recall because $S^{l2}$ "switches-in" at grammatically correct point as the new subject \citep{pratapa2018language}. 

\begin{table*}[ht!]
\centering
\begin{tabular}{@{}ll@{}}
\toprule
 & xlm-r input \\ 
 \midrule
$I_{mono}$ & Paris is the capital of  \textbf{$<$mask$>$}\\
$I_{cm}$ & \<باريس> is the capital of \textbf{$<$mask$>$} \\
$I_{\setminus (S^{l1} \cap O)}$ & $<$mask$>$  is the capital of \textbf{$<$mask$>$}\\
\bottomrule
\end{tabular}
\caption{Input sample for the evaluation task for xlm-r. We only predict the object in bold. $I_{\setminus (S^{l1} \cap O)}$ is the baseline input.}
\label{tab_apx_input_sampel_encoder}
\end{table*}

\begin{table*}[ht!]
\centering
\begin{tabular}{@{}ll@{}}
\toprule
 & mt0 input \\ 
 \midrule
$I_{mono}$ & Paris is the capital of \textbf{$<$extra\_id\_0$>$}\\
$I_{cm}$ & \<باريس> is the capital of  \textbf{$<$extra\_id\_0$>$}\\
$I_{\setminus (S^{l1} \cap O)}$ & $<$extra\_id\_0$>$  is the capital of  \textbf{$<$extra\_id\_1$>$}\\
\bottomrule
\end{tabular}
\caption{Input sample for the evaluation task for mt0. We only predict the object in bold. $I_{\setminus (S^{l1} \cap O)}$ is the baseline input.}
\label{tab_apx_input_sampel_encoder-decoder}
\end{table*}

\begin{table*}[ht!]
\centering
\scriptsize
\begin{tabular}{@{}ll@{}}
\toprule
 & Llama input \\ 
 \midrule
$I_{mono}$ & Finish the cloze question with words. Do not give additional comments. Question: Paris is the capital of \textunderscore. Answer:\\
$I_{cm}$ & Finish the cloze question with words. Do not give additional comments. \texttt{\textbackslash n} Question: \<باريس>  is the capital of \textunderscore. Answer:\\
$I_{\setminus (S^{l1} \cap O)}$ & Finish the cloze question with words. Do not give additional comments. Question: \textunderscore is the capital of \textunderscore.  Answer: \\
\bottomrule
\end{tabular}
\caption{Input sample for the evaluation task for llama 3. We only predict the object in bold. $I_{\setminus (S^{l1} \cap O)}$ is the baseline input.}
\label{tab_apx_input_sampel_decoder}
\end{table*}

\subsection{Metric Function and Interpretability Approach }
\label{apx_consistency_metric}
\paragraph{RankC} RankC \citep{qi-etal-2023-cross} is used to evaluate the cross-lingual knowledge consistency. Given a set of statements $S$ where each of the statement having each own $I_{mono}$ and $I_{cm}$, the number of candidates $Cand(O_{\in V}|I_{mono})$ of i-th statement $N_i$,  $mono^j$ stands for the j-th candidate of  $Cand(O_{\in V}|I_{mono})$, $cm^j$ stands for the j-th candidate of  $Cand(O_{\in V}|I_{cm})$, and the RankC score of $Cand(O_{\in V}|I_{mono})$ concerning $Cand(O_{\in V}|I_{cm})$ can be written as 
\begin{gather}
    RankC(cm, mono) \\=\frac{\sum_{i=1}^{|S|}\sum_{j=1}^{N_i}\frac{e^{N_i-j}}{\sum_{k=1}^{N_i}e^{k-j}}*P@j}{|I_{mono}|}, \\
    P@j = \frac{1}{j}|\{cm_i^1,cm_i^2,\cdots,cm_i^j\} \cap \\
    \{mono_i^1,mono_i^2,\cdots,mono_i^j\}|. 
\end{gather}
\paragraph{Top@1 Accuracy} The Top@1 accuracy is defined as the average number of exact matches between the top-1 predictions given $I_{mono}$ and $I_{cm}$.

\paragraph{$IG^2$ Score} If $w_j^{(l)}$ is the activation value of $j$-th neuron in the $l$-th layer of a particular input (either code-mixed or not), $m$ is the approximation step, and $t$ as a token of the whole ground truth object entity, the score for a given  $I_{mono}$ or $I_{cm}$ is defined as 
\begin{gather}
IG^2(w_j^{(l)}) = \sum_{t \in T}\frac{\frac{w_j^{(l)}}{m}\sum_{k=1}^m \frac{\partial P(t|\frac{k}{m}w_j^{(l)})}{\partial (\frac{k}{m}w_j^{(l)})}}{|T|}
\end{gather}

\subsection{Findings in Details}
\label{sess:findings_in_details}
\subsubsection{Layer-wise Consistency}
\label{sess:findings_in_details_Layer-wise_Consistency}
\begin{figure*}
    \centering
    \begin{subfigure}{0.45\linewidth}{
        \centering
        \includegraphics[trim=0cm 0cm 0cm 2cm,clip=true,width=\linewidth]{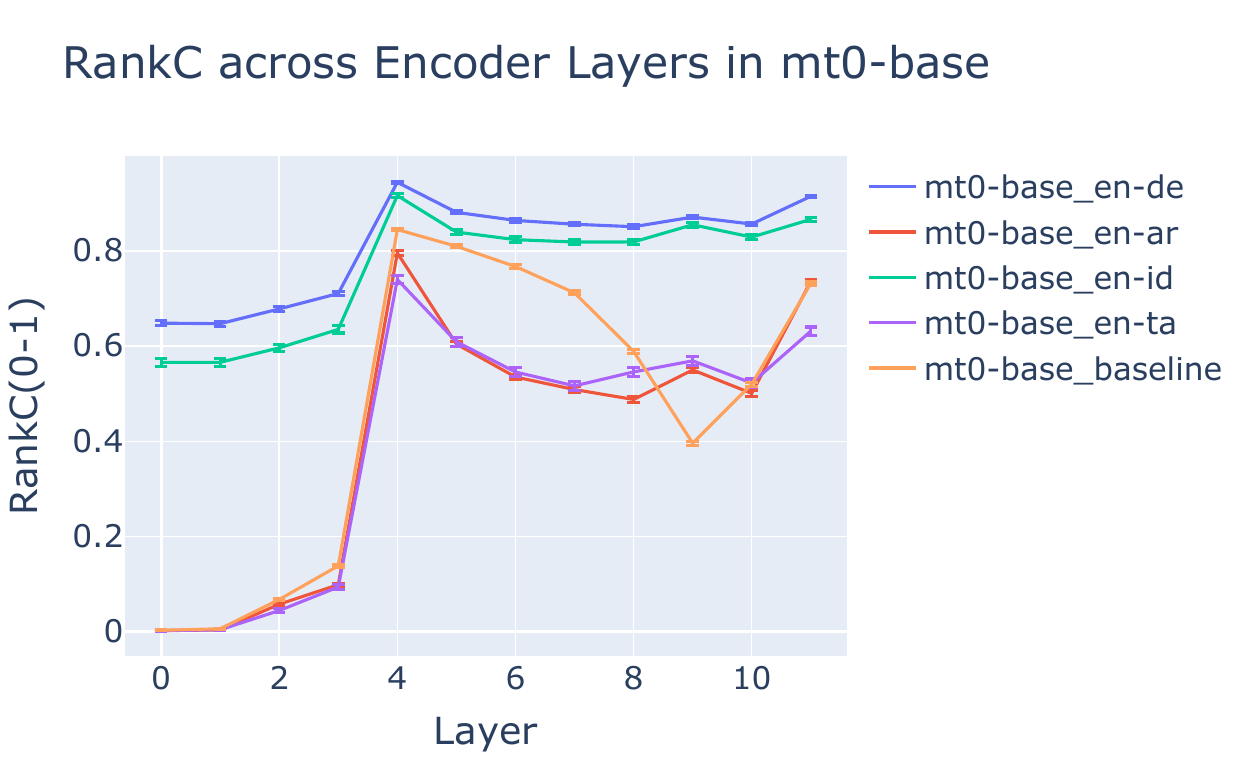}
    }\end{subfigure}
    \begin{subfigure}{0.45\linewidth}{
        \centering
        \includegraphics[trim=0cm 0cm 0cm 2cm,clip=true,width=\linewidth]{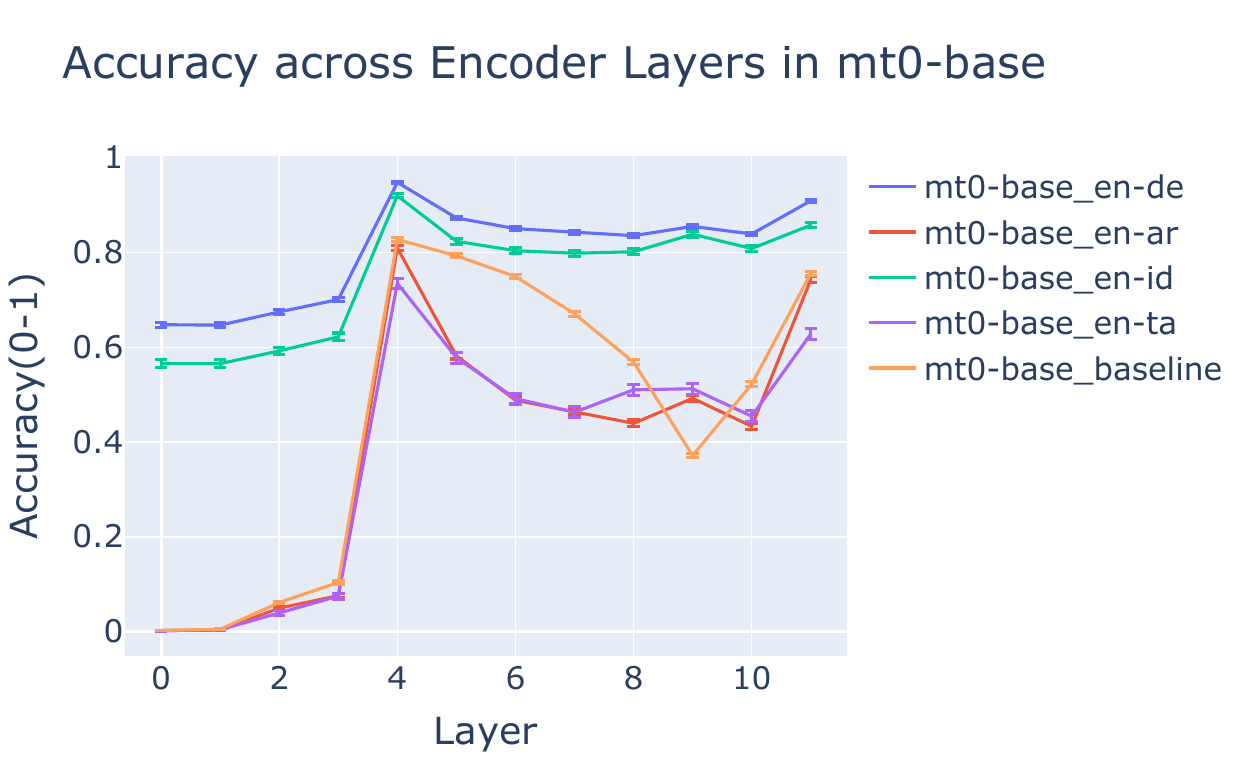}
    }\end{subfigure} \\
    \begin{subfigure}{0.45\linewidth}{
        \centering
        \includegraphics[trim=0cm 0cm 0cm 2cm,clip=true,width=\linewidth]{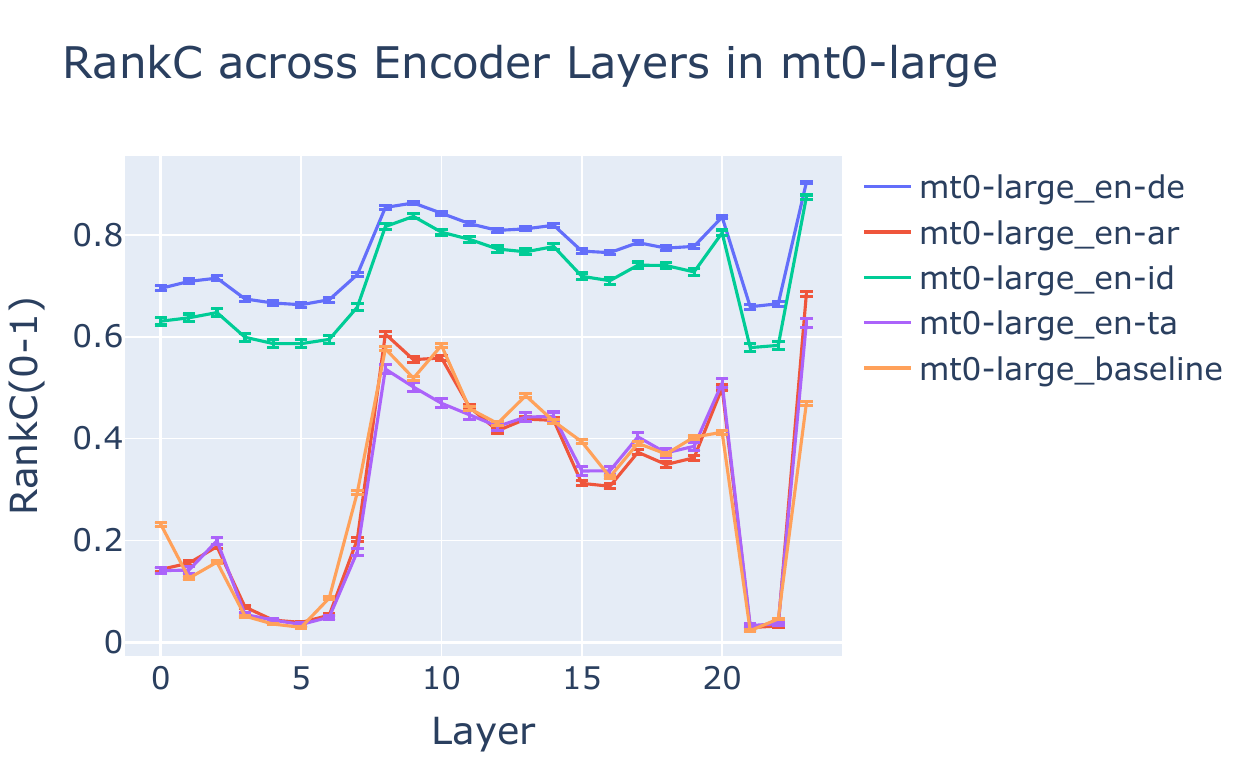}
    }\end{subfigure}
    \begin{subfigure}{0.45\linewidth}{
        \centering
        \includegraphics[trim=0cm 0cm 0cm 2cm,clip=true,width=\linewidth]{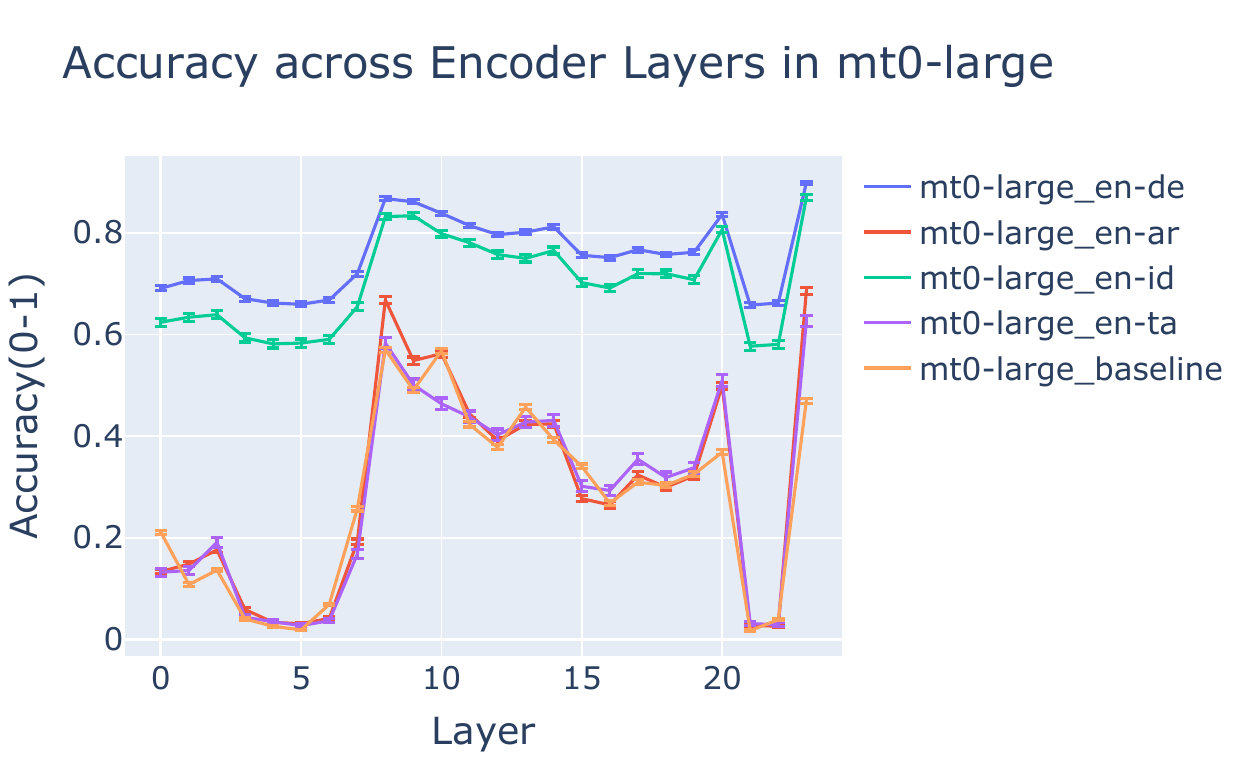}
    }\end{subfigure} \\
    \begin{subfigure}{0.45\linewidth}{
        \centering
        \includegraphics[trim=0cm 0cm 0cm 2cm,clip=true,width=\linewidth]{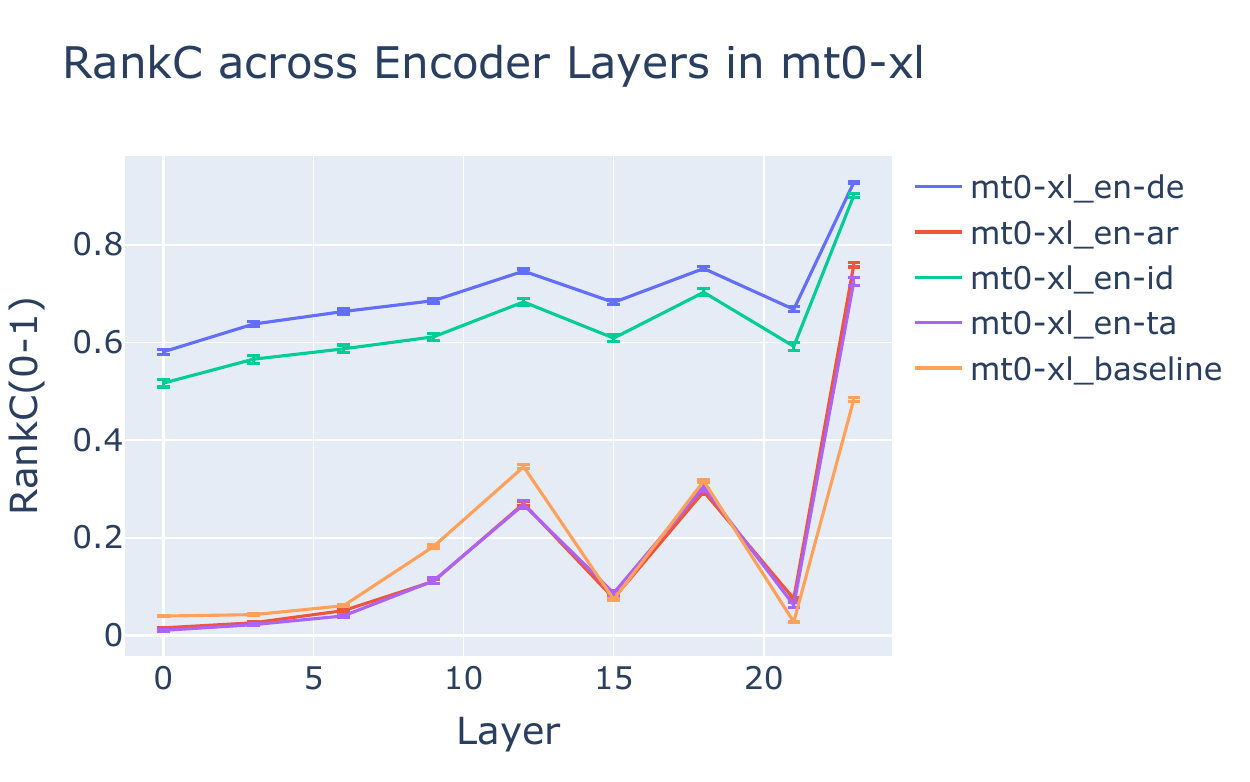}
    }\end{subfigure}
    \begin{subfigure}{0.45\linewidth}{
        \centering
        \includegraphics[trim=0cm 0cm 0cm 2cm,clip=true,width=\linewidth]{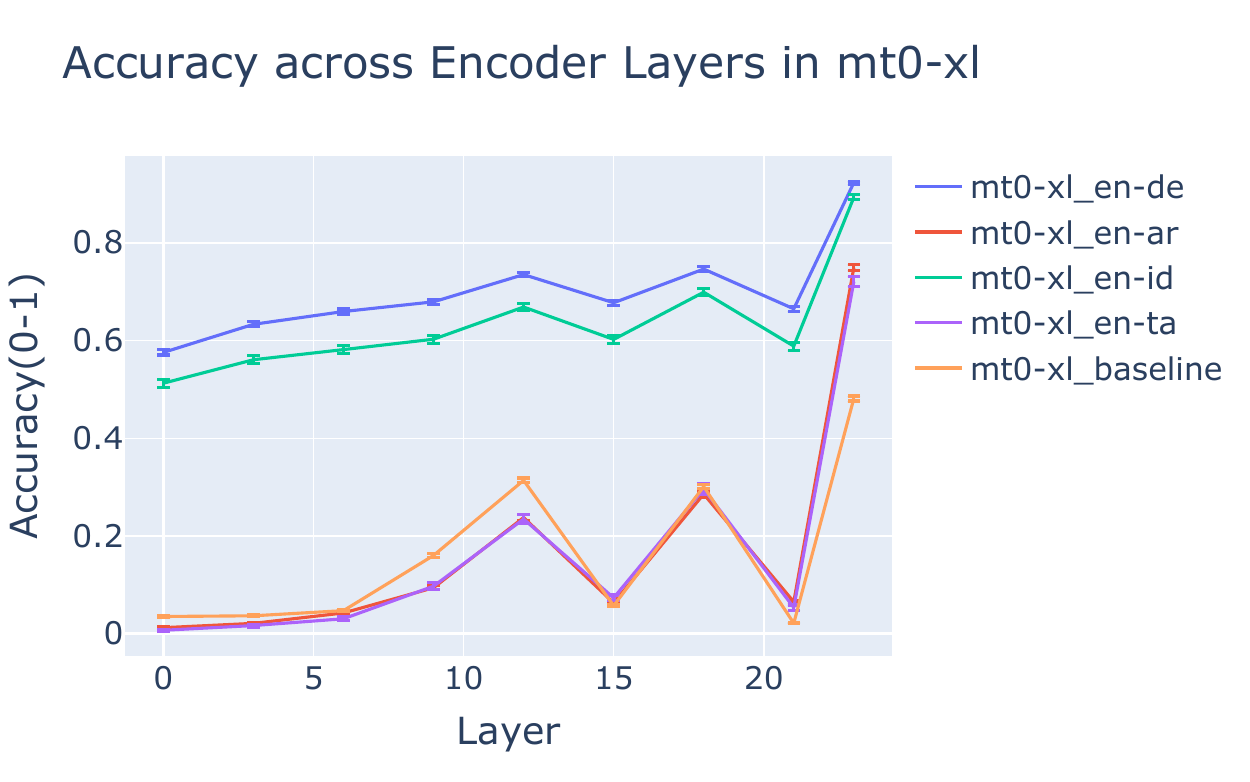}
    }\end{subfigure}
    \caption{mT0 (base, large, XL) layer-wise cross-lingual consistency scores (left: RankC, right: Top@1)}
    \label{fig:mt0-kc}
\end{figure*}

\begin{figure*}[ht!]
    \centering
    \begin{subfigure}{0.45\linewidth}{
        \centering
        \includegraphics[trim=0cm 0cm 0cm 2cm,clip=true,width=\linewidth]{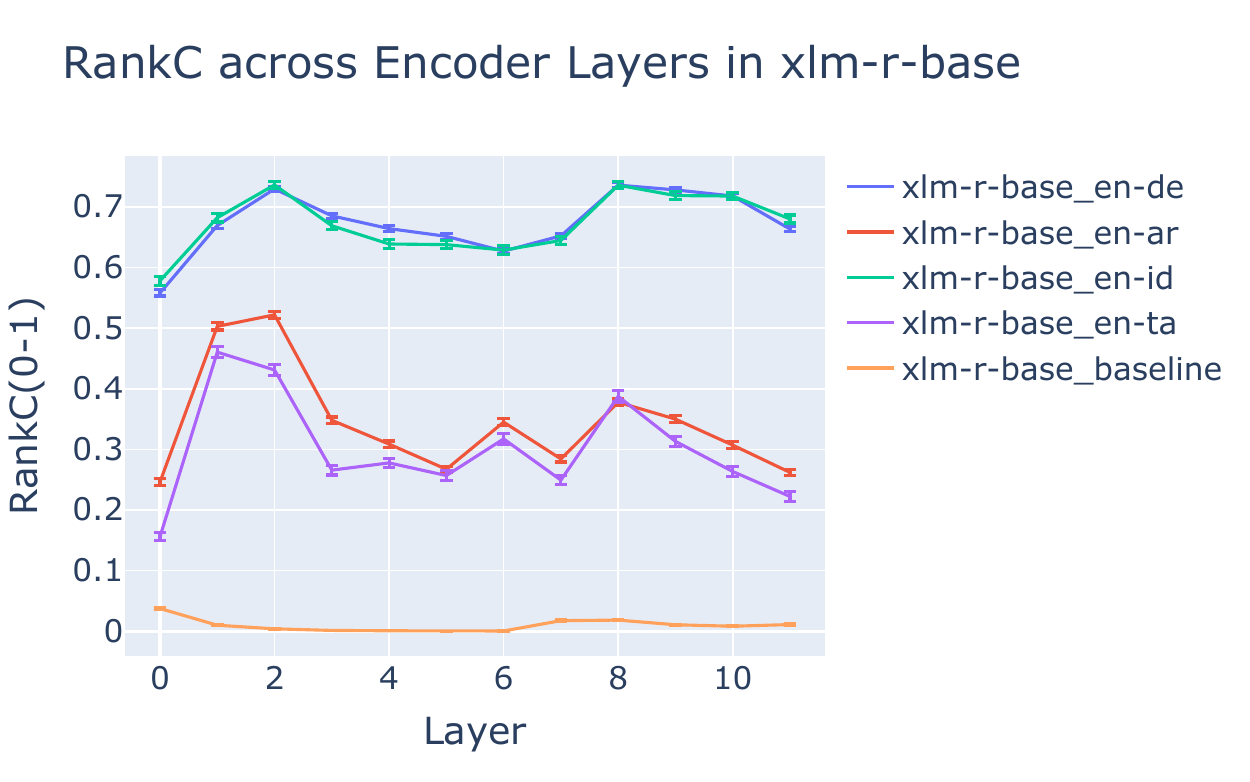}
    }\end{subfigure}
    \begin{subfigure}{0.45\linewidth}{
        \centering
        \includegraphics[trim=0cm 0cm 0cm 2cm,clip=true,width=\linewidth]{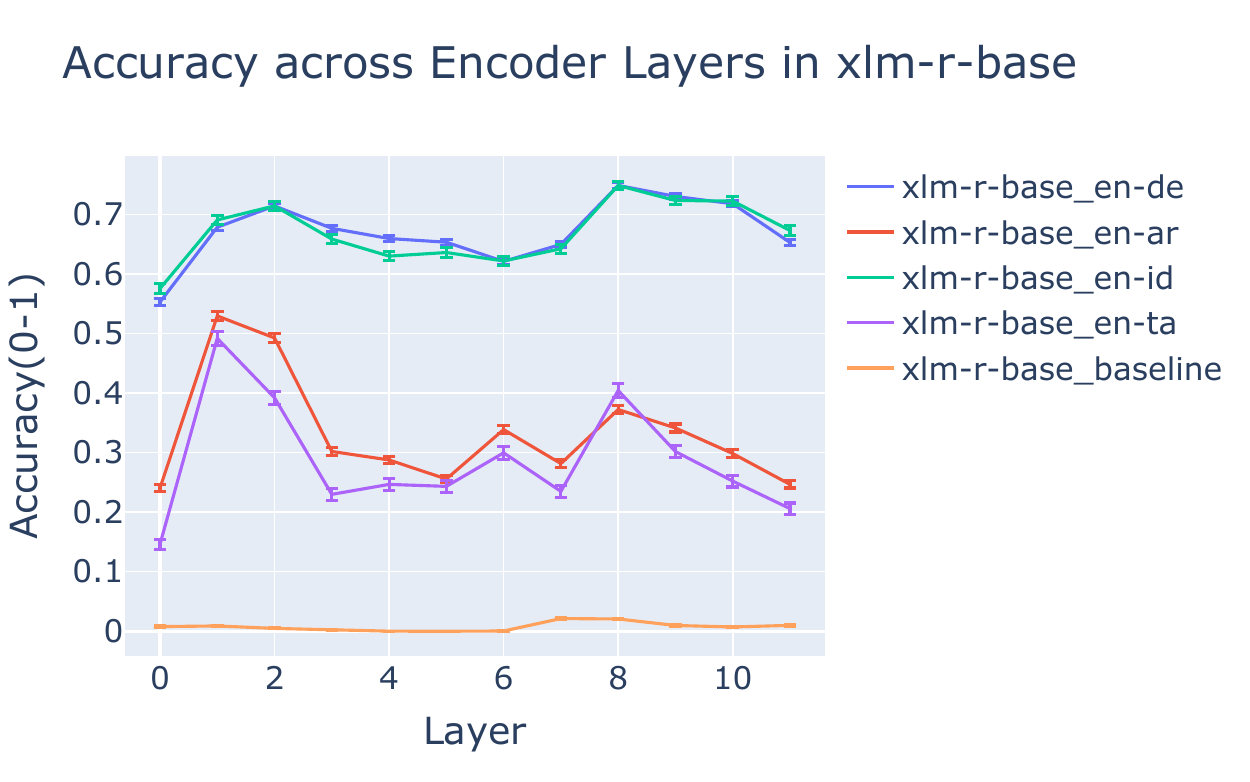} 
    }\end{subfigure} \\
    \begin{subfigure}{0.45\linewidth}{
        \centering
        \includegraphics[trim=0cm 0cm 0cm 2cm,clip=true,width=\linewidth]{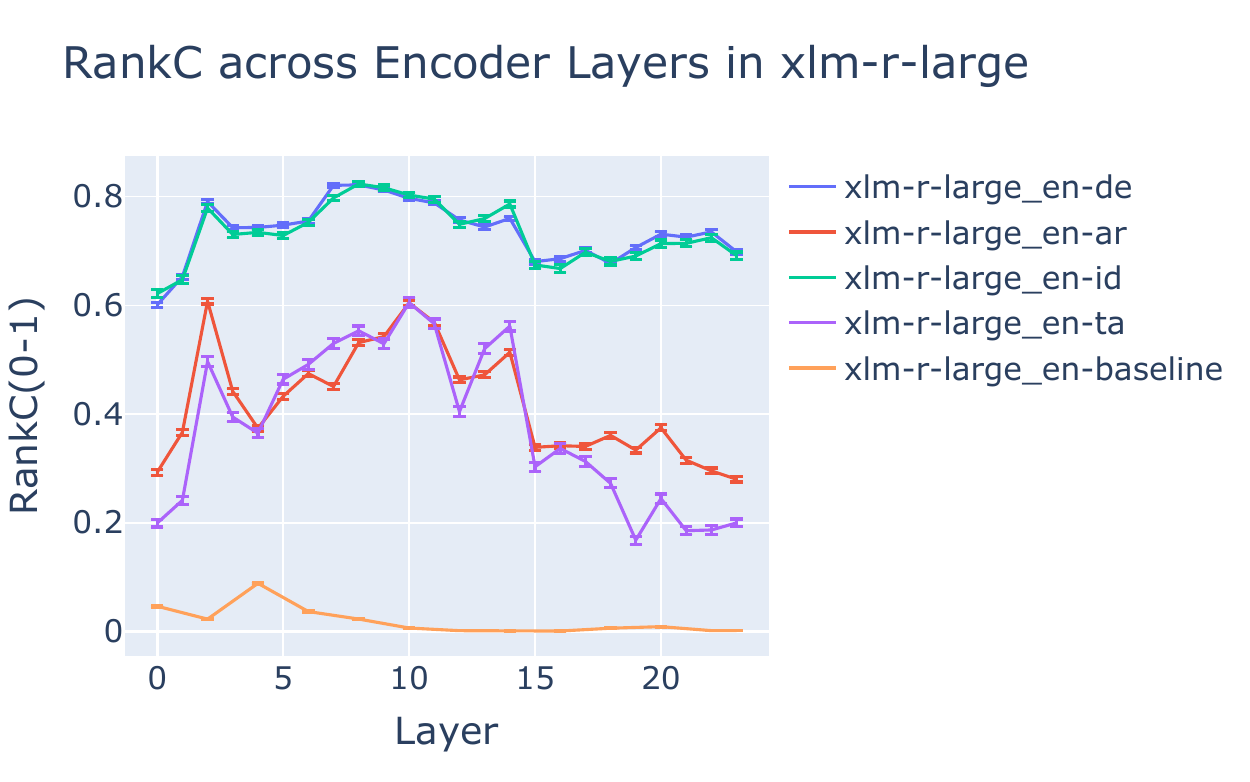}
    }\end{subfigure}
    \begin{subfigure}{0.45\linewidth}{
        \centering
        \includegraphics[trim=0cm 0cm 0cm 2cm,clip=true,width=\linewidth]{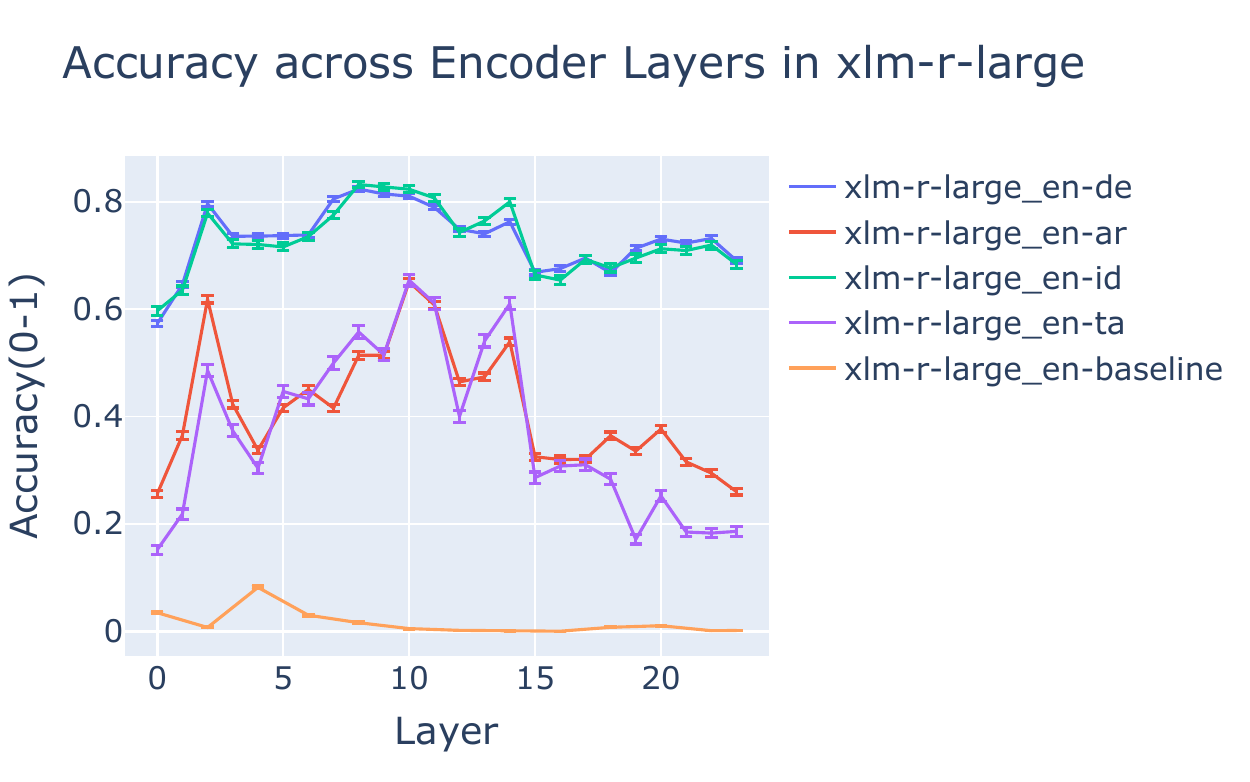}  
    }\end{subfigure} \\
    \begin{subfigure}{0.45\linewidth}{
        \centering
        \includegraphics[trim=0cm 0cm 0cm 2cm,clip=true,width=\linewidth]{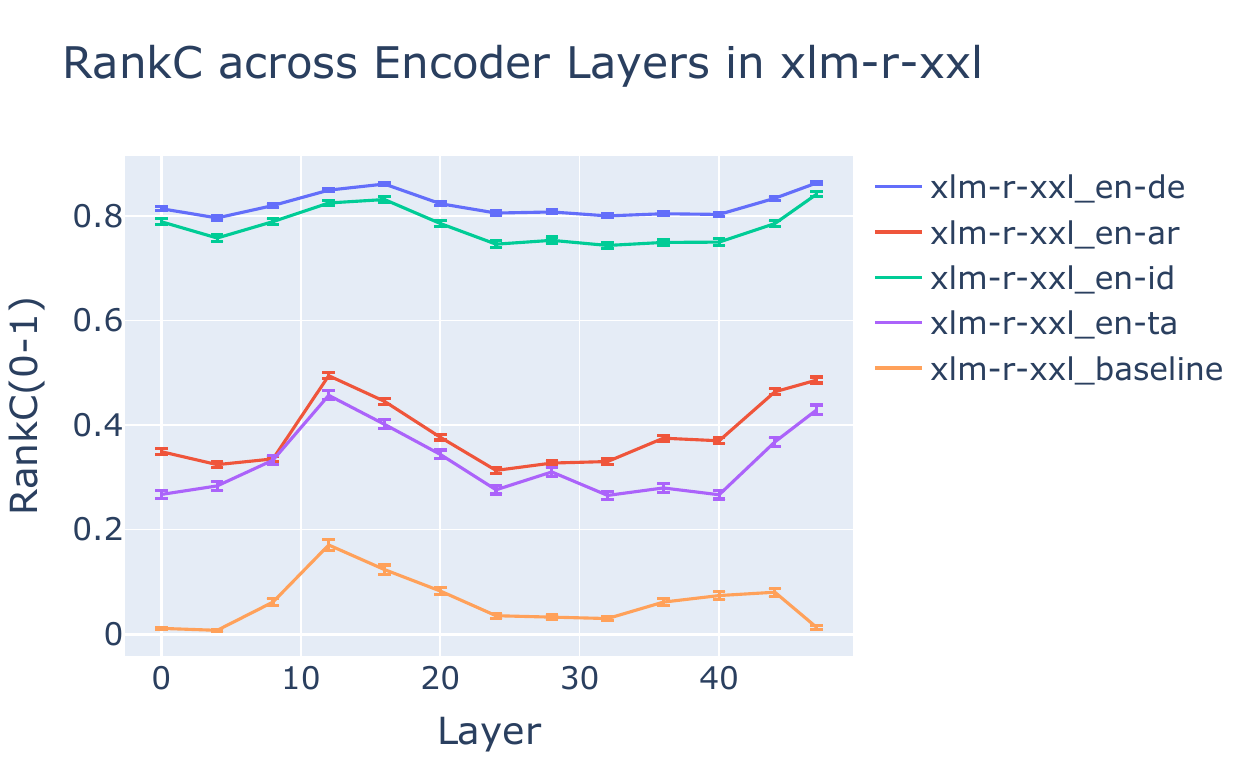}
    }\end{subfigure}
    \begin{subfigure}{0.45\linewidth}{
        \centering
        \includegraphics[trim=0cm 0cm 0cm 2cm,clip=true,width=\linewidth]{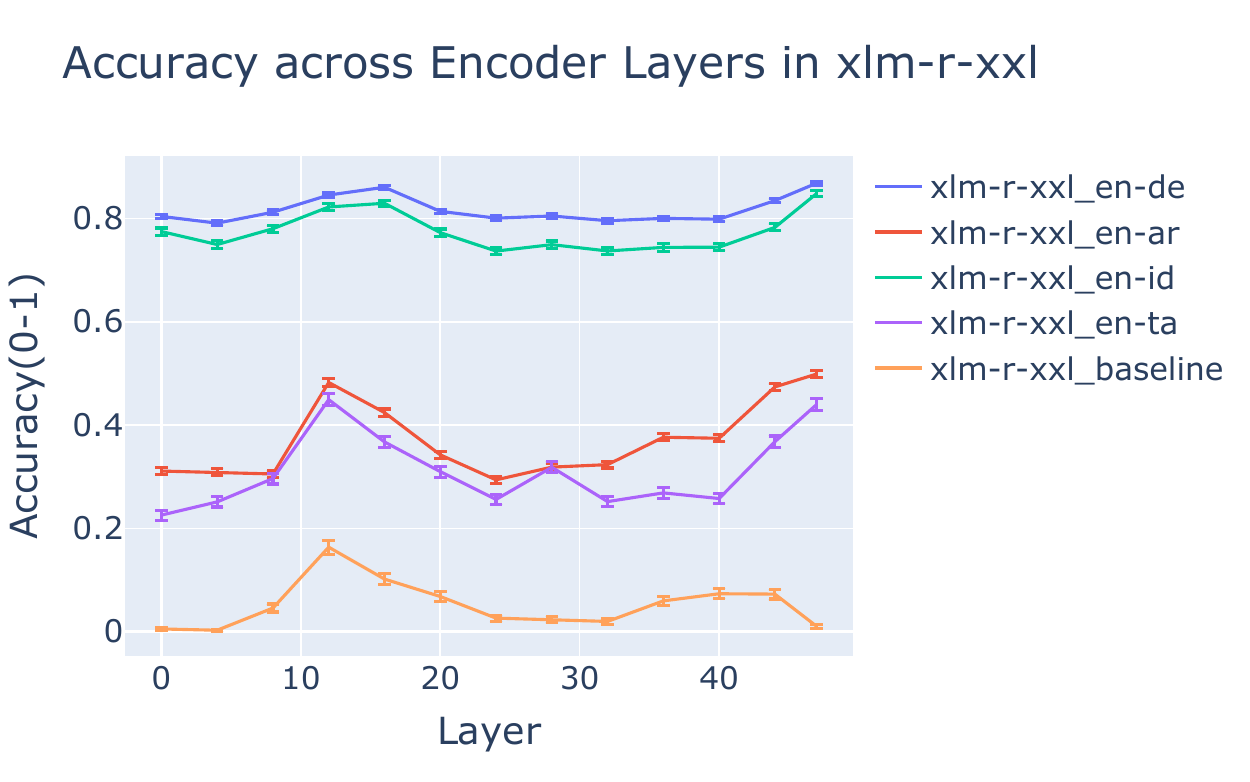}  
    }\end{subfigure}
    \caption{xlm-r (base, large, XXL) layer-wise crosslingual consistency scores (left: RankC, right: Top@1)}
    \label{fig:xlmr-kc}
\end{figure*}

\begin{figure*}[ht!]
    \centering
    \begin{subfigure}{0.45\linewidth}{
        \centering
        \includegraphics[trim=0cm 0cm 0cm 2cm,clip=true,width=\linewidth]{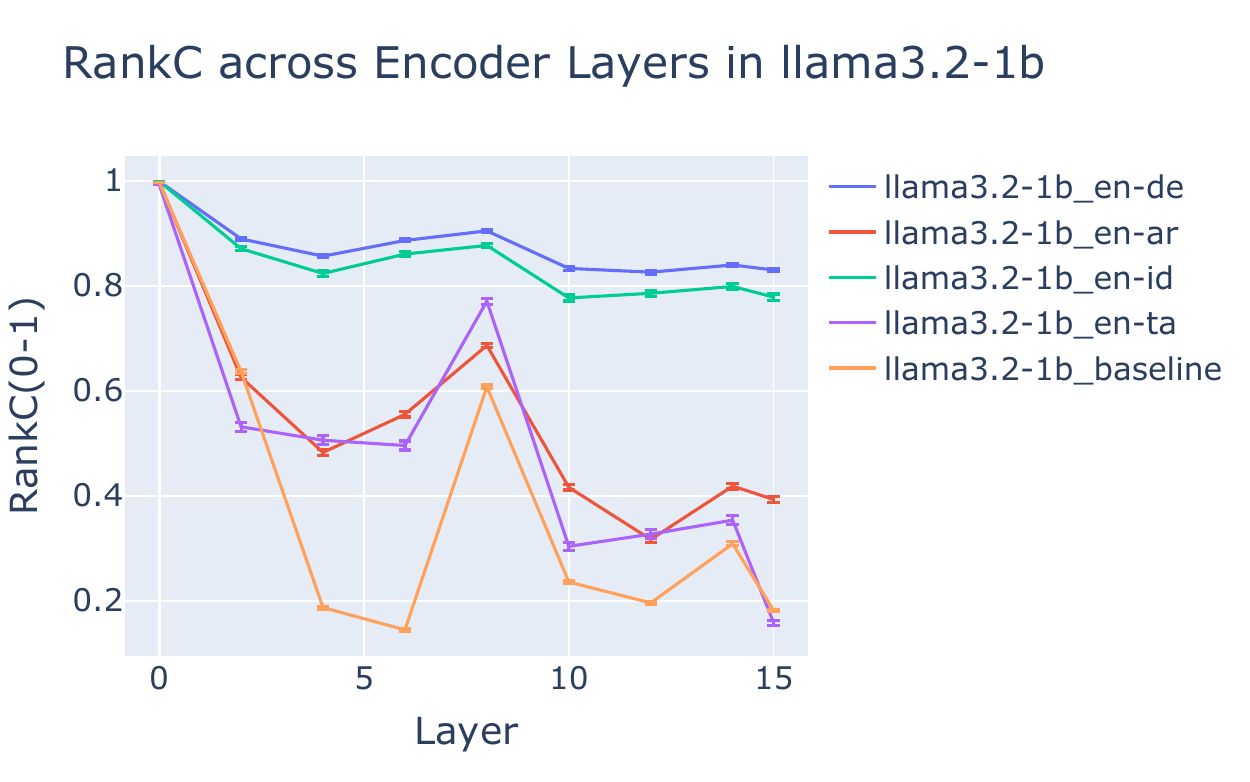}
    }\end{subfigure}
    \begin{subfigure}{0.45\linewidth}{
        \centering
        \includegraphics[trim=0cm 0cm 0cm 2cm,clip=true,width=\linewidth]{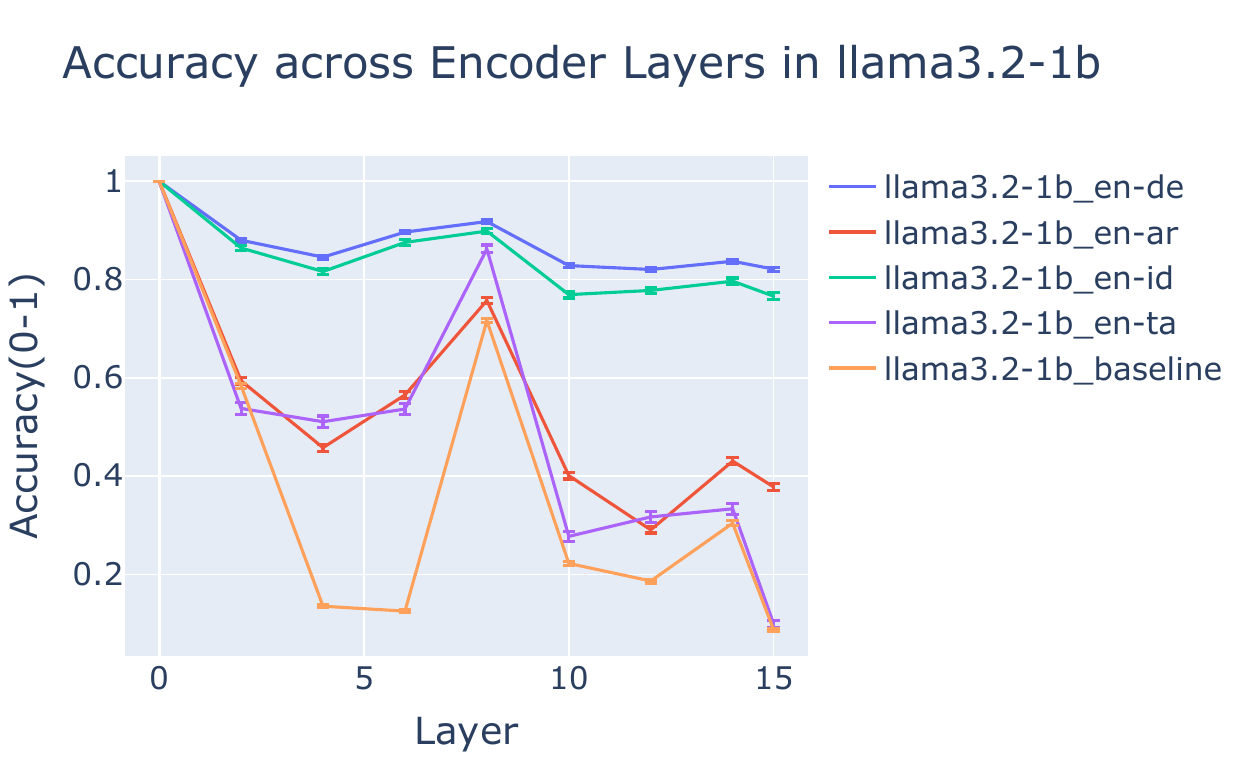} 
    }\end{subfigure} \\
    \begin{subfigure}{0.45\linewidth}{
        \centering
        \includegraphics[trim=0cm 0cm 0cm 2cm,clip=true,width=\linewidth]{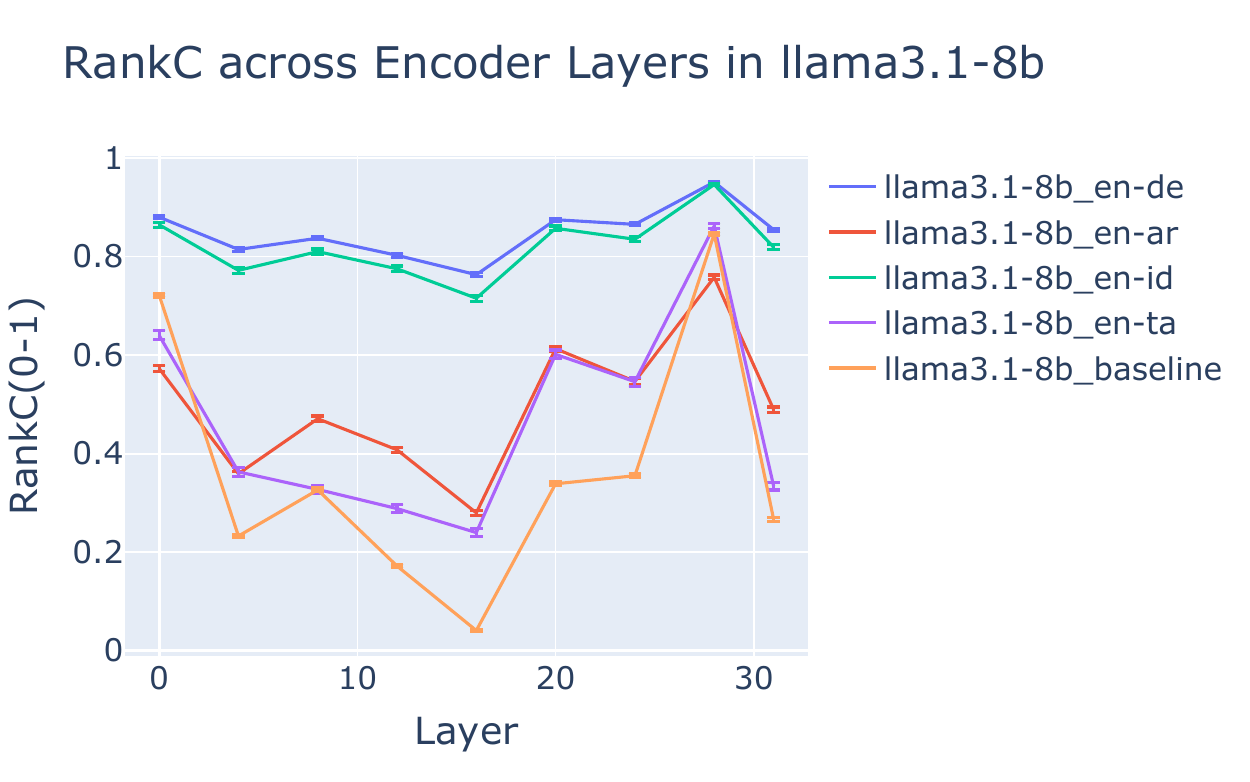}
    }\end{subfigure}
    \begin{subfigure}{0.45\linewidth}{
        \centering
        \includegraphics[trim=0cm 0cm 0cm 2cm,clip=true,width=\linewidth]{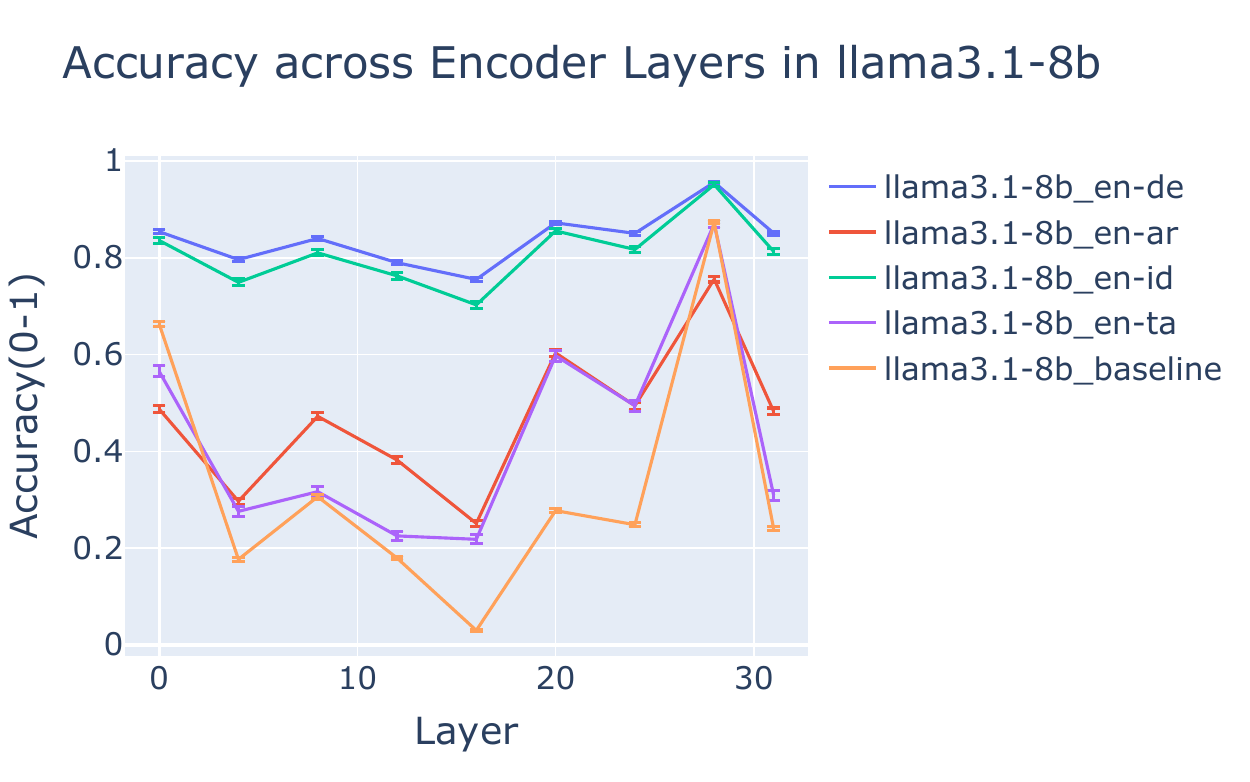}  
    }\end{subfigure}
    \caption{llama 3 (1B, 8B) layer-wise cross-lingual consistency scores (left: RankC, right: Top@1)}
    \label{fig:llama3-kc}
\end{figure*}

Refer to Figure \ref{fig:mt0-kc}, \ref{fig:xlmr-kc}, and \ref{fig:llama3-kc}.

\subsubsection{Overall Consistency of Output distribution}
\label{sess:findings_in_details_Overall_Consistency}
\begin{figure*}[ht!]
    \centering
    \begin{subfigure}{0.32\linewidth}{
        \centering
        \includegraphics[trim=0cm 0cm 0cm 2.5cm,clip=true,width=\linewidth]{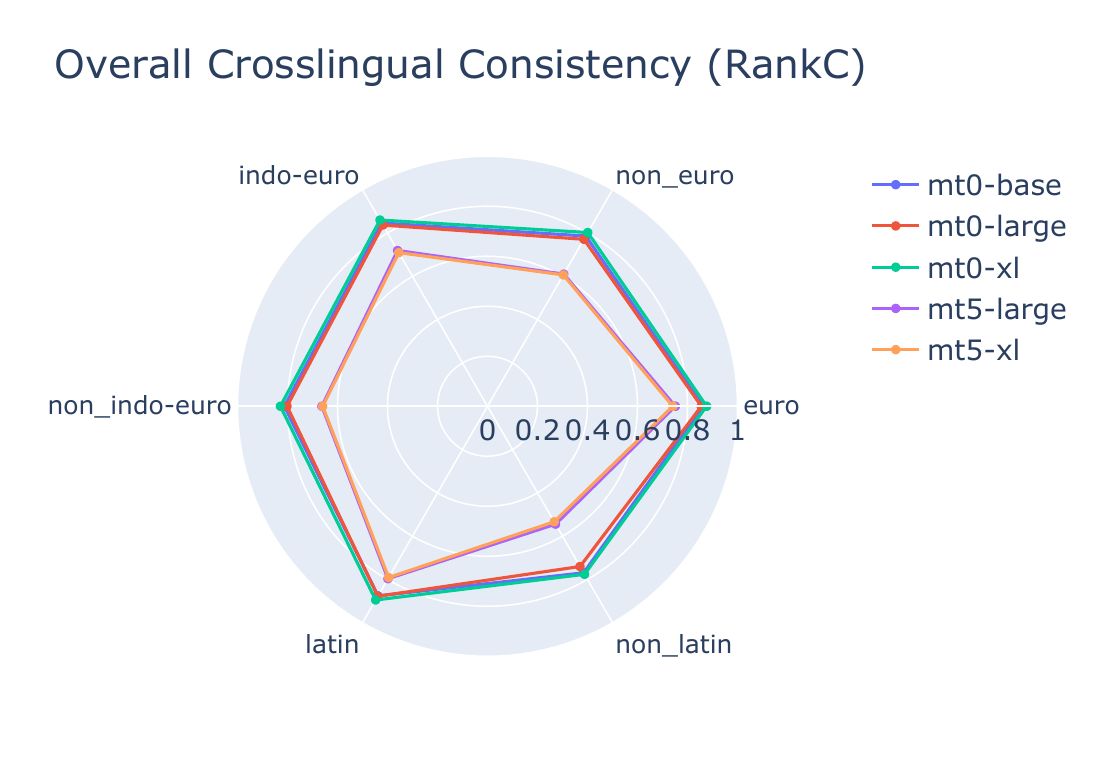}
        \label{app_fig:encoder-decoder-overall-rankc-scores-categorized}
    }\end{subfigure}
    \begin{subfigure}{0.32\linewidth}{
        \centering
        \includegraphics[trim=0cm 0cm 0cm 2.5cm,clip=true,width=\linewidth]{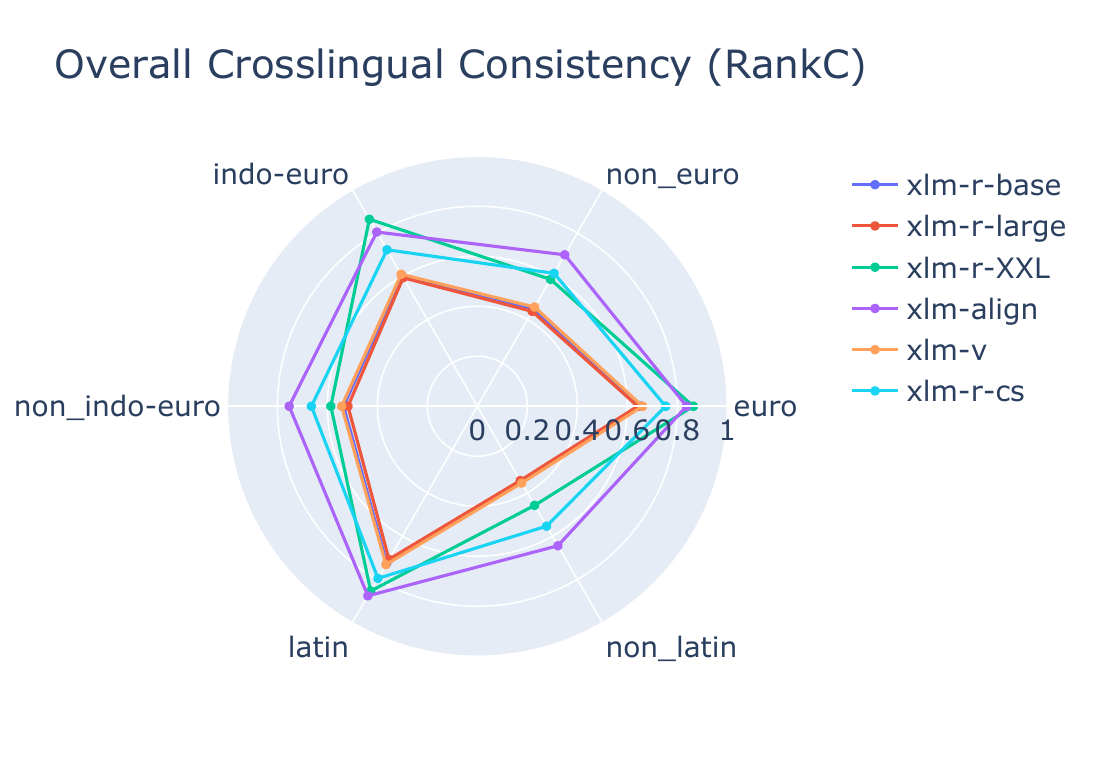}
        \label{app_fig:encoder-overall-rankc-scores-categorized}
    }\end{subfigure}
            \centering
     \begin{subfigure}{0.32\linewidth}{
        \centering
        \includegraphics[trim=0cm 0cm 0cm 2.5cm,clip=true,width=\linewidth]{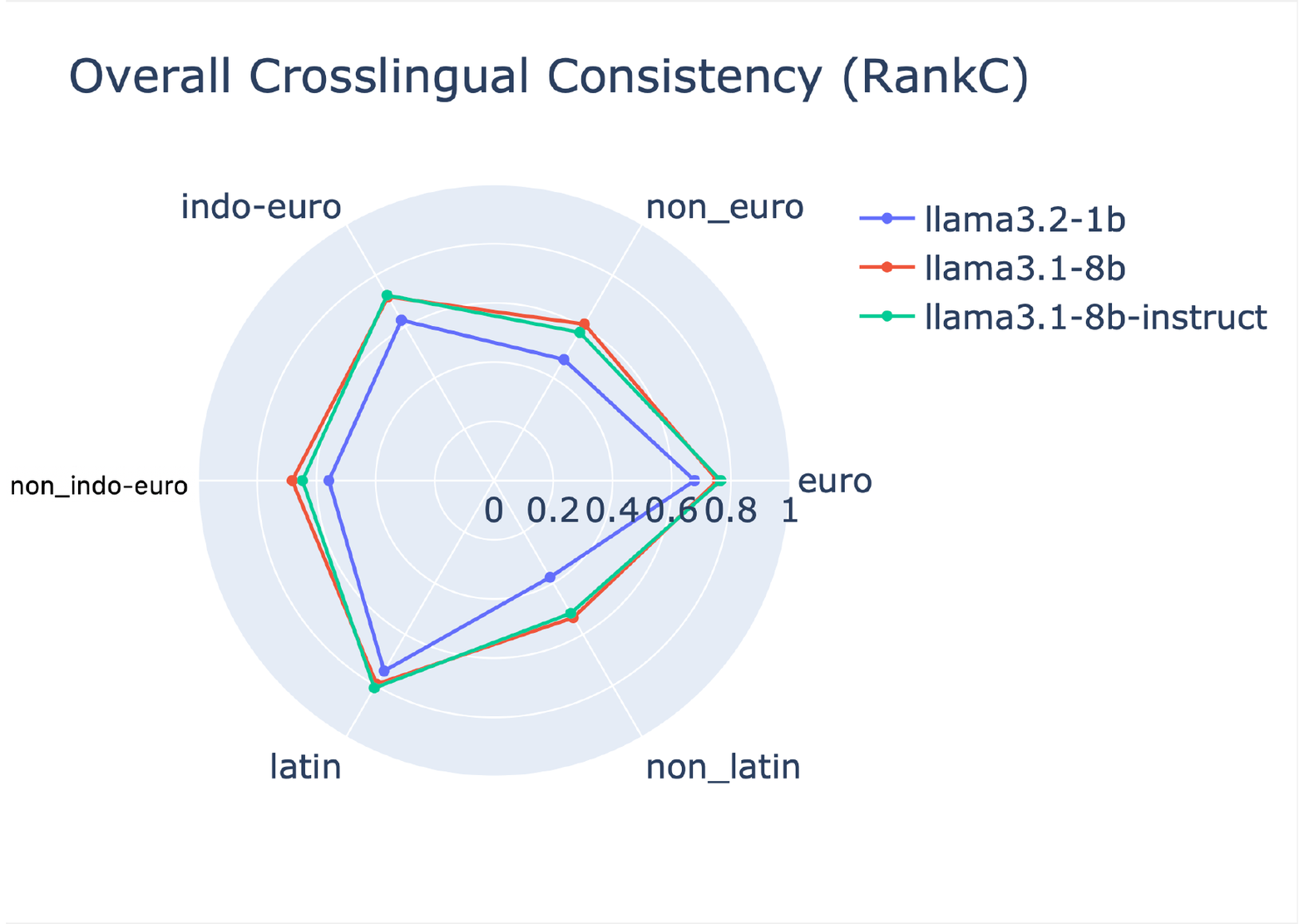}
        \label{app_fig:decoder-overall-rankc-scores-categorized}
    }\end{subfigure}
    \begin{subfigure}{0.32\linewidth}{
        \centering
        \includegraphics[trim=0cm 0cm 0cm 2.5cm,clip=true,width=\linewidth]{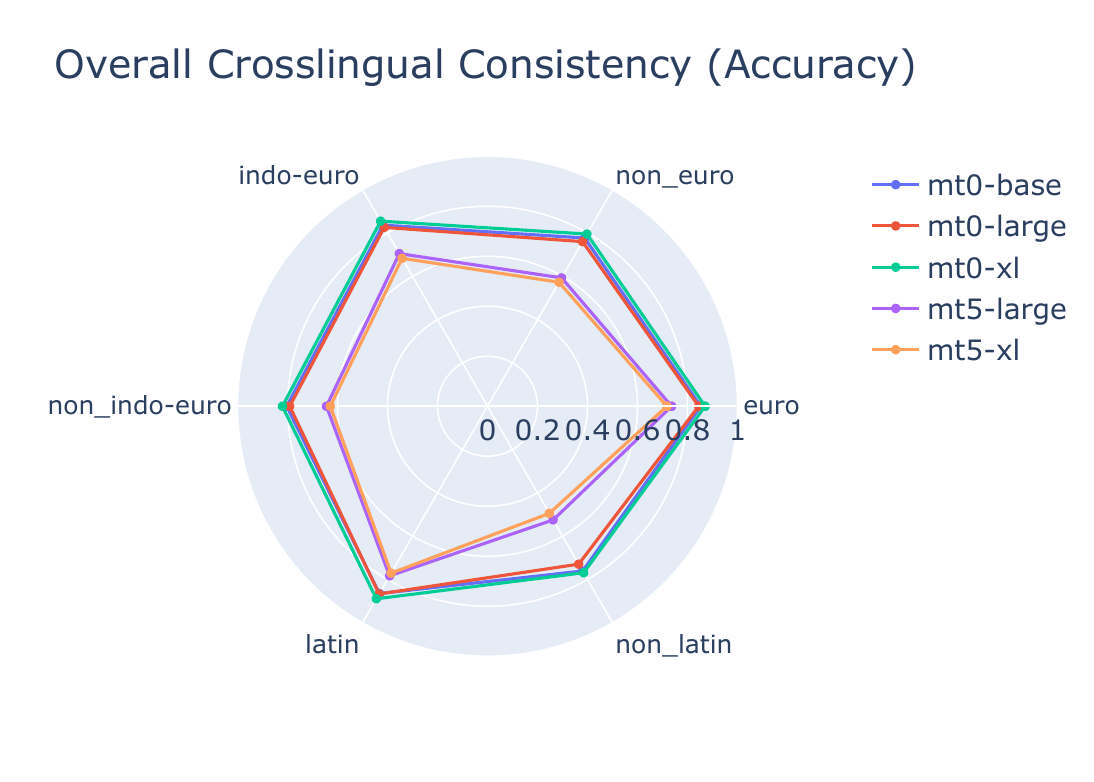}
        \label{app_fig:encoder-decoder-overall-acc-scores-categorized}
    }\end{subfigure}
    \begin{subfigure}{0.32\linewidth}{
        \centering
        \includegraphics[trim=0cm 0cm 0cm 2.5cm,clip=true,width=\linewidth]{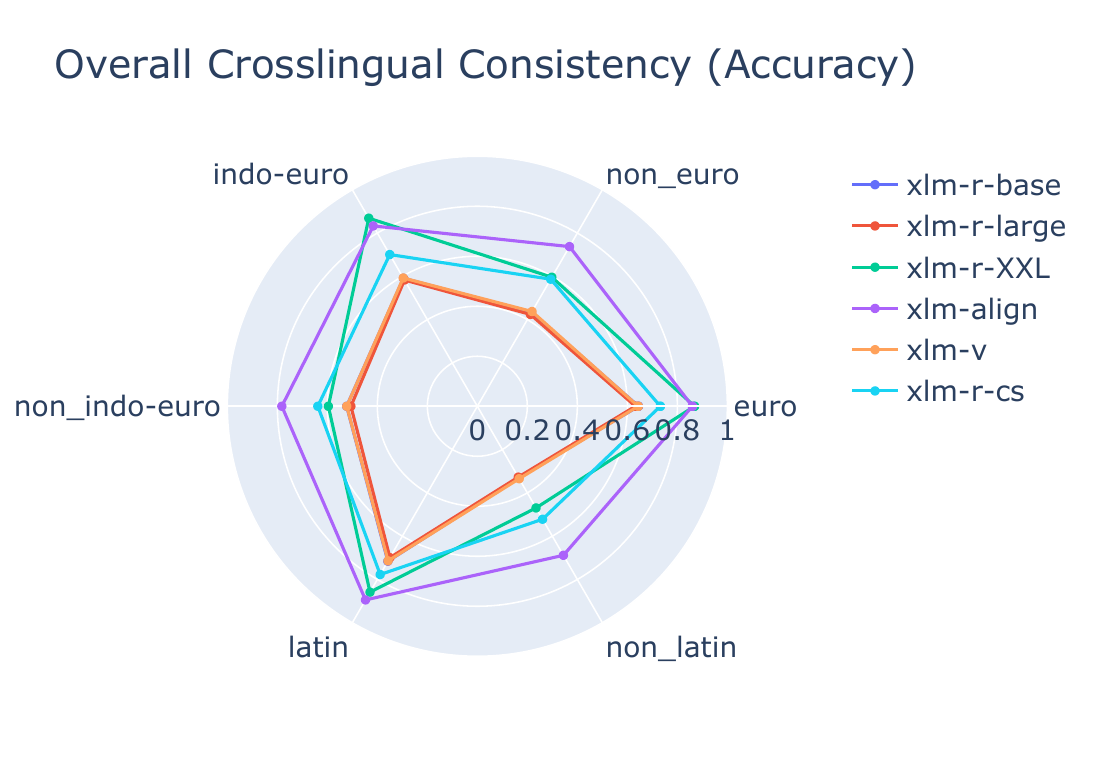}
        \label{app_fig:encoder-overall-accuracy-scores-categorized}
    }\end{subfigure}
    \begin{subfigure}{0.32\linewidth}{
        \centering
        \includegraphics[trim=0cm 0cm 0cm 2.5cm,clip=true,width=\linewidth]{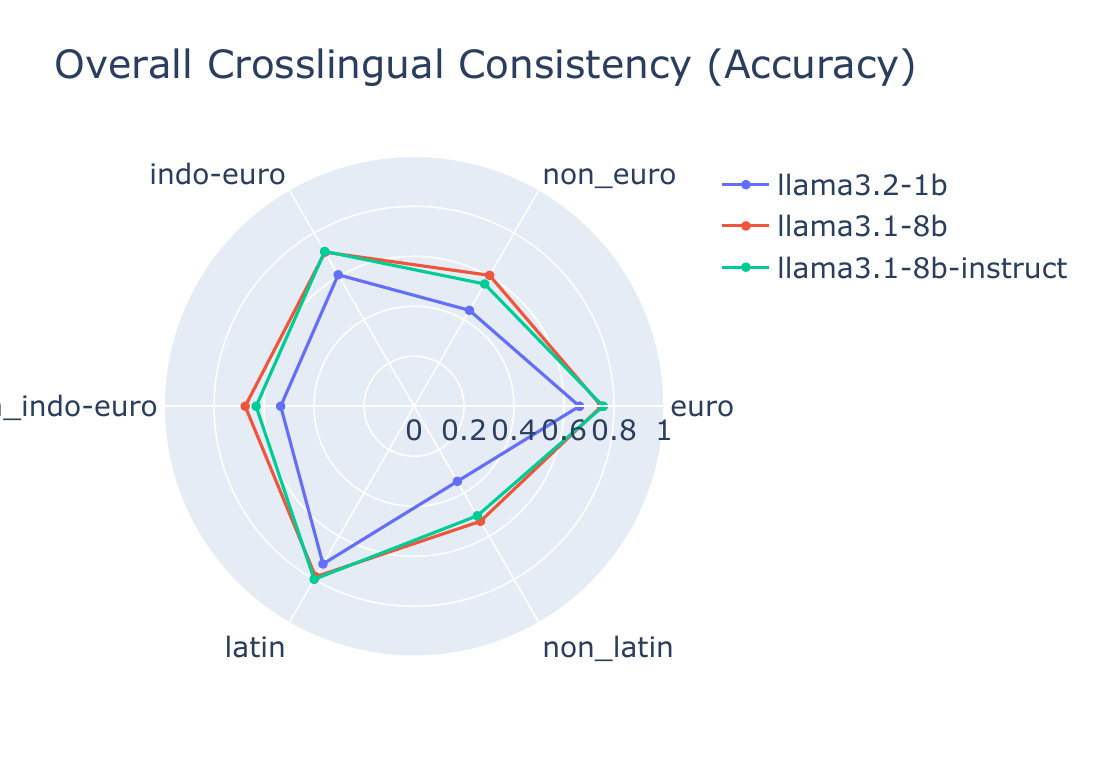}
        \label{app_fig:decoder-overall-acc-scores-categorized}
    }\end{subfigure}
    \caption{Overall cross-lingual consistency across different transformer types (left: encoder-decoder, middle: encoder, right: decoder)  grouped by 3 factors (geographics: europe \& non\_europe, language family: indo-european \& non\_indo-european, writing scripts: latin \& non\_latin).}
    \label{app_fig:overall-encoder-encoder-decoder-crosslingual-factors}
\end{figure*}

\begin{figure*}[ht!]
    \centering
    \begin{subfigure}{\linewidth}{
        \centering
        \includegraphics[ width=\linewidth]{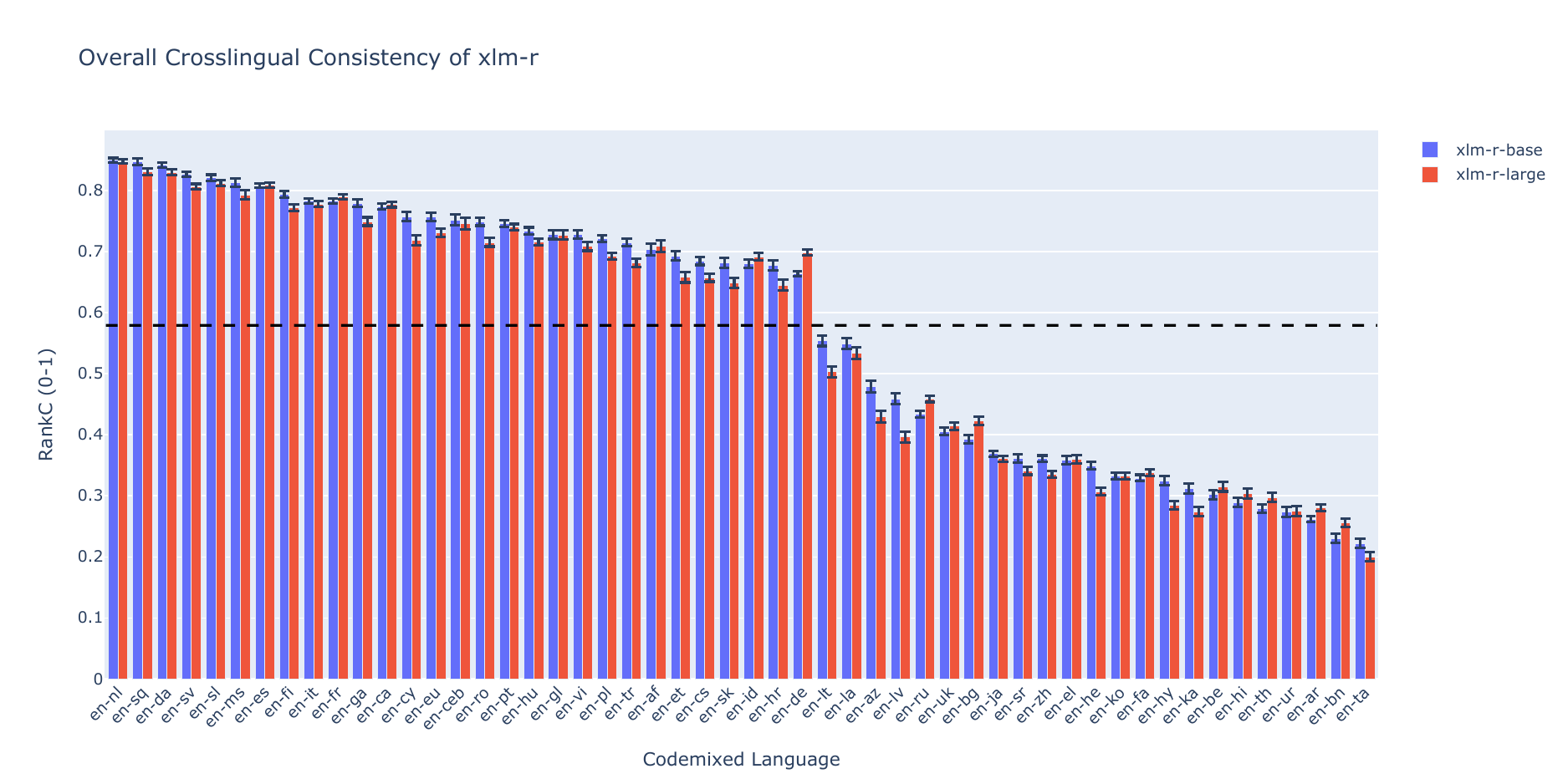}
        \label{fig:xlm-r-overall-crosslingual-consistency-rankC}
    }\end{subfigure}
    \\
    \begin{subfigure}{\linewidth}{
        \centering
        \includegraphics[ width=\linewidth]{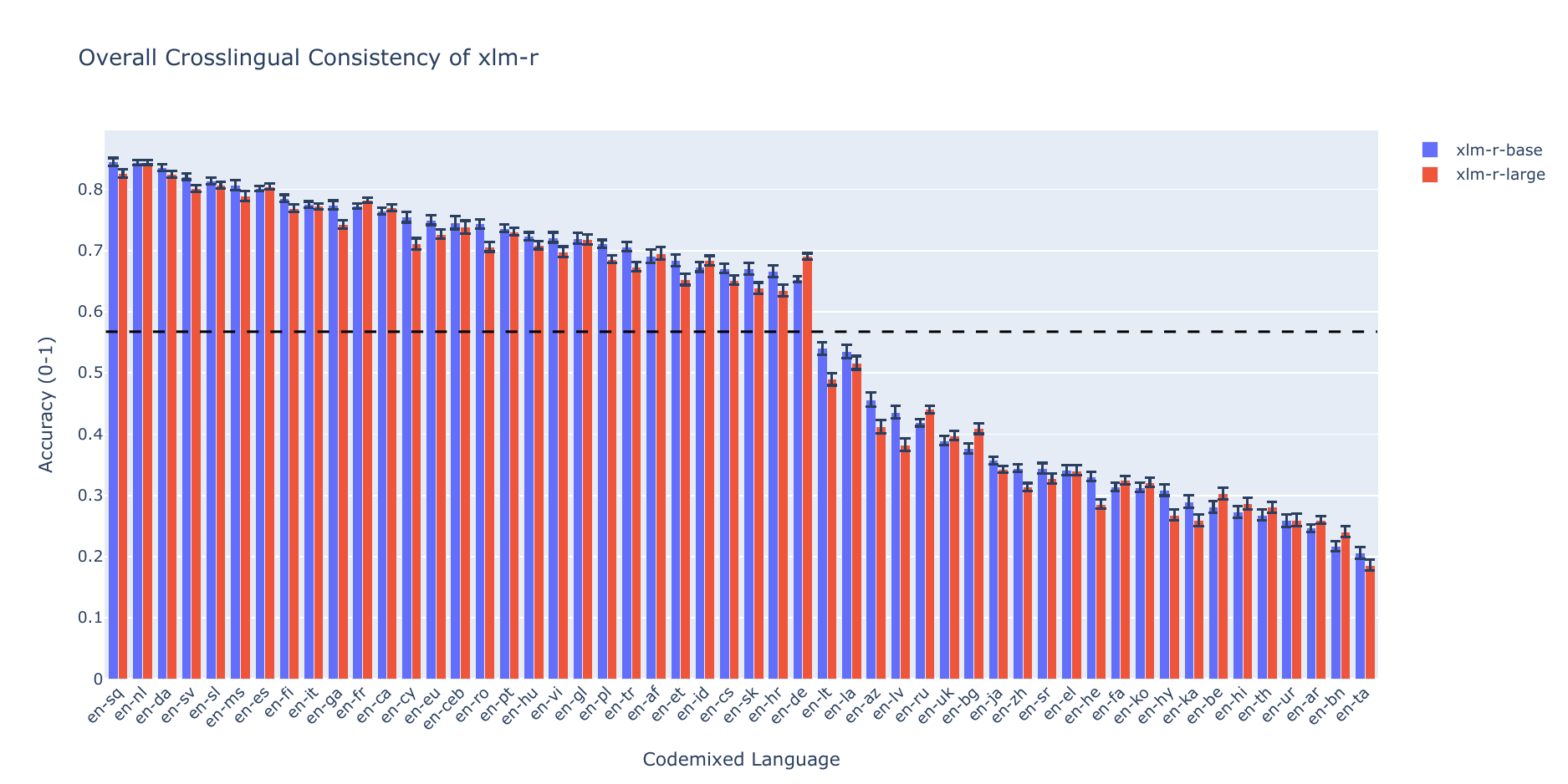}
        \label{fig:xlm-r-overall-crosslingual-consistency-acc}
    }\end{subfigure}
    \caption{Cross-lingual consistency scores across languages of xlm-r (top: RankC, bottom: Top@1 Accuracy). Note: The dashed line here is the average corresponding consistency scores of xlm-r-base across languages}
    \label{fig:xlm-r-overall-crosslingual-consistency}
\end{figure*}

\begin{figure*}[ht]
    \centering
    \begin{subfigure}{\linewidth}{
        \centering
        \includegraphics[ width=\linewidth]{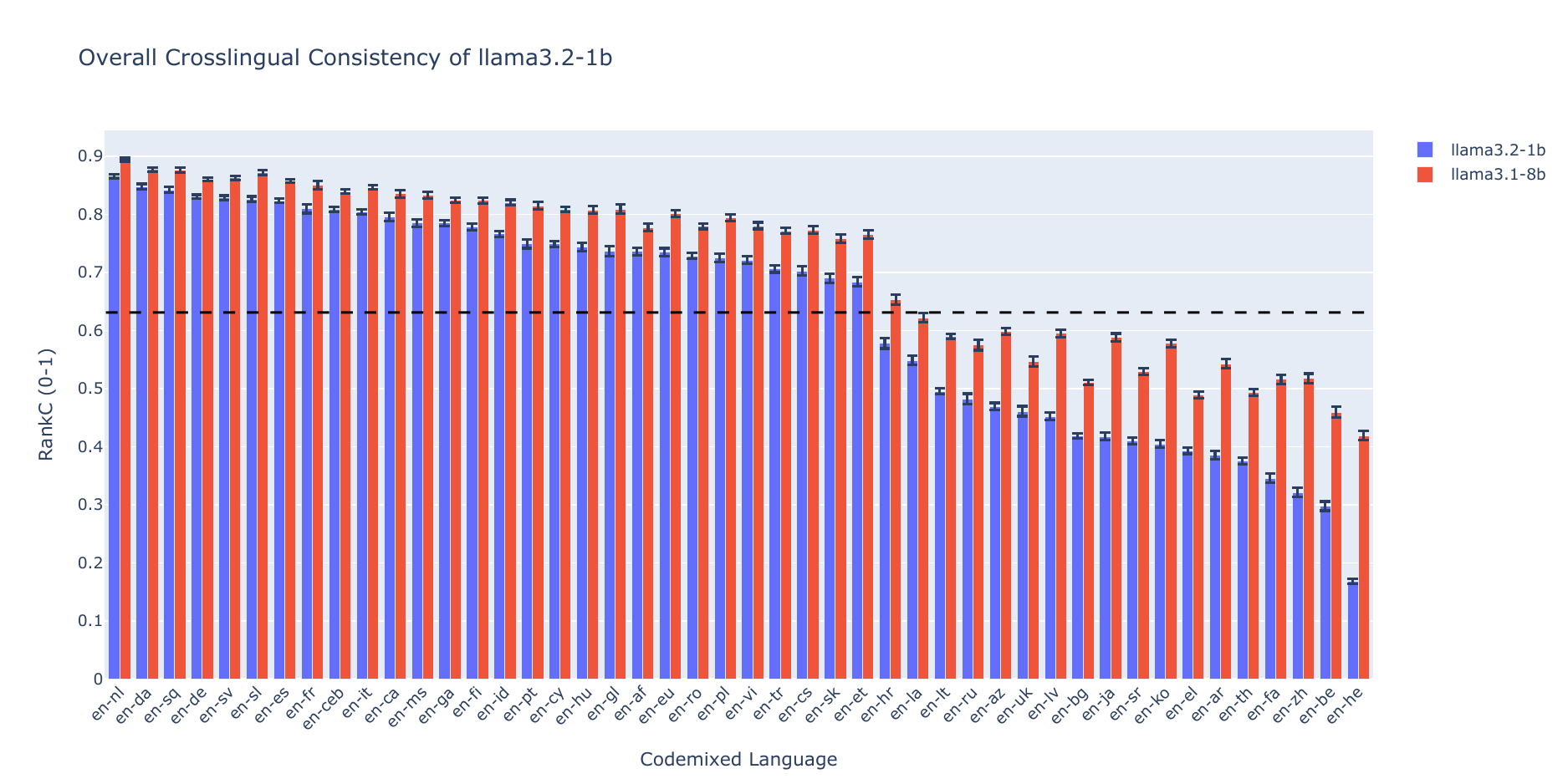}
        \label{fig:llama3-overall-crosslingual-consistency-rankC}
    }\end{subfigure}
    \\
    \begin{subfigure}{\linewidth}{
        \centering
        \includegraphics[ width=\linewidth]{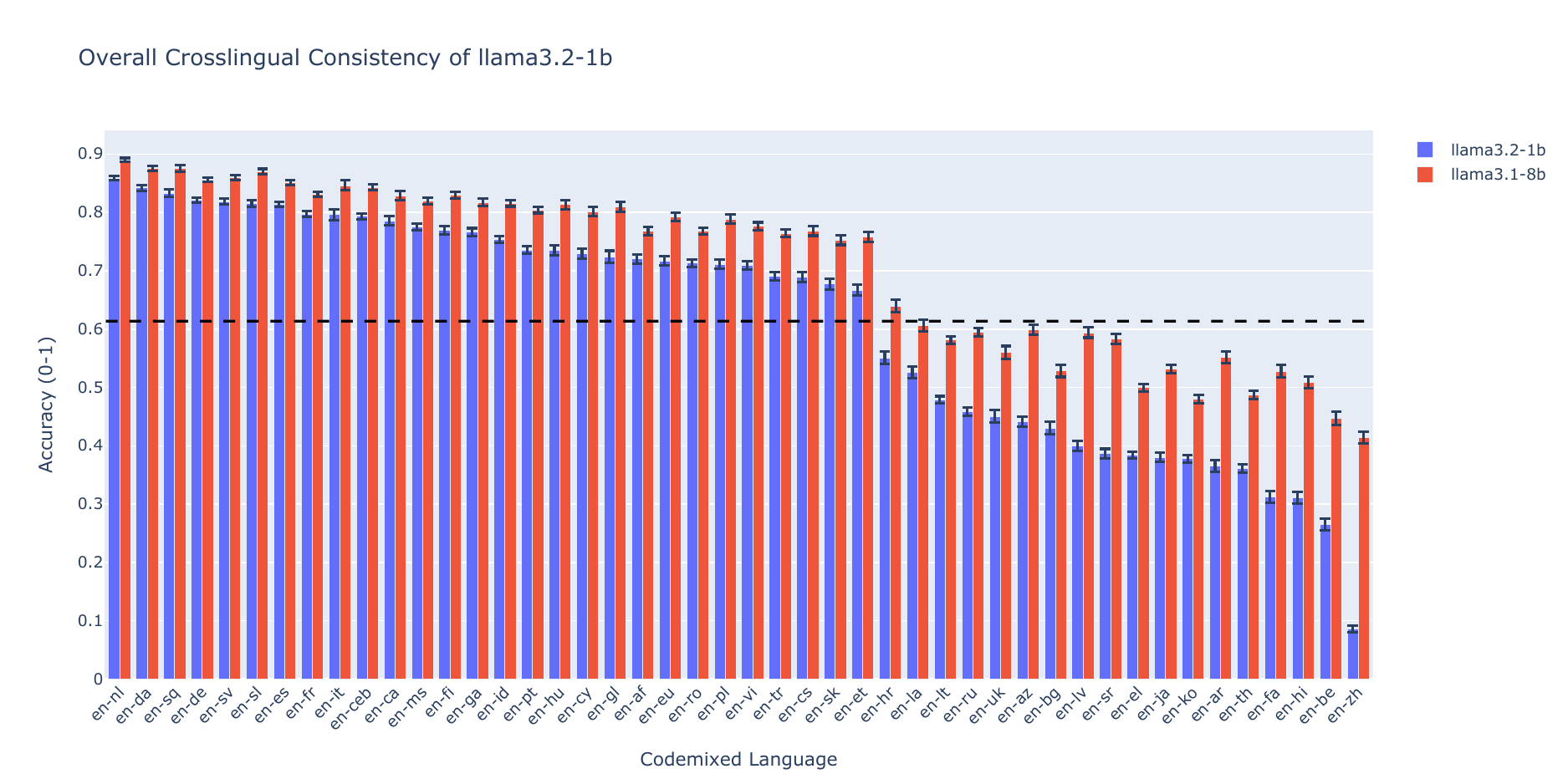}
        \label{fig:llama3-overall-crosslingual-consistency-acc}
    }\end{subfigure}
    \caption{Cross-lingual consistency scores across languages of llama 3 (top: RankC, bottom: Top@1 Accuracy). Note: The dashed line here is the average corresponding consistency scores of llama3.2-1b across languages}
    \label{fig:llama-overall-crosslingual-consistency}
\end{figure*}

Refer to Figure \ref{app_fig:overall-encoder-encoder-decoder-crosslingual-factors}, \ref{fig:xlm-r-overall-crosslingual-consistency}, and \ref{fig:llama-overall-crosslingual-consistency}.
\begin{figure*}[ht!]
    \centering
    \begin{subfigure}{\linewidth}{
        \centering
        \includegraphics[ width=\linewidth]{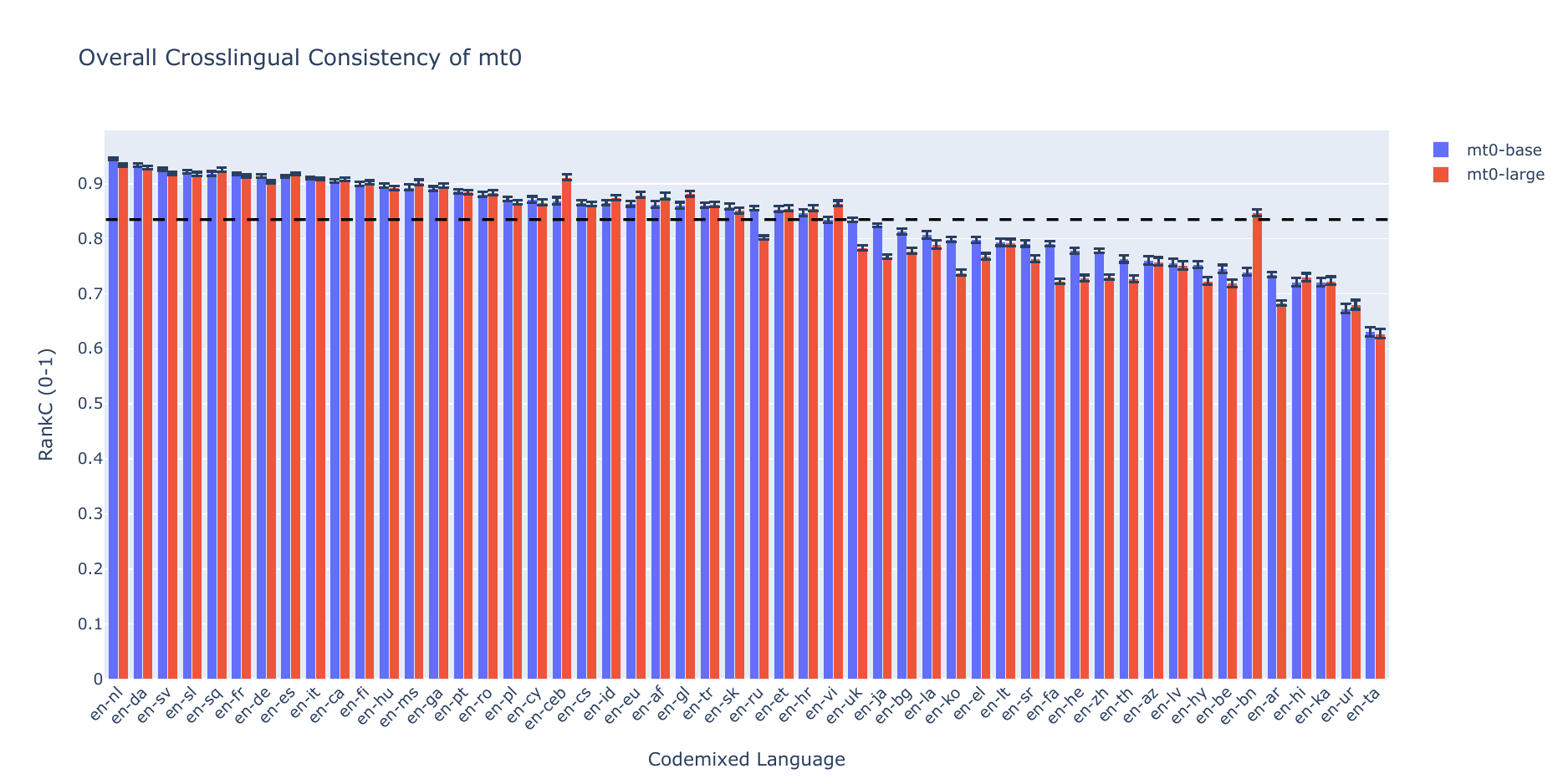}
        \label{fig:mt0-base-overall-crosslingual-consistency-rankC}
    }\end{subfigure}
    \\
    \begin{subfigure}{\linewidth}{
        \centering
        \includegraphics[ width=\linewidth]{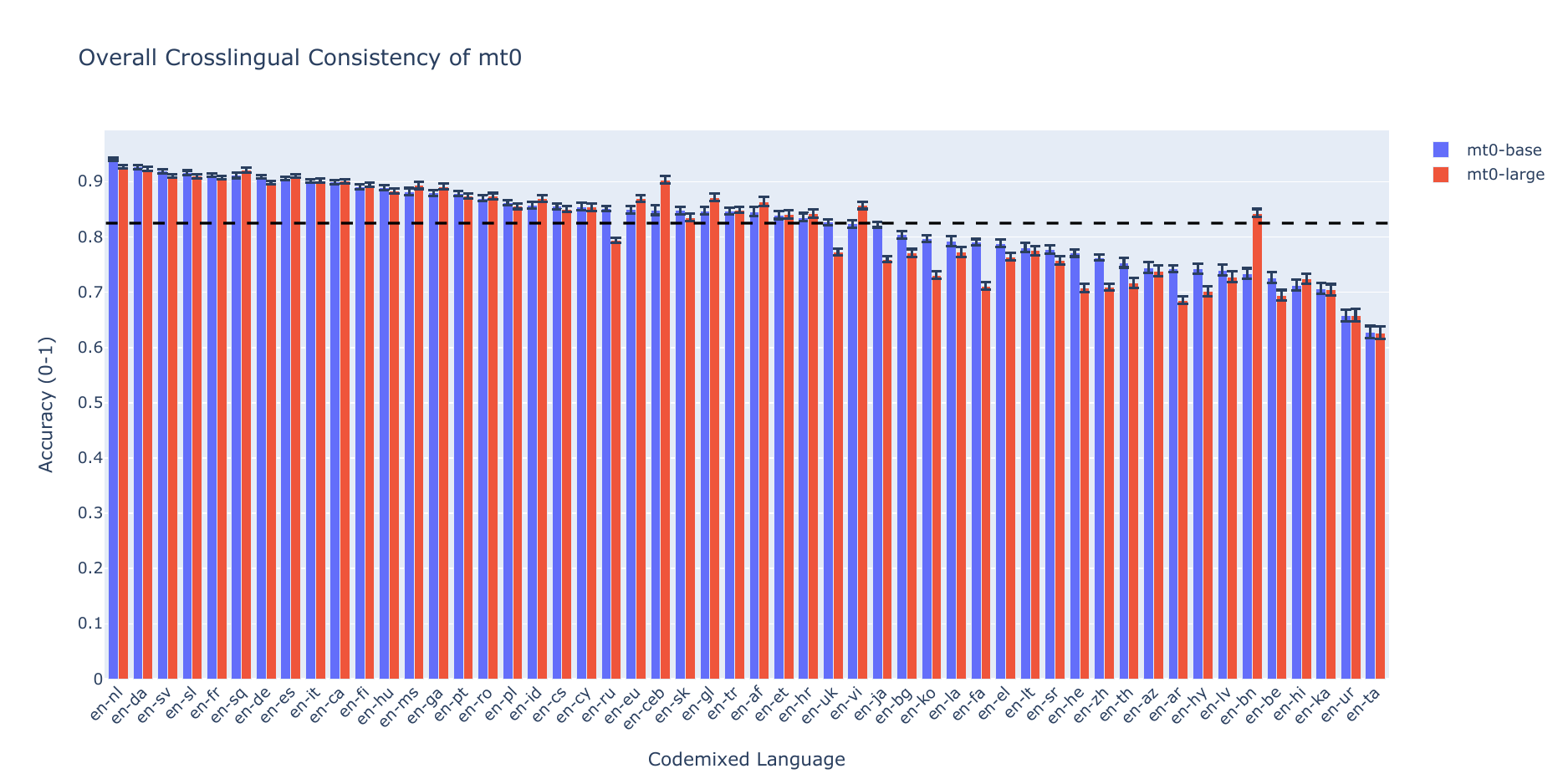}
        \label{fig:mt0-base-overall-crosslingual-consistency-acc}
    }\end{subfigure}
    \caption{Cross-lingual consistency scores across languages of mt0 (top: RankC, bottom: Top@1 Accuracy). Note: The dashed line here is the average corresponding consistency scores of mt0-base across languages}
    \label{fig:mt0-base-overall-crosslingual-consistency}
\end{figure*}

\subsubsection{Consistency of Non-English Matrix Languages}
\label{sess:matrix-non-english}
Refer to Figure \ref{app_fig:overall-encoder-encoder-decoder-crosslingual-factors}, \ref{fig:xlm-r-overall-crosslingual-consistency}, and \ref{fig:llama-overall-crosslingual-consistency}.
\begin{figure*}[ht!]
    \centering
    \begin{subfigure}{0.32\linewidth}{
        \centering
        \includegraphics[trim=0cm 0cm 0cm 2.5cm,clip=true,width=\linewidth]{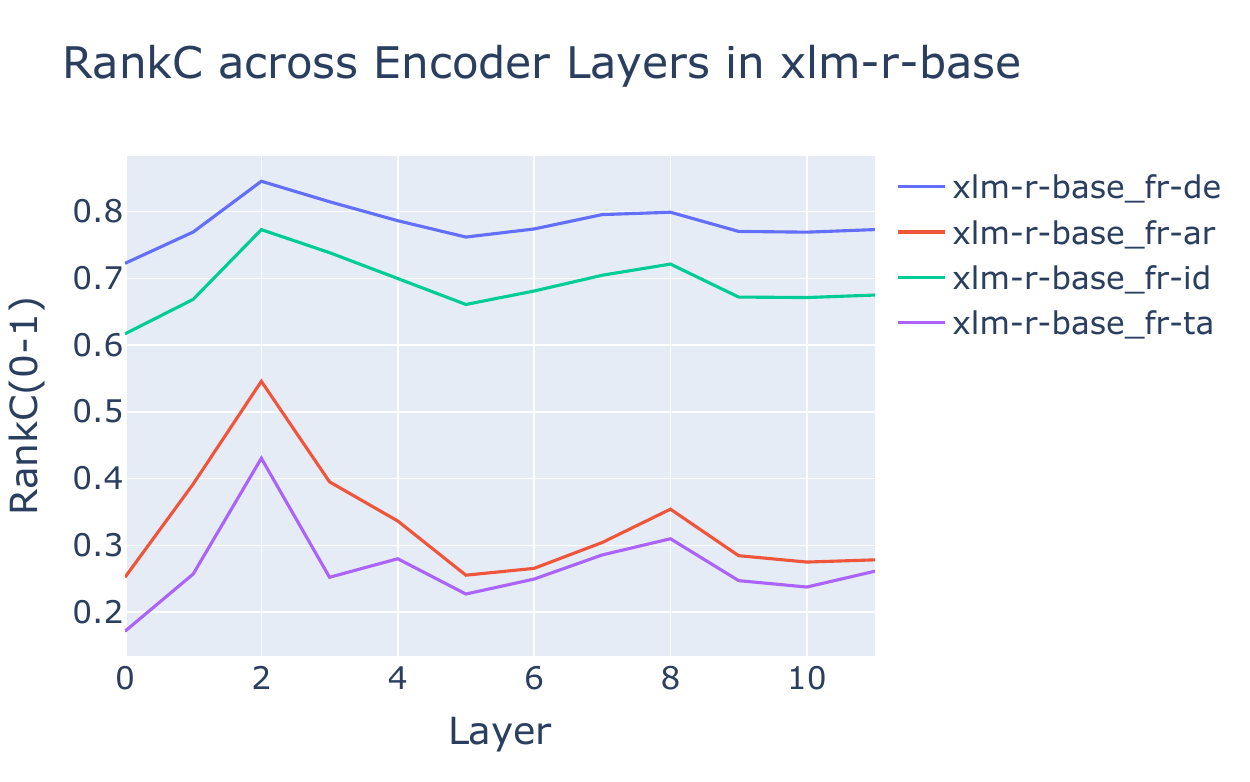}
        \label{app_fig:xlm-r-rankc-fr}
    }\end{subfigure}
    \begin{subfigure}{0.32\linewidth}{
        \centering
        \includegraphics[trim=0cm 0cm 0cm 2.5cm,clip=true,width=\linewidth]{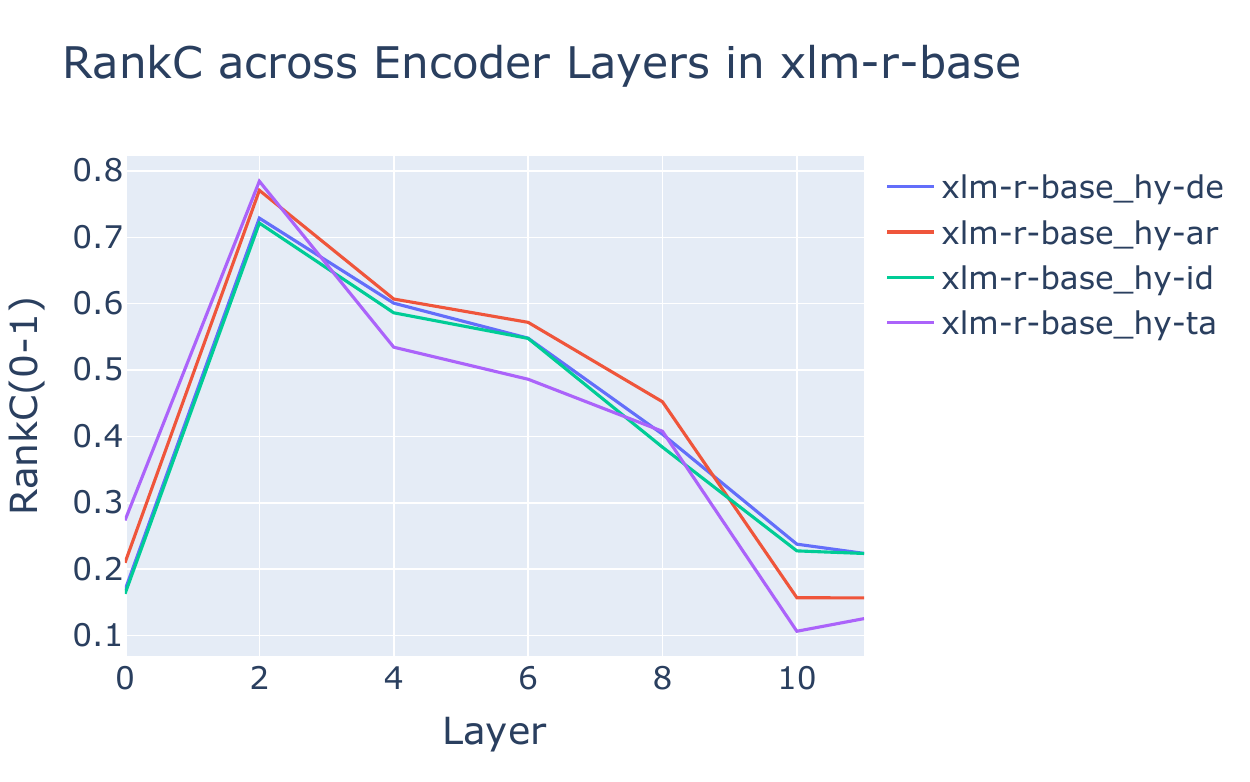}
        \label{app_fig:xlm-r-rankc-hy}
    }\end{subfigure}
            \centering
     \begin{subfigure}{0.32\linewidth}{
        \centering
        \includegraphics[trim=0cm 0cm 0cm 2.5cm,clip=true,width=\linewidth]{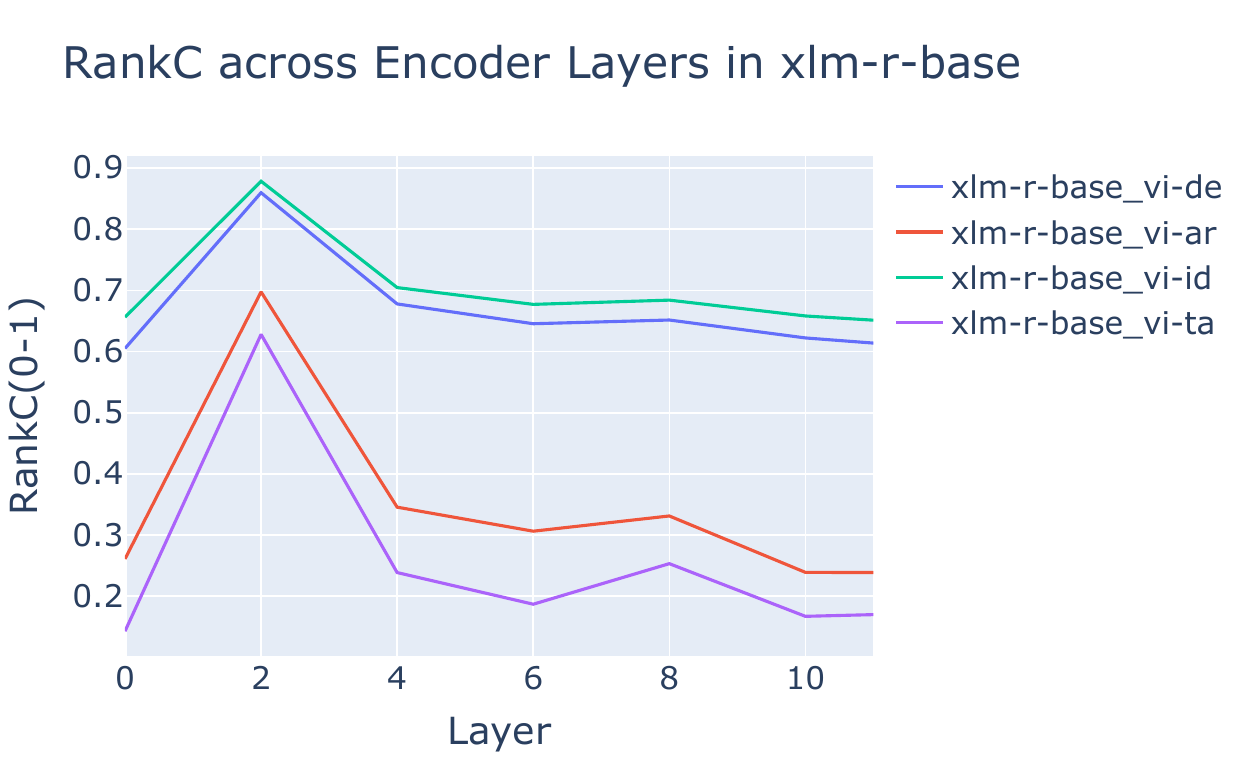}
        \label{app_fig:xlm-r-rankc-vi}
    }\end{subfigure}\\
    \begin{subfigure}{0.32\linewidth}{
        \centering
        \includegraphics[trim=0cm 0cm 0cm 2.5cm,clip=true,width=\linewidth]{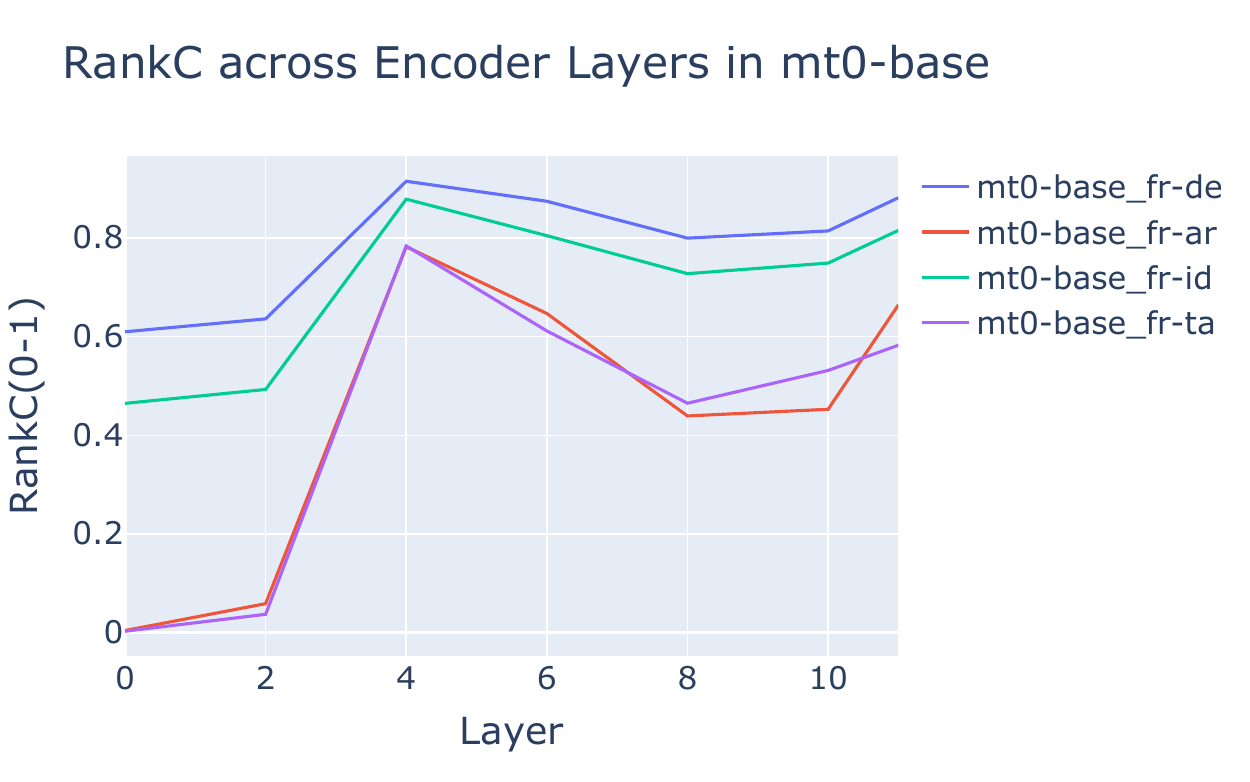}
        \label{app_fig:mt0-rankc-fr}
    }\end{subfigure}
    \begin{subfigure}{0.32\linewidth}{
        \centering
        \includegraphics[trim=0cm 0cm 0cm 2.5cm,clip=true,width=\linewidth]{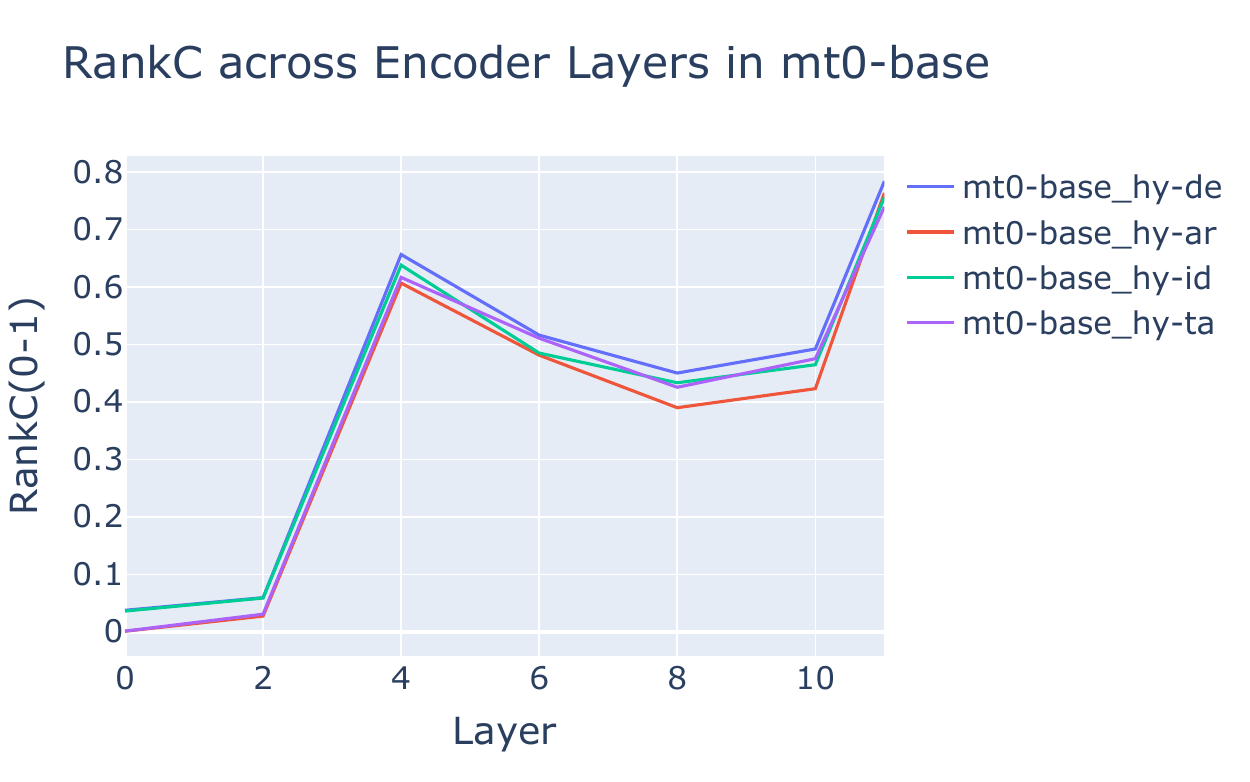}
        \label{app_fig:mt0-rankc-hy}
    }\end{subfigure}
    \begin{subfigure}{0.32\linewidth}{
        \centering
        \includegraphics[trim=0cm 0cm 0cm 2.5cm,clip=true,width=\linewidth]{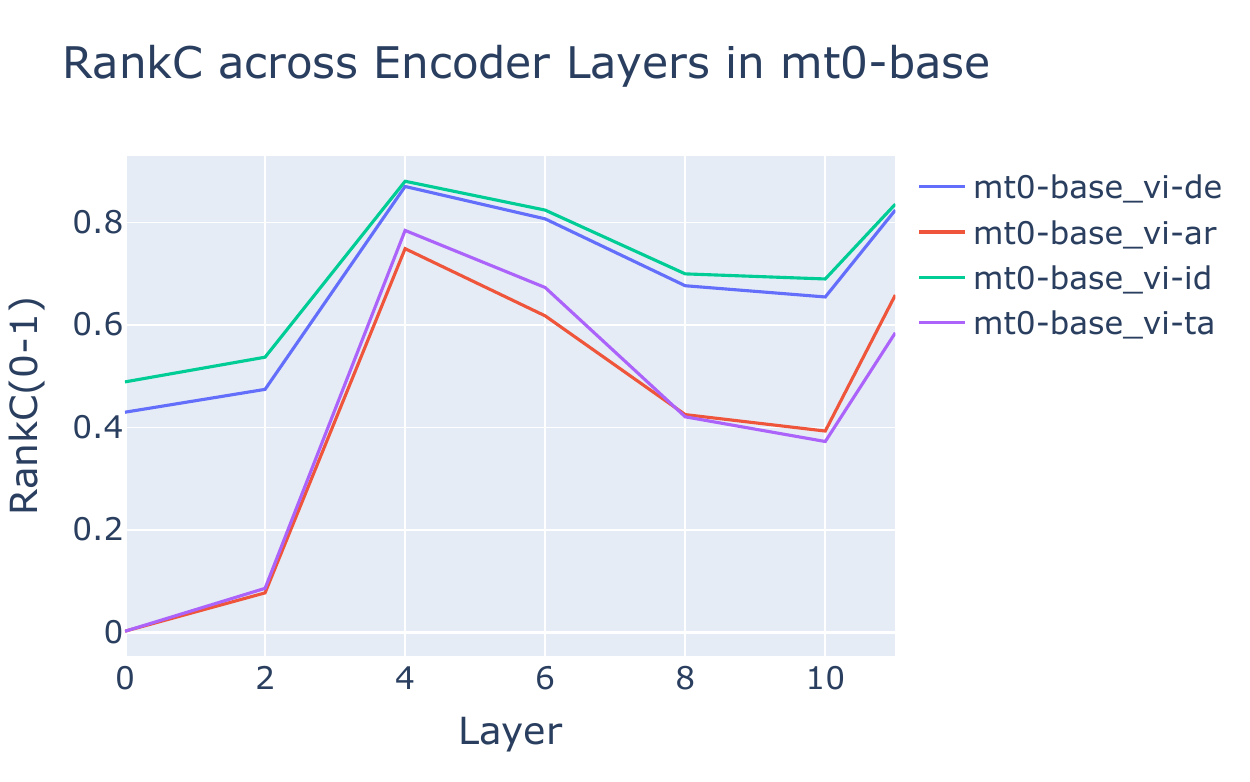}
        \label{app_fig:mt0-rankc-vi}
    }\end{subfigure}\\
        \begin{subfigure}{0.32\linewidth}{
        \centering
        \includegraphics[trim=0cm 0cm 0cm 2.5cm,clip=true,width=\linewidth]{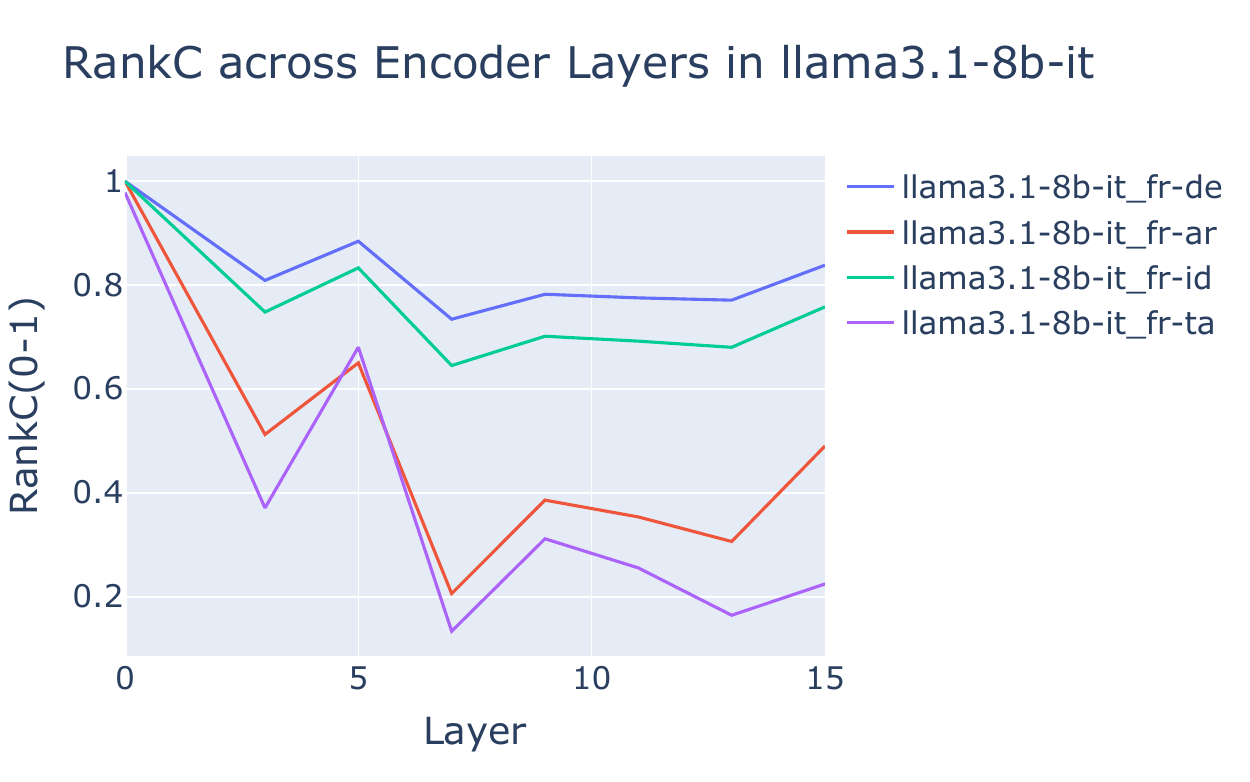}
        \label{app_fig:llama-rankc-fr}
    }\end{subfigure}
    \begin{subfigure}{0.32\linewidth}{
        \centering
        \includegraphics[trim=0cm 0cm 0cm 2.5cm,clip=true,width=\linewidth]{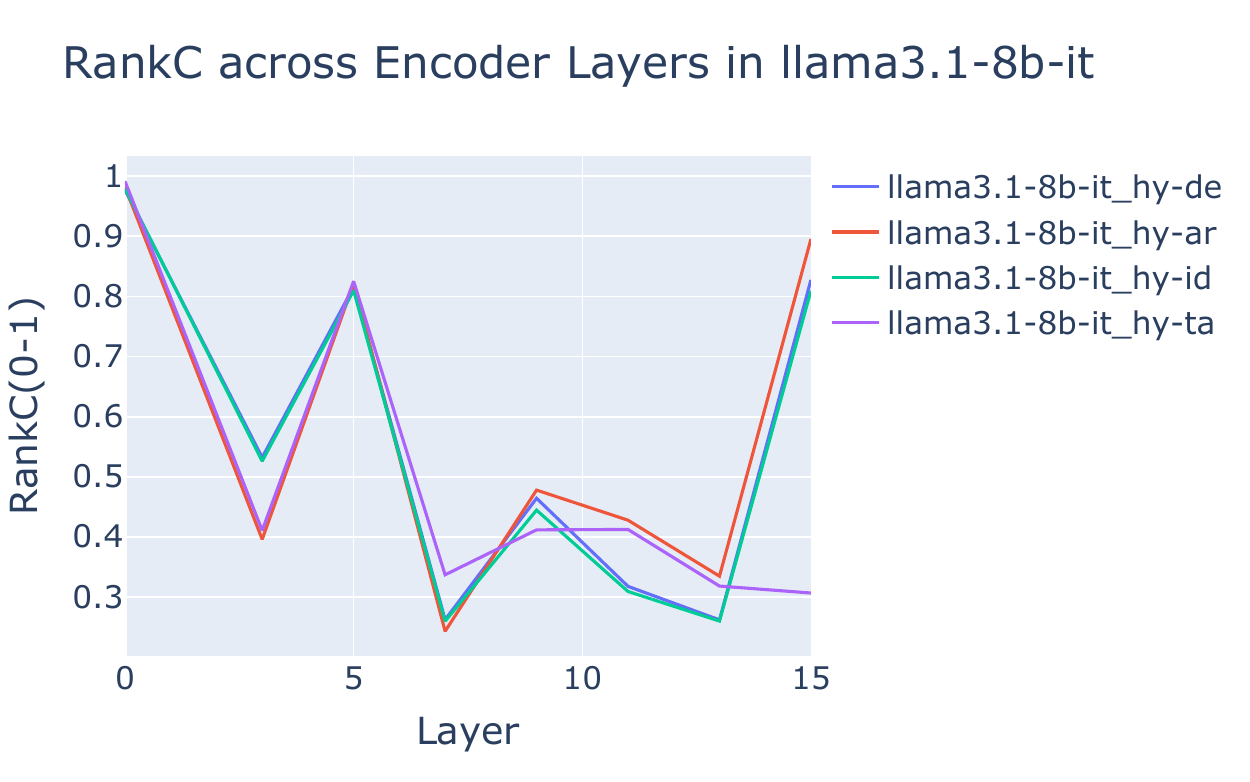}
        \label{app_fig:llama-rankc-hy}
    }\end{subfigure}
    \begin{subfigure}{0.32\linewidth}{
        \centering
        \includegraphics[trim=0cm 0cm 0cm 2.5cm,clip=true,width=\linewidth]{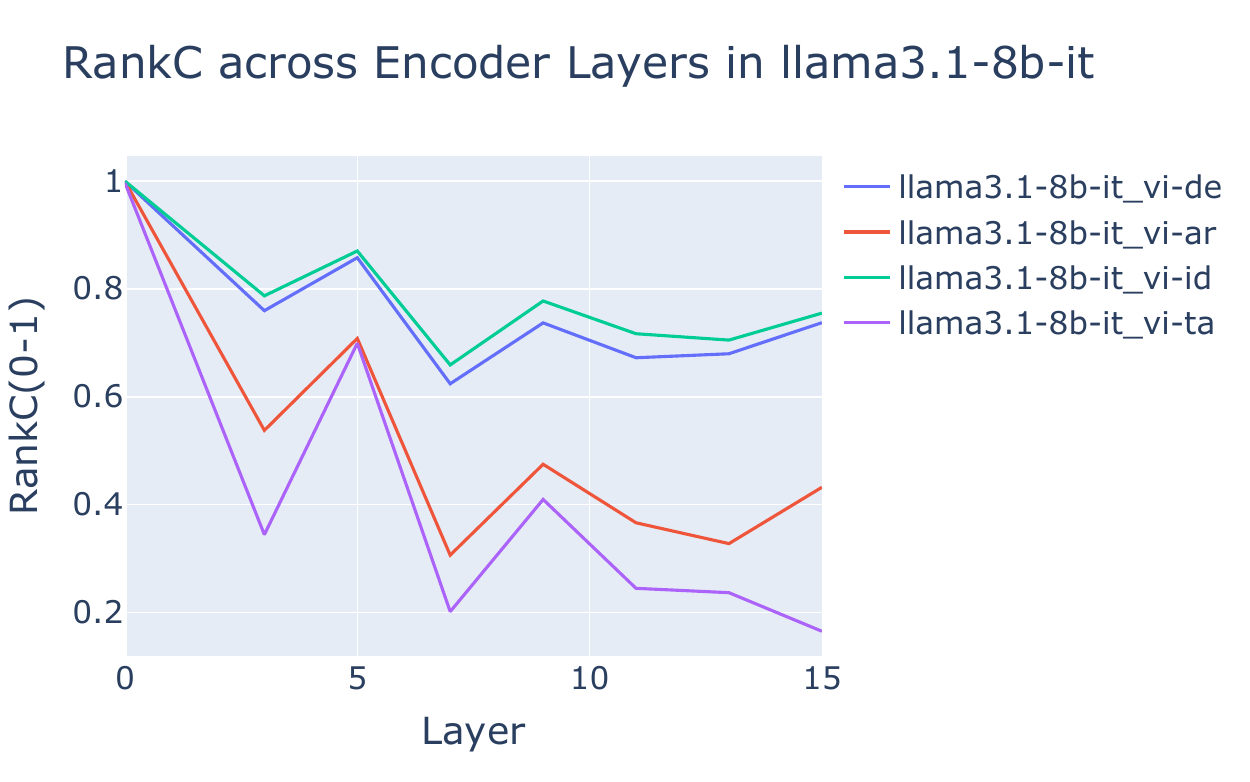}
        \label{app_fig:llama-rankc-vi}
    }\end{subfigure}\\
    \caption{Layerwise cross-lingual consistency (rankC) across different transformer types (top: encoder, middle: encoder-decoder, bottom: decoder)  on different non-english matrix languages (left: french, center: armenian, right: vietnamese).}
    \label{app_fig:layerwise-non-english-rankc}
\end{figure*}
\begin{figure*}[ht!]
    \centering
    \begin{subfigure}{0.32\linewidth}{
        \centering
        \includegraphics[trim=0cm 0cm 0cm 2.5cm,clip=true,width=\linewidth]{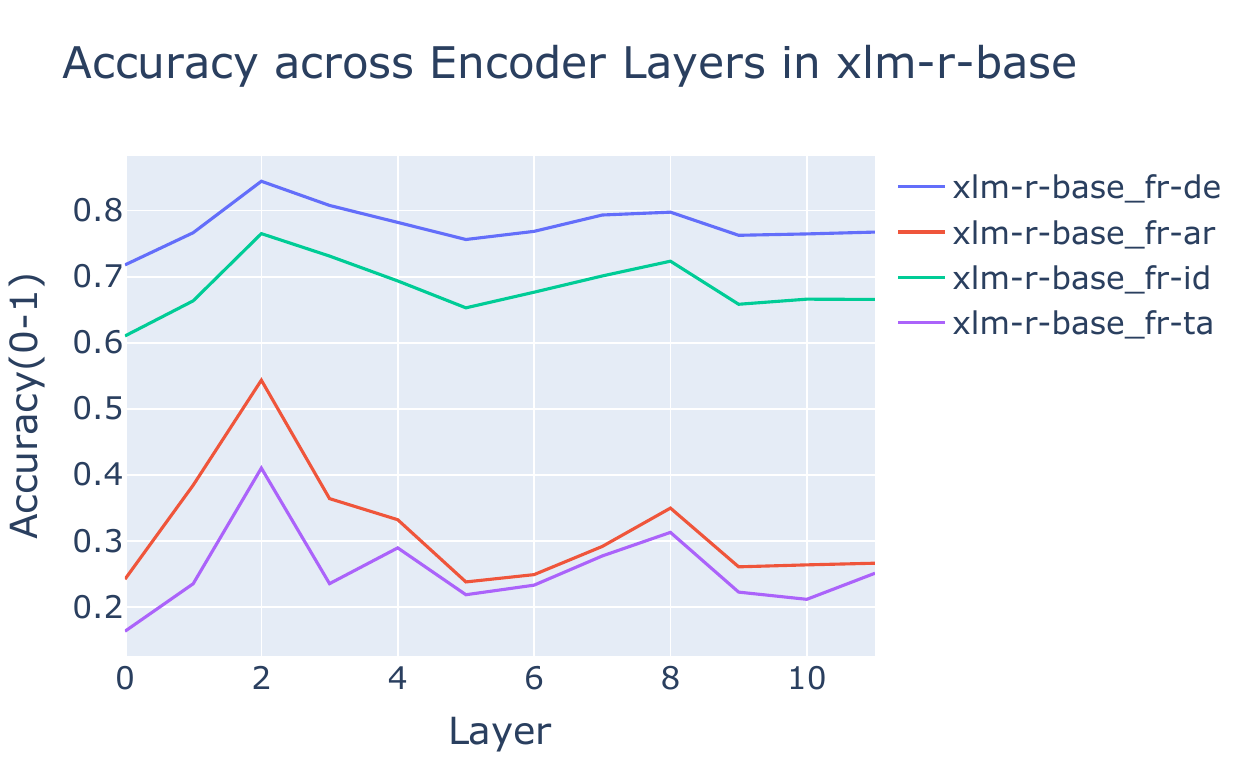}
        \label{app_fig:xlm-r-acc-fr}
    }\end{subfigure}
    \begin{subfigure}{0.32\linewidth}{
        \centering
        \includegraphics[trim=0cm 0cm 0cm 2.5cm,clip=true,width=\linewidth]{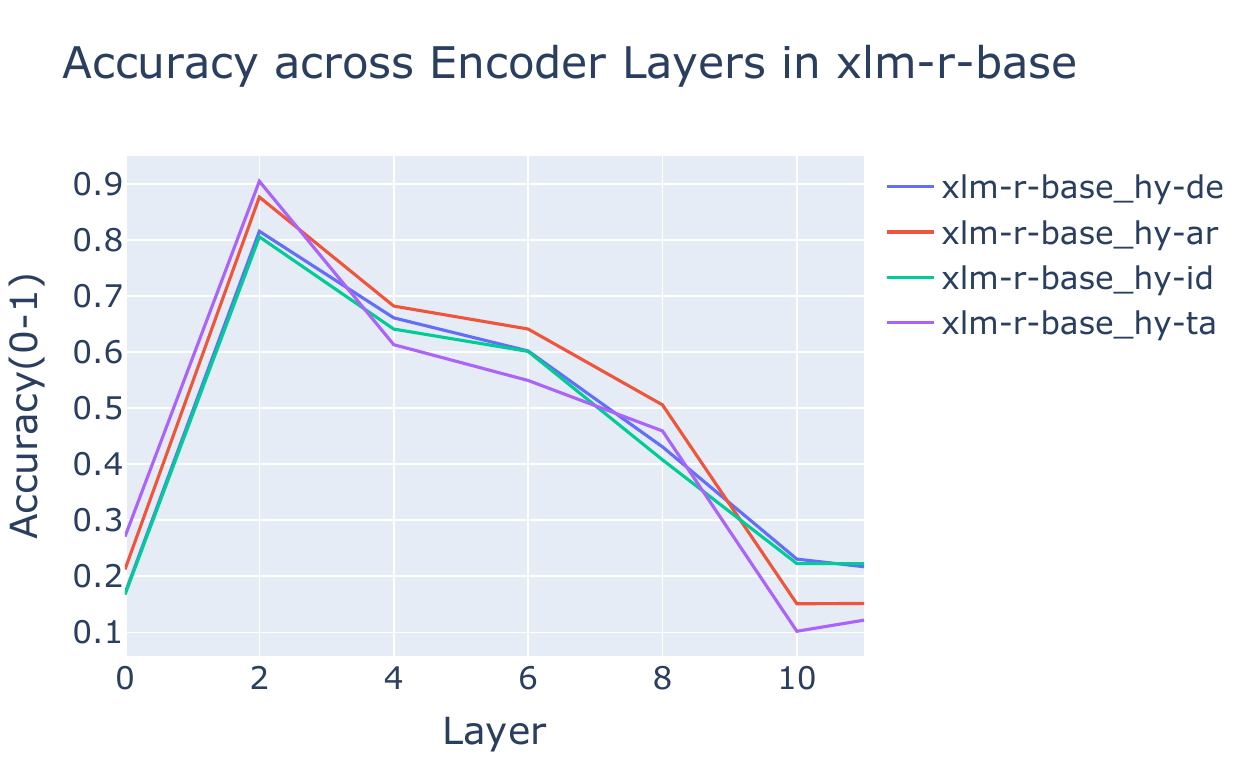}
        \label{app_fig:xlm-r-acc-hy}
    }\end{subfigure}
            \centering
     \begin{subfigure}{0.32\linewidth}{
        \centering
        \includegraphics[trim=0cm 0cm 0cm 2.5cm,clip=true,width=\linewidth]{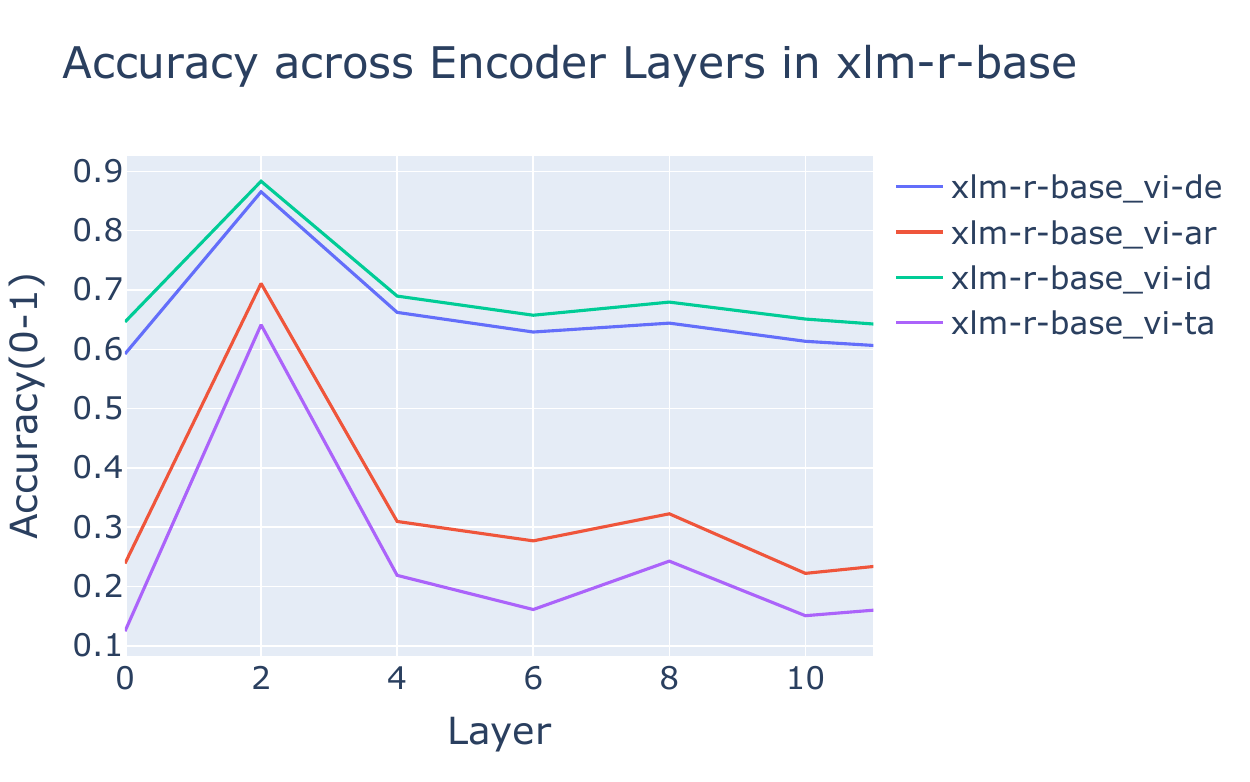}
        \label{app_fig:xlm-r-acc-vi}
    }\end{subfigure}\\
    \begin{subfigure}{0.32\linewidth}{
        \centering
        \includegraphics[trim=0cm 0cm 0cm 2.5cm,clip=true,width=\linewidth]{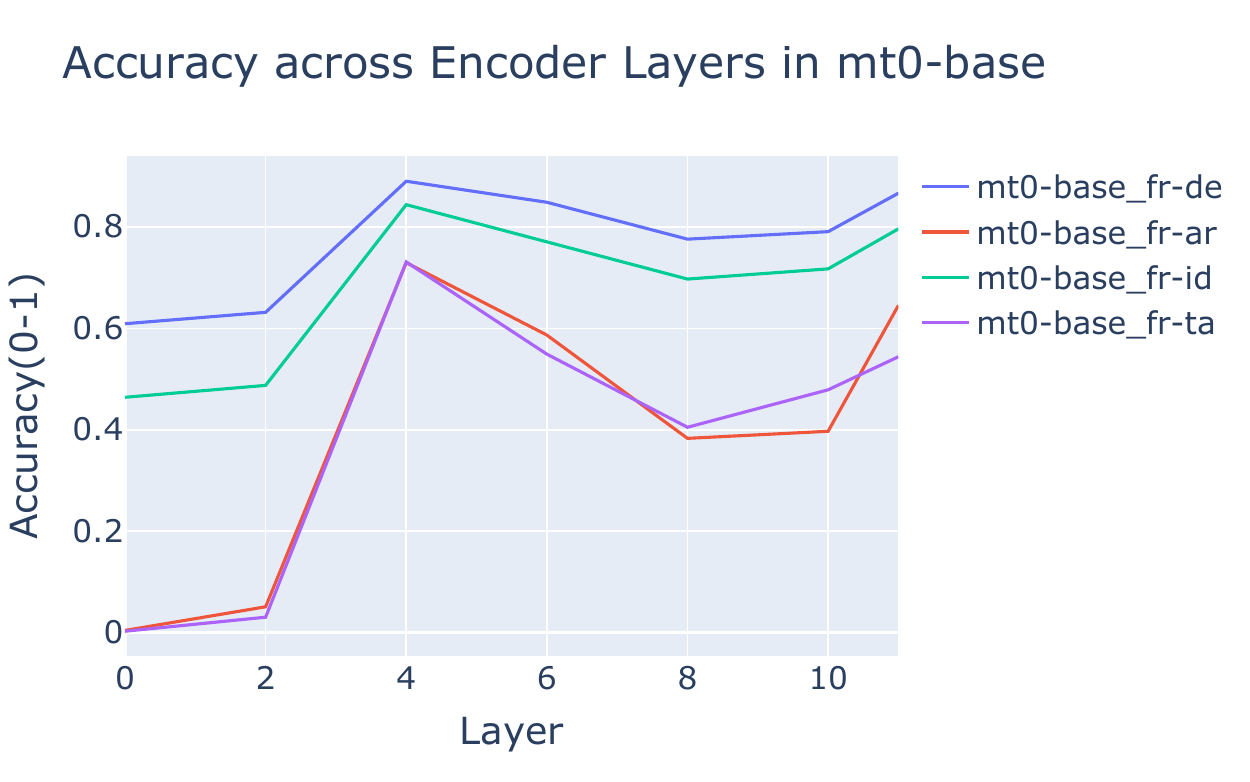}
        \label{app_fig:mt0-acc-fr}
    }\end{subfigure}
    \begin{subfigure}{0.32\linewidth}{
        \centering
        \includegraphics[trim=0cm 0cm 0cm 2.5cm,clip=true,width=\linewidth]{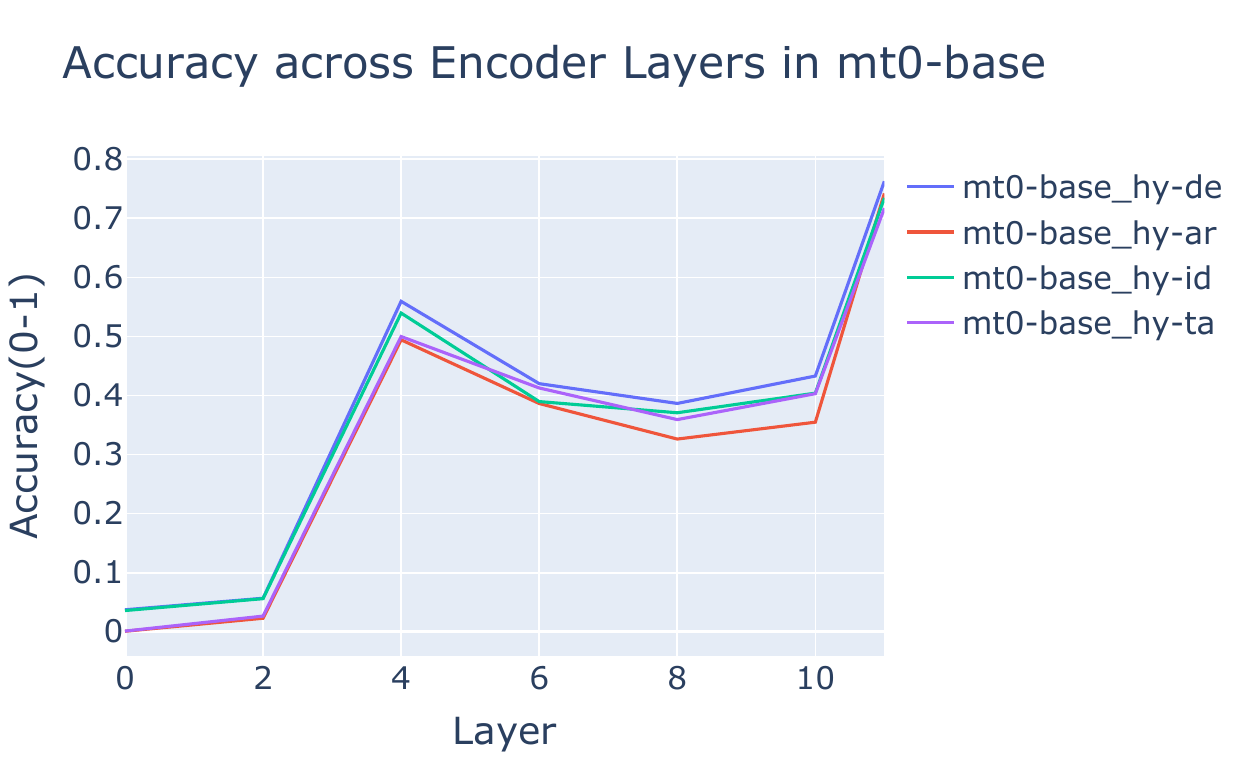}
        \label{app_fig:mt0-acc-hy}
    }\end{subfigure}
    \begin{subfigure}{0.32\linewidth}{
        \centering
        \includegraphics[trim=0cm 0cm 0cm 2.5cm,clip=true,width=\linewidth]{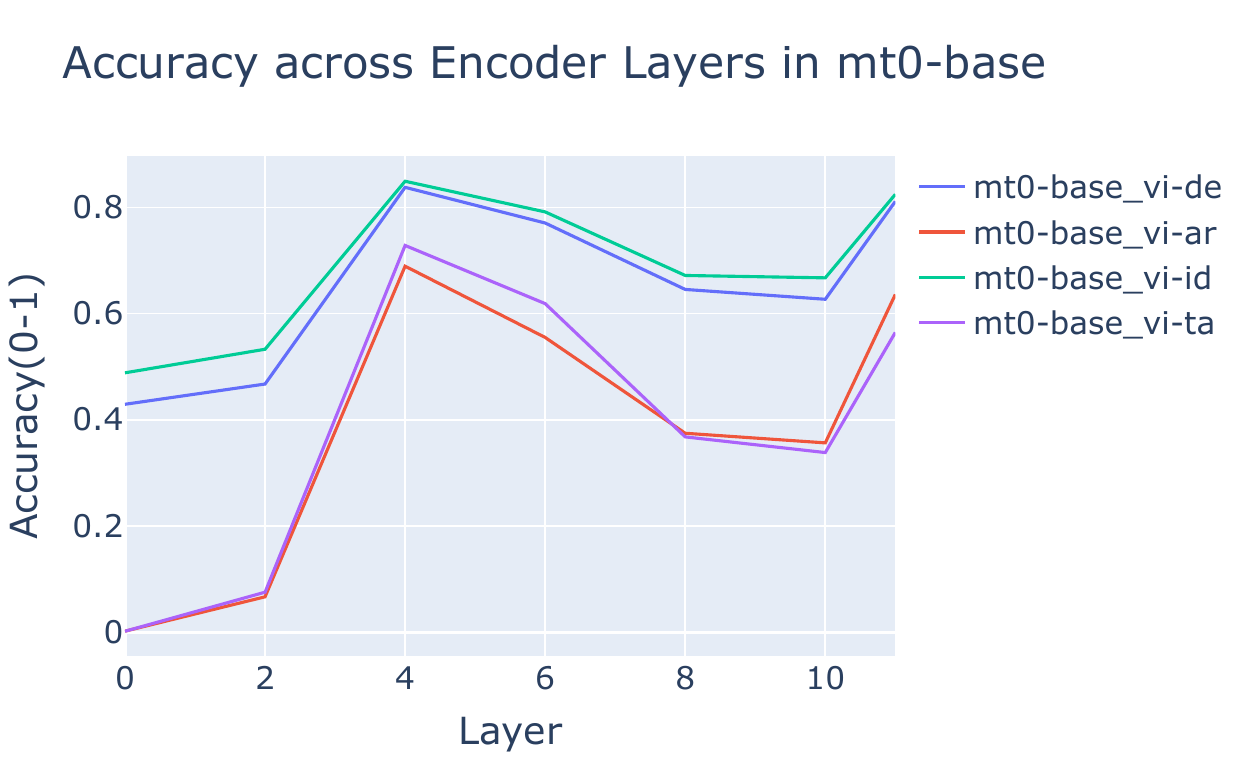}
        \label{app_fig:mt0-acc-vi}
    }\end{subfigure}\\
        \begin{subfigure}{0.32\linewidth}{
        \centering
        \includegraphics[trim=0cm 0cm 0cm 2.5cm,clip=true,width=\linewidth]{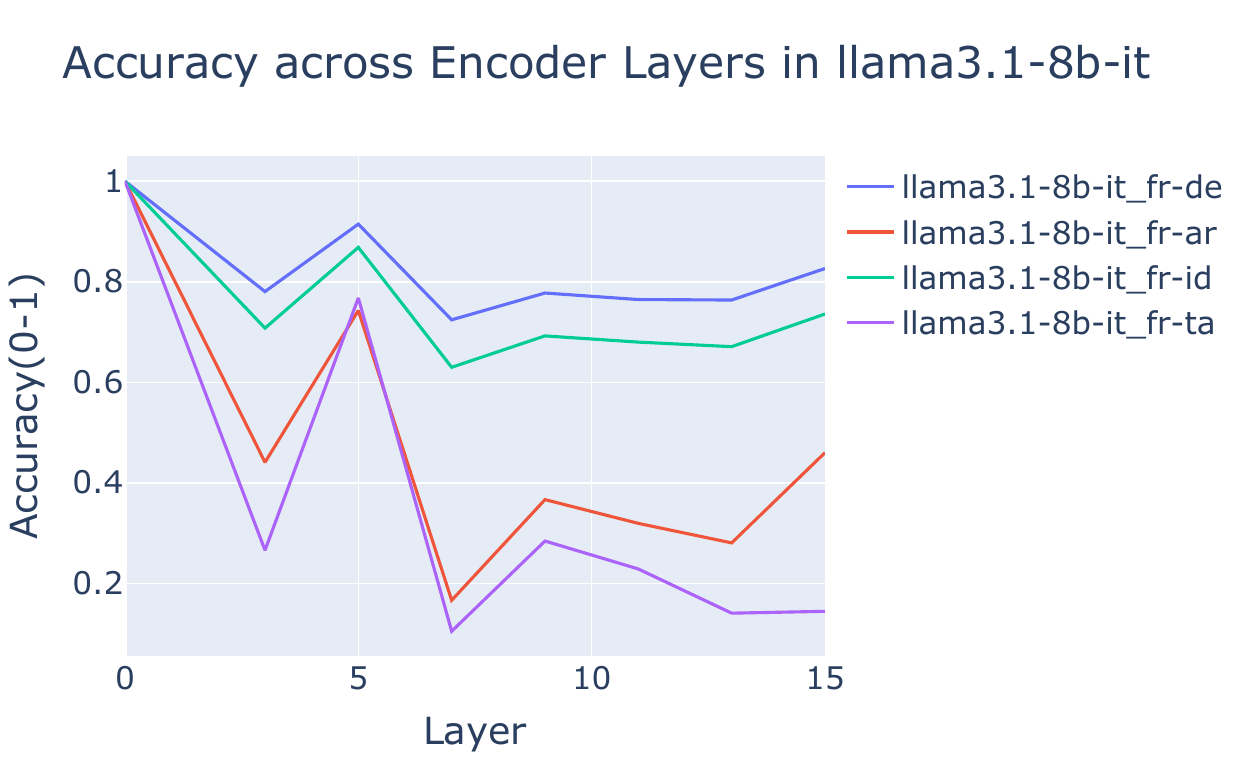}
        \label{app_fig:llama-acc-fr}
    }\end{subfigure}
    \begin{subfigure}{0.32\linewidth}{
        \centering
        \includegraphics[trim=0cm 0cm 0cm 2.5cm,clip=true,width=\linewidth]{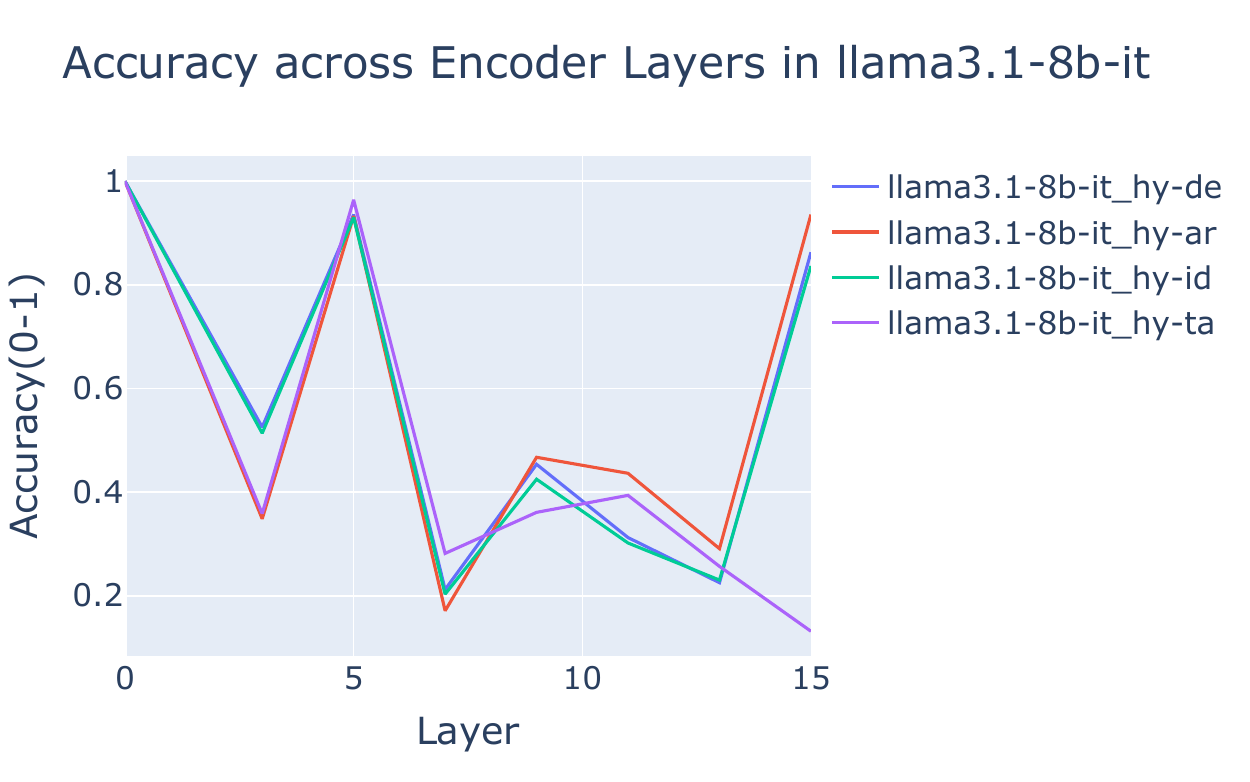}
        \label{app_fig:llama-acc-hy}
    }\end{subfigure}
    \begin{subfigure}{0.32\linewidth}{
        \centering
        \includegraphics[trim=0cm 0cm 0cm 2.5cm,clip=true,width=\linewidth]{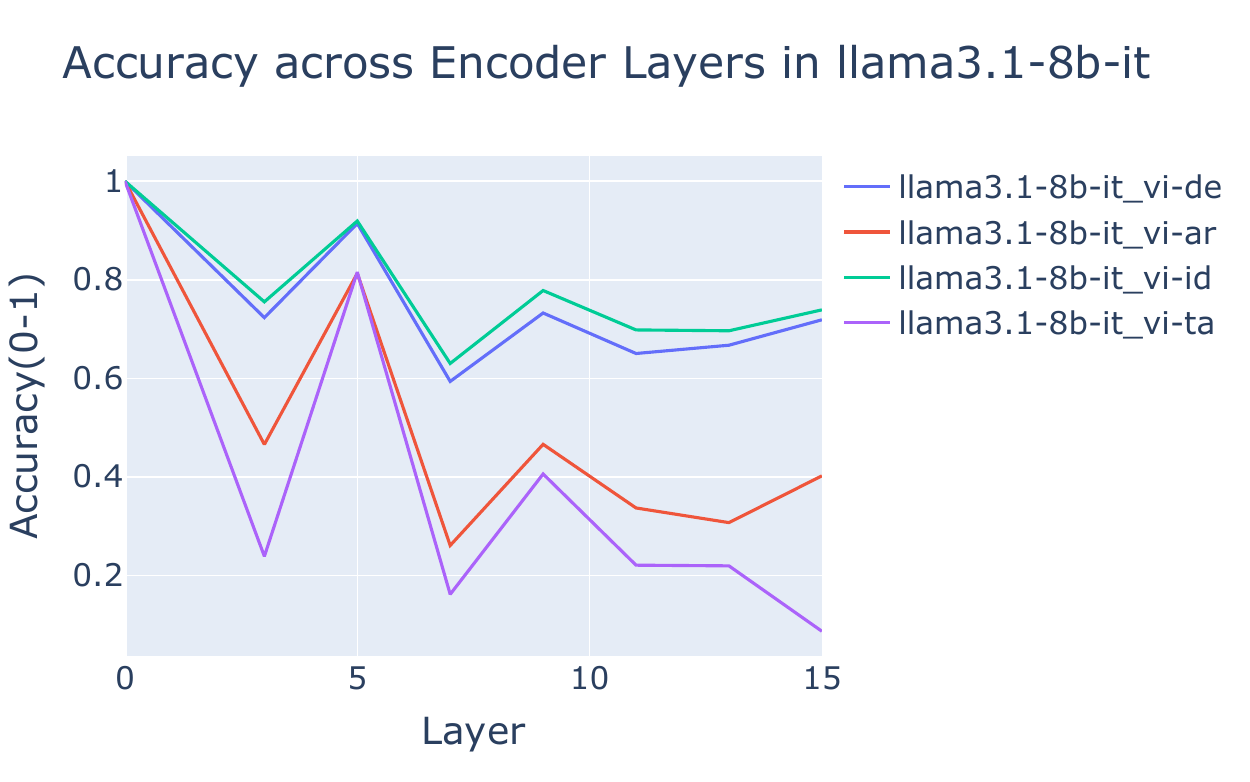}
        \label{app_fig:llama-acc-vi}
    }\end{subfigure}\\
    \caption{Layerwise cross-lingual consistency (rankC) across different transformer types (top: encoder, middle: encoder-decoder, bottom: decoder)  on different non-english matrix languages (left: french, center: armenian, right: vietnamese).}
    \label{app_fig:layerwise-non-english-acc}
\end{figure*}

\subsubsection{Feed-Forward Neurons' Gradients Sum}
\label{sess:findings_in_details_Feed-Forward Neurons}
\begin{figure*}[ht]
    \centering
    \begin{subfigure}{0.45\linewidth}{
        \centering
        \includegraphics[trim=0cm 0cm 0cm 2cm,clip=true,width=\linewidth]{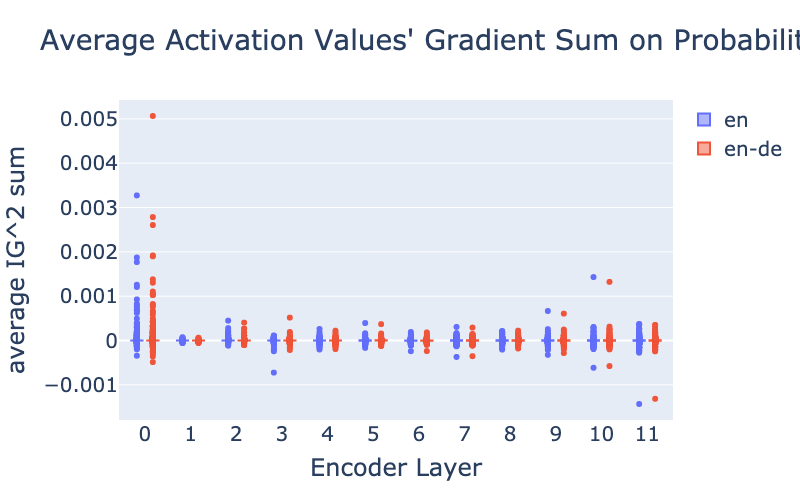}
        \label{fig:mt0-de-en-ffn-bias}
    }\end{subfigure}
    \begin{subfigure}{0.45\linewidth}{
        \centering
        \includegraphics[trim=0cm 0cm 0cm 2cm,clip=true,width=\linewidth]{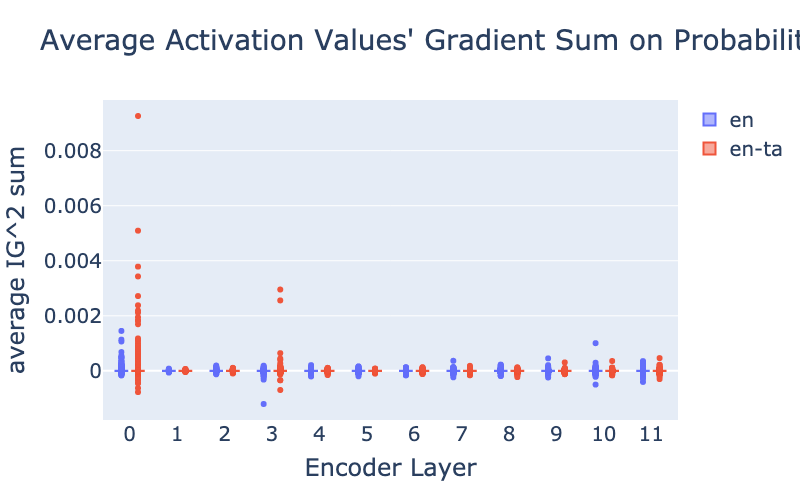}
        \label{fig:mt0-ta-en-ffn-bias}
    }\end{subfigure} \\
    \begin{subfigure}{0.45\linewidth}{
        \centering
        \includegraphics[trim=0cm 0cm 0cm 2cm,clip=true,width=\linewidth]{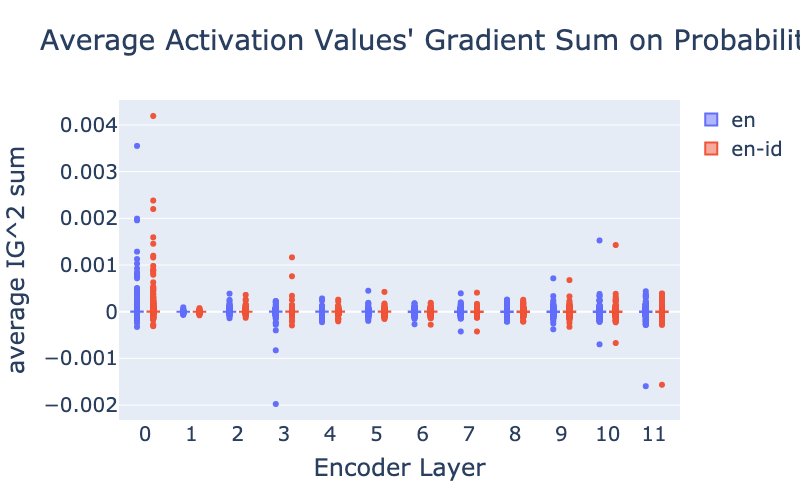}
        \label{fig:mt0-base-id-en-ffn-bias}
    }\end{subfigure}
    \begin{subfigure}{0.45\linewidth}{
        \centering
        \includegraphics[trim=0cm 0cm 0cm 2cm,clip=true,width=\linewidth]{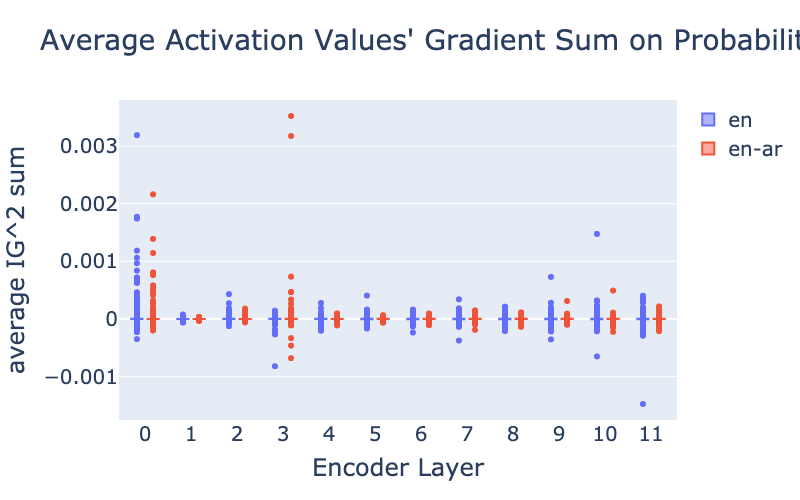}
        \label{fig:mt0-base-ar-en-ffn-bias}
    }\end{subfigure}\\
        \begin{subfigure}{0.45\linewidth}{
        \centering
        \includegraphics[trim=0cm 0cm 0cm 2cm,clip=true,width=\linewidth]{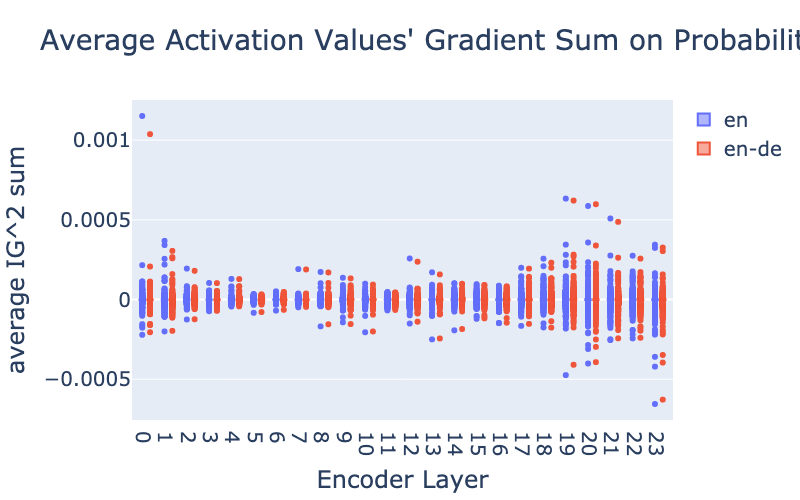}
        \label{fig:mt0-large-de-en-ffn-bias}
    }\end{subfigure}
    \begin{subfigure}{0.45\linewidth}{
        \centering
        \includegraphics[trim=0cm 0cm 0cm 2cm,clip=true,width=\linewidth]{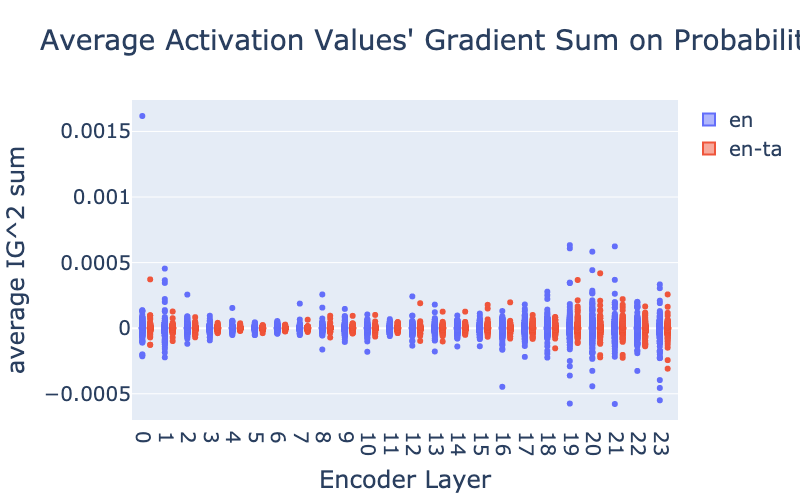}
        \label{fig:mt0-large-ta-en-ffn-bias}
    }\end{subfigure} \\
    \begin{subfigure}{0.45\linewidth}{
        \centering
        \includegraphics[trim=0cm 0cm 0cm 2cm,clip=true,width=\linewidth]{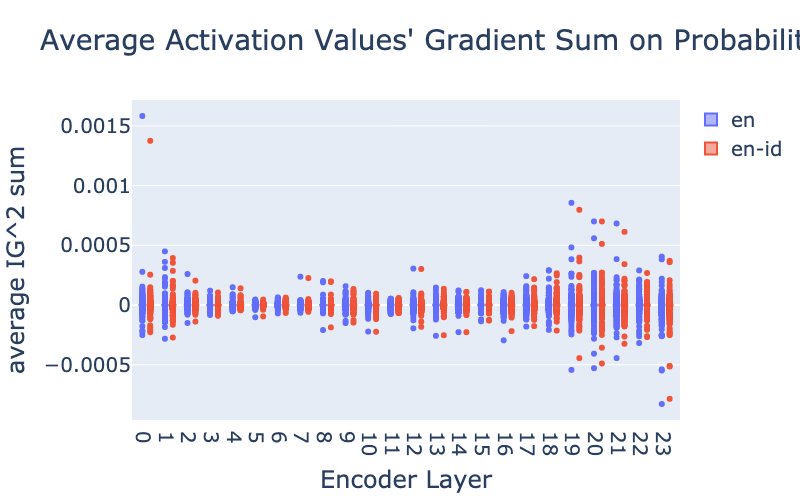}
        \label{fig:mt0-large-id-en-ffn-bias}
    }\end{subfigure}
    \begin{subfigure}{0.45\linewidth}{
        \centering
        \includegraphics[trim=0cm 0cm 0cm 2cm,clip=true,width=\linewidth]{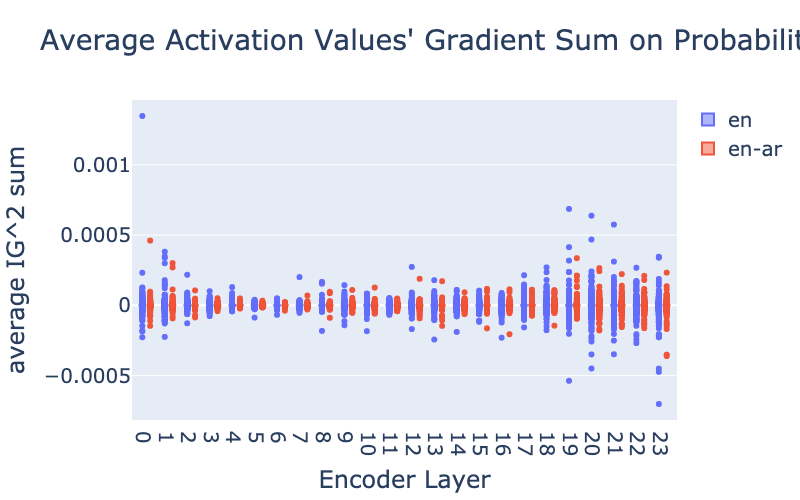}
        \label{fig:mt0-large-ar-en-ffn-bias}
    }\end{subfigure}
    \caption{ $IG^2$ scores in mt0 for en--de, en--ta, en--id, and en--ar. Models legend: upper two rows: mt0-base, lower two rows: mt0-large. }
    \label{fig:mt0-ffn-bias-de-id-ta-ar-en}
\end{figure*}

\begin{figure*}[ht]
    \centering
    \begin{subfigure}{0.45\linewidth}{
        \centering
        \includegraphics[trim=0cm 0cm 0cm 2cm,clip=true,width=\linewidth]{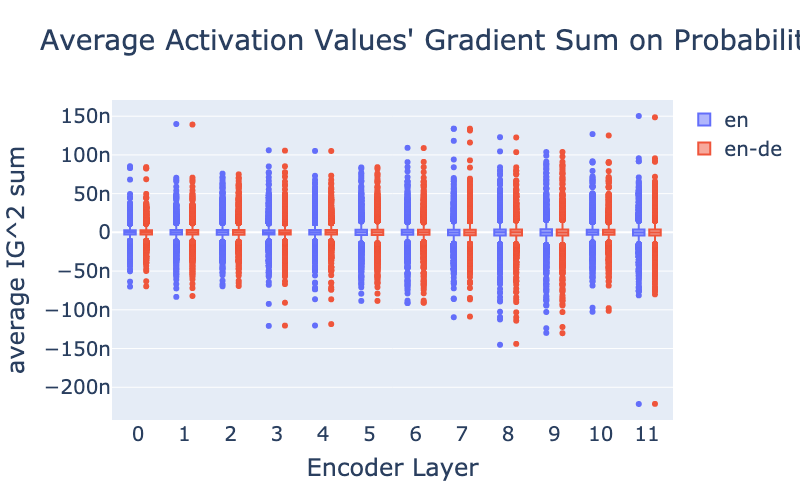}
        \label{fig:xlmr-de-en-ffn-bias}
    }\end{subfigure}
    \begin{subfigure}{0.45\linewidth}{
        \centering
        \includegraphics[trim=0cm 0cm 0cm 2cm,clip=true,width=\linewidth]{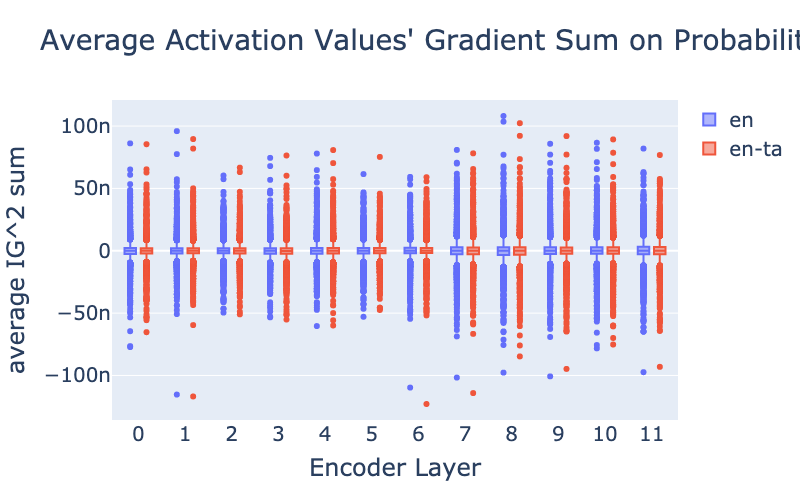}
        \label{fig:xlmr-ta-en-ffn-bias}
    }\end{subfigure}\\
    \begin{subfigure}{0.45\linewidth}{
        \centering
        \includegraphics[trim=0cm 0cm 0cm 2cm,clip=true,width=\linewidth]{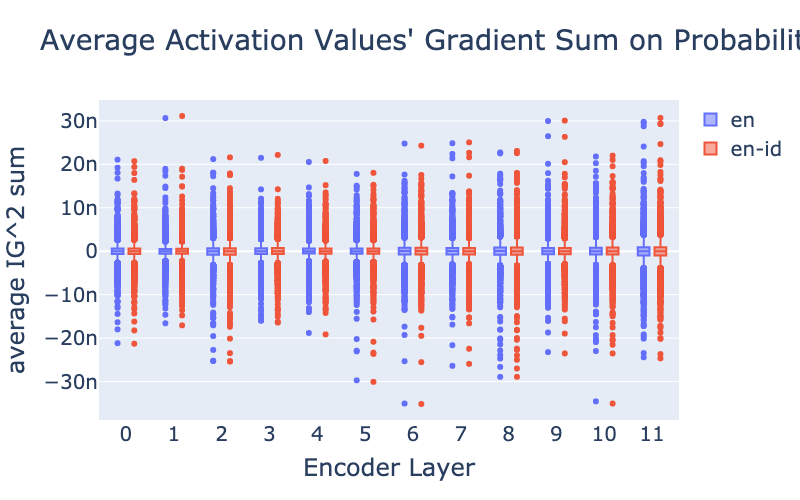}
        \label{fig:xlmr-id-en-ffn-bias}
    }\end{subfigure}
    \begin{subfigure}{0.45\linewidth}{
        \centering
        \includegraphics[trim=0cm 0cm 0cm 2cm,clip=true,width=\linewidth]{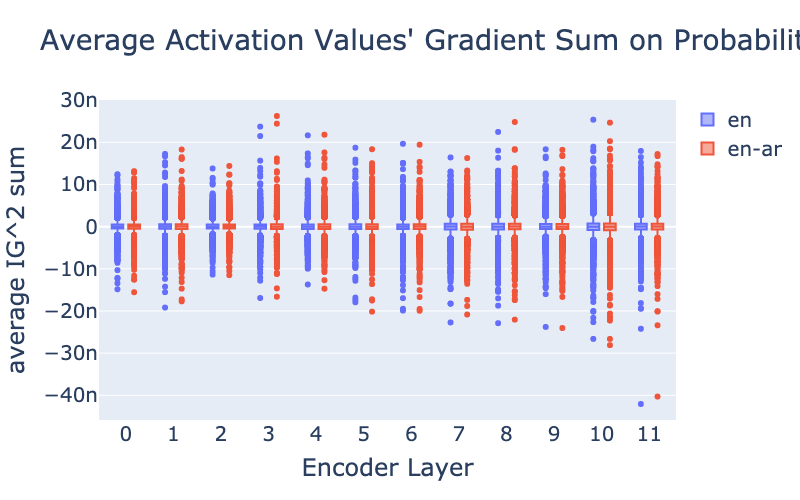}
        \label{fig:xlmr-ar-en-ffn-bias}
    }\end{subfigure}\\
    \begin{subfigure}{0.45\linewidth}{
        \centering
        \includegraphics[trim=0cm 0cm 0cm 2cm,clip=true,width=\linewidth]{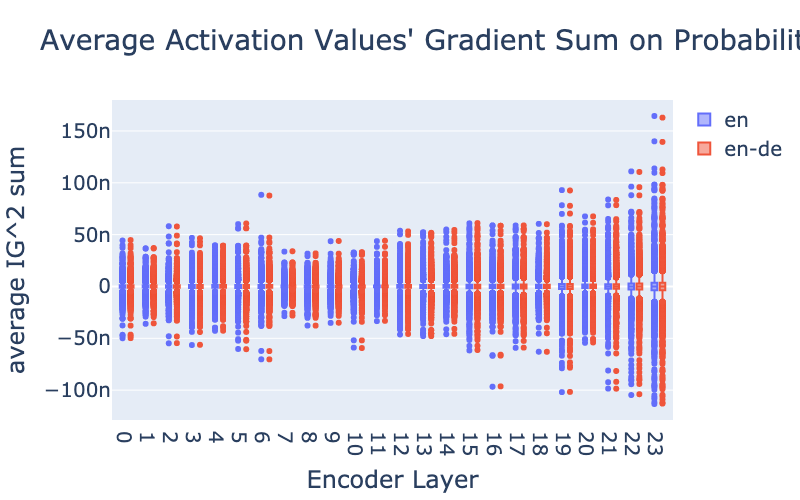}
        \label{fig:xlmr-large-de-en-ffn-bias}
    }\end{subfigure}
    \begin{subfigure}{0.45\linewidth}{
        \centering
        \includegraphics[trim=0cm 0cm 0cm 2cm,clip=true,width=\linewidth]{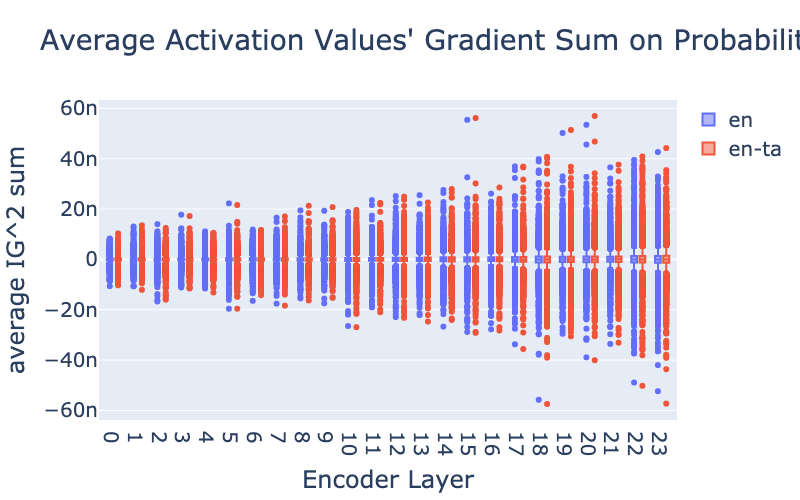}
        \label{fig:xlmr-large-ta-en-ffn-bias}
    }\end{subfigure}\\
    \begin{subfigure}{0.45\linewidth}{
        \centering
        \includegraphics[trim=0cm 0cm 0cm 2cm,clip=true,width=\linewidth]{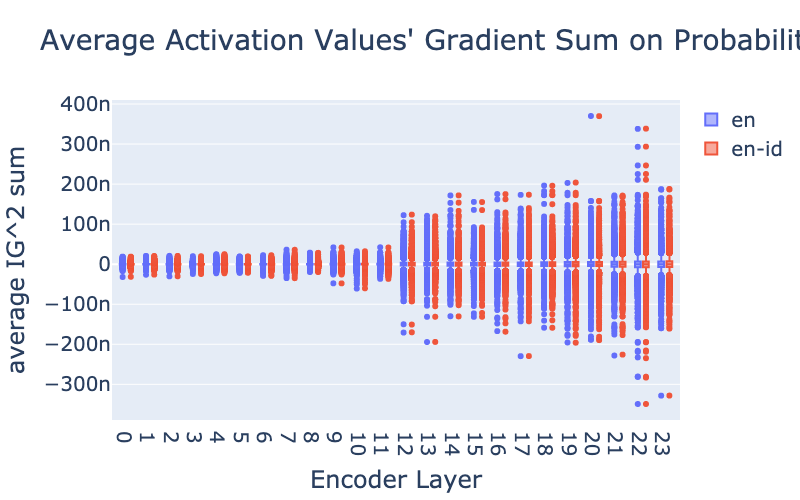}
        \label{fig:xlmr-large-id-en-ffn-bias}
    }\end{subfigure}
    \begin{subfigure}{0.45\linewidth}{
        \centering
        \includegraphics[trim=0cm 0cm 0cm 2cm,clip=true,width=\linewidth]{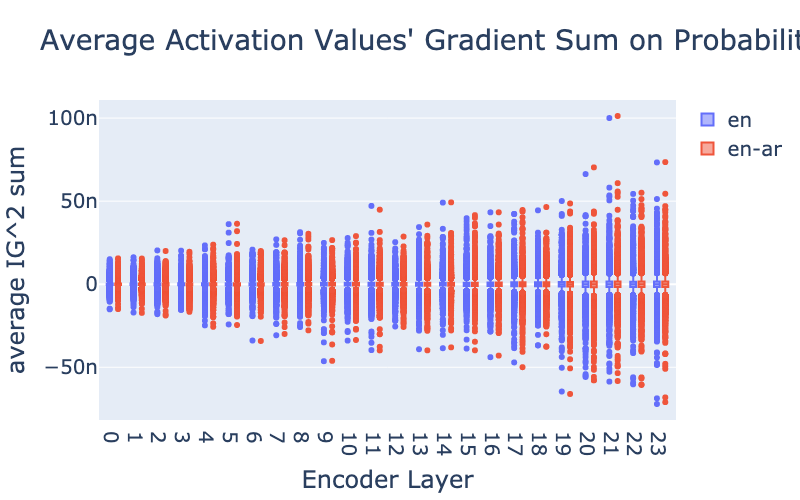}
        \label{fig:xlmr-large-ar-en-ffn-bias}
    }\end{subfigure}
    \caption{$IG^2$ scores in xlm-r for en--de, en--ta, en--id, and en--ar. Models legend: upper two rows: xlm-r-base, lower two rows: xlm-r-large}
    \label{fig:xlmr-ffn-bias-de-id-ta-ar-en}
\end{figure*}

Refer to Figure \ref{fig:mt0-ffn-bias-de-id-ta-ar-en}, \ref{fig:xlmr-ffn-bias-de-id-ta-ar-en}, and \ref{fig:xlmr-ffn-bias-de-id-ta-ar-en}
\begin{figure*}[ht]
    \centering
    \begin{subfigure}{0.45\linewidth}{
        \centering
        \includegraphics[trim=0cm 0cm 0cm 2cm,clip=true,width=\linewidth]{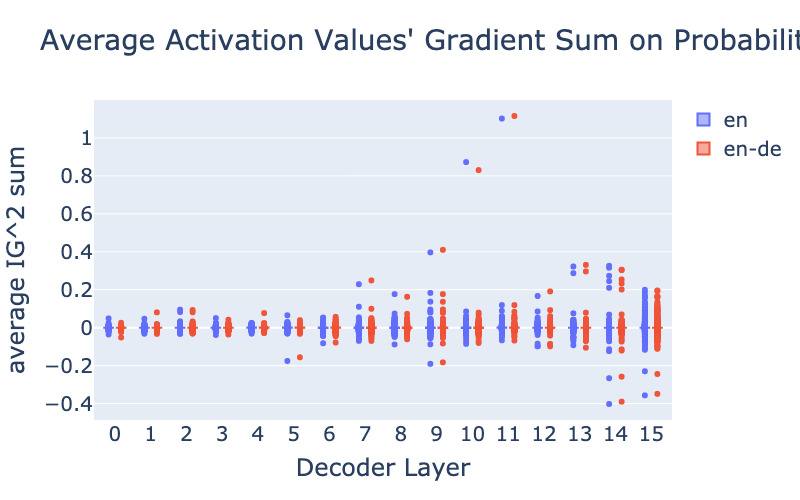}
        \label{fig:llama3.2-1b-de-en-ffn-bias}
    }\end{subfigure}
    \begin{subfigure}{0.45\linewidth}{
        \centering
        \includegraphics[trim=0cm 0cm 0cm 2cm,clip=true,width=\linewidth]{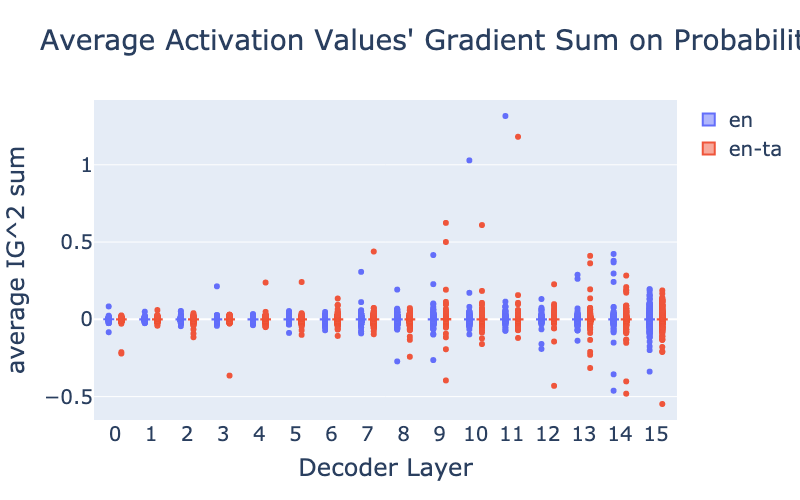}
        \label{fig:llama3.2-1b-ta-en-ffn-bias}
    }\end{subfigure}\\
    \begin{subfigure}{0.45\linewidth}{
        \centering
        \includegraphics[trim=0cm 0cm 0cm 2cm,clip=true,width=\linewidth]{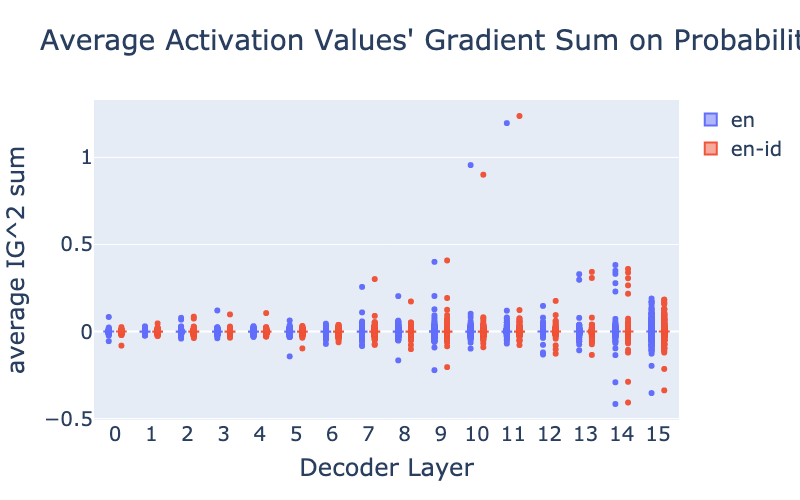}
        \label{fig:llama3.2-1b-id-en-ffn-bias}
    }\end{subfigure}
    \begin{subfigure}{0.45\linewidth}{
        \centering
        \includegraphics[trim=0cm 0cm 0cm 2cm,clip=true,width=\linewidth]{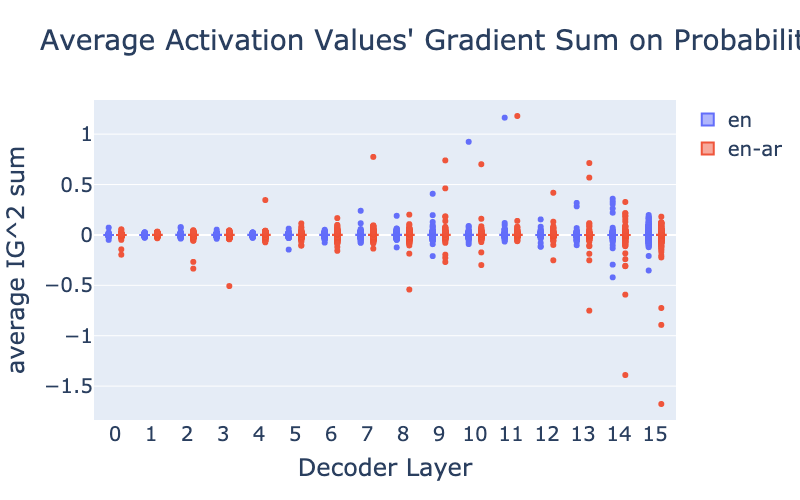}
        \label{fig:llama3.2-1b-ar-en-ffn-bias}
    }\end{subfigure}
    \caption{$IG^2$ scores in llama3.2-1b for en--de, en--ta, en--id, and en--ar.}
    \label{fig:llama3.2-1b-ffn-bias-de-id-ta-ar-en}
\end{figure*}

\subsection{Improving Consistency}
\subsubsection{Adding Monolingual Bias}
\label{sess:findings_in_details_Adding_Monolingual_Bias}
\begin{figure*}[ht!]
    \centering
    \begin{subfigure}{0.45\linewidth}{
        \centering
        \includegraphics[trim=0cm 0cm 0cm 2cm,clip=true,width=\linewidth]{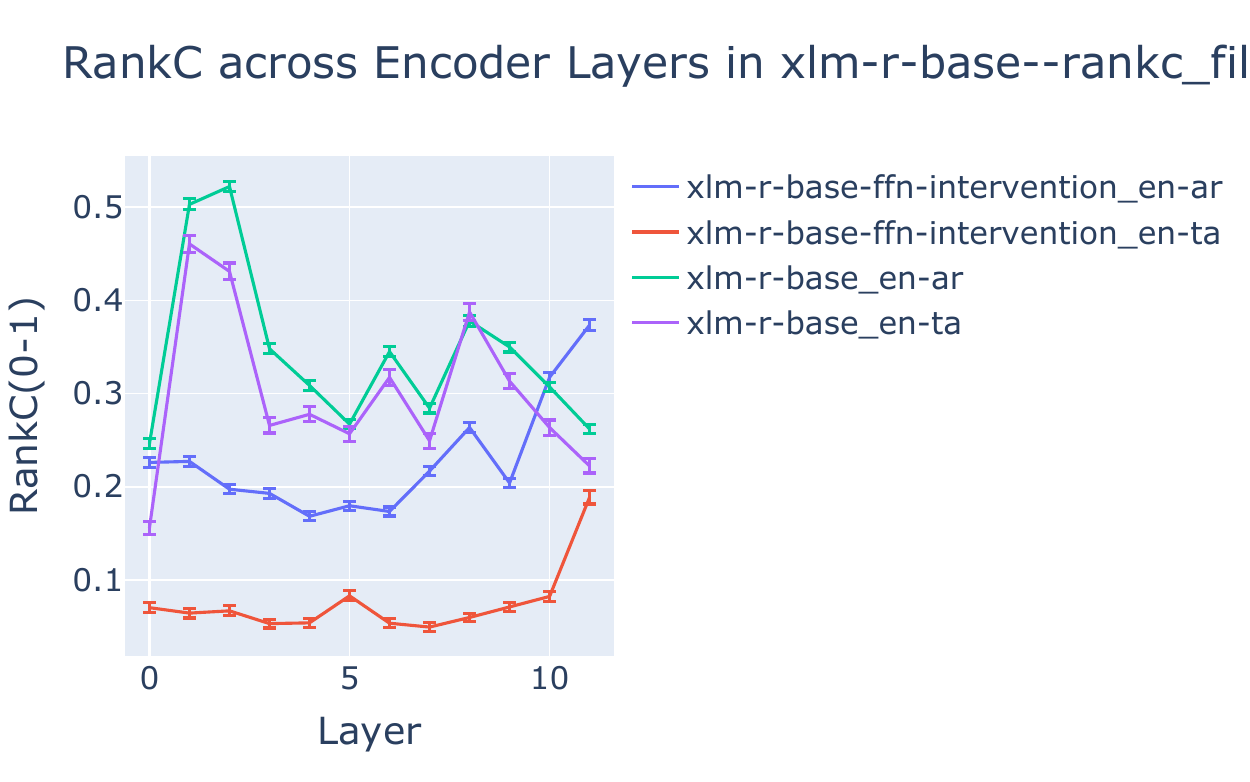}
        \label{fig:xlm-r-base-ffn-rankc}
    }\end{subfigure}
    \begin{subfigure}{0.45\linewidth}{
        \centering
        \includegraphics[trim=0cm 0cm 0cm 2cm,clip=true,width=\linewidth]{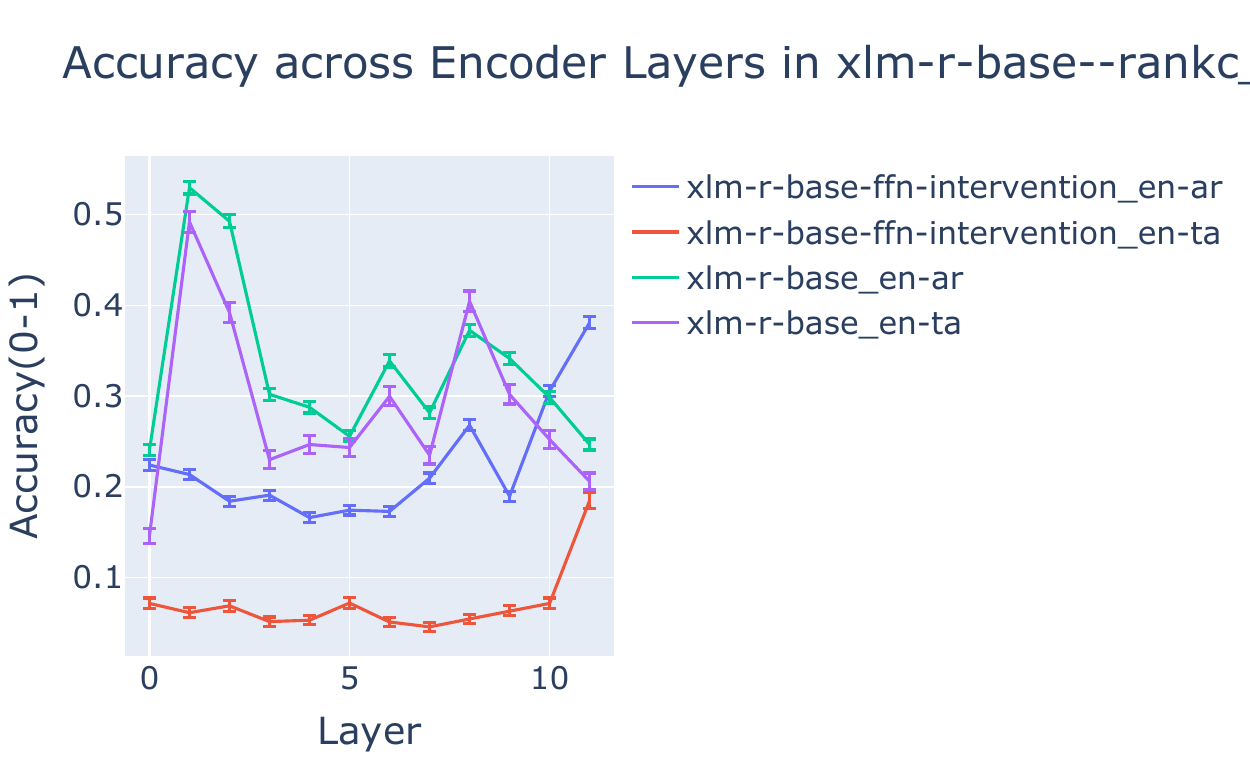}
        \label{fig:xlm-r-base-ffn-acc}
    }\end{subfigure}\\
    \begin{subfigure}{0.45\linewidth}{
        \centering
        \includegraphics[trim=0cm 0cm 0cm 2cm,clip=true,width=\linewidth]{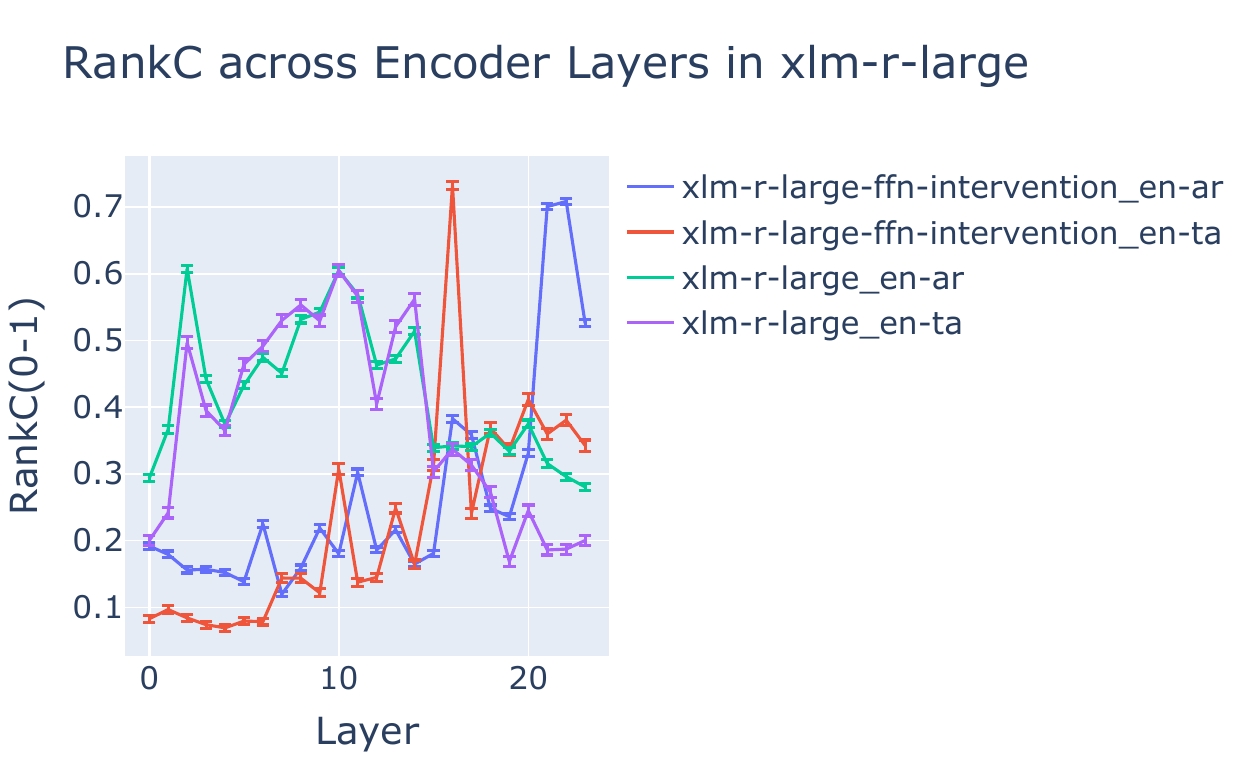}
        \label{fig:xlm-r-large-ffn-rankc}
    }\end{subfigure}
    \begin{subfigure}{0.45\linewidth}{
        \centering
        \includegraphics[trim=0cm 0cm 0cm 2cm,clip=true,width=\linewidth]{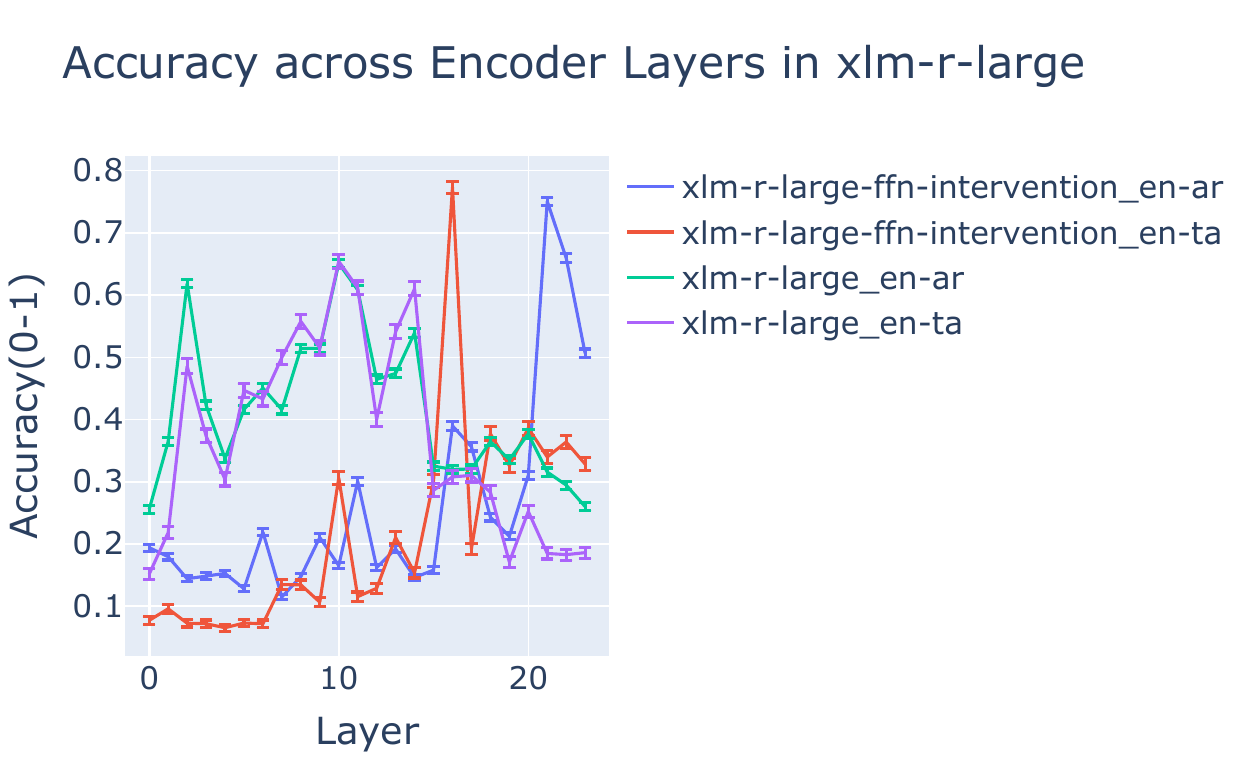}
        \label{fig:xlm-r-large-ffn-acc}
    }\end{subfigure}\\
    \caption{Intervention scores across all models. Metrics legend: left: RankC, right: Top@1 Accuracy. Model family: xlm-r}
    \label{fig:consistency-ffn-intervention-xlm-r}
\end{figure*}
\begin{figure*}[ht!]
    \begin{subfigure}{0.45\linewidth}{
        \centering
        \includegraphics[trim=0cm 0cm 0cm 2cm,clip=true,width=\linewidth]{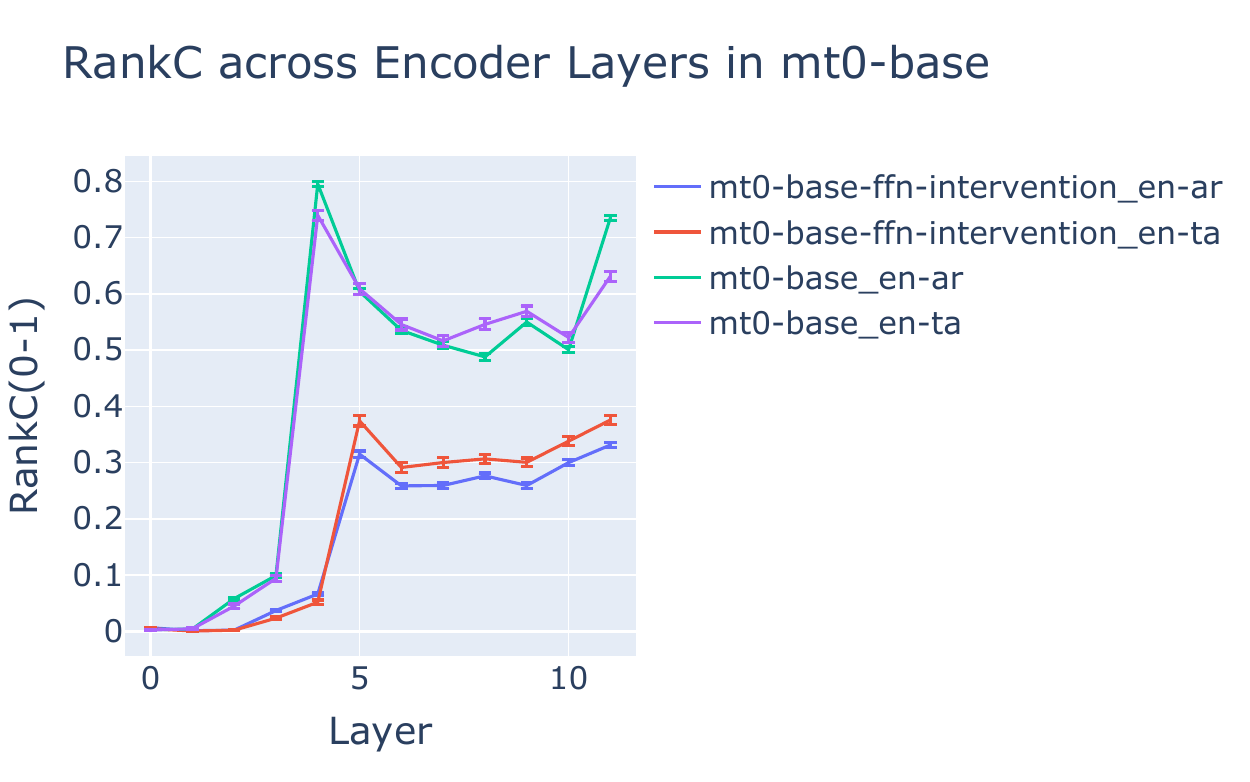}
        \label{fig:mt0-base-ffn-rankc}
    }\end{subfigure}
    \begin{subfigure}{0.45\linewidth}{
        \centering
        \includegraphics[trim=0cm 0cm 0cm 2cm,clip=true,width=\linewidth]{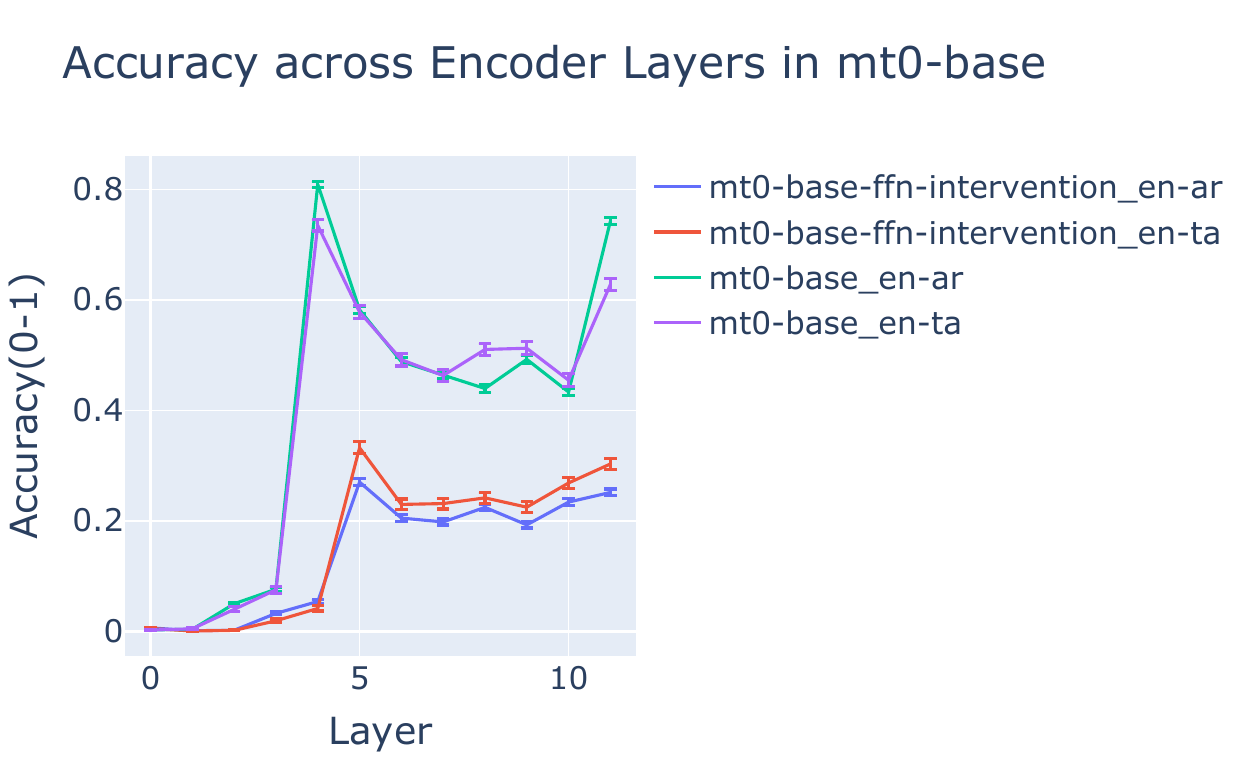}
    }\end{subfigure}\\
    \begin{subfigure}{0.45\linewidth}{
        \centering
        \includegraphics[trim=0cm 0cm 0cm 2cm,clip=true,width=\linewidth]{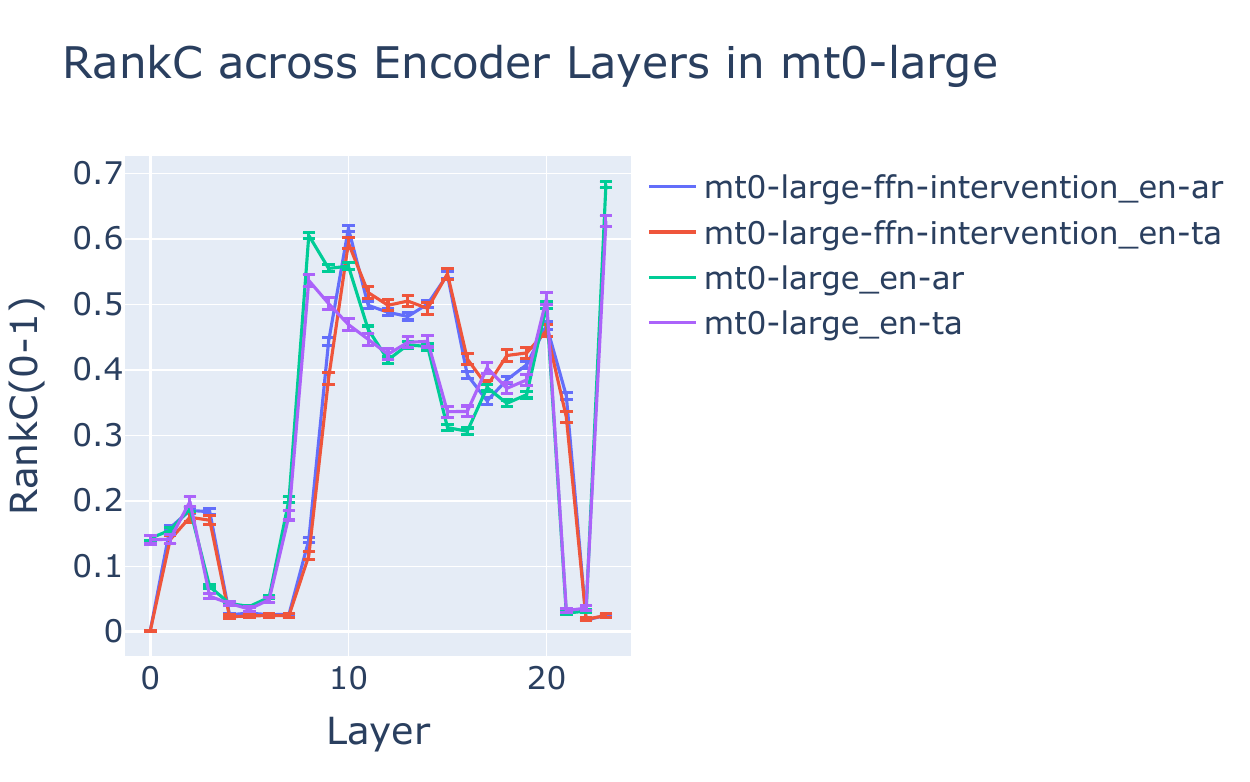}
        \label{fig:mt0-large-ffn-rankc}
    }\end{subfigure}
    \begin{subfigure}{0.45\linewidth}{
        \centering
        \includegraphics[trim=0cm 0cm 0cm 2cm,clip=true,width=\linewidth]{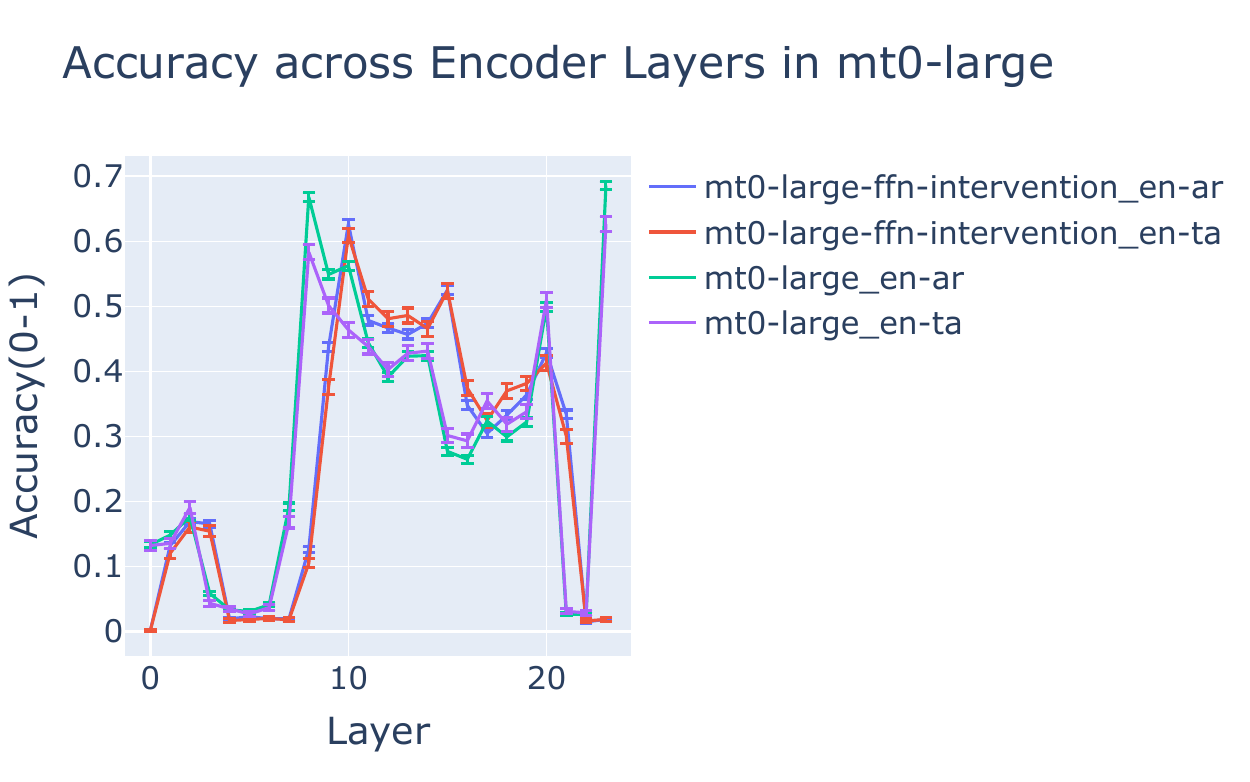}
    }\end{subfigure}
\caption{Intervention scores across all models. Metrics legend: left: RankC, right: Top@1 Accuracy. Model family: mt0.}
    \label{fig:consistency-ffn-intervention-mt0}
\end{figure*}
\begin{figure*}[ht!]
    \begin{subfigure}{0.45\linewidth}{
        \centering
        \includegraphics[trim=0cm 0cm 0cm 2cm,clip=true,width=\linewidth]{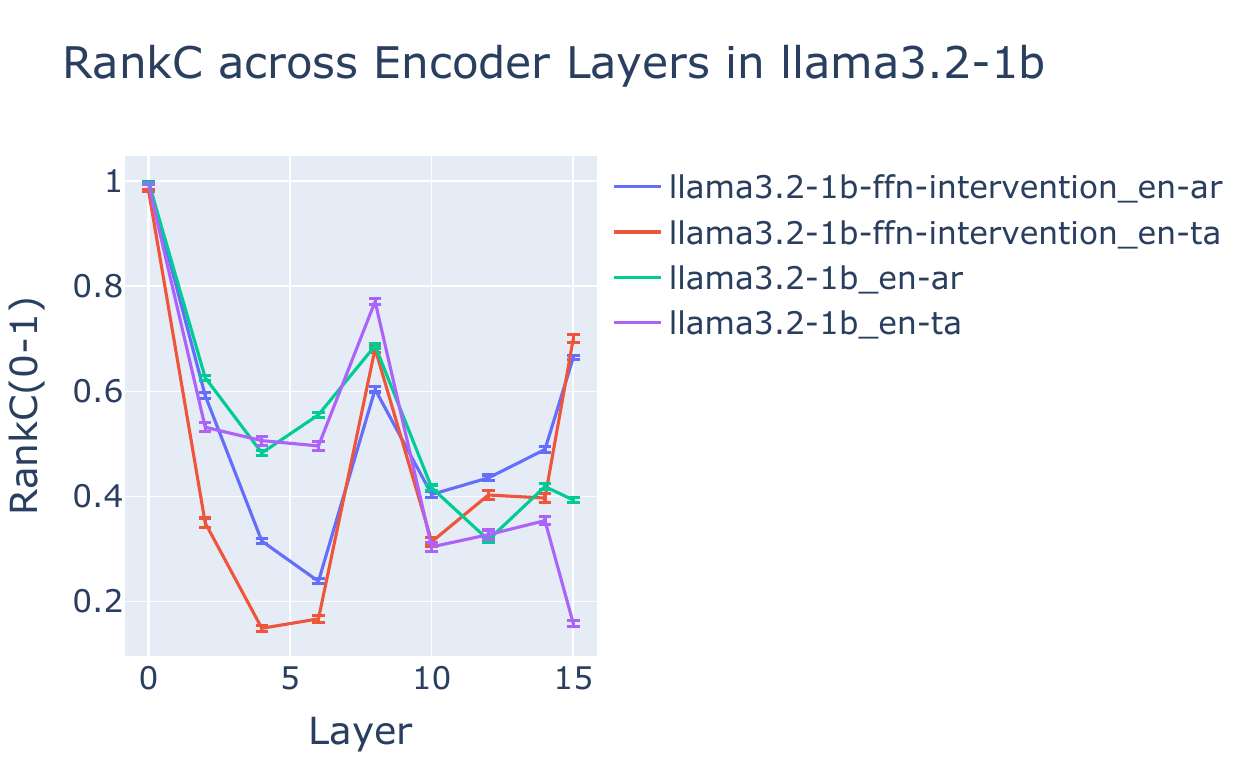}
        \label{fig:llama3.2-1b-ffn-rankc}
    }\end{subfigure}
    \begin{subfigure}{0.45\linewidth}{
        \centering
        \includegraphics[trim=0cm 0cm 0cm 2cm,clip=true,width=\linewidth]{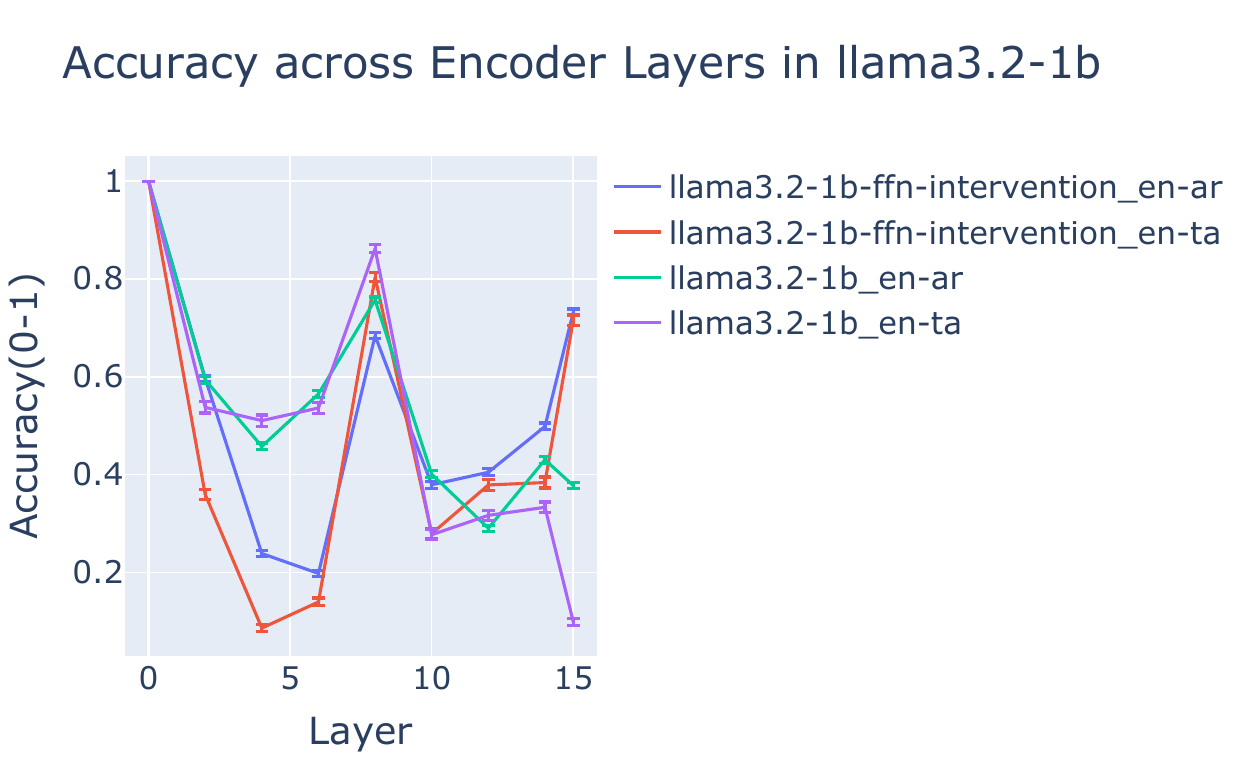}
        \label{fig:llama3.2-1b-ffn-acc}
    }\end{subfigure}
    \begin{subfigure}{0.45\linewidth}{
        \centering
        \includegraphics[trim=0cm 0cm 0cm 2cm,clip=true,width=\linewidth]{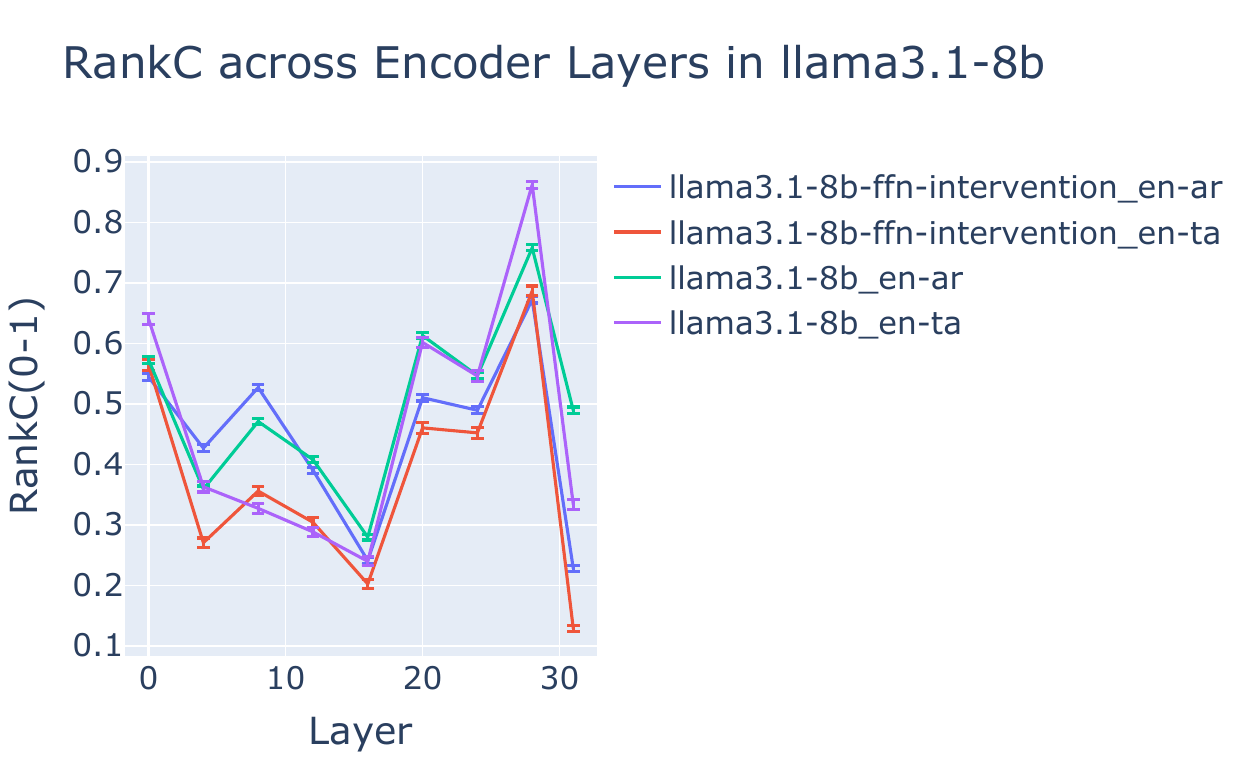}
        \label{fig:llama3.1-8b-ffn-rankc}
    }\end{subfigure}
    \begin{subfigure}{0.45\linewidth}{
        \centering
        \includegraphics[trim=0cm 0cm 0cm 2cm,clip=true,width=\linewidth]{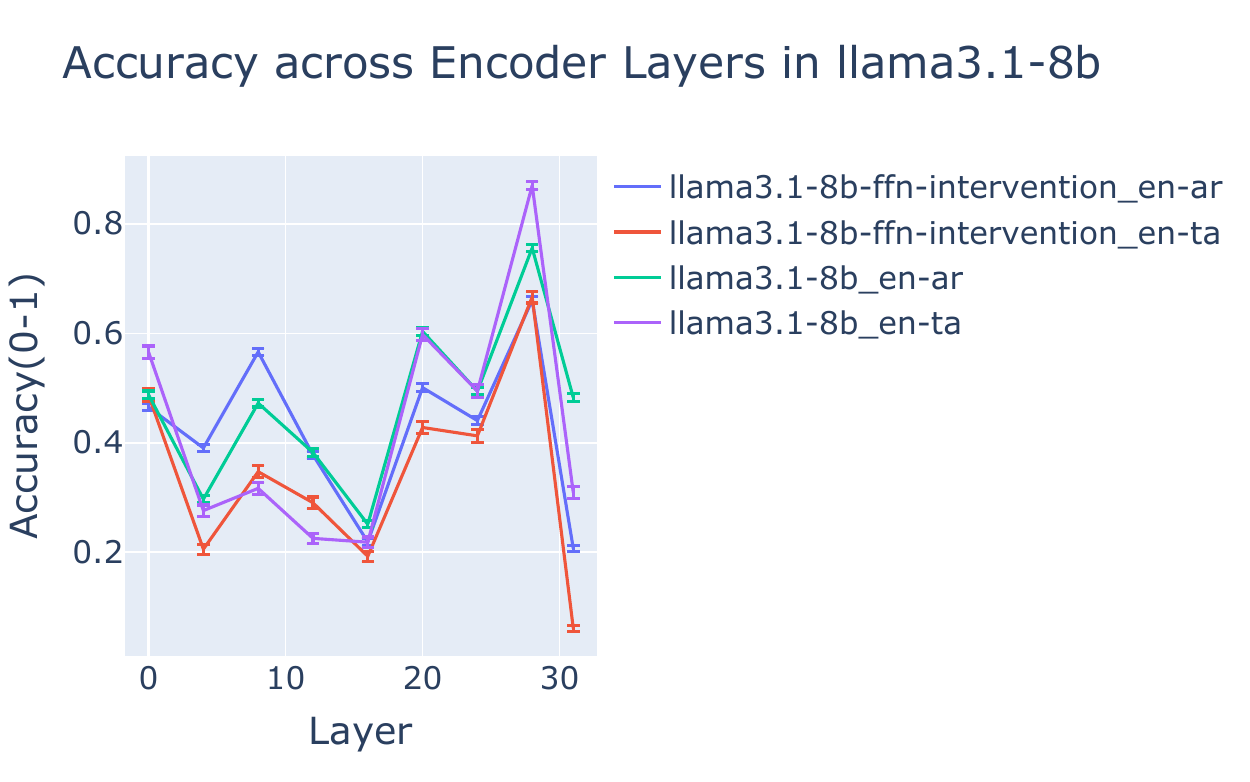}
        \label{fig:llama3.1-8b-ffn-acc}
    }\end{subfigure}
\caption{Intervention scores across all models. Metrics legend: left: RankC, right: Top@1 Accuracy. Model family: Llama 3.}
    \label{fig:consistency-ffn-intervention-xlm-r}
\end{figure*}

\subsubsection{Impact of Larger Vocabulary}
\label{sess:findings_in_details_Impact_of_Larger_Vocabulary}
\begin{figure}[ht!]
    \centering
    \includegraphics[width=\columnwidth]{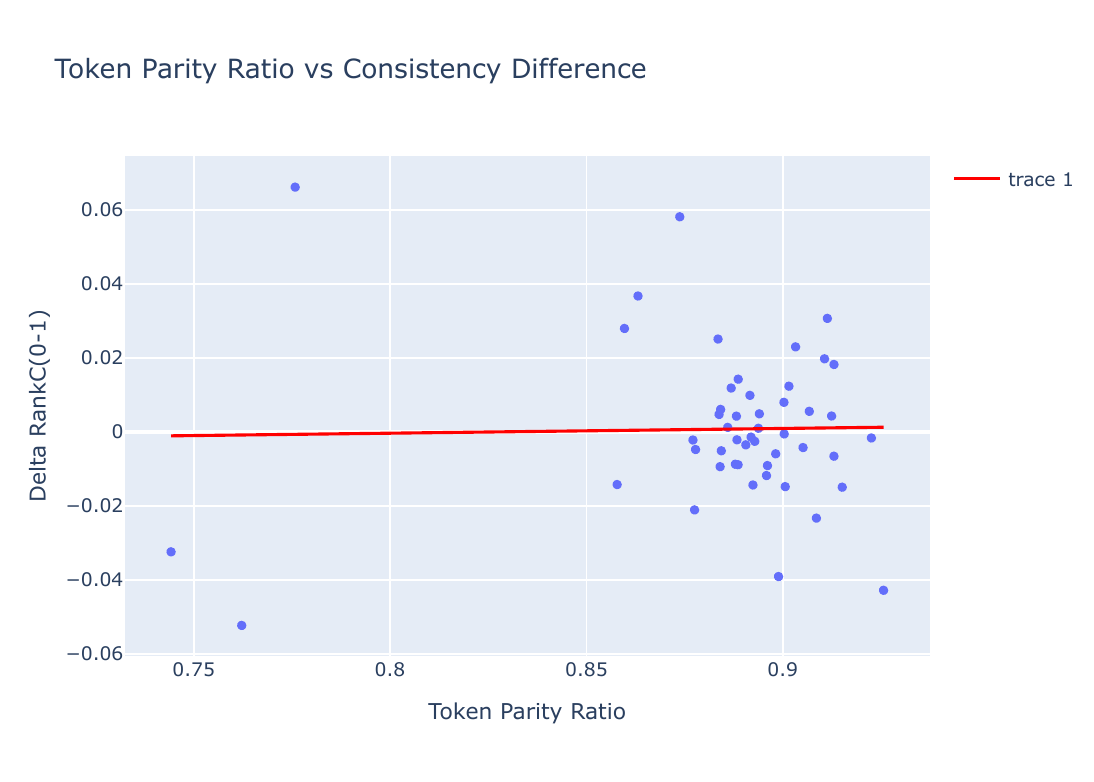}
    \caption{Regression analysis between parity ratio and RankC improvement offered by xlm-v to xlm-r. Spearman $\rho$ = 0.06. 
We define parity ratio as the token length ratio between tokenized subjects for xlm-v-base and xlm-r-base. Our analysis discovers that many languages have a token parity ratio average within 0.8-1, which means that many of the subject entities are known on both tokenizers of the models.}
    \label{fig:parity-ratio-correlation-rankC}
\end{figure}

\begin{figure*}[ht!]
    \centering
    \begin{subfigure}{\linewidth}{
        \centering
        \includegraphics[trim=0cm 0cm 0cm 2cm,clip=true,width=\linewidth]{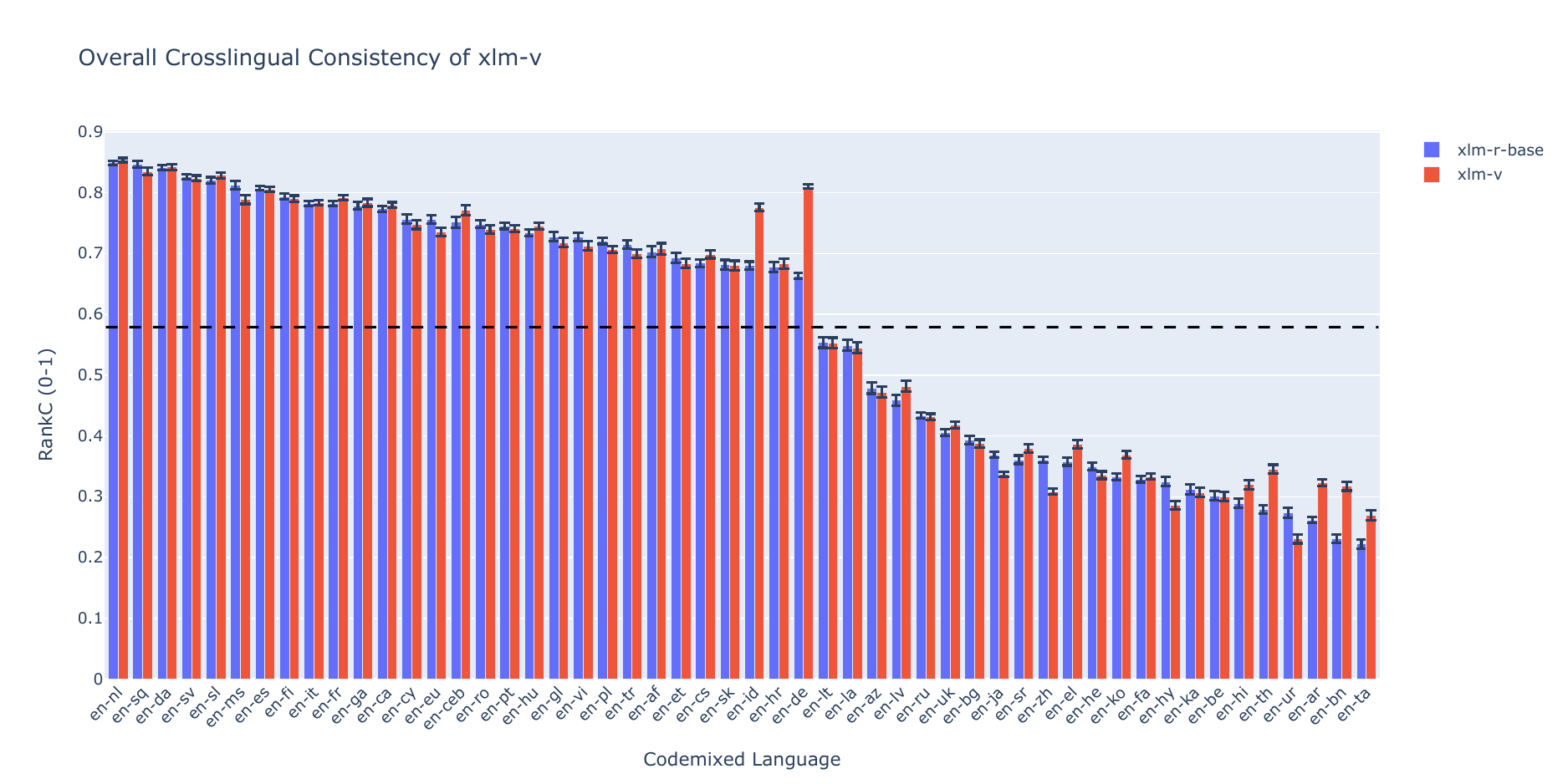}
        \label{fig:xlm-v-base-overall-crosslingual-consistency-rankC}
    }\end{subfigure}
    \begin{subfigure}{\linewidth}{
        \centering
        \includegraphics[trim=0cm 0cm 0cm 2cm,clip=true,width=\linewidth]{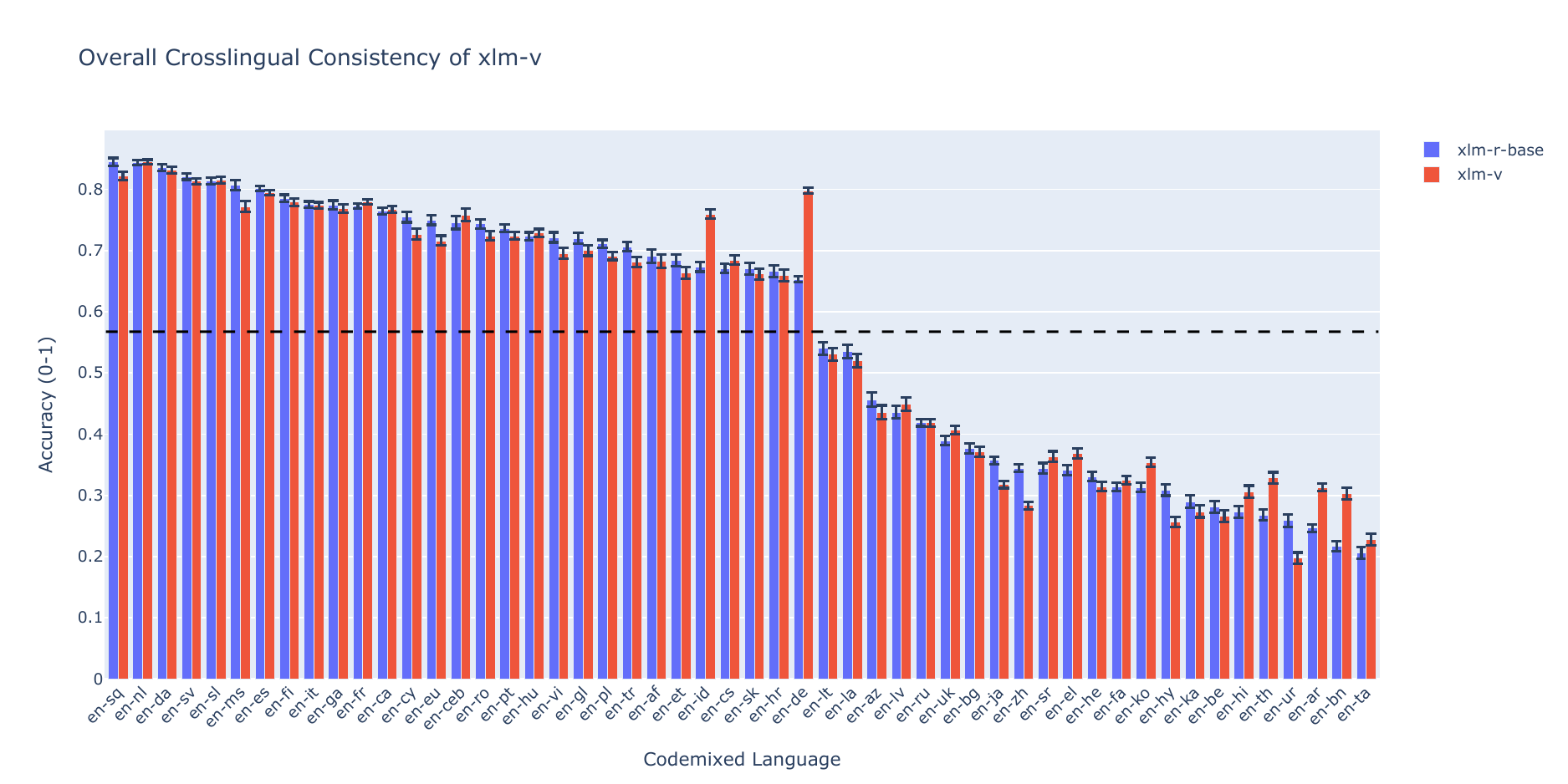}
        \label{fig:xlm-v-base-overall-crosslingual-consistency-acc}
    }\end{subfigure}
    \caption{Effects of vocabulary expansion to overall cross-lingual consistency (top: RankC, bottom: Top@1 Accuracy). Note: The dashed line here is the average corresponding consistency scores of xlm-r-base
across languages}
    \label{fig:vocab-expansion-overall-crosslingual-consistency}
\end{figure*}

When expanding the vocabulary size, we found on Figure \ref{fig:vocab-expansion-overall-crosslingual-consistency} that such method causes marginal improvement. Furthermore we conducted correlation analysis and based on Figure \ref{fig:parity-ratio-correlation-rankC}, we discovered no correlation between token parity seen in table  and consistency improvement and this explains why we observed such limited improvement.

\begin{figure}[ht!]
    \centering
    \begin{subfigure}{\linewidth}{
        \centering
        \includegraphics[trim=0cm 0cm 0cm 2cm,clip=true,width=\linewidth]{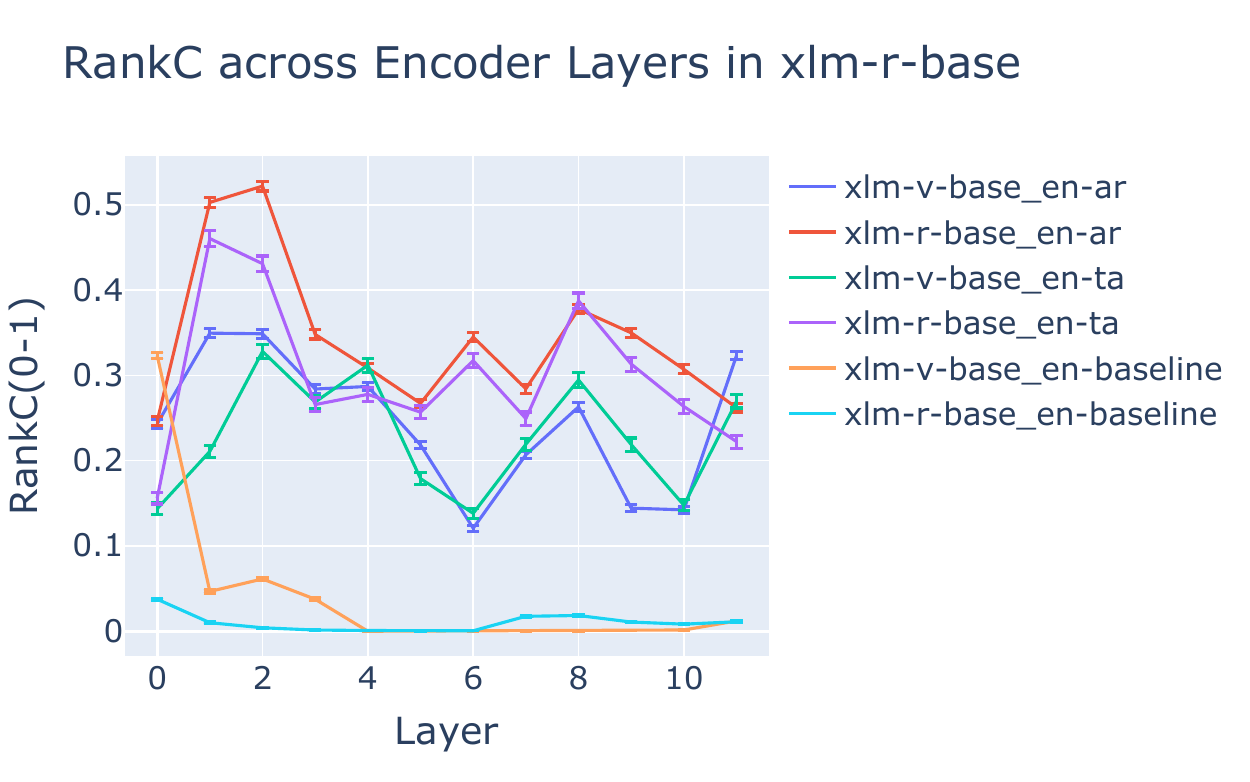}
        \label{fig:xlmv-rankc}
    }\end{subfigure}
    \begin{subfigure}{\linewidth}{
        \centering
        \includegraphics[trim=0cm 0cm 0cm 2cm,clip=true,width=\linewidth]{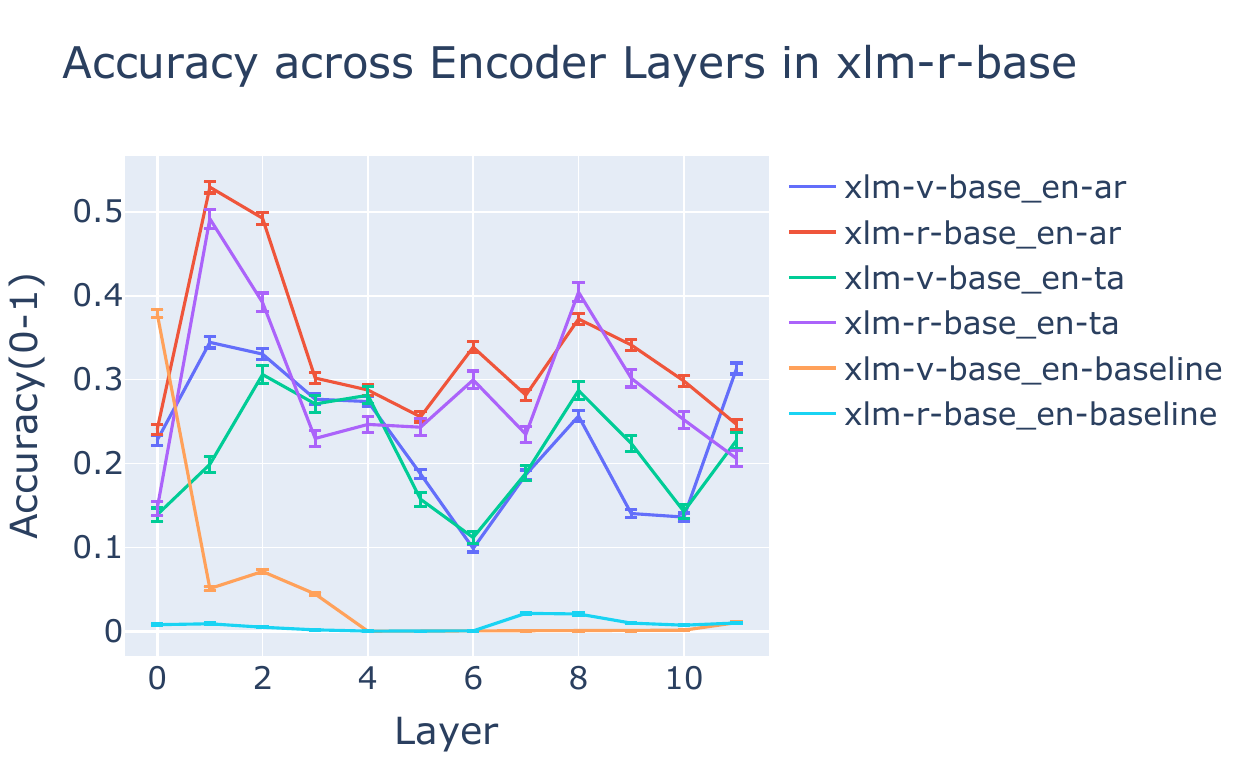}
        \label{fig:xlmv-acc}
    }\end{subfigure}
    \caption{Layer-wise cross-lingual knowledge consistency of xlm-v vs xlm-r-base}
    \label{fig:xlmv-xlmr-kc}
\end{figure}
\subsubsection{The Effect of Cross-Lingual Word Alignment Training Objective to The Cross-lingual Consistency}
\label{ssecsec:word-align-effect}
Another possible hypothesis is that there might be an entanglement of features between linguistic and knowledge features. \cite{elhage2022toy} discovered that a neural network could fit multiple features into one dimension at the price of more entangled features, and this entanglement could cause tokens not cross-lingually aligned, as there may be an entanglement between syntactic and semantic features within one dimension. Inspired by that, we suspect that this might hinder the consistency of language models. To test this assumption, we evaluated two similar language models in which one model is trained solely on MLM objective (xlm-r), and another similar model is trained on one additional objective to align word translations (xlm-align \citep{chi2021improving}), where this word alignment could be helpful in aligning references across languages. 
\begin{figure}[ht!]
        \centering
        \begin{subfigure}{\linewidth}{
        \centering
        \includegraphics[trim=0cm 0cm 0cm 2cm,clip=true,width=\linewidth]{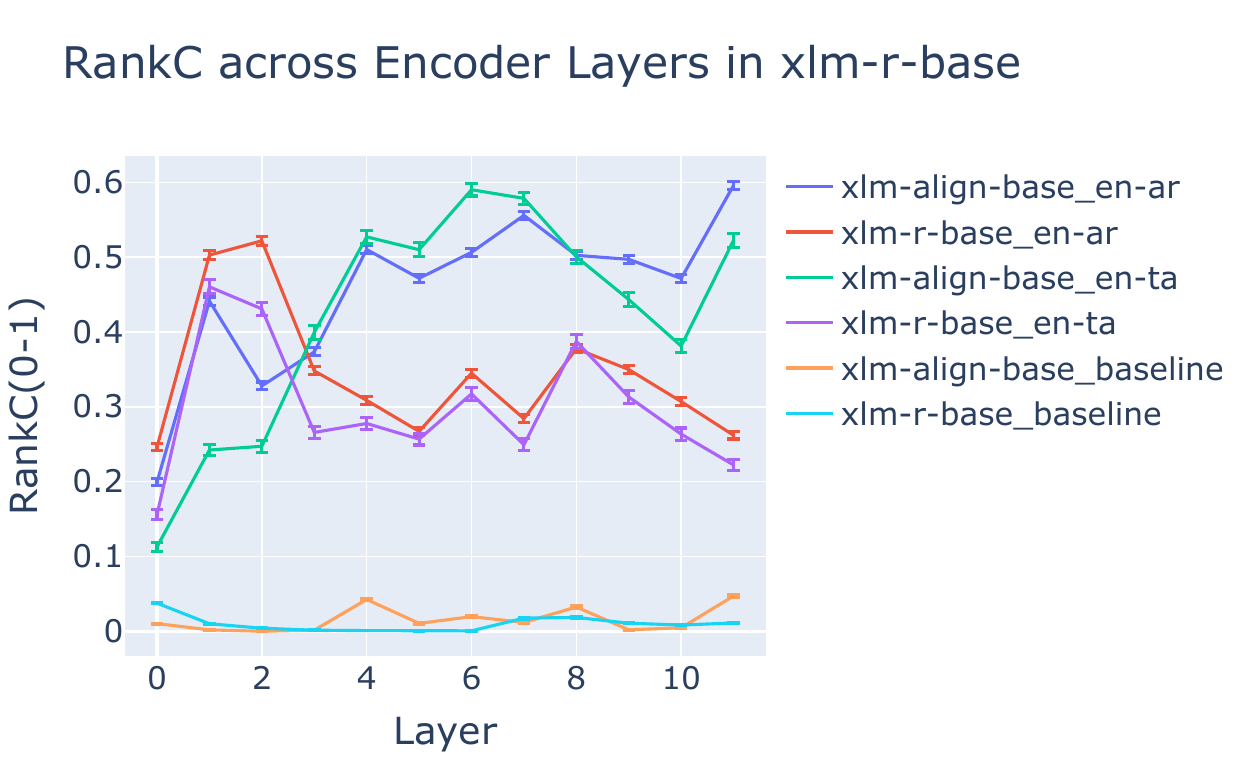}
        \label{app_fig:xlm-align-base-layerwise-crosslingual-consistency-rankC}
    }\end{subfigure}\\
    \begin{subfigure}{\linewidth}{
        \centering
        \includegraphics[trim=0cm 0cm 0cm 2cm,clip=true,width=\linewidth]{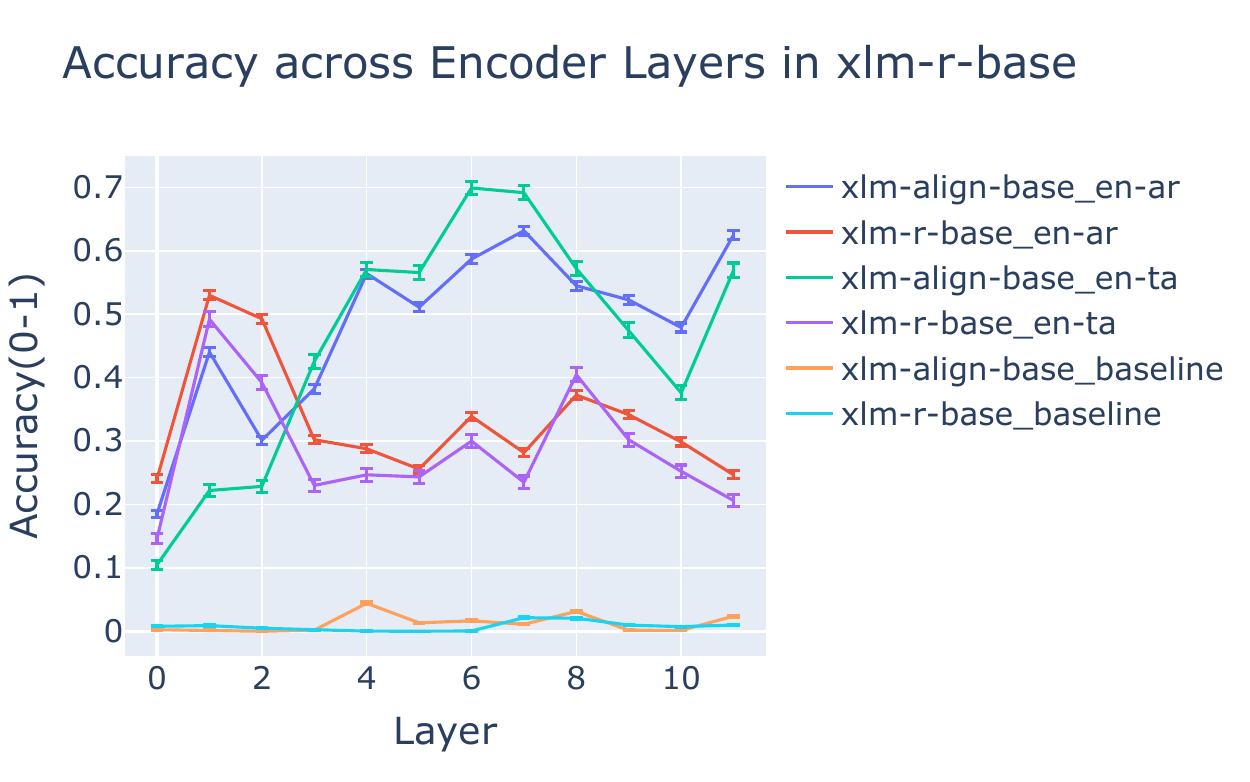}
        \label{app_fig:xlm-align-base-layerwise-crosslingual-consistency-acc}
    }\end{subfigure}
    \caption{Effects of cross-lingual word-alignment  training on the layer-wise consistency.}
    \label{app_fig:xlm-align-kc}
\end{figure}
\begin{figure*}[ht!]
    \centering
    \begin{subfigure}{\linewidth}{
        \centering
        \includegraphics[ width=\linewidth]{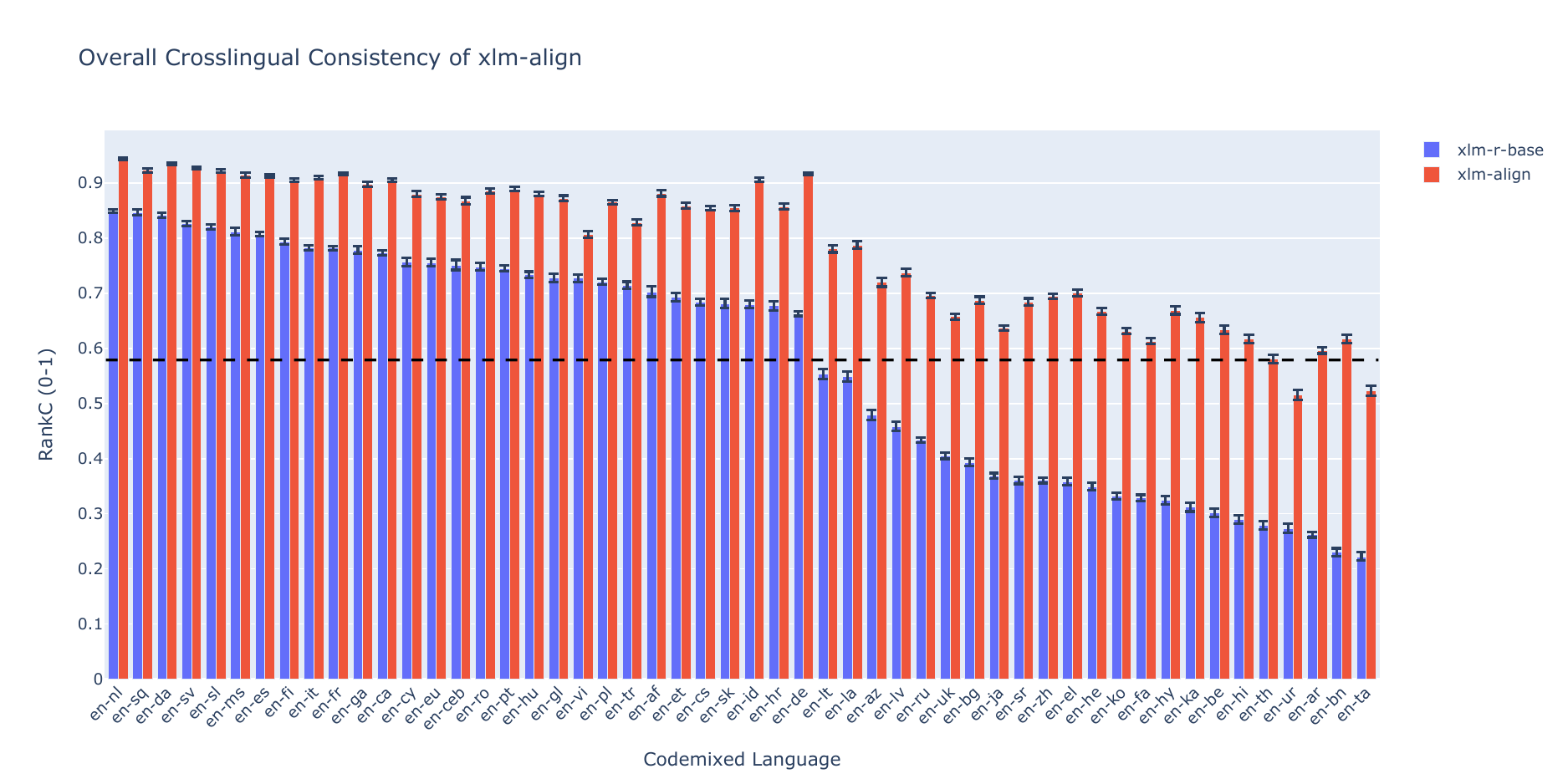}
        \label{app_fig:xlm-align-base-overall-crosslingual-consistency-rankC}
    }\end{subfigure}
    \\
    \begin{subfigure}{\linewidth}{
        \centering
        \includegraphics[ width=\linewidth]{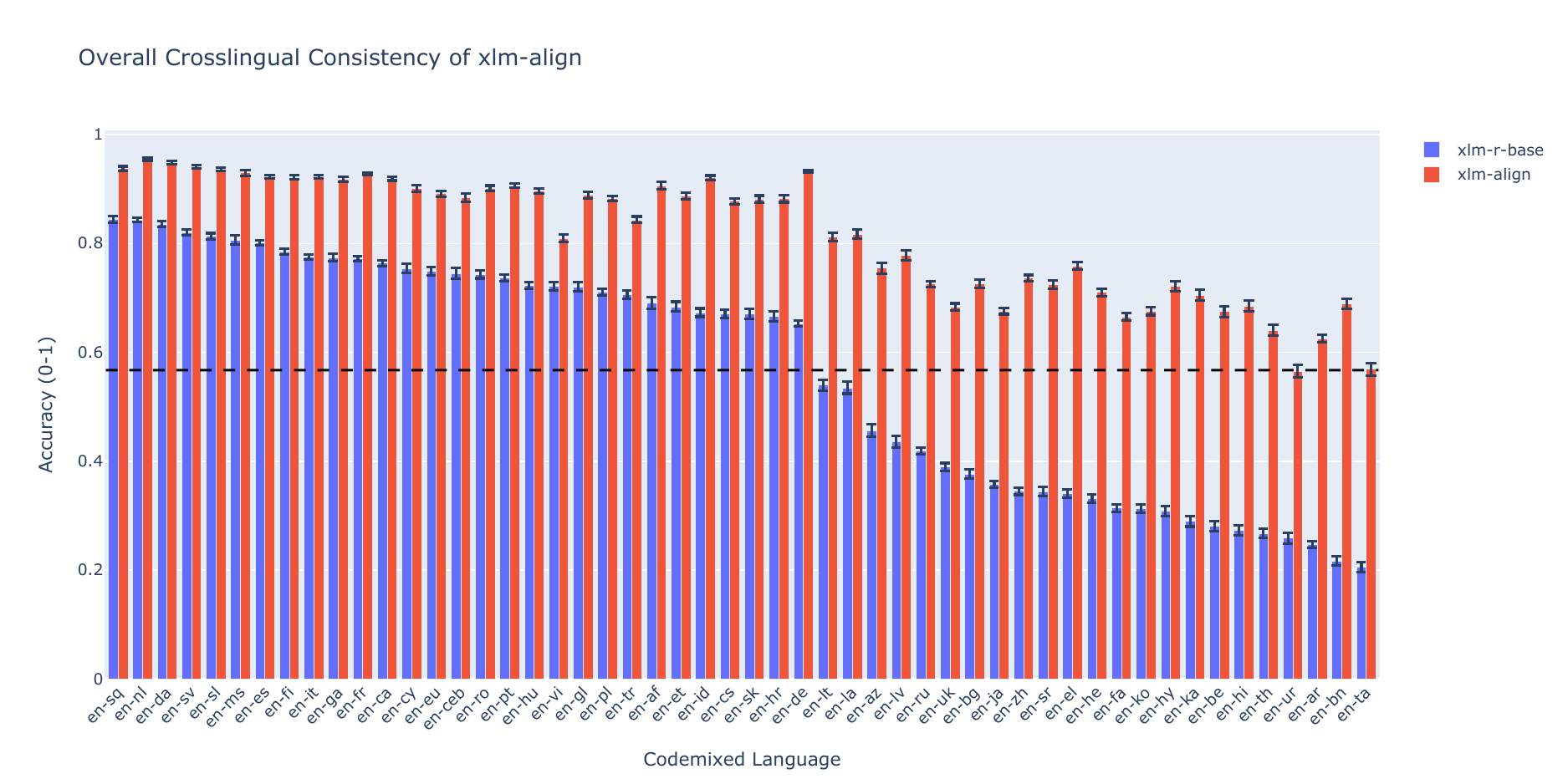}
        \label{app_fig:xlm-align-base-overall-crosslingual-consistency-acc}
    }\end{subfigure}
    \caption{Effects of additional cross-lingual word alignment to overall cross-lingual consistency (top: RankC, bottom: Top@1 Accuracy). Note: The dashed line here is the average corresponding consistency scores of xlm-r-base
across languages}
\end{figure*}

Word alignment increases cross-lingual consistency monotonically to alleviate the cross-lingual bottleneck. Similar to the vocabulary expansion, this strategy does not improve the consistency for the baseline as we would expect. The aligned model outperforms the baseline starting from the middle layers in Figure \ref{app_fig:xlm-align-kc}. Multiple pre-training objectives that could approximately disentangle different features can help preserve the model's knowledge of different languages. We could also confirm this finding by observing the overall cross-lingual consistency result in. In addition, word alignments improve consistency for transliterations or similar orthographical forms, contributing to model's robustness against orthographic variations and non-standard spellings, but vocabulary expansion can not offer such gains. 

\subsubsection{The Effect of Code-switching Training to The Cross-lingual Consistency}
\label{ssecsec:cs-effect}
\begin{figure}[ht!]
    \centering
    \begin{subfigure}{\linewidth}{
        \centering
        \includegraphics[trim=0cm 0cm 0cm 2cm,clip=true,width=\linewidth]{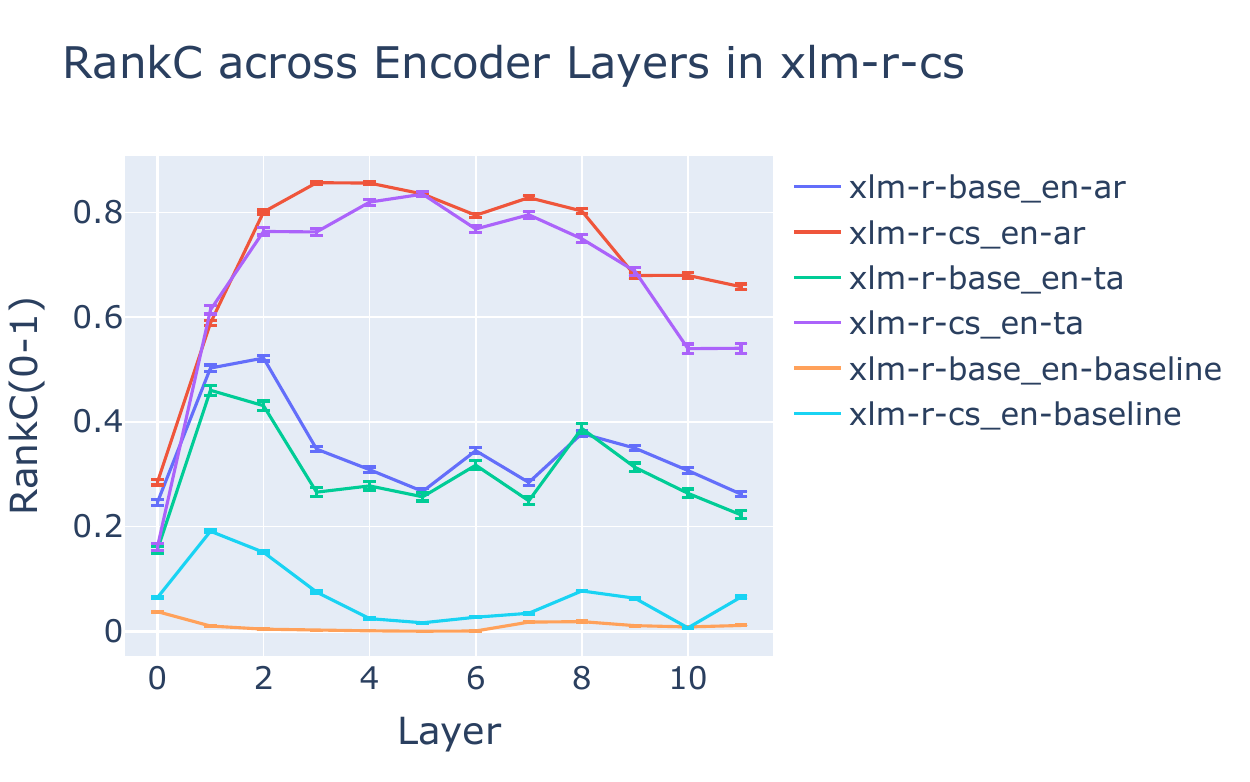}
        \label{app_fig:xlm-r-cs-layerwise-crosslingual-consistency-rankC}
    }\end{subfigure}
    \\
    \begin{subfigure}{\linewidth}{
        \centering
        \includegraphics[trim=0cm 0cm 0cm 2cm,clip=true,width=\linewidth]{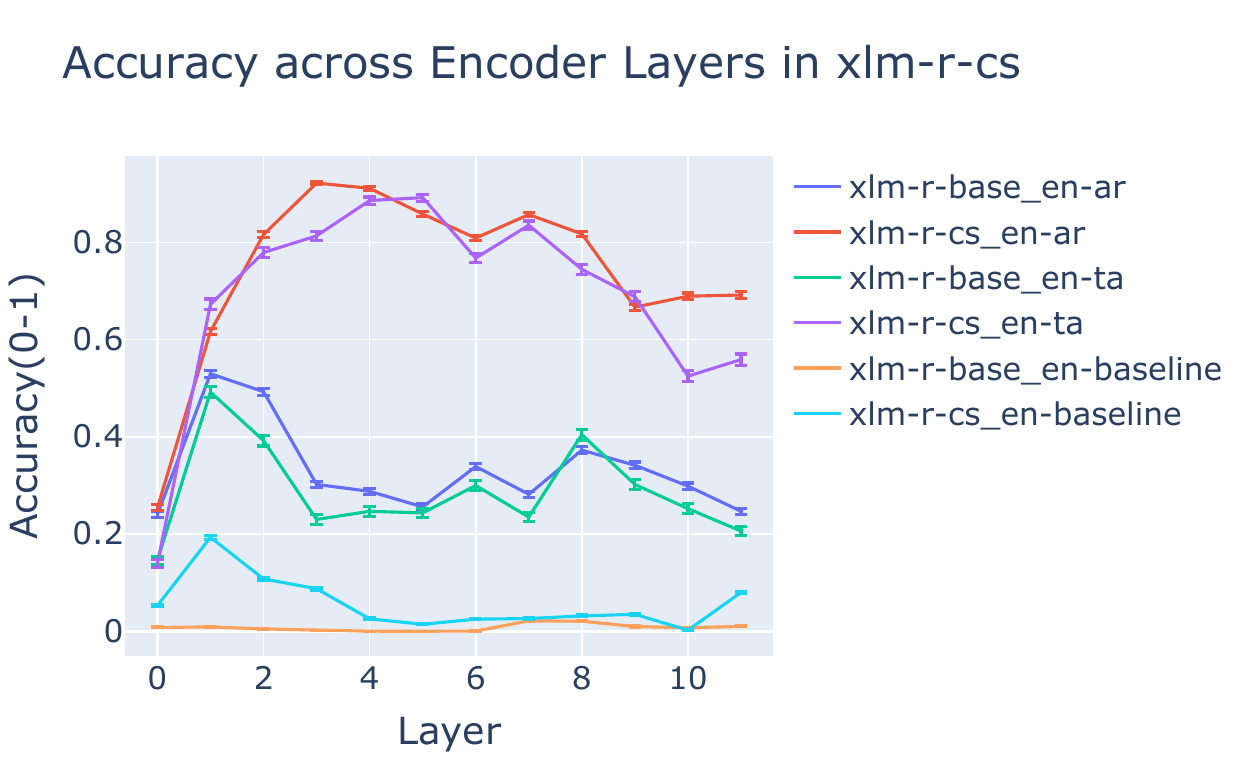}
        \label{app_fig:xlm-r-cs-layerwise-crosslingual-consistency-acc}
    }\end{subfigure}
    \caption{Effects of code-switching training to layer-wise cross-lingual consistency (top: RankC, bottom: Top@1 Accuracy).}
    \label{app_fig:xlmrCS-kc}
\end{figure}
\begin{figure*}[ht!]
    \centering
    \begin{subfigure}{\linewidth}{
        \centering
        \includegraphics[trim=0cm 0cm 0cm 2cm,clip=true,width=\linewidth]{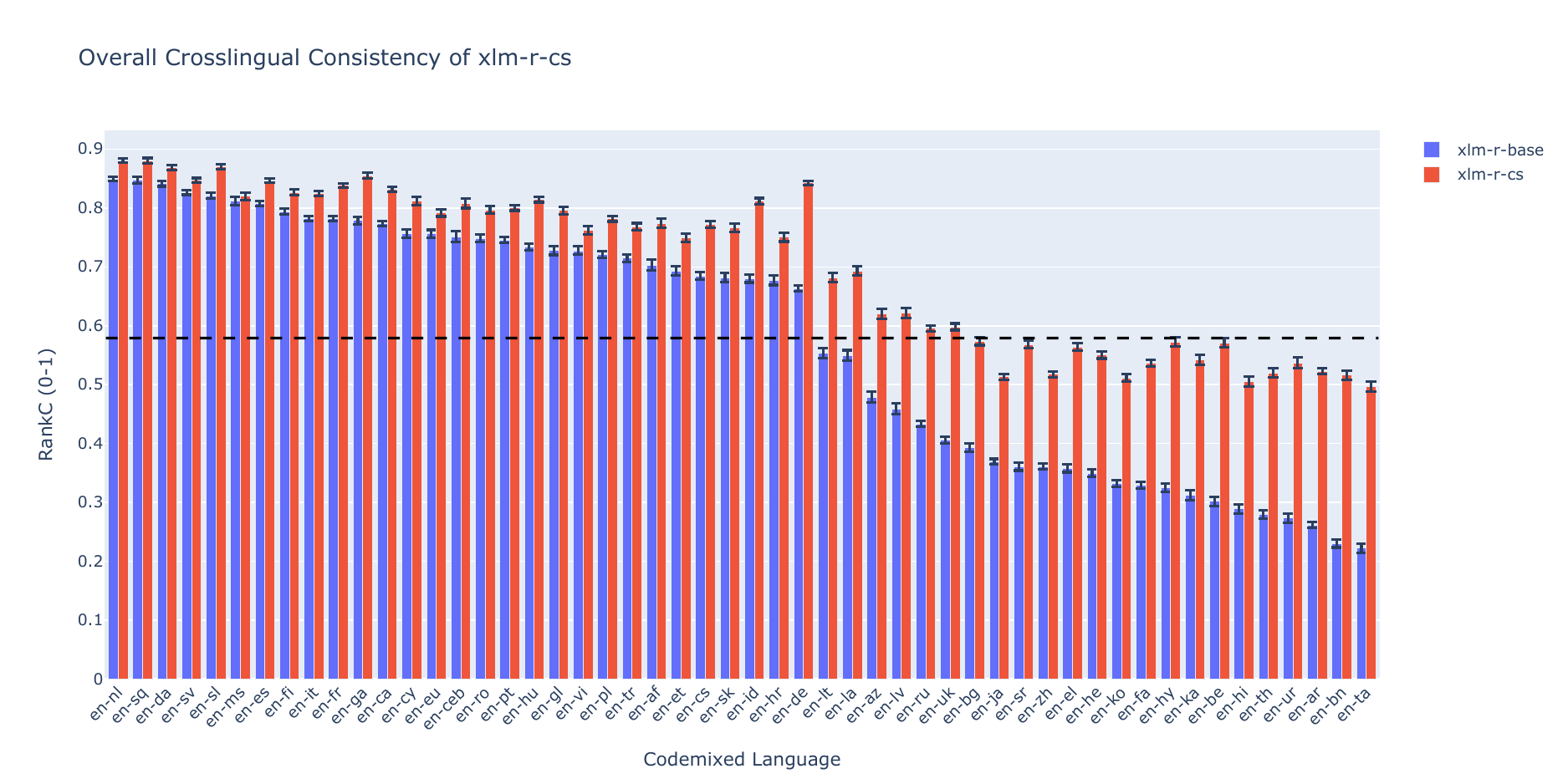}
        \label{app_fig:xlm-r-cs-overall-crosslingual-consistency-rankC}
    }\end{subfigure}
    \\
    \begin{subfigure}{\linewidth}{
        \centering
        \includegraphics[trim=0cm 0cm 0cm 2cm,clip=true,width=\linewidth]{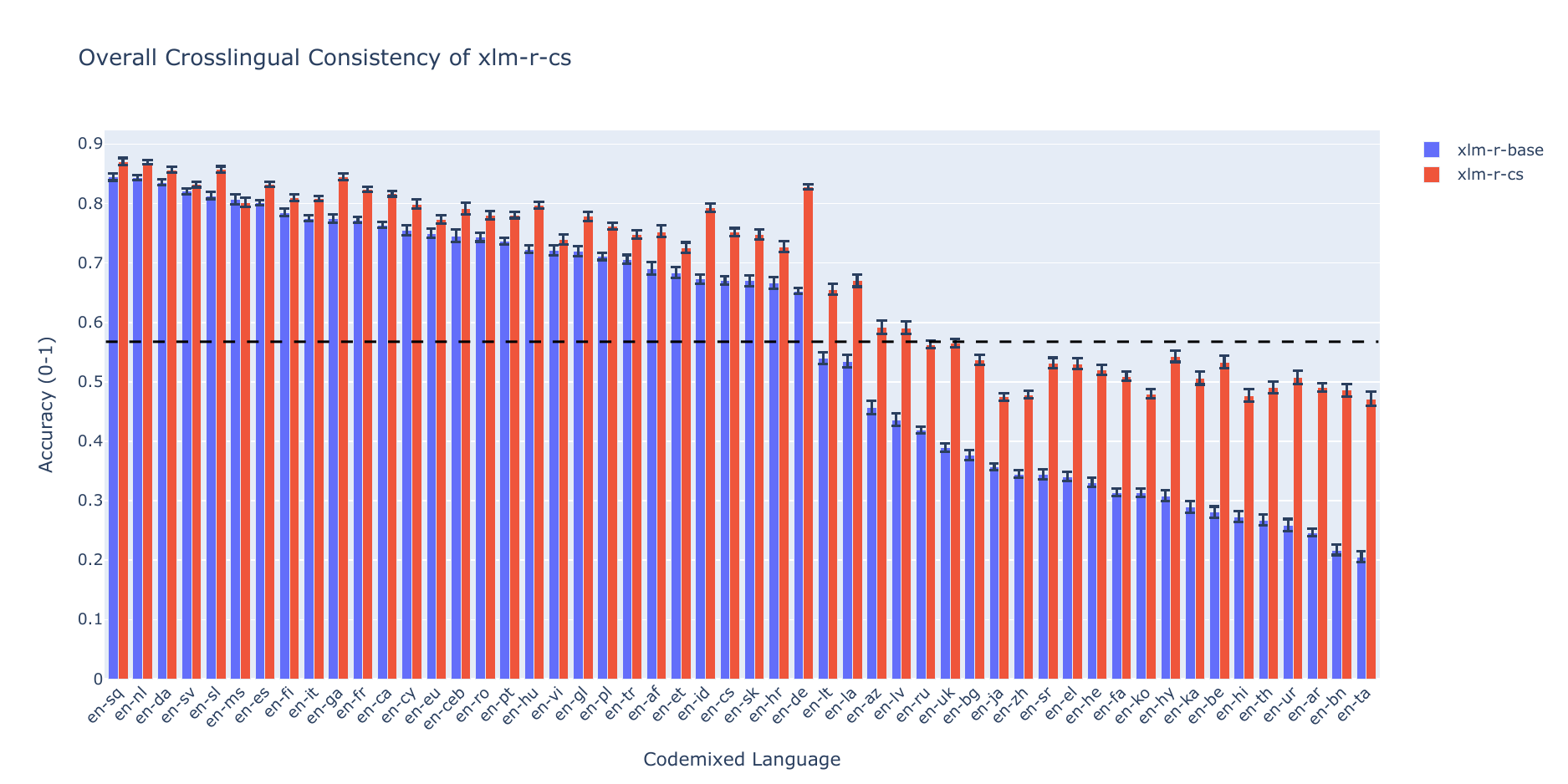}
        \label{app_fig:xlm-r-cs-overall-crosslingual-consistency-acc}
    }\end{subfigure}
    \caption{Effects of code-switching training to overall cross-lingual consistency (top: RankC, bottom: Top@1 Accuracy). Note: The dashed line here is the average corresponding consistency scores of xlm-r-base
across languages}
\end{figure*}
Inspired by the experiment on cross-lingual supervision, we further hypothesize that code-switching training, which substitutes an entity with alternatives from other languages for intra-sentential alignments in cross-lingual settings, can help the model understand common knowledge across languages for cross-lingual consistency to some extent. To evaluate this hypothesis, we study xlm-r and xlm-r-cs \citep{whitehouse-etal-2022-entitycs}, where xlm-r-cs is continuously trained on code-switching corpus from xlm-r-base and shows high performance in multilingual fact-checking. From Figure \ref{app_fig:xlmrCS-kc}, we observe a shift in the consistency bottleneck from the middle layers to the later layers of xlm-r-cs, where the consistency gap between dissimilar and similar languages narrows in xlm-r-cs compared to xlm-r in the middle layers. Overall, code-switching can offer significant gains to the cross-lingual consistency, even without additional objectives. 

\subsubsection{The Effect of Multi-task Fine-tuning to The Cross-lingual Consistency}
\label{ssecsec:mutlitask-effect}
\begin{figure}[ht!]
    \centering
    \begin{subfigure}{\linewidth}{
        \centering
        \includegraphics[trim=0cm 0cm 0cm 2cm,clip=true,width=\linewidth]{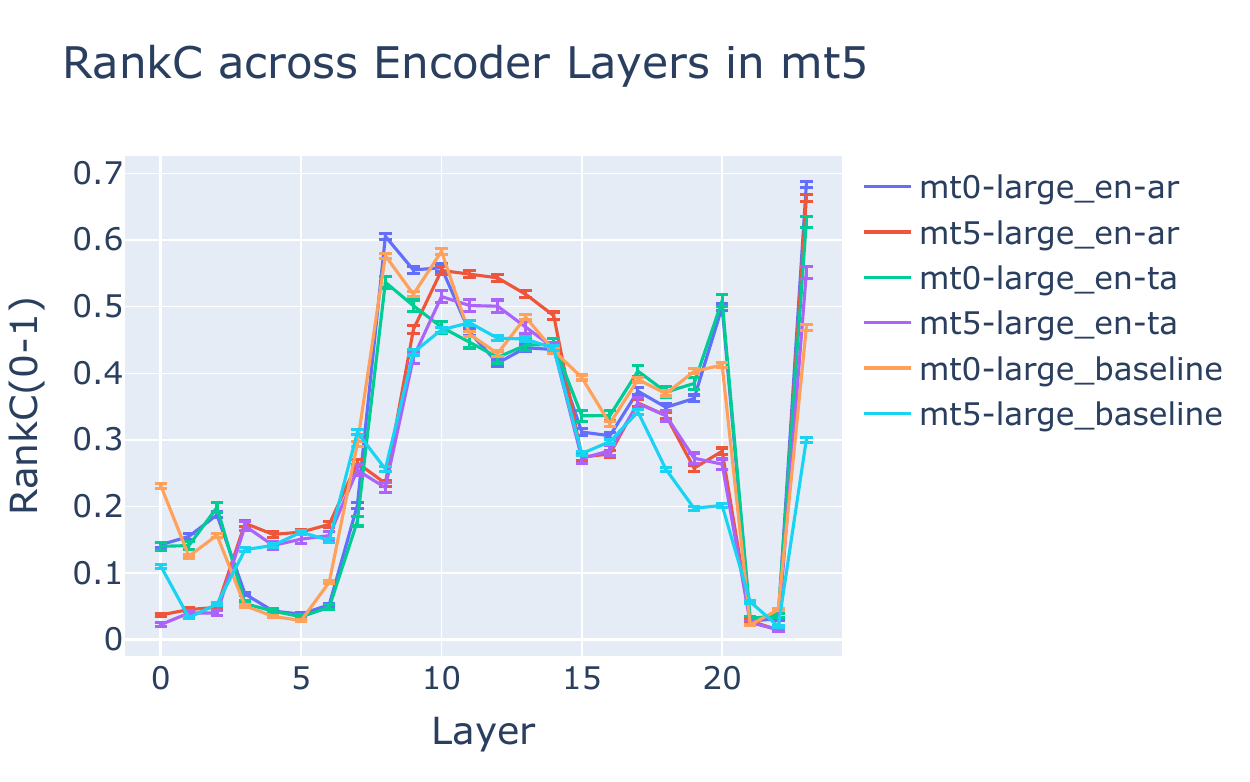}
        \label{app_fig:mt5-layerwise-crosslingual-consistency-rankC}
    }\end{subfigure} \\ 
        \begin{subfigure}{\linewidth}{
        \centering
        \includegraphics[trim=0cm 0cm 0cm 2cm,clip=true,width=\linewidth]{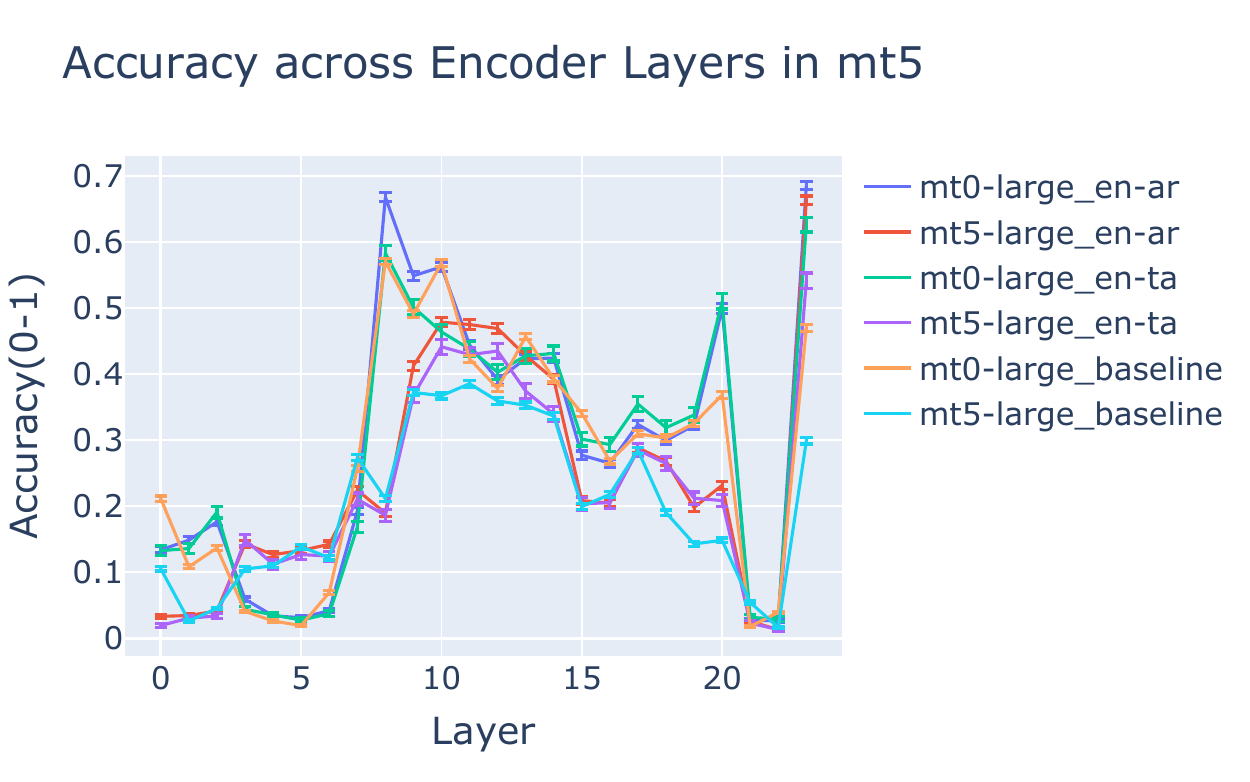}
        \label{app_fig:mt5-layerwise-crosslingual-consistency-rankC}
    }\end{subfigure} \\ 
    \caption{Effects of multi-task instruction tuning on the layer-wise consistency in encoder-decoder models.}
    \label{app_fig:multi-task-instruction-tuning_encoder-decoder-layerwise}
\end{figure}
\begin{figure}[ht!]
    \centering
    \begin{subfigure}{\linewidth}{
        \centering
        \includegraphics[trim=0cm 0cm 0cm 2cm,clip=true,width=\linewidth]{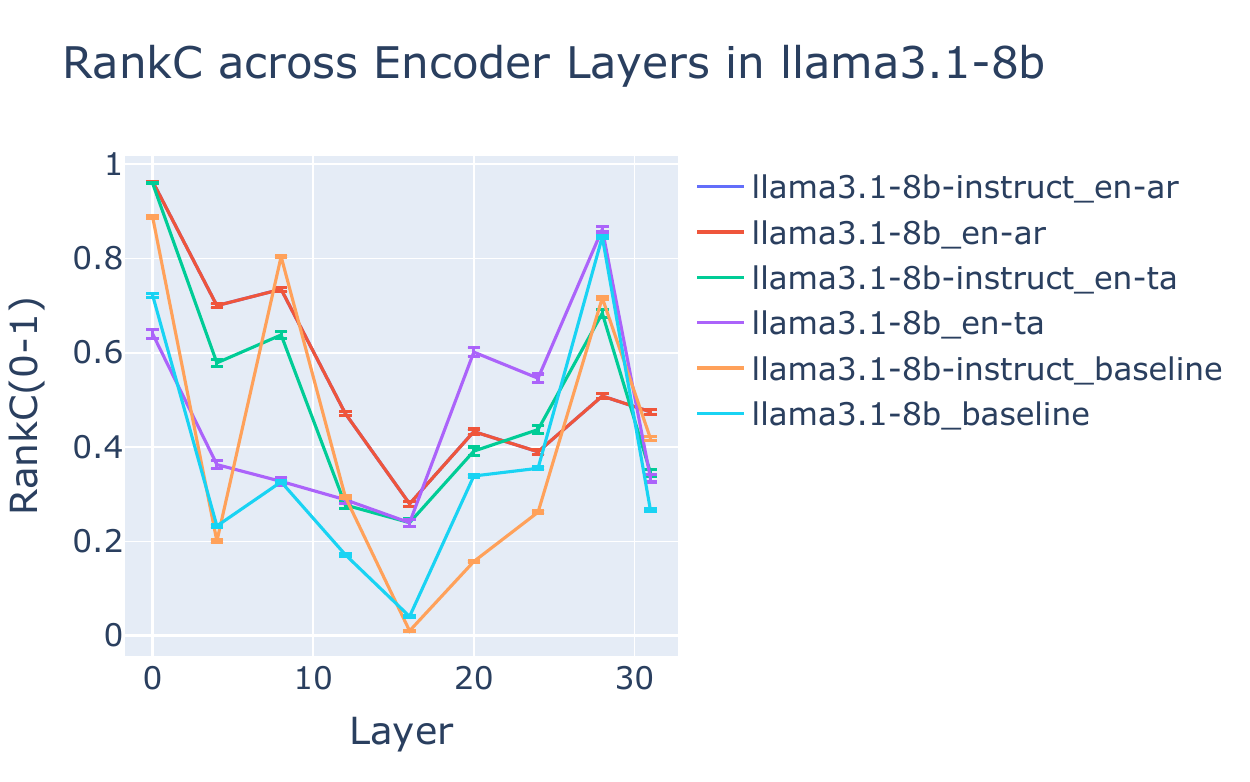}
        \label{app_fig:llama3.1-8b-instruct-layerwise-crosslingual-consistency-rankC}
    }\end{subfigure} \\ 
        \begin{subfigure}{\linewidth}{
        \centering
        \includegraphics[trim=0cm 0cm 0cm 2cm,clip=true,width=\linewidth]{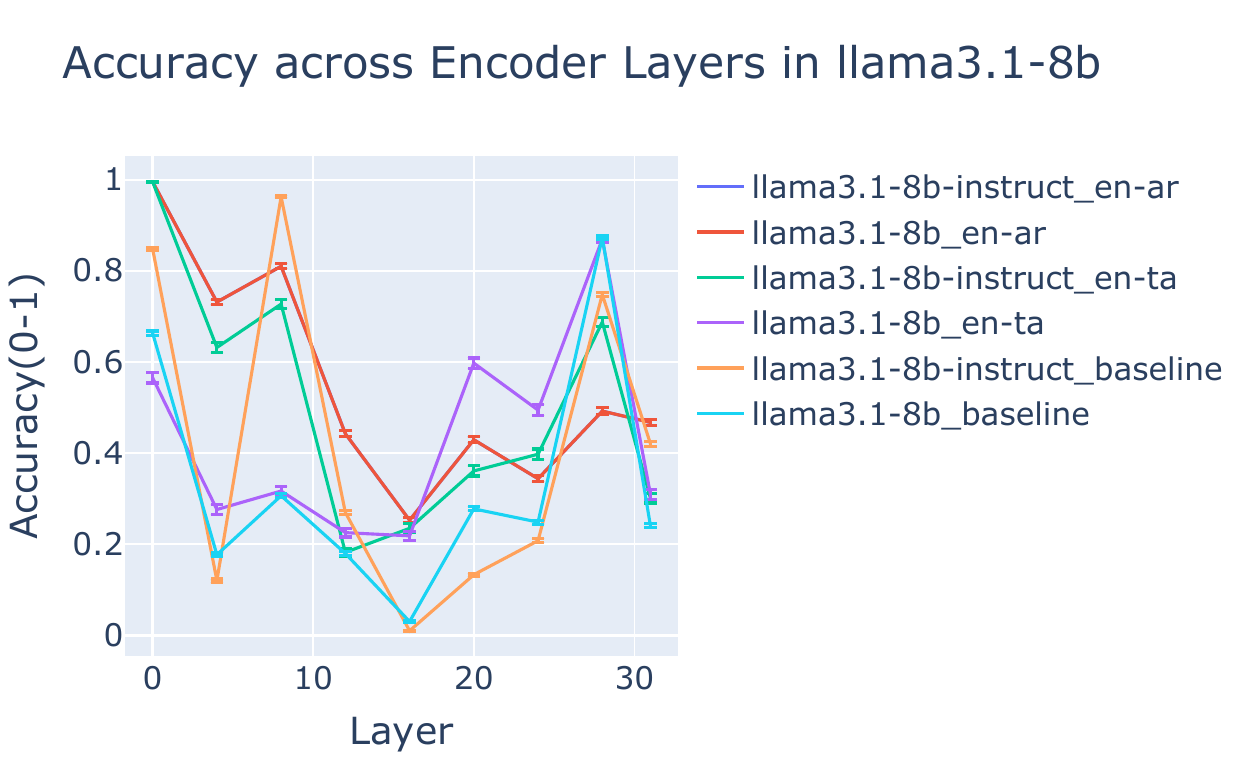}
        \label{app_fig:llama3.1-8b-instruct-layerwise-crosslingual-consistency-accuracy}
    }\end{subfigure} \\ 
    \caption{Effects of multi-task instruction tuning on the layer-wise consistency of decoder models.}
    \label{app_fig:multi-task-instruction-tuning_decoder-layerwise}
\end{figure}
\begin{figure*}[ht!]
    \centering
    \begin{subfigure}{\linewidth}{
        \centering
        \includegraphics[trim=0cm 0cm 0cm 2cm,clip=true,width=\linewidth]{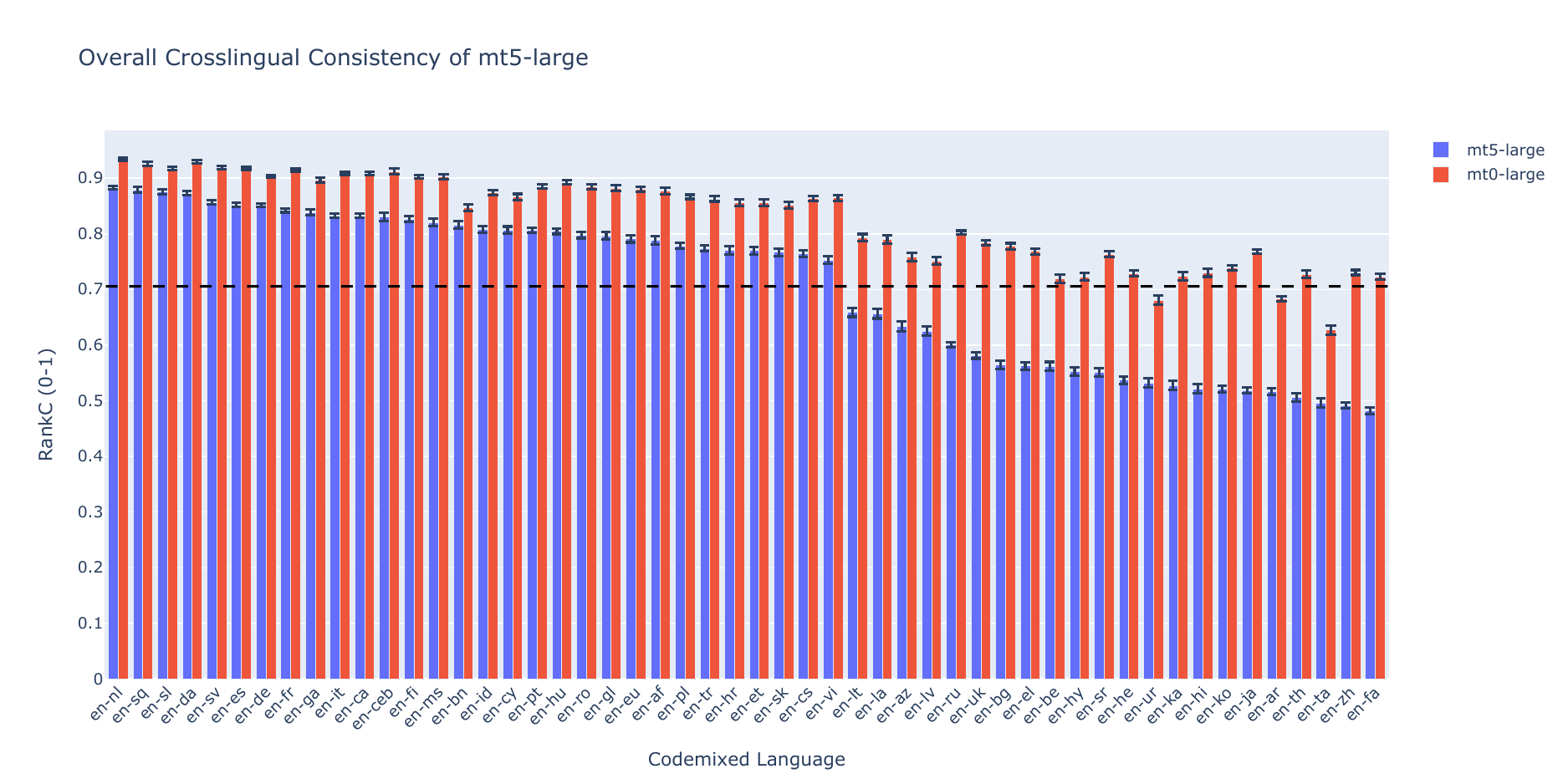}
        \label{app_fig:mt5-large-overall-crosslingual-consistency-rankC}
    }\end{subfigure}
    \\
    \begin{subfigure}{\linewidth}{
        \centering
        \includegraphics[trim=0cm 0cm 0cm 2cm,clip=true,width=\linewidth]{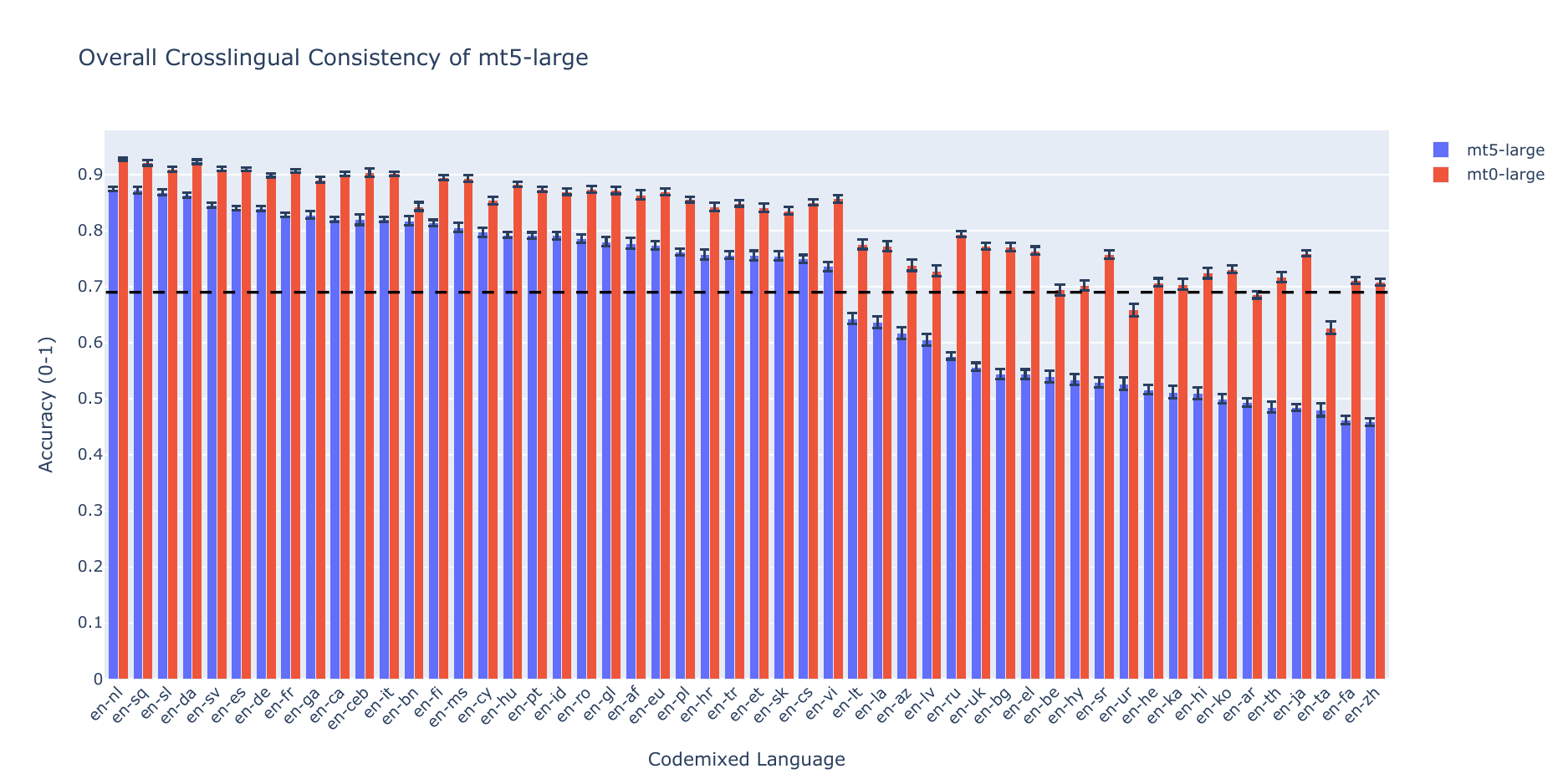}
        \label{app_fig:mt5-large-overall-crosslingual-consistency-acc}
    }\end{subfigure}
    \caption{Effects of multi-task instruction tuning to overall cross-lingual consistency (top: RankC, bottom: Top@1 Accuracy). Note: The dashed line here is the average corresponding consistency scores of mt5-large
across languages}
\label{app_fig:multi-task-instruction-tuning_encoder-decoder-overall}
\end{figure*}

\begin{figure*}[ht]
    \centering
    \begin{subfigure}{\linewidth}{
        \centering
        \includegraphics[trim=0cm 0cm 0cm 2cm,clip=true,width=\linewidth]{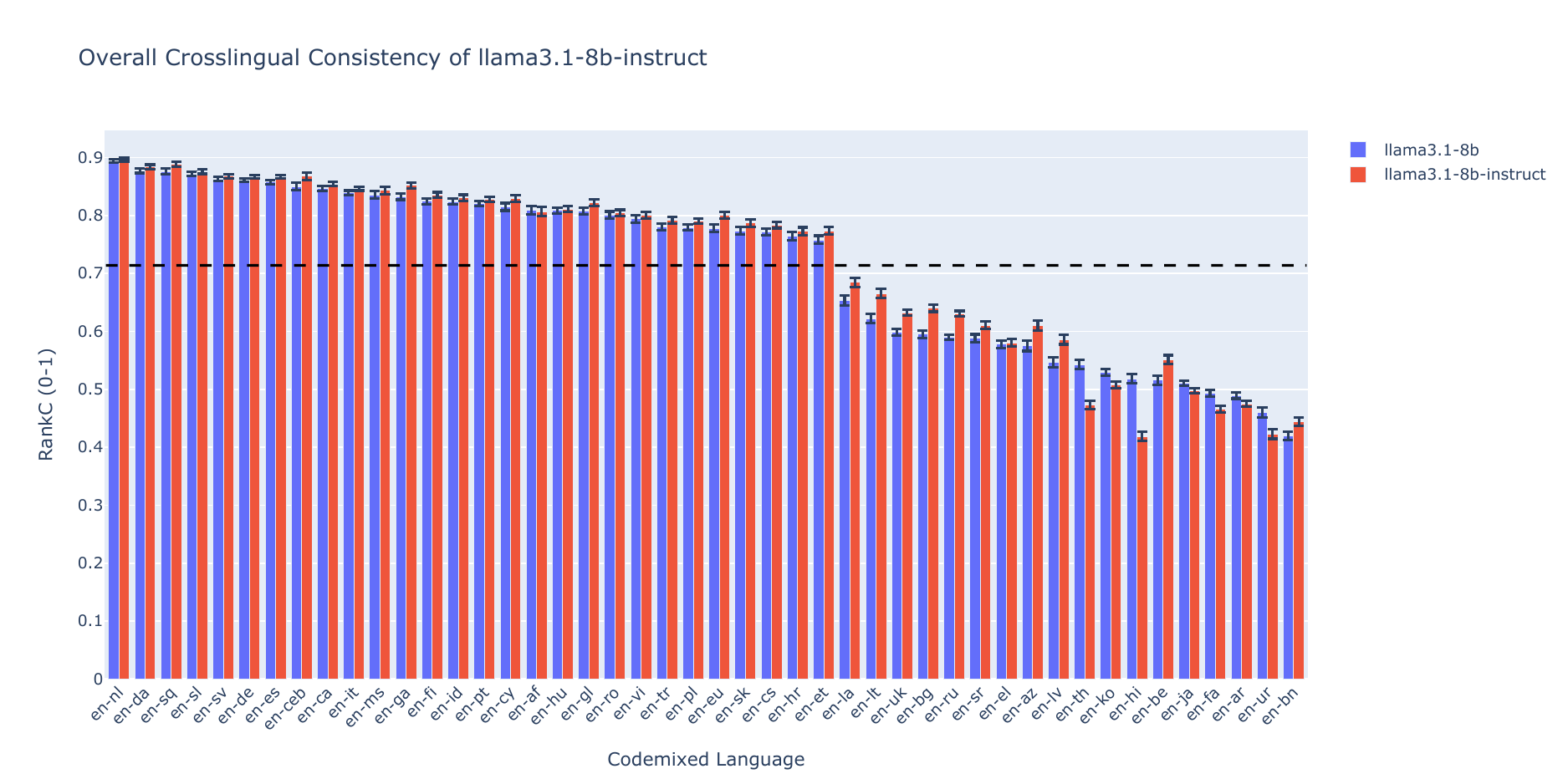}
        \label{app_fig:llama3.1-8b-instruct-overall-crosslingual-consistency-rankC}
    }\end{subfigure}
    \\
    \begin{subfigure}{\linewidth}{
        \centering
        \includegraphics[trim=0cm 0cm 0cm 2cm,clip=true,width=\linewidth]{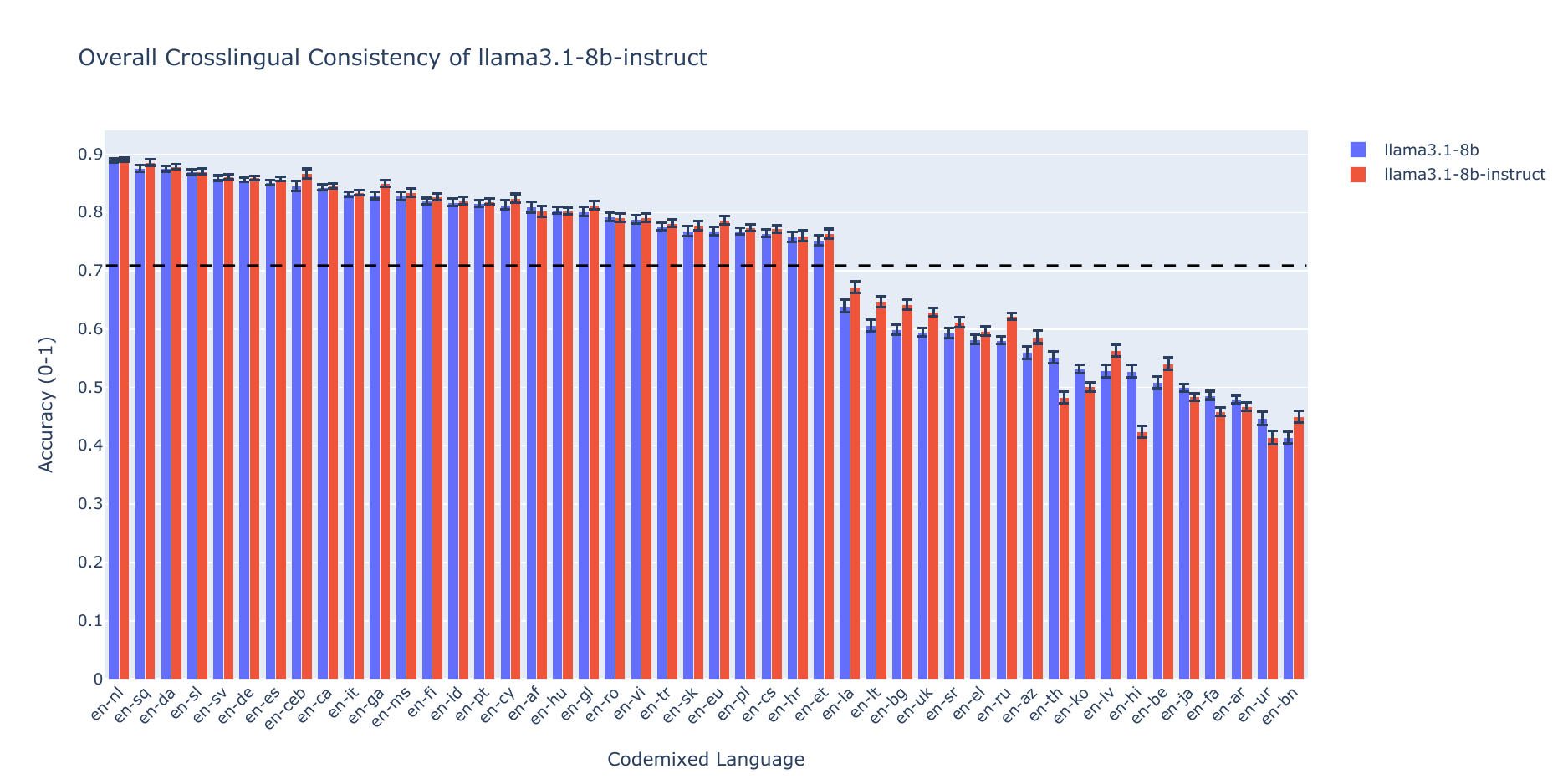}
        \label{app_fig:llama3.1-8b-instruct-overall-crosslingual-consistency-acc}
    }\end{subfigure}
    \caption{Effects of multi-task instruction training to overall cross-lingual consistency (top: RankC, bottom: Top@1 Accuracy). Note: The dashed line here is the average corresponding consistency scores of llama3.1-8b
across languages}
\label{app_fig:multi-task-instruction-tuning_decoder-overall}
\end{figure*}

We hypothesize that method of fine-tuning can improve the cross-lingual consistency due to improved cross-lingual generalization across similar tasks in different languages, as opposed to word-level alignments discussed in previous sections. Surprisingly, multi-task fine-tuning can not offer significant gains to the layerwise cross-lingual consistency. As presented in Figure \ref{app_fig:multi-task-instruction-tuning_encoder-decoder-layerwise} and Figure \ref{app_fig:multi-task-instruction-tuning_decoder-layerwise}, the consistency patterns are quite similar for both type of model families (decoder is represented by llama3.1-8b-instruct, encoder-decoder is represented by mt0-large). Intriguingly, we can more salient enhancement on encoder-decoder models as shown in Figure \ref{app_fig:multi-task-instruction-tuning_encoder-decoder-overall} than decoder models as evident in Figure \ref{app_fig:multi-task-instruction-tuning_decoder-overall} which might suggest the possibility of ratio of multilingual examples in the pretraining corpora could play role on such improvement. 

\subsubsection{Case Study for Transliteration}
\label{app_Transliteration}
\begin{figure}[ht]
    \centering
    \begin{subfigure}{\linewidth}{
        \centering
        \includegraphics[trim=0cm 0cm 0cm 2cm,clip=true,width=\linewidth]{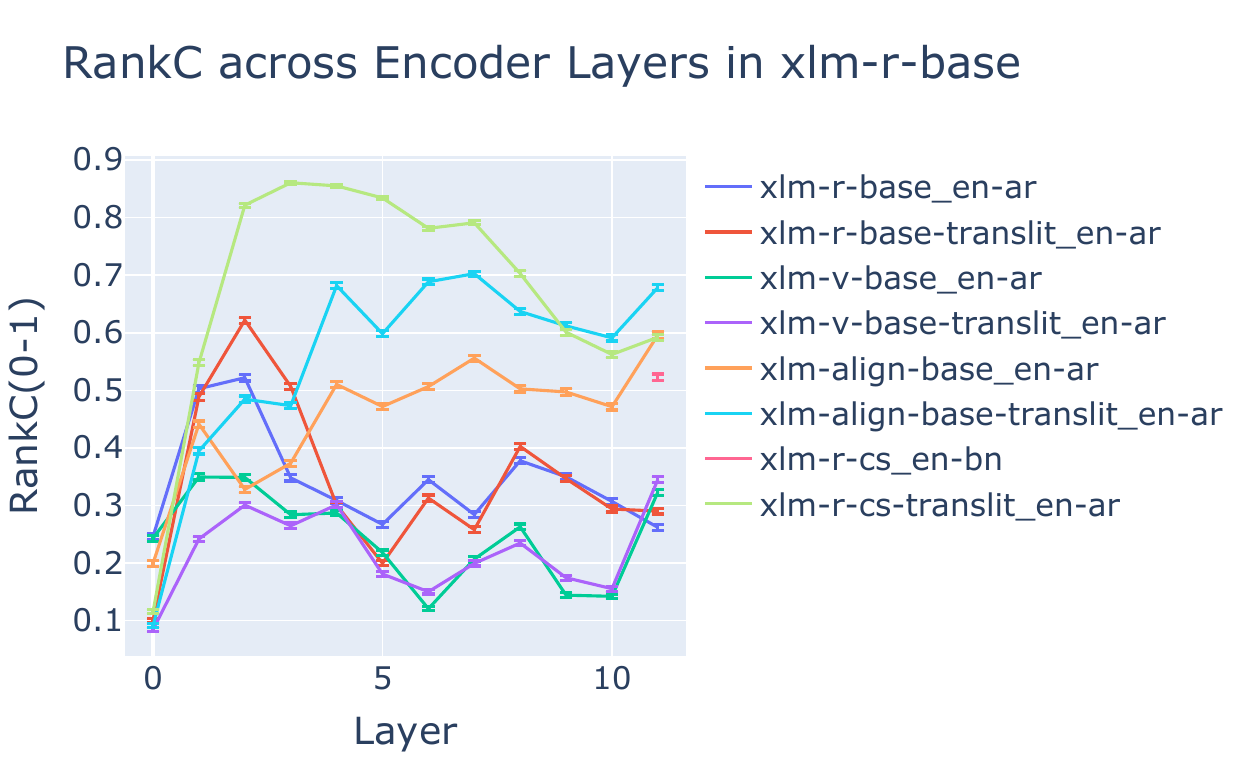}
        \label{fig:xlm-r-transliteration-ar}
    }\end{subfigure}
    \begin{subfigure}{\linewidth}{
        \centering
        \includegraphics[trim=0cm 0cm 0cm 2cm,clip=true,width=\linewidth]{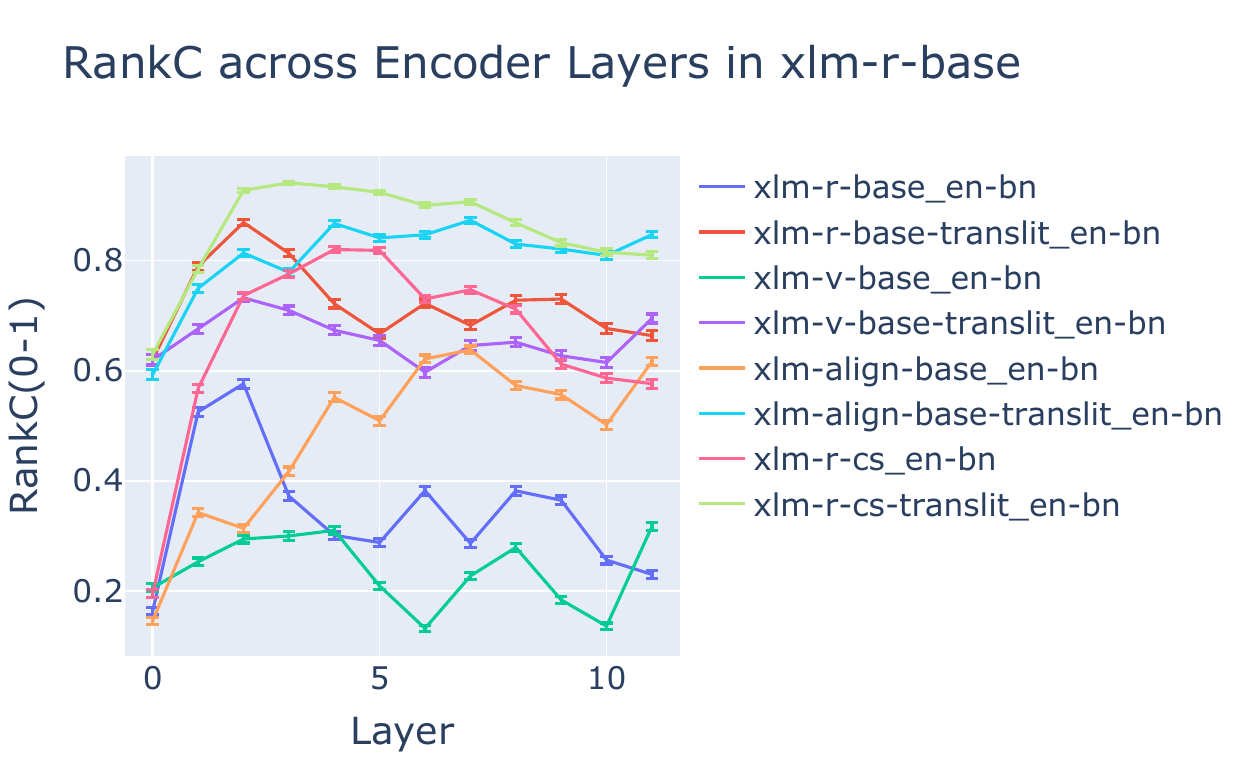}
        \label{fig:xlm-r-transliteration-bn}
    }\end{subfigure}
    \caption{Impact of Transliterations.}
    \label{fig:xlm-r-transliteration}
\end{figure}

Instead of using translations, we transliterate bn\footnote{https://github.com/shhossain/BanglaTranslationKit} and ar\footnote{https://github.com/hayderkharrufa/arabic-buckwalter-transliteration} to understand the impact of writing systems, particularly transliterations. As presented in Figure  \ref{fig:xlm-r-transliteration}, word alignments (or the similar effect from CS training) contribute to the model’s cross-lingual consistency against writing systems because xlm-align and xlm-r-cs show similar performance in both original and transliteration settings. Meanwhile, we can observe that xlm-align and xlm-r-cs significantly improve the overall performance for non-Latin scripts in \S \ref{ssecsec:word-align-effect} \& \S \ref{ssecsec:cs-effect}. This is reasonable as word alignments or CS training help the model link original words with their translations or transliterations, depending on the training corpus, thereby enhancing cross-lingual consistency. We suspect that these word alignments might also improve robustness for handling non-standard spellings and orthographic variations. However, xlm-v-base and xlm-r-base without word alignment benefit from transliterations, which means that xlm-v-base and xlm-r-base do not sufficiently align original words with their transliterations to main cross-lingual consistency. It is also confirmed by the overall performance of vocabulary expansions in \S \ref{sess:findings_in_details_Impact_of_Larger_Vocabulary}, where vocabulary expansions can not offer significant gains for cross-lingual consistency. Overall,  the evaluation task does not inadequately boost consistency for languages using Latin script because word alignments resulting in cross-lingual consistency are the main factor. 

\end{document}